\documentclass[10pt,twocolumn,letterpaper]{article}

\usepackage[pagenumbers]{iccv} 
\definecolor{iccvblue}{rgb}{0.21,0.49,0.74}
\usepackage[pagebackref,breaklinks,colorlinks,allcolors=iccvblue]{hyperref}
\usepackage[accsupp]{axessibility}
\usepackage{url}
\usepackage{booktabs}
\usepackage{multirow}
\usepackage{xspace}
\usepackage{graphicx}
\usepackage{capt-of}
\usepackage{adjustbox}
\usepackage{pifont}
\usepackage{epigraph}
\usepackage{tcolorbox}
\usepackage{amsmath}
\usepackage{float}
\usepackage{stfloats}

\def\paperID{100} 
\def\confName{ICCV}
\def\confYear{2025}
 
\title{\cure: Cultural Gaps in the Long Tail of Text-to-Image Systems}

\author{Aniket Rege$\rm^1$\thanks{Corresponding author: \texttt{aniketr@cs.wisc.edu}} \enspace   Zinnia Nie$\rm ^1$ \enspace Mahesh Ramesh$\rm ^1$ \enspace Unmesh Raskar$\rm ^1$ \enspace Zhuoran Yu$\rm^1$ \\ Aditya Kusupati$\rm ^2$\thanks{Equal Advising} \enspace  Yong Jae Lee$\rm ^1$\footnotemark[2] \enspace Ramya Korlakai Vinayak$\rm ^1$\footnotemark[2] \vspace{1mm}\\
$\rm^1$University of Wisconsin-Madison \enspace $\rm ^2$University of Washington
}

\newcommand{\cure}{\ensuremath{{\rm CuRe}}\xspace}

\usepackage{amsmath,amsfonts,bm}

\def\eqref#1{equation~\ref{#1}}

\def\1{\bm{1}}

\DeclareMathAlphabet{\mathsfit}{\encodingdefault}{\sfdefault}{m}{sl}
\SetMathAlphabet{\mathsfit}{bold}{\encodingdefault}{\sfdefault}{bx}{n}

\begin{document}
\maketitle
\begin{abstract}

Popular text-to-image (T2I) systems are trained on web-scraped data, which is heavily Amero and Euro-centric, underrepresenting the cultures of the Global South. To analyze these biases, we introduce \cure, a novel and scalable benchmarking and scoring suite for cultural representativeness that leverages the marginal utility of attribute specification to T2I systems as a proxy for human judgments. Our \cure benchmark dataset has a novel categorical hierarchy built from the crowdsourced Wikimedia knowledge graph, with 300 cultural artifacts across 32 cultural subcategories grouped into six broad cultural axes (food, art, fashion, architecture, celebrations, and people). Our dataset's categorical hierarchy enables \cure scorers to evaluate T2I systems by analyzing their response to increasing the informativeness of text conditioning, enabling fine-grained cultural comparisons. We empirically observe much stronger correlations of our class of scorers to human judgments of perceptual similarity, image-text alignment, and cultural diversity across image encoders (SigLIP 2, AIMV2 and DINOv2), vision-language models (OpenCLIP, SigLIP 2, Gemini 2.0 Flash) and state-of-the-art text-to-image systems, including three variants of Stable Diffusion (1.5, XL, 3.5 Large), FLUX.1 [dev], Ideogram 2.0, and DALL-E 3. The code and dataset is open-sourced and available at \url{https://aniketrege.github.io/cure/}.
\end{abstract}    
\section{Introduction}
\label{sec:intro}
\epigraph{Cultural diversity brings a collective strength that can benefit all of humanity.}{\textit{Robert Alan Arthur}}

\begin{figure}[t]
    \centering
    \includegraphics[width=\linewidth]{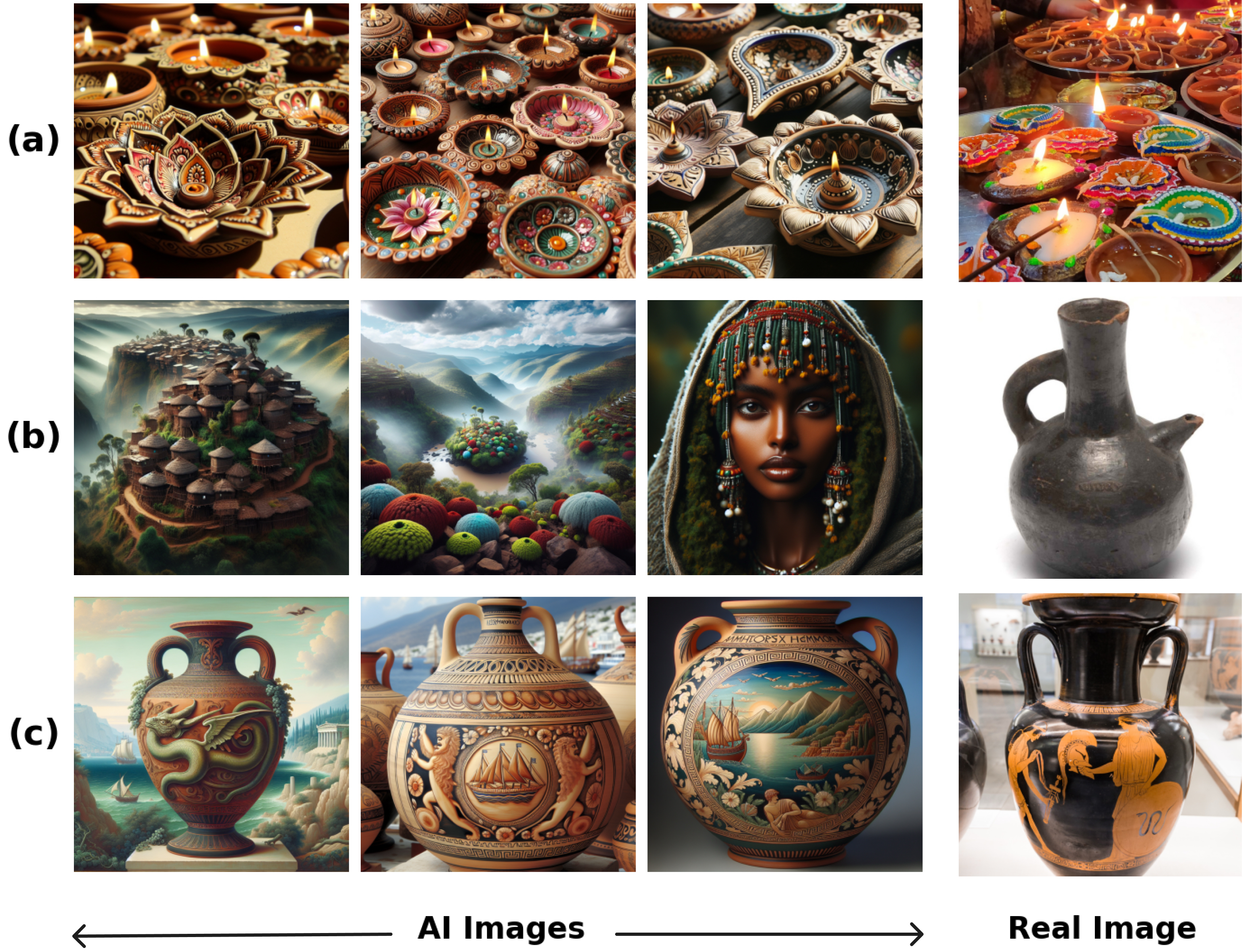}
    \caption{Three images of types of pottery generated by a state-of-the-art T2I system, DALL-E 3, with varying amounts of information in the prompt. The T2I system is consistently accurate at generating \textit{``ceramic diyas''} (a), but hallucinates incorrect details for \textit{``jebena, from Ethiopia''} (b) and \textit{``amphora of Hermonax, a type of pottery from Greece''} (c). Specifying more information in the prompt \textit{can} help (\ie in c but not in b), but is an unreliable method to make the T2I system more culturally representative.}
    \label{fig:teaser}
    \vspace{-5mm}
\end{figure}

\begin{table*}[t]
\centering
\caption{We tabulate existing cultural benchmarks for Text-to-Image Systems, organized by their contributions towards dataset design, quantitative metrics to measure cultural representativeness, and towards the user study. We also tabulate statistics of the \cure dataset compared to existing cultural benchmark datasets. Here $|\mathcal{R}|$ is the number of cultural regions (countries), $|\mathcal{S}|$ the number of cultural axes (supercategories), $|\mathcal{C}|$ the number of cultural categories, and $|\mathcal{N}|$ the total number of cultural artifacts. We also note the number of T2I systems evaluated via a user study, and the number of T2I systems evaluated only quantitatively.}
\resizebox{\textwidth}{!}{
\begin{tabular}{@{}l|ccc|cccc|cccc|cc@{}}
\toprule
\multicolumn{1}{c|}{\multirow{2}{*}{\textbf{Work}}} & \multicolumn{7}{|c|}{\textbf{Dataset}}                                                                                                                                                                                                                                      & \multicolumn{4}{c|}{\textbf{Quantitative Metrics}}                                                                                                                                                                    & \multicolumn{2}{l}{\textbf{\# T2I Systems}}                                                                                               \\ \cmidrule(l){2-14} 
\multicolumn{1}{c|}{}                               & \textbf{\begin{tabular}[c]{@{}c@{}}Crowd-\\ sourced\end{tabular}} & \textbf{Scalable} & \textbf{\begin{tabular}[c]{@{}c@{}}Category\\ Hierarchy\end{tabular}} & \textbf{$|\mathcal{R}|$} & \textbf{$|\mathcal{S}|$} & \textbf{$|\mathcal{C}|$} & \textbf{$|\mathcal{N}|$} & \textbf{\begin{tabular}[c]{@{}c@{}}New\\ Metric\end{tabular}} & \textbf{\begin{tabular}[c]{@{}c@{}}Img-Txt\\ Sim\end{tabular}} & \textbf{\begin{tabular}[c]{@{}c@{}}Img-Img\\ Sim\end{tabular}} & \textbf{Diversity} & \textbf{\begin{tabular}[c]{@{}c@{}}In User\\ Study\end{tabular}} & \textbf{\begin{tabular}[c]{@{}c@{}}Only\\ Quant\end{tabular}} \\ \midrule
\citet{liu2023towards}                             & \checkmark                                                        &                   &                                                                       & 8                        & 9                        & -                        & 1095                     &                                                               &                                                                &                                                                &                    & 1                                                                & 1                                                             \\
\citet{basu2023inspecting}                         &                                                                   &                   &                                                                       & 27                       & -                        & 10                       & -                        &                                                               & \checkmark                                                     & \checkmark                                                     &                    & 2                                                                & 2                                                             \\
\citet{ventura2023navigating}                      &                                                                   &                   &                                                                       & 10                       & 8                        & 200                      & -                        & \checkmark                                                    & \checkmark                                                     & \checkmark                                                     &                    & 3                                                                & 6                                                             \\
\citet{jha2024visage}                              &                                                                   &                   &                                                                       & 135                      & -                        & -                        & -                        & \checkmark                                                    & \checkmark                                                     &                                                                &                    & 1                                                                & 1                                                             \\
\citet{khanuja2024image}                           & \checkmark                                                        &                   &                                                                       & 7                        & -                        & 17                       & 580                      &                                                               & \checkmark                                                     & \checkmark                                                     &                    & 2                                                                & 2                                                             \\
\citet{kannen2024beyond}                           & \checkmark                                                        & \checkmark        &                                                                       & 8                        & 3                        & -                        & 1000                     & \checkmark                                                    &                                                                &                                                                & \checkmark         & 2                                                                & 2                                                             \\
\citet{zhang2024partiality}                        &                                                                   &                   &                                                                       & 10                       & 9                        & -                        & 595                      & \checkmark                                                    &                                                                & \checkmark                                                     & \checkmark         & 3                                                                & 3                                                             \\
\citet{bayramli2025diffusion}                      & \checkmark                                                        & \checkmark        &                                                                       & 10                       & 3                        & -                        & 150                      &                                                               & \checkmark                                                     &                                                                &                    & 3                                                                & 3                                                             \\ \midrule
\cure (Ours)                                             & \checkmark                                                        & \checkmark        & \checkmark                                                            & 64                       & 6                        & 32                       & 300                      & \checkmark                                                    & \checkmark                                                     & \checkmark                                                     & \checkmark         & 3                                                                & 6                                                             \\ \bottomrule
\end{tabular}
}
\vspace{-3mm}
\label{tab:contributions}
\end{table*}
Text-to-Image (T2I) systems~\cite{ramesh2022hierarchical, rombach2022high, podell2024sdxl, saharia2022photorealistic} are trained on web-scale data~\cite{schuhmann2021laion, schuhmann2022laion, gadre2023datacomp, chenpali}, which is long-tail in nature~\cite{parashar2024neglected}. This translates to hallucinations when generating images of data in the tail, as the model has not seen enough examples during training.   This training paradigm has been shown to amplify societal biases and stereotypes encoded in the training data~\cite{bianchi2023easily, cho2023dall}, including harmful and offensive content~\cite{birhane2021multimodal, birhane2024into}. We show a simple example of T2I system bias using a state-of-the-art T2I system~\cite{dalle3} to generate images of culturally specific pottery in~\cref{fig:teaser}. Generating images with the text prompt ``\textit{ceramic diyas}" (a) gives \textbf{diverse, consistent and factual} outputs, while prompting with ``\textit{jebena, from Ethiopia}" (b) and ``\textit{amphora of Hermonax, a type of pottery from Greece}'' (c) generate \textbf{diverse but low quality and inaccurate} images that do not resemble pottery (b), or miss culture-specific details of the artifact (c). Building generative models that faithfully represent the diversity of human preferences, values, and experiences across global cultures in this long tail requires a culturally-aware benchmarking and reliable bias measurement, which is non-trivial. A typical method to measure bias is through large-scale user studies on crowdsourcing platforms~\cite{kannen2024beyond, jha2024visage}, leveraging human judgments for assessment. While this approach accurately reflects feedbacks from humans, it is expensive and difficult to scale. To mitigate this cost, prior works propose proxy scorers to estimate human judgments, such as computing similarity of generated images to real images with large neural encoders~\cite{khanuja2024image}, realism metrics~\cite{heusel2017gans, binkowski2018demystifying}, alignment of images to some desired attributes specified through text~\cite{khanuja2024image, ventura2023navigating} and cultural diversity~\cite{kannen2024beyond}. We find that these proxy scorers empirically do not correlate well with human judgments of cultural representativeness and similarity to ground truth across popular T2I systems (\cref{sec:experiments}). We also highlight the ``\textit{generative entanglement}'' of existing proxy scorers: a miscalibrated estimate of T2I system quality due to overlapping training data between the scorer and T2I system (\cref{subsec:scorer_percepsim}).

\begin{tcolorbox}[colback=lightgray!10, colframe=black, title={Research Aim}]
Our goal is to measure the \textit{cultural representativeness capability} of T2I systems across global cultures, which we call \cure.
\end{tcolorbox}

\begin{figure}[t]
\centering
\includegraphics[width=\linewidth]{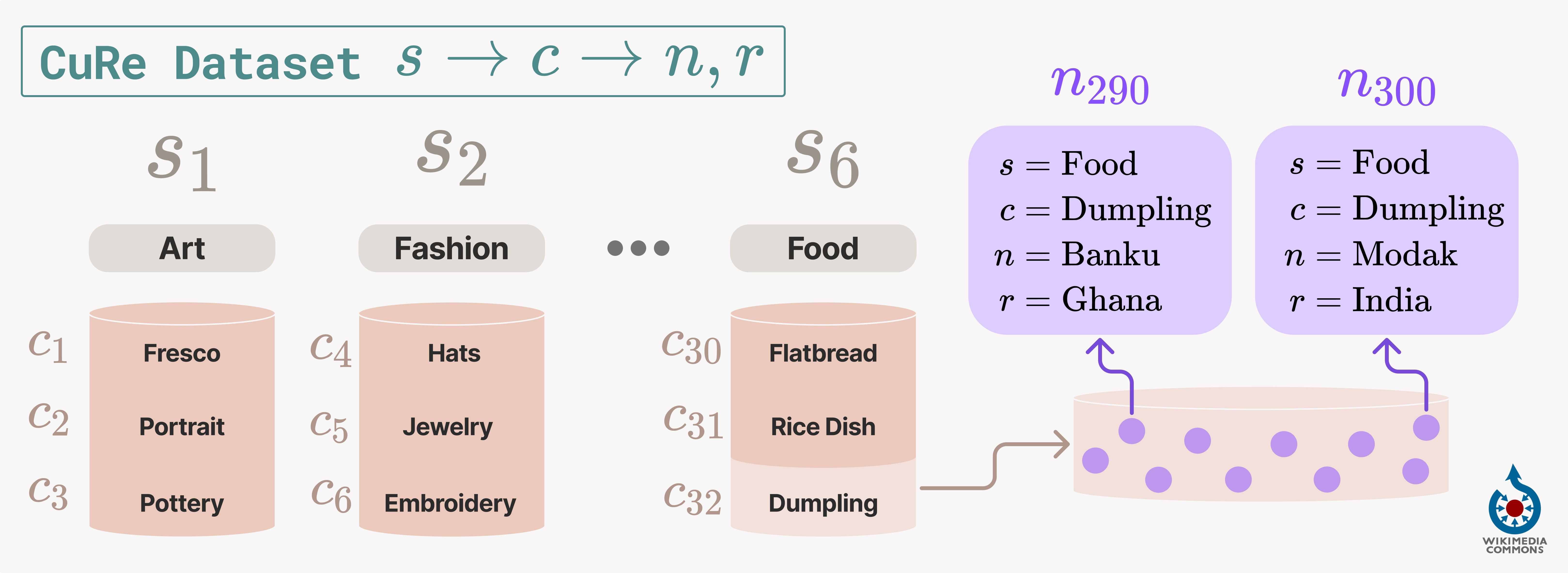}
\caption{The \cure dataset is constructed in a hierarchical fashion ($s\to c\to n,r$) from Wikimedia with 300 cultural artifacts $n\in\mathcal{N}$ grouped into 6 cultural axes $s\in\mathcal{S}$, 32 cultural categories $c\in\mathcal{C}$,  described by a name $n$ and associated region $r$. For example, $s =$ ``food", $c =$ ``dumpling", $n =$ ``banku", $r =$ ``Ghana''.}
\vspace{-4mm}
\label{fig:teaser-cure-dataset}
\end{figure}

To overcome these shortcomings, we propose \cure, a dataset and scorer suite to accurately and efficiently benchmark the cultural representativeness of T2I systems, \ie \textit{how equipped are state-of-the-art T2I systems at accurately generating samples across global cultures that make up their long-tail training data}? We propose a novel framework for scoring cultural representativeness through the lens of \textbf{\emph{marginal utility of increasing attribute specification}}  (\cref{subsec:metric_mia}), which correlates better to human judgments than proxy scorers across T2I systems (see \cref{tab:results_ps}~-~\ref{tab:results_div} in \cref{sec:experiments}). To enable measuring this marginal utility of information, we create a benchmark dataset with a novel categorical hierarchy of attributes, as seen in~\cref{fig:teaser-cure-dataset}.
We compare our dataset and metric design to prior work in~\cref{tab:contributions} and summarize our contributions below:
\begin{itemize}
    \item A \textbf{new dataset} constructed in a scalable fashion directly from the large crowdsourced Wikimedia knowledge graph~\cite{wikimedia} with a \textbf{novel coarse-to-fine categorical hierarchy} of 300 cultural artifacts across six cultural axes, 32 cultural categories and 64 countries (see~\cref{fig:teaser-cure-dataset}).
    \item A novel scoring of cultural representativeness through the lens of \textbf{marginal utility of specifying more information} to the T2I system across cultural attributes (\eg cultural axis, cultural category, and cultural region).
    \item A \textbf{large-scale user study} asking real humans to rate the perceptual similarity, cultural representativeness, offensiveness, and stereotypicalness of T2I systems, alongside detailed freeform feedback about culturally specific failures. We query workers who explicitly identify with the culture of their country  of nationality, which is largely an unverified assumption made in prior works. 
    \item A detailed analysis of how \cure \textbf{scorers correlate to real human judgments} of cultural representativeness and factuality, which highlights the misleading takeaways of popular status quo scorers (\cref{sec:experiments}). 
    \item For the first time, we evaluate the cultural capabilities of a natively multimodal large language model (\cref{subsec:mllm_judge}).
\end{itemize}

\section{Related Work}
\label{sec:related_work}

\begin{figure*}[t]
\centering
\includegraphics[width=\linewidth]{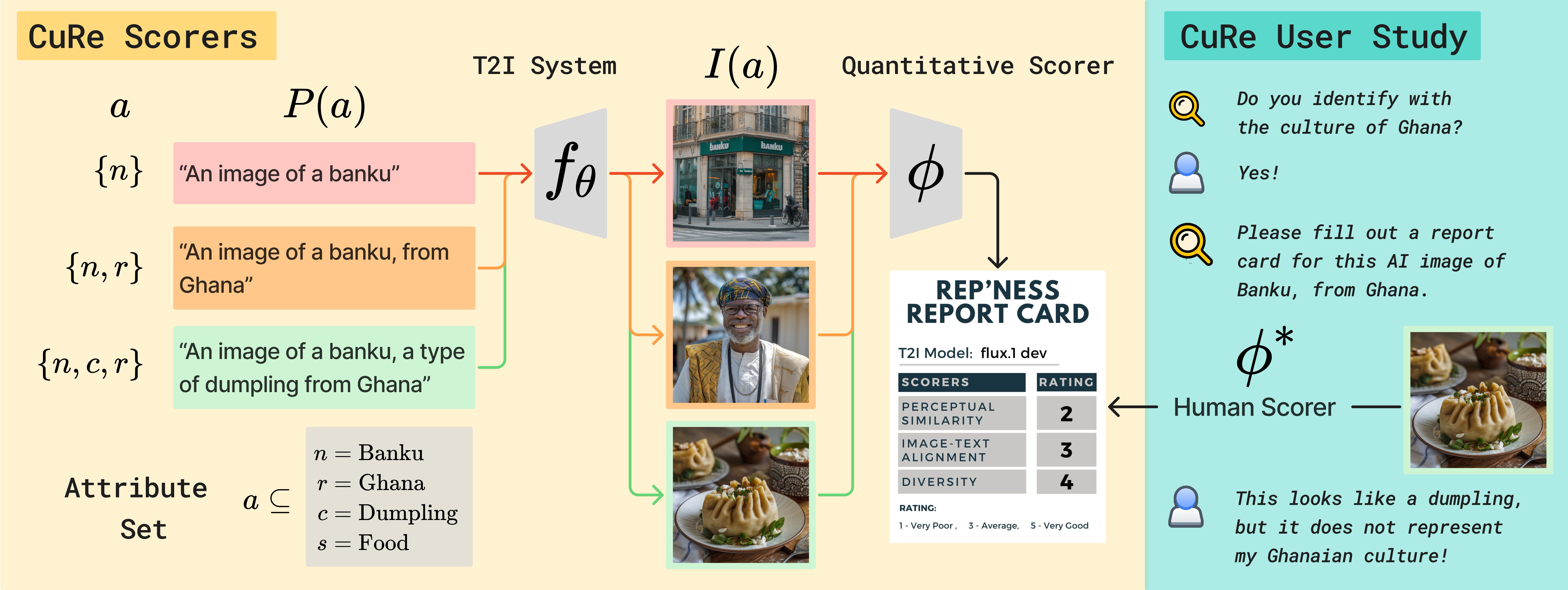}
\caption{An overview of the scorers and user study of \cure. Through the lens of marginal utility of attributes, we analyze images $I(a)$ generated from prompts templates $P(a)$ over attribute subsets $a\subseteq \{s, c, n, r\}$ of the 300 cultural artifacts $n\in\mathcal{N}$ in our benchmark dataset. For example, we generate an image from name and region attributes as $I(\{n,r\}) = f_\theta\bigl(P(\{n,r\})\bigr)$. This is the middle image (orange) in the figure above. These images are then scored by quantitative scorers $\phi$ (see \cref{sec:metrics}) and user judgments from a large-scale user study $\phi^*$ (see \cref{subsec:user_study}). Crucially, cultural artifacts from region $r$ are rated only by workers who identify with the culture of region $r$.} 
\label{fig:teaser-cure-system}
\vspace{-3mm}
\end{figure*}

\paragraph{Dataset Biases}  Datasets used to train generative models are known to have biases, including the Amerocentric and Eurocentric \textit{geographical distribution} bias~\cite{shankar2017geodiversity, de2019does} of crowd-labeled datasets such as ImageNet~\cite{russakovsky2015imagenet}, Open Images~\cite{krasin2017openimages} and MS COCO~\cite{lin2014microsoft}. Several works also highlight the \textit{data collection bias}; \eg, on YFCC100m~\cite{thomee2016yfcc100m}, the data from underrepresented countries is often taken by tourists ($\sim 47$\%~\cite{wang2022revise}), and does not capture the true local distribution of objects, people, and language. These dataset biases propagate to T2I systems after training.

\vspace{-10pt}
\paragraph{Text-to-Image System Biases}
T2I systems are predominantly built on diffusion models ~\cite{rombach2022high, saharia2022photorealistic}, autoregressive Transformer models~\cite{ramesh2022hierarchical, ding2022cogview2, yu2022scaling} or GANs~\cite{kang2023scaling, sauer2023stylegan}. The pretraining datasets of these models are web-scale~\cite{schuhmann2022laion} and long-tail~\cite{parashar2024neglected}, and there have been several recent works examining the biases present within T2I systems. Several recent works analyze the biases of T2I systems from a geographical~\cite{basu2023inspecting}, gender occupation~\cite{seshadri2024bias}, cultural~\cite{bianchi2023easily, turk2023how} and social~\cite{luccioni2023stable, naik2023social} perspective. \citet{luccioni2023stable} provide an excellent overview of T2I system biases, from the data collection~\cite{birhane2021multimodal} and filtering~\cite{rando2022red} to model training~\cite{wolfe2022american}. 

\vspace{-10pt}
\paragraph{Cultural Representativeness Metrics}
To measure \cure on benchmark datasets, prior works use proxy scorers such as using deep image encoders to compute similarity of generated images to real images~\cite{khanuja2024image}, which contain similar dataset biases as T2I systems themselves~\cite{seshadri2024bias}. Another class of scorers use realism metrics~\cite{heusel2017gans, binkowski2018demystifying}, which ignore culture-specific nuance. Lastly, prior works evaluate image-text alignment to carefully chosen prompts~\cite{khanuja2024image, ventura2023navigating} and cultural diversity as a proxy for representativeness~\cite{kannen2024beyond}, which often empirically do not correlate well human judgments across global cultures (see \cref{sec:experiments}).

\vspace{-10pt}
\paragraph{Cultural Benchmarks}
To create cultural benchmarks to measure T2I biases, previous works typically crowdsource data, either directly from workers on online platforms~\cite{khanuja2024image, prolific} or with massive inter-organization efforts~\cite{romero2025cvqa}. Some works also rely on cultural experts to create the data~\cite{zhang2024partiality, liu2023towards}. While these methods can give high quality data, they are expensive and inscalable, as to add new data to the benchmark, new workers or experts must be hired each time. To measure cultural biases, prior works rely on querying users across global cultures for their judgment on images generated by T2I systems~\cite{khanuja2024image, basu2023inspecting, kannen2024beyond}. We overcome these limitations by constructing our benchmark \cure directly from the live Wikimedia graph~\cite{wikimedia}, by traversing parent nodes (cultural axes) and grouping child nodes (cultural categories) by region. This methodology is both cheap and scalable, as new categories can be added to our benchmark on-the-fly by crawling Wikimedia.

\vspace{-10pt}
\paragraph{Multimodal Language Models}
There has been a significant recent effort towards extending language models, which only understand text, to multimodal large language models (MLLMs) that can understand both text and images~\cite{liu2023visual, gemini, wang2024emu3, yu2023scaling, tong2024metamorph}. While the details of the pretraining data of state-of-the-art MLLMs is typically hidden or proprietary, they are significantly larger than previous Vision-Language Models like CLIP \cite{radford2021learning, schuhmann2022laion} and perform very well on complex visual question answering benchmarks~\cite{antol2015vqa, romero2025cvqa}. They can thus be directly queried for culture-specific knowledge, similar to existing works that use MLLMs as a judge~\cite{chen2024mllm}. Even at this scale of pretraining data, we show that MLLMs still fall short at evaluating cultural representativeness (\cref{subsec:mllm_judge}).
\section{\cure Dataset}
\label{sec:dataset}

Creating a high quality cultural benchmark is non-trivial for two primary reasons: a) it requires significant crowdsourcing efforts~\cite{khanuja2024image, kannen2024beyond, romero2025cvqa} or hiring domain experts~\cite{liu2023towards, zhang2024partiality} b) it requires good ``cultural coverage", \ie collecting cultural artifacts across a large number of cultural regions $r$ and cultural categories. We address the first difficulty by designing a scalable dataset construction methodology that enables democratic scaling, as any cultural artifact of interest can easily be added to the benchmark by querying Wikimedia\footnote{We open-source our dataset at \url{https://github.com/aniketrege/cure-bench}}. We address the second difficulty by collecting cultural artifacts across 64 countries, which is higher than all existing cultural benchmarks (\cref{tab:contributions}).

\vspace{-10pt}
\paragraph{The necessity of Categorical Hierarchy} \textit{Culture} has a sense of shared values through lived experiences and one's surroundings (intra-culture), which differ greatly across geographies (inter-culture)~\cite{hofstede2001culture}. An important goal towards accurately measuring cultural representativeness of T2I systems is to capture their behavior at both inter and intra-culture levels. For instance, \textit{Cuisine} is considered an important axis of culture~\cite{counihan2013food}, yet it is difficult to compare how well T2I systems do at generating cuisine of, for example, the United States compared to Nigeria, as cuisine has very high \textit{intra-class variance} (diversity). We show that aggregating bias measurement at such coarse levels can lead to misleading takeaways about T2I performance (see diversity measured by Vendi scores in \cref{subsec:results_diversity}). To mitigate this, for each cultural axis, we propose comparing T2I performance at a finer granularity of cultural categorization. For example, many cultures around the world have their own form of dumpling. While all these forms have the same core structure (\ie much lower intra-class variance than cuisine as a whole), they vary greatly in their ingredients, preparation, presentation, etc. specific to each culture. We design the \cure dataset with a novel coarse to fine hierarchy to capture these nuances of cultural categorization.

\cure consists of a set of 300 cultural artifacts $n\in\mathcal{N}$ organized in a categorical hierarchy of attributes: a name $n$ (\eg ``modak"), a cultural category $c$ (\eg ``dumpling"), a cultural supercategory $s$ (\eg ``food"), and a cultural region of origin $r$ (\eg India). The \cure dataset has six cultural axes (or supercategories) $\mathcal{S}$, \ie architecture, art, celebrations, fashion, food, people, and each supercategory $s\in\mathcal{S}$ contains exactly 50 artifacts.
We show an illustration of the \cure dataset ($s\to c\to n, r$) in \cref{fig:teaser-cure-dataset} and tabulate the $s\to c$ hierarchy of the dataset in~\cref{tab:dataset_structure} (\cref{app:dataset}). 
\section{Measuring \cure}
\label{sec:metrics}
We describe our proposed measure of cultural representativeness capabilities of a T2I system, \cure, through the novel lens of marginal utility of information in \cref{subsec:metric_mia}. We introduce this lens into three prominent existing classes of quantitative scorers: approximating human \textbf{perceptual similarity} via cosine distance between the generated image and the ground truth 
(\cref{subsec:scorer_percepsim}), evaluating \textbf{image-text alignment} with a desired attribute specified through text as a proxy for visual question-answering (\cref{subsec:scorer_imtxt_align}), and measuring the \textbf{diversity} of images generated by the T2I system across global cultures (\cref{subsec:scorer_diversity}).  We begin by setting up some useful notation.

\vspace{-10pt}
\paragraph{Notation:}

To generate an image of a cultural artifact $n$, we must choose how to describe it to a T2I system using some subset of artifact attributes $a \subseteq \{s, c, n, r\}$. To this end, we construct a text prompt $P$ with a template over these attributes $a$. We summarize an exhaustive list of templates we use in~\cref{tab:prompt_templates} (\cref{app:dataset}). We then pass this text prompt $P(a)$ to a T2I system $f$ parametrized by weights $\theta$, which generates an image $I$, \ie
\begin{align*}
    I(a)  &= f_\theta\bigl(P(a)\bigr), \text{where} \\
    a &\subseteq \{s, c, n, r\}
\end{align*}

We show an example of three prompt templates and their corresponding images over three attribute subsets for banku, a type of dumpling from Ghana in~\cref{fig:teaser-cure-system}. As seen in the figure, for this cultural artifact, the choice of $a$ has a large impact on the generated image. Choosing $a = n$ or $a = \{n, r\}$ generates images from incorrect classes, \ie a bank (in red) and a man (in orange). When we add category information informing the T2I system that banku is a type of dumpling, \ie $a = \{n, c, r\}$, the image generated (in green) much more resembles real images of banku. To measure cultural representativeness capability of T2I system \( f_\theta \) for region \( r \), we need a method of scoring the quality of generated images $I$. Let \(\phi : I \to \mathbb{R}\) be a quality scorer for \cure (\eg similarity of $I$ to ground-truth images of artifact $n$).

\paragraph{Choosing a Scorer:}
The gold standard for quality scorer $\phi$ is to survey a large number of people from region $r$ to rate T2I system performance along pre-defined rubrics such as realism and image-text alignment, typically with a 1-5 scale Likert score~\cite{likert1932technique}, which we denote $\phi^*$ (see~\cref{fig:teaser-cure-system}). We interchangeably refer to Likert scores from the user study $\phi^*$ as ``\textbf{gold scores}'', as they are rated by humans who identify with the culture of $r$. Designing surveys to collect these human perceptual scores in this manner is non-trivial: eliciting calibrated scores is difficult and launching surveys is expensive~\cite{lazar2017research}. If enough people are not queried in the survey, there is also the possibility their scores may not correlate well with the opinions of people who will use the T2I system after deployment~\cite{dillman2014internet}. To get around these difficulties, existing works design automated quantitative proxy scorers for cultural representativeness based on related but distinct goals like image perceptual similarity, image-text alignment, and diversity. These class of scorers have unique strengths and weaknesses, and we find they empirically do not correlate strongly to human judments of quality (\cref{sec:experiments}).

\subsection{Finding a \cure through Information}
\label{subsec:metric_mia}
Our insight towards designing a more accurate, reliable scorer for \cure across global cultures was motivated by a question: \emph{how many attributes of a cultural artifact does a T2I system need to know in order to faithfully generate an image?} We illustrate our insight through an example in~\cref{fig:teaser-cure-system}. We observe that simply specifying the name $n$ and cultural region $\{n, r\}$ associated with banku, a type of dumpling from Ghana, is insufficient for current state-of-the-art T2I systems (Stable Diffusion 3.5 Large, FLUX.1 [dev], Ideogram 2.0) to produce an image faithful to its real-world counterpart. When also specifying the category $\{n,c,r\}$, the T2I system is able to generate an image that appears more faithful to banku. When this image is passed to existing quantitative scorers $\phi$, they tend to overestimate its cultural representativeness (see examples in ~\cref{fig:metric_comparisons}). However, when this image is shown to a real human who identifies as culturally Ghanaian ($\phi^*$), they highlight that this image looks like a generic dumpling and does not reflect their cultural context of Ghanaian cuisine. Inspired by this, we propose a hypothesis complementary to these existing quantitative scorers:
\begin{tcolorbox}[colback=lightgray!10, colframe=black, title={Key Insight}]
Evaluating how the change in information \textit{explicitly provided} to a T2I system changes its behavior reveals valuable insights into its cultural representativeness capabilities.
\end{tcolorbox}

Unlike existing methods, this approach measures how well a T2I system internalizes cultural knowledge by analyzing the \textbf{marginal utility of each additional attribute} specified during generation. For example, with the ``banku'' artifact in \cref{fig:teaser-cure-system}, we need to marginally increase attributes specified to the T2I system via the prompt $a: n\to \{n,r\}\to \{n,c,r\}$ for culturally accurate T2I generation. We refer to this class of scorers as \textbf{Marginal Information Attribution} (MIA) scorers.  For each class of proxy scorer, we show qualitative examples of the capability of MIA-based scorers to differentiate cultural artifacts with vastly different human-rated scores for perceptual similarity and representativeness, while existing scorers are unable to do so, in ~\cref{fig:metric_comparisons}. We demonstrate this behavior quantitatively over the entire \cure dataset in~\cref{sec:experiments}. We formally define the variant of MIA scorer for each class of proxy scorer in the relevant section below.

\begin{figure}[t]
\centering
\includegraphics[width=\linewidth]{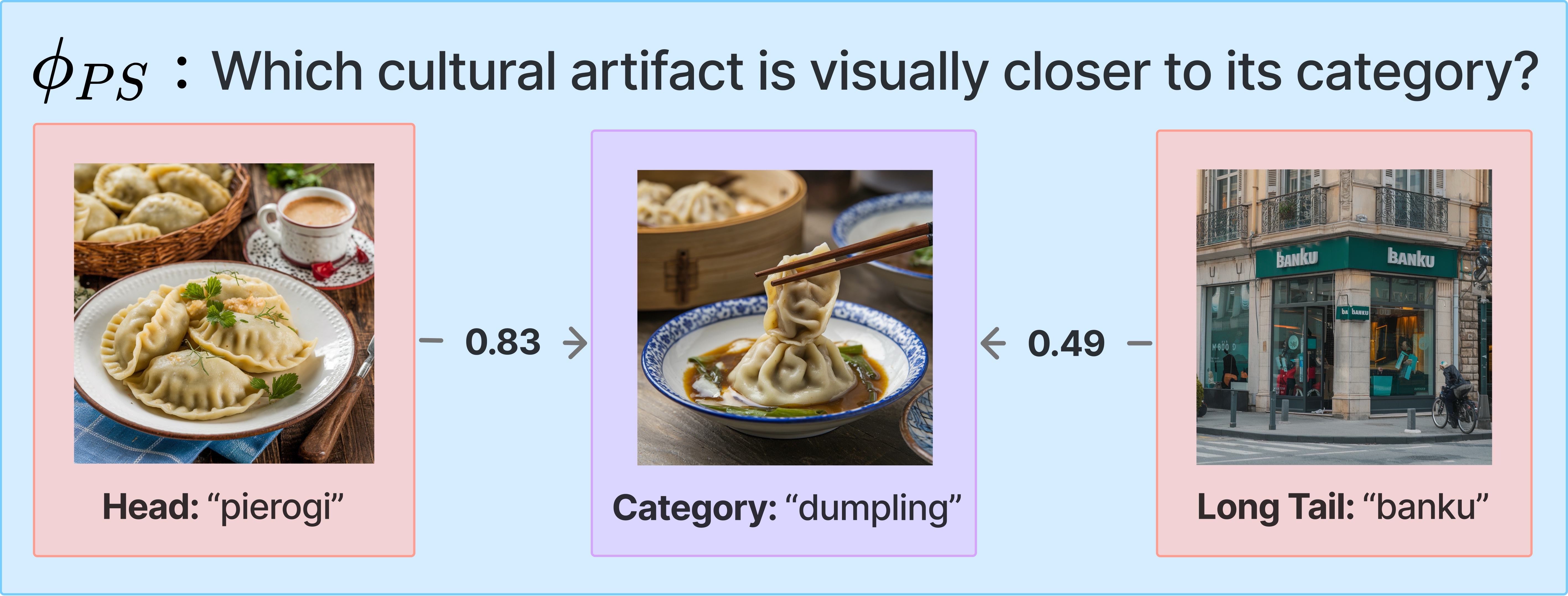}
\caption{There is a large drop in $\phi_{PS}\in[0,1]$, \ie how visually similar images of cultural artifacts $I(n)$ are to images of their cultural category $I(c)$, from the \textit{head} $\to$ the \textit{long tail} of the T2I system (\textit{pierogi} = 0.83 $\to$ \textit{banku} = 0.49).}
\vspace{-4mm}
\label{fig:ps_scorer_fig}
\end{figure}

\subsection{Perceptual Similarity Scorers}
\label{subsec:scorer_percepsim}

\begin{figure*}[!t]
\centering
\begin{tabular}{ccc}
    \begin{subfigure}[t]{0.30\textwidth}
        \centering
        \begin{adjustbox}{width=\columnwidth, height=2cm}
        {\fontsize{11}{11}\selectfont
        \begin{tabular}{@{}clc@{}}
        \toprule
        \textbf{Images} & \textbf{Metric} & \textbf{$\phi$} \\ \midrule
        \multirow{5}{*}{\Large $n_1$}
        \multirow{5}{*}{\includegraphics[width=0.8\textwidth]{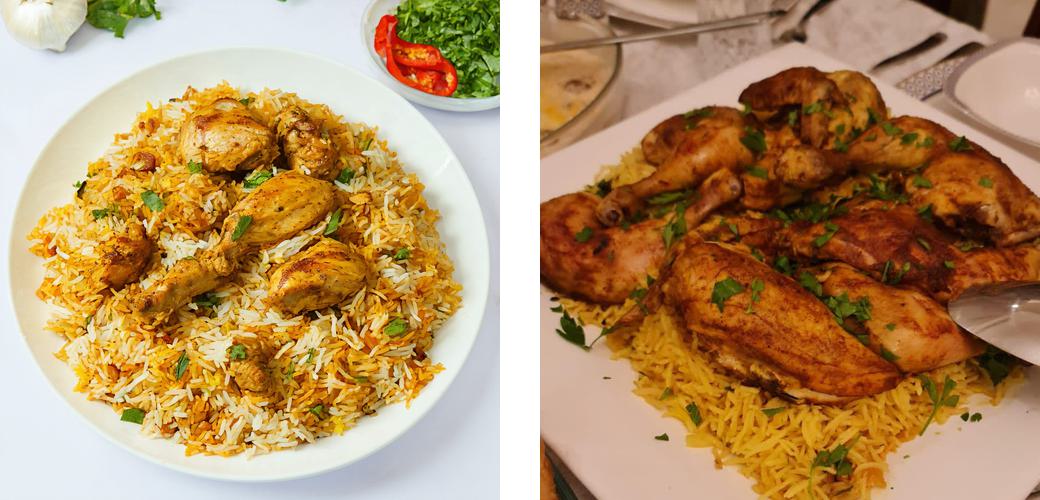}}
        & \cite{oquabdinov2}   & 0.67 \\
        & \cite{tschannen2025siglip}    & 0.79 \\
        & $\phi^*_{\cure}$                     & 0.83 \\
        & $\phi^*_{PS}$                        & 0.75 \\
        \cmidrule(lr){2-3}
        & $\phi_{PS}\downarrow$                 & 0.49 \\
        \cmidrule(lr){2-3}
        \multirow{5}{*}{\Large$n_2$} 
        \multirow{5}{*}{\includegraphics[width=0.8\textwidth]{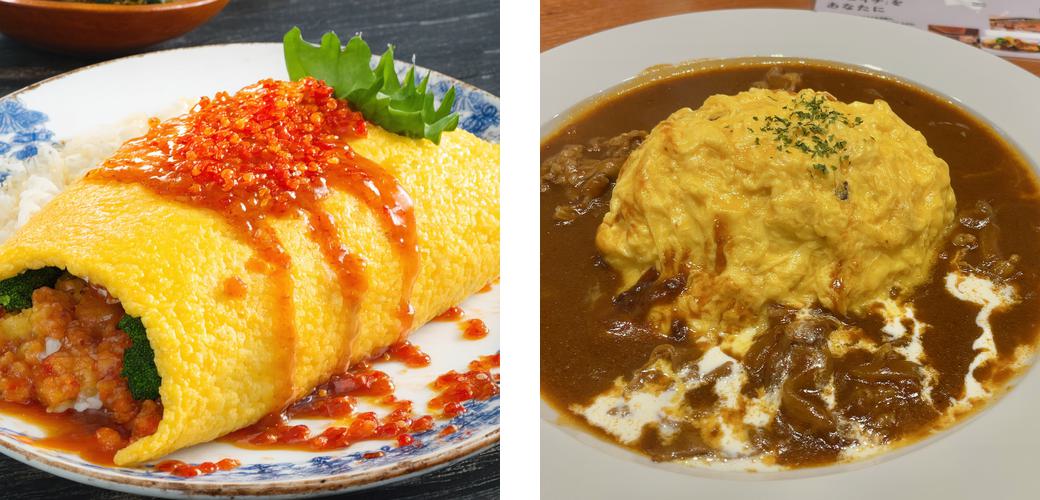}}  
        & \cite{oquabdinov2}  & 0.62 \\
        & \cite{tschannen2025siglip}  & 0.71 \\
        & $\phi^*_{\cure}$   & 0.31 \\
        & $\phi^*_{PS}$ & 0.44 \\ 
        \cmidrule(lr){2-3}
        & $\phi_{PS}\downarrow$  & 0.65 \\ 
        \cmidrule(lr){2-3}
        \multirow{1}{*}{%
          \begin{tabular}{c c}
            \textbf{~~~~~~~AI : $I(n)$} & \hspace{3mm} \textbf{~~Real : $G(n)$}
          \end{tabular}
        }
        \end{tabular}
        }
        \end{adjustbox}
        \vspace{2mm}
        \caption{Perceptual Similarity Scorer ($\phi_{PS}$). \textbf{$n_1$:} ``Omurice'', \textbf{$n_2$:} ``Chicken Biryani''.  Images were generated with SD 3.5 Large.}
        \label{tab:ai_vs_real_1}
    \end{subfigure}
    &
    \begin{subfigure}[t]{0.30\textwidth}
        \centering
        \begin{adjustbox}{width=\columnwidth, height=2cm}
        {\fontsize{11}{11}\selectfont
        \begin{tabular}{@{}clc@{}}
        \toprule
        \textbf{Images} & \textbf{Metric} & \textbf{$\phi$} \\ \midrule
        \multirow{5}{*}{\includegraphics[width=0.8\textwidth]{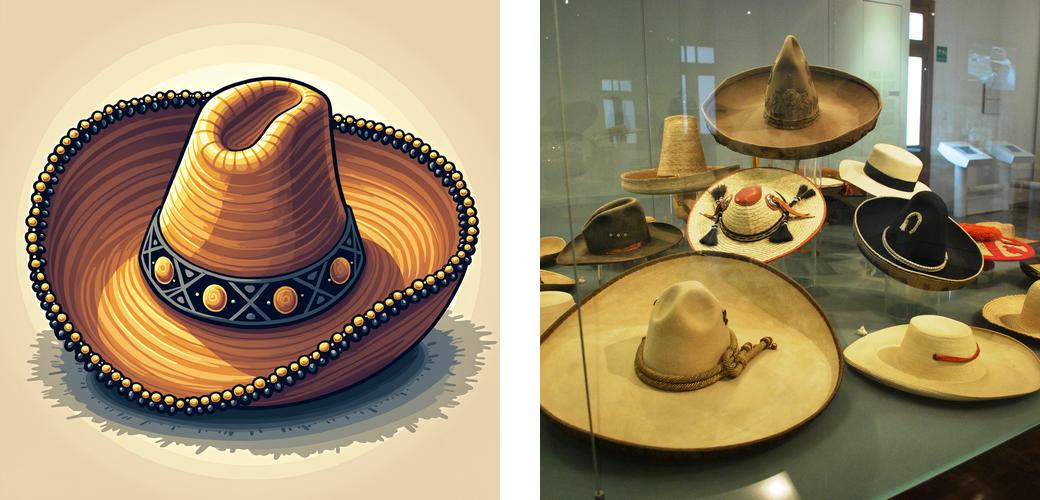}}  
        & \cite{khanuja2024image}       & 0.13 \\
        & \cite{ventura2023navigating}  & 0.11 \\
        & $\phi^*_{\cure}$                       & 1.00 \\
        & $\phi^*_{PS}$                     & 0.75 \\
        \cmidrule(lr){2-3}
        & $\phi_{ITA}\uparrow$                & 0.14 \\
        \cmidrule(lr){2-3}
        \multirow{5}{*}{\includegraphics[width=0.8\textwidth]{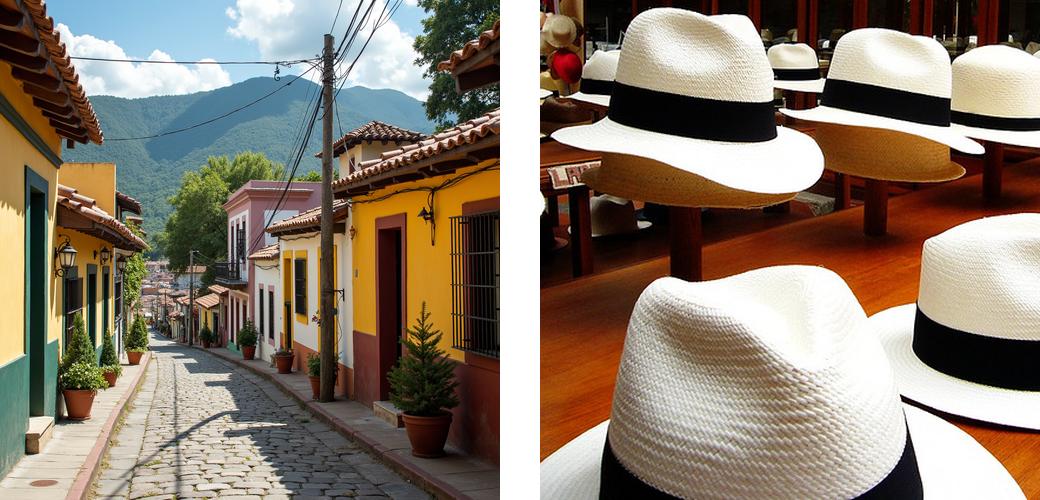}}  
        & \cite{khanuja2024image}       & 0.11 \\
        & \cite{ventura2023navigating}  & 0.09 \\
        & $\phi^*_{\cure}$                       & 0.17 \\
        & $\phi^*_{PS}$                     & 0.15 \\
        \cmidrule(lr){2-3}
        & $\phi_{ITA}\uparrow$                 & 0.01 \\
        \cmidrule(lr){2-3}
        \multirow{1}{*}{%
          \begin{tabular}{c c}
            \textbf{AI : $I(n)$} & \hspace{3mm} \textbf{Real : $G(n)$}
          \end{tabular}
        }
        \end{tabular}
        }
        \end{adjustbox}
        \vspace{2mm}
        \caption{Image-Text Alignment Scorer ($\phi_{ITA}$). \textbf{$n_1$:} ``Sombrero'', \textbf{$n_2$:} ``Toquilla''. Images were generated with FLUX.1 [dev].}
        \label{tab:ai_vs_real_2}
    \end{subfigure}
    &
    \begin{subfigure}[t]{0.30\textwidth}
        \centering
        \begin{adjustbox}{width=\columnwidth, height=2cm}
        {\fontsize{11}{11}\selectfont
        \begin{tabular}{@{}clc@{}}
        \toprule
        \textbf{Images} & \textbf{Metric} & \textbf{$\phi$} \\ \midrule
        \multirow{5}{*}{\includegraphics[width=0.8\textwidth]{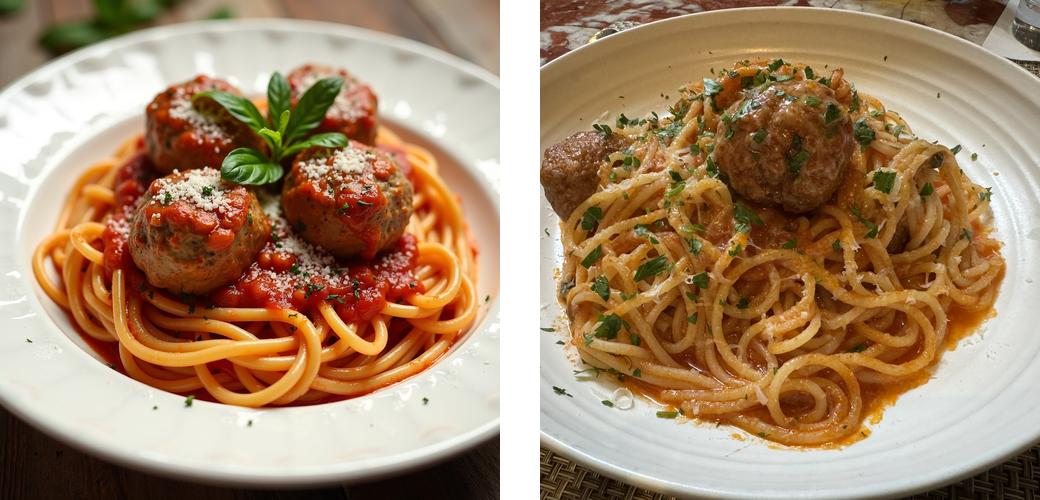}}  
        & LPIPS(n)       & 0.72 \\
        & VS(c)  & 0.24 \\
        & $\phi^*_{\cure}$                       & 0.93 \\
        & $\phi^*_{PS}$                     & 0.66 \\
        \cmidrule(lr){2-3}
        & $\phi_{DIV}\downarrow$                  & 0.57 \\  
        \cmidrule(lr){2-3}
        \multirow{5}{*}{\includegraphics[width=0.8\textwidth]{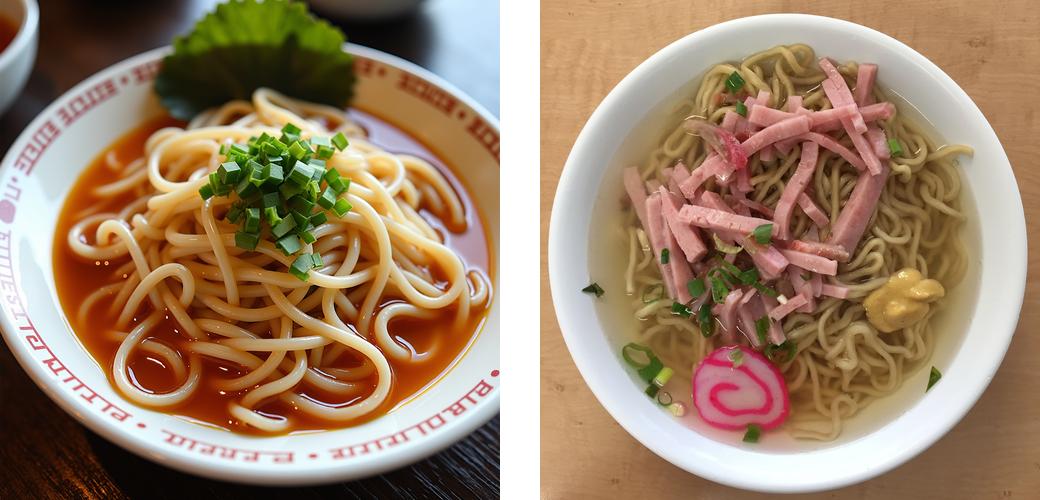}}  
        & LPIPS(n)       & 0.70 \\
        & VS(c)  & 0.24 \\
        & $\phi^*_{\cure}$                       & 0.46 \\
        & $\phi^*_{PS}$                     & 0.40 \\
        \cmidrule(lr){2-3}
        & $\phi_{DIV}\downarrow$                  & 0.79 \\
        \cmidrule(lr){2-3}
        \multirow{1}{*}{%
          \begin{tabular}{c c}
            \textbf{AI : $I(n)$} & \hspace{3mm} \textbf{Real : $G(n)$}
          \end{tabular}
        } 
        \end{tabular}
        }
        \end{adjustbox}
        \vspace{2mm}
        \caption{Diversity Scorer ($\phi_{DIV}$). \textbf{$n_1$:} ``Spaghetti and meatballs'', \textbf{$n_2$:} ``Saimin''. Images were generated with FLUX.1 [dev].}
        \label{tab:ai_vs_real_3}
    \end{subfigure}
\end{tabular}
\caption{A qualitative comparisons of our proposed MIA scorers compared to baselines for three scorer classes: a) Perceptual Similarity (\cref{subsec:scorer_percepsim}) ; b) Image-Text Alignment (\cref{subsec:scorer_imtxt_align}) ; c) Diversity (\cref{subsec:scorer_diversity}). In the figure above, $\phi^*_{\cure}$ and $\phi^*_{PS}$ are human judgments of cultural representativeness and perceptual similarity from our user study (\cref{subsec:user_study}) normalized from their original 1 - 5 Likert scale to a 0 - 1 scale. The top row of images ($n_1$) represent T2I generations of artifacts rated highly by humans ($\phi^*_{\cure}, \phi^*_{PS} \to 1$), while the bottom row ($n_2$) represents low rated artifact generations ($\phi^*_{\cure}, \phi^*_{PS} \to 0$). A lower divergence in perceptual similarity with marginally increasing information captured by our scorer ($\phi_{PS}\downarrow$) and a higher alignment of image features to textual queries for representativeness ($\phi_{ITA} \uparrow$) correspond to better human judgments $\phi^*$. While diversity in image generation is desirable, we empirically confirm prior observations of an inverse relationship~\cite{kannen2024beyond} between human judgments and quantitative measures of cultural diversity ($\phi_{DIV}\downarrow$).  We observe that across scorer classes, our proposed scorers can differentiate between $n_1$ and $n_2$, while baseline methods treat them nearly the same.}
\vspace{-3mm}
\label{fig:metric_comparisons}
\end{figure*}

The goal of perceptual similarity (\textbf{PS}) scorers is to compute the similarity between a generated image $I$ and another set of images representing the same artifact $n$, which are typically manually-curated ground-truth (GT) images $G$. While these scorers correlate reasonably well with human perceptual similarity judments, collecting appropriate ground-truth is expensive and occasionally infeasible. The gold scorer for perceptual similarity is $\phi^*_{PS}(a) = \text{Likert}(I(a), G)$, where a Likert score~\cite{likert1932technique} of 1 indicates very low perceptual similarity and 5 indicates very high similarity. As getting these gold scores is inscalable and expensive, proxy quantitative scorers are used~\cite{he2016deep, dosovitskiy2020image}, which we call $\phi_{GT}$. The proxy scorers often ignore when two images are semantically similar in favor of spatial and textural consistency~\cite{geirhos2018imagenet}, which causes a discrepancy between real human perceptual similarity ratings (see~\cref{tab:ai_vs_real_1}). 

To overcome these limitations, we propose a \textbf{marginal information attribution} scorer, $\phi_{PS}$, that compares T2I image features of cultural artifacts specified only by their name $a = n$ to images generated with only their categorical and region information $a = c$ or $a = \{c, r\}$. We hypothesize that if these images are similar, the T2I system has learned the artifact's cultural association well (\eg $a: n\to \{c, r\}$) and the artifact likely lies in the head of the T2I system's distribution (\cref{fig:ps_scorer_fig} for $n =$``\textit{pierogi}''). If these images are very different (\cref{fig:ps_scorer_fig} for  $n =$``\textit{banku}''), the T2I system has not learned categorical or region-specific associations well, and we hypothesize it lies in the long tail and shows poor \cure performance. For example, while $\phi^*_{PS}$ and $\phi_{GT}$ compute similarity of T2I generations of banku $I(n)$ to ground-truth images collected from a web database $G$, $\phi_{PS}(n)$ instead computes similarity of T2I generations of banku $I(n)$ to T2I generations of dumplings $I(c)$:
\begin{align}
    \phi_{GT}(a) &= sim(I(a), G) \\
    \phi_{PS}(a) &= sim \bigl(I(a), I(c)\bigr)
    \label{eq:percep_sim}
\end{align}

\begin{figure}[t]
\centering
\includegraphics[width=\linewidth]{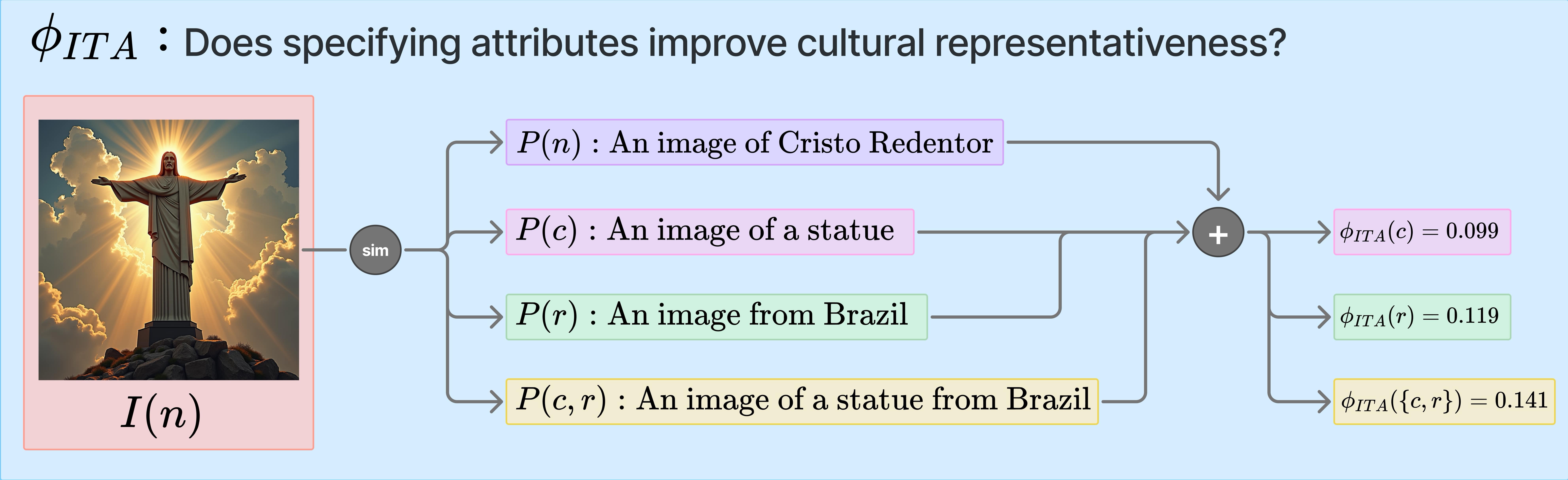}
\caption{Computation of image-text alignment score $\phi_{ITA}(a)$ on $n = \text{``Cristo Redentor''}$ when changing attribute $n\to a$. As in \cref{eq:ita}, to compute $\phi_{ITA}(a)$, a similarity score $sim$ is taken between image $I(n)$ and text prompt $P(a)$, and this is added to the similarity score between $I(n)$ and $P(n)$.}
\vspace{-4mm}
\label{fig:ita_scorer_fig}
\end{figure}

Following the status quo for semantic similarity~\cite{johnson2019billion}, $sim$ is a cosine distance between embeddings from state-of-the-art large vision encoders (SigLIP 2~\cite{tschannen2025siglip}, AIMV2~\cite{fini2024multimodal}, and DINOv2~\cite{oquabdinov2}). We note that since $\phi_{PS}$ compares only generated images, it is much cheaper to compute than $\phi_{GT}$ and $\phi^*_{PS}$, which require high-quality ground-truth data $G$ and for the latter, also finding and hiring workers to provide $\phi^*$. A drawback of this class of scorer is in the presence of ambiguity, \ie if a cultural artifact has multiple meanings. For example ``damper'' is both a culturally Australian homemade bread, as well as a device used to suppress vibrations in mechanical systems. $\phi_{PS}$ cannot differentiate these easily, as it has no access to ground-truth information. We avoid these cases by discarding ambiguous prompts during our \cure dataset design. We show the capability of $\phi_{PS}$ at correlating to real human perceptual similarity judgments compared to these strong baselines qualitatively in~\cref{tab:ai_vs_real_1} and quantitatively in~\cref{tab:results_ps}.

\subsection{Image-Text Alignment Scorers}
\label{subsec:scorer_imtxt_align}

The goal of image-text alignment (\textbf{ITA}) scorers is to compute an ``alignment" or similarity between an image and a piece of text. CLIP~\cite{radford2021learning} popularized using textual descriptions of ImageNet classes as zero-shot labels for image classification via unsupervised contrastive learning. In a similar vein, to evaluate cultural representativeness, previous works evaluate how ``close" generated images $I(a)$ are to textual descriptions of desired attributes $P(a)$ in the aligned latent space with vision-language models (VLMs) like CLIP~\cite{radford2021learning} and SigLIP~\cite{tschannen2025siglip}. We denote this class of image-text scorers by $sim(I(n), P(a))$. For example, to evaluate country-specific representativeness, \citet{khanuja2024image} check CLIP similarity of $I(n)$ with $P(r) = $ ``\textit{This image is culturally relevant to [r]}". We note that prior works typically use CLIP trained on LAION-2B~\cite{schuhmann2021laion} as their VLM of choice for image-text alignment, which has been shown to have an Amero and Euro-centric bias~\cite{basu2023inspecting, luccioni2023stable} and is also part of the pretraining datasets of many popular T2I systems~\cite{rombach2022high, podell2024sdxl, saharia2022photorealistic, yu2022scaling, kang2023scaling}. This overlap causes misleading over-estimations of quality, which we call \textbf{generative entanglement} (see \cref{subsec:results_ita} and \cref{tab:clip_correlations} in \cref{app:ita}).

\begin{figure}[t]
\centering
\includegraphics[width=\linewidth]{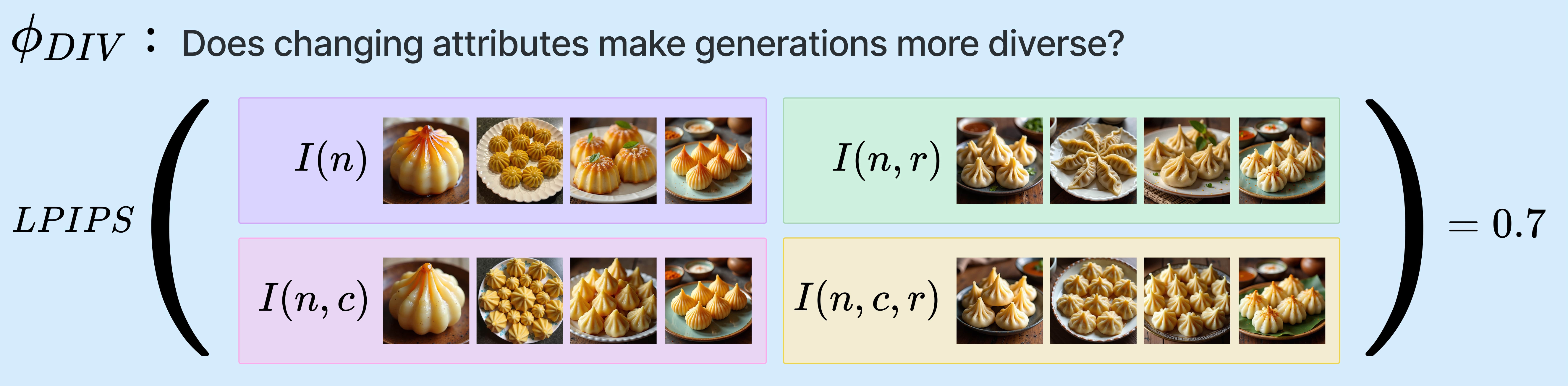}
\caption{Computation of diversity score $\phi_{DIV}$ on $n=\text{``modak''}$, a type of dumpling from India. $\phi_{DIV}$ computes an aggregate pairwise dissimilarity (LPIPS~\cite{zhang2018unreasonable}) between multiple seeds of images generated with varying attribute specification, \ie $I(n), I(n,c), I(n,r), I(n,c,r)$. In the example above, this encompasses $16c2=120$ pairs.}
\vspace{-4mm}
\label{fig:div_scorer_fig}
\end{figure}

These image-text scorers assume that embeddings of images containing attribute $a = n$ are clustered close in the VLM latent space to embeddings of textual descriptions of $n$. In other words, they rely on the VLM's ability to distinguish cultural relevance to different regions by seeing enough artifact-region associations ($n,c \to r$) during training~\cite{romero2025cvqa}. We show in \cref{tab:results_ita} that VLM knowledge of this association is difficult to query explicitly as $sim(I(n), P(r))$, and indirectly querying this knowledge through the impact of changing $a: n \to r$ correlates better with human judgments, \ie adding $sim(I(n), P(r))$ to $sim(I(n), P(n))$. We thus define $\phi_{ITA}$ as:
\begin{equation}
    \phi_{ITA}(a) = \dfrac{sim\bigl(I(n), P(n)\bigr) + sim\bigl(I(n), P(a)\bigr)}{2}
    \label{eq:ita}
\end{equation}

In contrast to baselines, when assessing an image of an artifact \eg toquilla in \cref{tab:ai_vs_real_2}, our scorer $\phi_{ITA}(r)$ evaluates both the \textit{visual correctness} of toquilla being a hat, \ie $sim(I(n), P(n))$ and its \textit{cultural linkage} to $r = $ Ecuador, \ie $sim(I(n), P(r))$.

\subsection{Diversity Scorers}
\label{subsec:scorer_diversity}

The goal of diversity (\textbf{DIV}) scorers is to capture the heterogeneity or intra-class variance of images generated by T2I systems~\cite{kannen2024beyond}. In this work, we view diversity through the lens of culture, \ie how culturally diverse are T2I systems, and can diversity be a predictor of cultural representativeness? We define \textbf{intra-category diversity} as the extent to which T2I images generated with underspecified prompts~\cite{basu2023inspecting} of cultural artifacts $I(c)$ or $I(s)$ reflect culture-specific nuances of regional variants $I(n)$ or $I(\{n,r\})$. In other words, if we sample $I(c) =$ ``An image of a house"~\cite{basu2023inspecting} from a T2I system 100 times, how many of these 100 images will be $r =$ Japanese or Canadian houses? We also measure \textbf{intra-artifact diversity}, \ie how diverse are multiple seeds of images of the same artifact $I(n)$, \eg when sampling 100 images with $P(n) = $``an image of chicken biryani", is there heterogeneity in the images of chicken biryani, or do they all appear visually homogeneous?

Existing works differ in the granularity at which they compute and aggregate diversity. LPIPS~\cite{zhang2018unreasonable} computes an average over pair-wise dissimilarity of images across all cultural artifacts $n$ associated with category $c$ using deep features extracted from convolutional neural networks~\cite{krizhevsky2012imagenet}. Similar to Perceptual Similarity scorers (\cref{subsec:scorer_percepsim}), LPIPS (which uses encoders trained on ImageNet~\cite{russakovsky2015imagenet}) can ignore image semantics in favor of spatial, color, and textural consistency~\cite{geirhos2018imagenet}. Another recent approach to scoring diversity is via Vendi Scores (VS)~\cite{friedman2023vendiscorediversityevaluation, zhang2024partiality}, which quantify diversity by estimating the entropy of a kernel similarity matrix computed over all pairs of artifacts ($a_i, a_j$) belonging to same category $c$. While Vendi Scores capture intra-category heterogeneity, they lack any sense of each individual artifact's image quality. \citet{kannen2024beyond} propose quality-weighting the Vendi score (qVS) with a human preference reward model~\cite{wu2023human} to address this limitation. 

We propose a modification to LPIPS that captures marginal information attribution, which we show in \cref{fig:div_scorer_fig}. For a given cultural artifact $n$, we consider a set of images generated with incrementally changing attributes, \ie \{$I(n)$, $I(\{n,c\})$, $I(\{n,r\})$, $I(\{n,c,r\})$\}. We compute LPIPS over each pair in this set and take an average, which we denote by:
\begin{equation}
    \phi_{DIV} = LPIPS\bigl(n, \{n,c\}, \{n,r\}, \{n,c,r\}\bigr) 
    \label{eq:diversity}
\end{equation}
         
If $\phi_{DIV} \sim LPIPS(n)$, diversity is relatively unaffected as we increase information specified to the T2I system from $n \to a$ (\cref{fig:lpips_gn_gs}, \cref{app:diversity}) and we hypothesize that artifact $n$ lies in the head of the T2I system distribution. We demonstrate qualitatively (\cref{tab:ai_vs_real_3}) and quantitatively (\cref{tab:results_div}) that LPIPS, VS, and QVS do not correlate well with human judgments of \cure when compared to our scorer $\phi_{DIV}$.
\section{Experiments}
\label{sec:experiments}

In this section, we discuss user study design (\cref{subsec:user_study}) and how we measure the capability of our proposed scorers to approximate real human judgments in \cref{subsec:scorer_rho}. We then discuss empirical setup and observations for each class of scorer in the relevant section below, \ie Perceptual Similarity scorers $\phi_{PS}$ in \cref{subsec:results_ps}, Image-Text Alignment scorers $\phi_{ITA}$ in \cref{subsec:results_ita}, and Diversity scorers $\phi_{DIV}$ in \cref{subsec:results_diversity}. Lastly, we benchmark popular state-of-the-art T2I systems on our dataset with our proposed scorers in \cref{subsec:benchmark_results}. Details on T2I inference and seeding used to compute all scores is provided in \cref{app:t2i_inference_details}.

\subsection{User Study}
\label{subsec:user_study}
To measure user judgments of T2I systems across global cultures, we hire workers on the crowdsourcing tool Prolific\footnote{https://www.prolific.com/}. We hire three workers per region (by country of nationality) to answer survey questions about the cultural artifacts from the \cure dataset (\cref{sec:dataset}) specific to their region. We ask each worker from region $r$ to rate on a 1-5 Likert scale a generated image of artifact $n$ for

\begin{enumerate}
    \item Cultural representativeness $\phi^*_{\cure}$, \ie ``How likely can this image be found in your country?''
    \item Perceptual similarity to ground-truth images $\phi^*_{PS}$, \ie ``How similar is this AI image to these four real images?''
    \item Likelihood that the image belongs to its ground-truth class $\phi^*_{GT}$, \ie ``How likely is this an image of \{n\}''
\end{enumerate}

We expect rational users to provide a similar rating for $\phi^*_{GT}$ and $\phi^*_{PS}$ - the distinction being that workers are not provided ground-truth reference images while rating $\phi^*_{GT}$ and must rely on their prior knowledge of artifact $n$. We discuss the UI, survey design choices, and survey questions asked to workers in \cref{app:user_study}. We also compare our study to prior works in \cref{tab:user_study_literature} (\cref{app:us_analytics}). We pay workers the platform set minimum of \$8 per hour.

\subsection{Scorer Correlation to Human Judgments}
\label{subsec:scorer_rho}

To validate the alignment of our proposed scorers (PS, ITA, DIV) to real human judgments, we compute a Spearman rank correlation $\rho$ to three gold scores from the user study, \ie $\phi^*_{\cure}$, $\phi^*_{PS}$, and $\phi^*_{GT}$. Spearman's $\rho$ is a nonparametric measure of rank correlation, \ie how well can the relationship between quantitative proxy scorers and real human judgments be described by a monotonic function? A $\rho \to 1$ indicates a monotonically non-decreasing relationship (\eg in \cref{tab:results_ps}, as our proposed $\phi_{PS}$ scores increase, human judments of \cure and perceptual similarity also increase). On the other hand, $\rho \to -1$ indicates a monotonically non-increasing relationship (\eg in \cref{tab:results_div}, as our proposed $\phi_{DIV}$ scores increase, human judments of \cure and perceptual similarity decrease). A $\rho \sim 0$ indicates a very weak correlation, \ie the scorer contains very little predictive signal over human judgments. We show qualitatively in \cref{fig:metric_comparisons} that our proposed scorers can differentiate between images treated differently by real humans (\ie $\phi^*$ are different), while baseline scorers treat these images the same. We discuss quantitative results over our entire \cure dataset in each corresponding section below.

\subsection{Perceptual Similarity}
\label{subsec:results_ps}

\begin{table}[t!]
\caption{Spearman rank correlation $\rho$ between perceptual similarity (\textbf{PS}) scorers and user judgments $\phi^*_{\cure}$, $\phi^*_{PS}$, and $\phi^*_{GT}$ across state-of-the-art large image encoders and T2I Systems on the \cure dataset. $\uparrow$ indicates that $\rho\to1$ is better, and $\downarrow$ indicates $\rho\to-1$ is better. The highest magnitude of $\rho$ in each column is bolded for the scorer that a) uses ground-truth images ($\phi_{GT}$ and Gemini) and b) does not use ground-truth images ($\phi_{PS}$ and $\Delta\phi_{PS}$).}
\begin{adjustbox}{max width=\columnwidth}
\begin{tabular}{@{}c|l|ccc|ccc|ccc@{}}
\toprule
\multirow{2}{*}{\textbf{Encoder}} & \multicolumn{1}{c|}{\multirow{2}{*}{\textbf{Scorer}}} & \multicolumn{3}{c|}{\textbf{FLUX.1 [dev]}} & \multicolumn{3}{c|}{\textbf{SD 3.5 Large}} & \multicolumn{3}{c}{\textbf{SD 1.5}} \\ \cline{3-11} 
 & \multicolumn{1}{c|}{} & $\phi^*_{\cure}$ & $\phi^*_{GT}$ & $\phi^*_{PS}$ & $\phi^*_{\cure}$ & $\phi^*_{GT}$ & $\phi^*_{PS}$ & $\phi^*_{\cure}$ & $\phi^*_{GT}$ & $\phi^*_{PS}$ \\ 
 \midrule
 - & Gemini 2.0 Flash $\uparrow$        &-  &-  & 0.40  &-  &-  &0.39  &-  &- &~0.40  \\ \midrule
\multirow{4}{*}{\textbf{SigLIP 2} \cite{tschannen2025siglip}} & $\phi_{GT}(n)\uparrow$ & \textbf{~0.25} & \textbf{~0.36} & \textbf{~0.44} & \textbf{~0.27} & \textbf{~0.33} & \textbf{~0.45} & \textbf{~0.25} & \textbf{~0.39} & \textbf{~0.51} \\
\cline{2-11} 
 & $\phi_{PS}(n)\uparrow$ & \textbf{~0.18} & ~0.25 & \textbf{~0.32} & \textbf{~0.22} & ~0.29 & \textbf{~0.38} & \textbf{~0.18} & \textbf{~0.27} & \textbf{~0.30} \\
 & $\Delta \phi_{PS}(\{n,c\})\downarrow$ & -0.16 & \textbf{-0.27} & -0.31 & -0.21 & -0.32 & -0.37 & -0.02 & ~0.03 & ~0.04 \\
 & $\Delta \phi_{PS}(\{n,c,r\})\downarrow$ & -0.17 & -0.25 & -0.30 & -0.21 & \textbf{-0.33} & \textbf{-0.38} & ~0.00 & ~0.09 & ~0.08 \\
 \midrule
\multirow{4}{*}{\textbf{AIMV2} \cite{fini2024multimodal}} & $\phi_{GT}(n)\uparrow$ & \textbf{~0.20} & ~0.28 & \textbf{~0.39} & \textbf{~0.24} & \textbf{~0.32} & \textbf{~0.44} & \textbf{~0.20} & \textbf{~0.35} & \textbf{~0.45} \\
\cline{2-11} 
 & $\phi_{PS}(n)\uparrow$ & ~0.08 & ~0.17 & ~0.25 & ~0.06 & ~0.14 & ~0.24 & \textbf{~0.09} & \textbf{~0.11} & \textbf{~0.12} \\
 & $\Delta \phi_{PS}(\{n,c\})\downarrow$ & -0.17 & \textbf{-0.30} & \textbf{-0.35} & \textbf{-0.19} & \textbf{-0.32} & \textbf{-0.31} & -0.01 & -0.02 & ~0.02 \\
 & $\Delta \phi_{PS}(\{n,c,r\})\downarrow$ & \textbf{-0.18} & -0.28 & -0.32 & -0.16 & -0.30 & \textbf{-0.33} & -0.01 & ~0.04 & ~0.05 \\
 \midrule
\multirow{4}{*}{\textbf{DINOv2} \cite{oquabdinov2}} & $\phi_{GT}(n)\uparrow$ & ~\textbf{0.17} & \textbf{~0.29} & \textbf{~0.40} & \textbf{~0.25} & \textbf{~0.36} & \textbf{~0.46} & \textbf{~0.24} & \textbf{~0.39} & \textbf{~0.52} \\
\cline{2-11} 
 & $\phi_{PS}(n)\uparrow$ & ~0.13 & ~0.22 & ~0.30 & ~0.15 & ~0.29 & \textbf{~0.35} & \textbf{~0.15} & \textbf{~0.27} & \textbf{~0.30} \\
 & $\Delta \phi_{PS}(\{n,c\})\downarrow$ & \textbf{-0.19} & \textbf{-0.27} & \textbf{-0.32} & \textbf{-0.21} & \textbf{-0.32} & \textbf{-0.35} & -0.07 & -0.05 & -0.01 \\
 & $\Delta \phi_{PS}(\{n,c,r\})\downarrow$ & \textbf{-0.19} & -0.25 & -0.31 & -0.20 & -0.31 & -0.34 & -0.08 & ~0.00 & -0.02 \\
 \bottomrule
\end{tabular}
\vspace{-3mm}
\end{adjustbox}
\label{tab:results_ps}
\end{table}

In this section, we compare our proposed PS scorers to human judgments of cultural representativeness $\phi^*_{\cure}$, perceptual similarity $\phi^*_{PS}$ and ground-truth label likelihood $\phi^*_{GT}$, as motivated in \cref{subsec:scorer_rho}. We tabulate a Spearman rank correlation across all artifacts $n$ in the \cure dataset in \cref{tab:results_ps}.

First, we compare how perceptually similar images of cultural artifacts $I(n)$ are to generated images of their cultural category $I(c)$, \eg T2I generated images of $n =$``banku" and T2I generated images of $c =$``dumpling" respectively: 
$$
    \phi_{PS}(n) = sim\bigl(I(n), I(c)\bigr)
$$
We find that our $\phi_{PS}(n)$ scorer is comparable to strong baseline $\phi_{GT}(n) = sim\bigl(I(n), G\bigr)$ in Spearman's $\rho$ with the gold scores $\phi^*_{GT}$ across all T2I systems and image encoders, despite using no ground-truth informaton (\cref{tab:results_ps}).

Next, recall that we claim that the relative importance or marginal utility of increasing attribute specification $a \subseteq \{n, c, r, s\}$ to the T2I system can provide valuable signal towards its cultural representativeness capabilities (\cref{subsec:metric_mia}). To more directly measure this marginal utility, we evaluate how much the \textit{change} in attribute specification ($n\to \{n, c\}$ and $n\to \{n, c, r\}$) affects perceptual similarity by computing a divergence $\Delta$ of $\phi_{PS}(a)$ from $\phi_{PS}(n)$ evaluated at attribute subsets $a = \{n, c\}$ and $a = \{n, c, r\}$: 
\begin{align*}
    \Delta\phi_{PS}(\{n, c\}) &= 0.5 + \phi_{PS}(\{n,c\}) - \phi_{PS}(n) \\ 
    \Delta\phi_{PS}(\{n, c, r\}) &= 0.5 + \phi_{PS}(\{n,c,r\}) - \phi_{PS}(n)
    \label{eq:ps_divergence}
\end{align*}

We add a $0.5$ scaling constant to our divergence scorers to bring them to a similar scale as $\phi_{GT}$ and $\phi_{PS}$.

On FLUX.1 [dev] and Stable Diffusion 3.5 Large, our divergence scorers $\Delta\phi_{PS}(\{n,c\})$ and $\Delta\phi_{PS}(\{n,c,r\})$ match or outperform $\phi_{GT}(n)$ in Spearman's $\rho$ across gold scores with AIMV2 and DINOv2 encoders. We note that rank correlations of divergence scorers with gold scores are always negative, since a low divergence with marginally increasing information indicates high perceptual similarity (denoted by $\downarrow$ in \Cref{tab:results_ps}). Overall, for all image encoders, all quantitative PS scorers $\phi_{GT}$, $\phi_{PS}$ and $\Delta\phi_{PS}$ correlate weaker with gold \cure scores $\phi^*_{\cure}$ than gold ground truth likelihood scores $\phi^*_{GT}$ and gold perceptual similarity scores $\phi^*_{PS}$, which they directly attempt to approximate. 

Lastly, we also evaluate Gemini 2.0 Flash~\cite{gemini}, a strong natively multimodal large language model (see \cref{subsec:mllm_judge}). We query Gemini to score perceptual similarity between $I(n)$ and ground truth images $G$ on a 1 to 5 scale, similar to the setup for $\phi_{GT}(n)$ and $\phi^*_{PS}$ from the user study. We observe that Gemini correlates almost identically well with $\phi^*_{PS}$ across T2I systems, and slightly lags behind $\phi_{GT}(n)$ in Spearman's $\rho$. 

We highlight that while $\phi_{GT}(n)$ and Gemini in general show the highest correlation with human perceptual judgments, our \textit{proposed scorers are capable of nearly matching them in rank correlation with real human judgments while using no ground-truth information}, and are thus much cheaper to compute. We note that no quantitative scorer surpasses $\rho = 0.51$, indicating that using cosine similarity with dense vector embeddings from large vision encoders still fall short at accurately approximating human judgements, even with large state-of-the-art encoders~\cite{tschannen2025siglip, fini2024multimodal, oquabdinov2}. 

\vspace{-10pt}
\paragraph{Discussion: Failure Mode of our Scorer:} While our PS scorers approximate gold scores well for FLUX.1 [dev] and SD 3.5 Large, divergence scorers $\Delta\phi_{PS}$ correlate very weakly with gold scores for SD 1.5, an older lower-resolution T2I system trained on much less data. In contrast, the baseline scorer $\phi_{GT}(n)$ maintains similar correlation across all T2I systems. This suggests that while our scorer, while not requiring ground truth images, relies more on the capacity of the T2I system to learn artifact - region associations than baselines. In other words, it is more sensitive to the coverage of the pretraining data, or the quality of the base model (\eg Image Arena Quality ELO in \cref{tab:benchmark_results}).

\subsection{Image-Text Alignment}
\label{subsec:results_ita}

\begin{table}[t!]
\caption{Spearman rank correlation $\rho$ between image-text alignment (\textbf{ITA}) scorers and user judgments $\phi^*_{\cure}$, $\phi^*_{PS}$, and $\phi^*_{GT}$  across T2I systems on the \cure dataset. All scorers except Gemini and human preference scorers (first block below) compute a cosine distance with SigLIP 2 embeddings. The highest magnitude of $\rho$ in each column is bolded, and $\rho\to1$ is the optimum scorer.}
  \centering
  \resizebox{\columnwidth}{!}{%
    \begin{tabular}{l |c c c| c c c| c c c}
      \toprule
      \multicolumn{1}{c|}{\multirow{2}{*}{\textbf{Scorer}}} & \multicolumn{3}{c|}{\textbf{FLUX.1 [dev]}} & \multicolumn{3}{c|}{\textbf{SD 3.5 Large}} & \multicolumn{3}{c}{\textbf{SD 1.5}} \\
      \cmidrule(lr){2-4} \cmidrule(lr){5-7} \cmidrule(lr){8-10}
      & \textbf{$\phi^*_{\cure}$} & \textbf{$\phi^*_{GT}$}  & \textbf{$\phi^*_{PS}$} & \textbf{$\phi^*_{\cure}$} & \textbf{$\phi^*_{GT}$}  & \textbf{$\phi^*_{PS}$}  & \textbf{$\phi^*_{\cure}$} & \textbf{$\phi^*_{GT}$} & \textbf{$\phi^*_{PS}$} \\
      \midrule
      PickScore~\cite{kirstain2023pick}        &0.20 &0.29  &0.34  &0.23  &\textbf{0.37}  &\textbf{0.40}  &0.23  &~0.38  &~\textbf{0.45}  \\
      Imagereward~\cite{xu2023imagereward}        &0.19  &0.21  &0.26  &0.16  &0.24  &0.31  &0.23  &~0.30  &~0.35  \\
      HPS v2~\cite{wu2023human}        &0.23  &0.29  &0.33  &0.18  &0.35  &0.37  &0.24  &\textbf{~0.39}  &~0.43  \\
      Gemini 2.0 Flash        &0.23  &\textbf{0.41}  & -  &\textbf{0.27}  &\textbf{0.37}  & -  &0.17  &~0.38  & -  \\ \midrule
      \citet{khanuja2024image}  &0.13  &0.08  &0.11  &0.05  &0.04  &0.00  & 0.06 &-0.02  &-0.04  \\
      \citet{ventura2023navigating}  &0.19  &0.15  &0.14  &0.10  &0.07  &0.05  &0.11  &~0.02  &~0.01  \\
      o3-mini~\cite{o3mini}  &0.17  &0.13  &0.14  &0.06  &0.03  &0.00  &0.06  &-0.01  &-0.03  \\
      $sim\bigl(I(n),\, P(n)\bigr)$       &0.24  &0.35  &0.38  &0.18  &0.31  &0.35  &0.22  &~0.34  &~0.44  \\
      $sim\bigl(I(n),\, P(c)\bigr)$       &0.20  &0.32  &0.34  &0.17  &0.30  &0.35  &0.19  &~0.34  &~0.37  \\
      $sim\bigl(I(n),\, P(r)\bigr)$       &0.20  &0.11  &0.12  &0.09  &0.03  &0.02  &0.07  &-0.04  &-0.04  \\ 
      $sim\bigl(I(n),\, P(\{c,r\})\bigr)$        &0.25  &0.35  &0.38  &0.22  &0.31  &0.36  &\textbf{0.24}  &~0.34  &~0.38  \\ \midrule
      $\phi_{ITA}(c)$  &0.24  &0.37  &0.40  &0.20  &0.34  &\textbf{0.40}  &0.22  &~0.37  &~\textbf{0.45}  \\
      $\phi_{ITA}(r)$   &\textbf{0.28}  &0.33  &0.35  &0.20  &0.27  &0.29  &0.23  &~0.29  &~0.39  \\
      $\phi_{ITA}(\{c,r\})$  &0.27  &0.38  &\textbf{0.42}  &0.23  &0.34  &0.39  &\textbf{0.24}  &~0.36  &~0.44  \\
      \bottomrule
    \end{tabular}%
  }
  \vspace{-3mm}
  \label{tab:results_ita}
\end{table}

Similar to PS scorers, we evaluate all quantitative ITA scorers via a Spearman rank correlation with gold scorers ($\phi^*_{\cure}$, $\phi^*_{GT}$ and $\phi^*_{PS}$) in \cref{tab:results_ita}. Recall that to compute $\phi_{ITA}$ in \cref{eq:ita}, we require a similarity measure between images and prompts $sim(I, P)$. To evaluate the \textbf{generative entanglement} between T2I system and scorer, we use a suite of vision-language models (VLMs) with different pretraining datasets as our similarity measure: four versions of OpenCLIP (trained on LAION-2B \cite{schuhmann2022laion}, Datacomp-1B \cite{gadre2023datacomp}, DataFN-5B \cite{fangdata}, and OpenAI WIT \cite{radford2021learning}) and SigLIP 2 \cite{tschannen2025siglip}). We begin by discussing results only with SigLIP 2, the current state-of-the art VLM.

First, we compare our scorer to three baseline scorers: two prior works evaluating cultural relevance~\cite{khanuja2024image} and cultural identity~\cite{ventura2023navigating}, and the best performing prompt $P(r)$ suggested by o3-mini~\cite{o3mini}. We also evaluate alignment of image $I(n)$ to text prompts with marginally increased attribution specification, \ie $P(n)$, $P(c)$, $P(r)$, and $P(\{c,r\})$. 

Our evaluation reveals that comparing $I(n)$ to $P(n)$, which specifies only the artifact’s name in the prompt (e.g., “Qingming festival”), consistently has a stronger positive rank correlation with human judgments than all methods that compare to only $P(r)$ (baseline scorers~\cite{khanuja2024image, ventura2023navigating, o3mini} and $sim(I(n), P(r)$) across all T2I systems (\cref{tab:results_ita} block 2). This supports our claim in \cref{subsec:scorer_imtxt_align} that explicitly querying VLMs for alignment to prompts describing the image by name $P(n)$ is more effective at assessing region-specific cultural relevance than directly querying for cultural relevance with region-specific prompts $P(r)$. Replacing $P(n)$ with category-specific prompts $P(c)$ (\eg “spring festival” for an image of Qingming festival), slightly reduces rank correlation with all gold scores across all T2I systems, which suggests that state-of-the-art VLMs learn $n \to c$ associations reasonably well. Lastly, using both category and region specific prompts $P(\{c,r\})$ tends to match and occasionally slightly outperform name-specific prompts $P(n)$ across T2I systems, indicating that switching to categorical and regional information ($n \to \{c, r\}$) is important to capture culture-specific human judgments.

Next, we compare our scorers to state-of-the art human preference reward model scorers~\cite{chen2025pal} trained on a million human-labeled pairwise preferences over T2I system generations, \ie PickScore~\cite{kirstain2023pick}, Imagereward~\cite{xu2023imagereward} and HPS v2~\cite{wu2023human}. Lastly, similar to PS scorers, we also evaluate Gemini 2.0 Flash (details in \cref{subsec:mllm_judge}).  

Recall that our scorers $\phi_{ITA}$ directly measure the marginal utility of increasing attributes specified to the T2I system in two parts, \ie the T2I system's ability to a) generate images that match textual descriptions of the artifact ($sim(I(n), P(n)$), b) capture an artifacts categorical ($sim(I(n), P(c)$) and regional associations ($sim(I(n), P(r)$). We observe that \(\phi_{ITA}(\{c, r\})\) has higher positive rank correlation than all baselines (\cref{tab:results_ita} block 2) across T2I systems, and is slightly outperformed by preference reward models trained on hundreds of thousands of human-labeled preferences and Gemini on Stable Diffusion T2I systems for $\phi^*_{PS}$ and $\phi^*_{GT}$ ($\phi_{ITA}(\{c,r\})\geq$ PickScore $>$ HPS v2 $>$ Imagereward). Our scorers outperform all methods including Gemini for FLUX.1 [dev].
\vspace{-10pt}

\paragraph{Remark: Generative Entanglement.} So far we have discussed ITA scorer results only with SigLIP 2 as a base VLM, the current state-of-the-art for image-text similarity (\cref{tab:results_ita}). For a more rounded examination, we also evaluate the impact of changing the base VLM as a similarity measure in the context of \textit{generative entanglement}. In \cref{tab:clip_correlations} (\cref{app:ita}), we show that with FLUX.1 [dev], all baseline methods have a high variance in Spearman's $\rho$ to human judgments of perceptual similarity ($\phi^*_{PS}$) with respect to the choice of similarity measure (OpenCLIP variant or SigLIP 2). In contrast, our scorers have lower variance across VLMs. As these VLMs differ primarily in their pretraining data, this result indicates that our scorers $\phi_{ITA}$ are much more robust to the choice of pretraining data and are less entangled with the T2I system.

\subsection{Diversity}
\label{subsec:results_diversity}

We next show a Spearman's rank correlation between all quantitative scorers $\phi_{DIV}$, LPIPS$(n)$ and VS$(c)$ and all gold scorers ($\phi^*_{\cure}(a)$, $\phi^*_{GT}(a)$, and $\phi^*_{PS}(a)$) in \cref{tab:results_div}. Consistent with our observations with PS and ITA scorers, we show that $\phi_{DIV}$ correlates more to human judgments than baselines across all T2I systems. This difference is more pronounced for modern, larger, high quality T2I systems (FLUX.1 [dev] and SD 3.5 Large) compared to a smaller, low quality T2I system (SD 1.5). See remark on Image Arena ELO remark in~\cref{subsec:benchmark_results} for a discussion on overall T2I system quality.

First, we consider LPIPS$(n)$, \ie a pairwise dissimilarity between seeds of a single cultural artifact $n$ (\textbf{intra-artifact}), averaged over all artifacts in the \cure dataset $n\in \mathcal{N}$. LPIPS$(n)$ shows a weak negative correlation with gold scores, but outperfoms Vendi Score~\cite{friedman2023vendiscorediversityevaluation} based scorers VS$(c)$ and qVS$(c)$ substantially, which are nearly uncorrelated ($\rho \sim 0$). As VS$(c)$ is computed as an aggregate at a coarse category level (\textbf{intra-category}) on static assignments over cultural artifacts and has no sense of image quality (see \cref{app:diversity} for details on VS computation), it shows the weakest correlation with human judgments for strong T2I systems (SD 3.5 Large and FLUX.1 [dev]. The recent quality-weighted Vendi Score~\cite{kannen2024beyond} fails to resolve these limitations, as it simply scales VS$(c)$ by a single scalar quality measure $q$ averaged over items (HPSv2~\cite{wu2023human}), leaving the rank correlation with human judgments unchanged.

In contrast, our MIA-based scorer $\phi_{DIV}$ achieves stronger negative correlations with human judgments than baselines for all T2I systems, especially those of higher quality (SD 3.5 Large and FLUX.1 [dev]). In other words, if by mixing images generated by the T2I system $I(n)$ with images generated with more information specification in the prompt $I(\{n, c\})$, $I(\{n, r\})$, and $I(\{n, c, r\})$, we observe a drop in diversity, human judgments of \cure and perceptual similarity improve. In this case, the cultural artifact $n$ is homogeneous across marginally increasing attribute specification and likely lies in the head of the T2I system's distribution. This aligns with our hypotheses in \cref{subsec:scorer_diversity} and indicates the potential of marginal utilty of information specification as a measure of diverse cultural representativeness.

\begin{table}[t!]
\caption{Spearman rank correlation $\rho$ between diversity (\textbf{DIV}) scorers and user judgments $\phi^*_{\cure}$, $\phi^*_{PS}$, and $\phi^*_{GT}$  across T2I systems on the \cure dataset. The highest magnitude of $\rho$ in each column is bolded, and $\rho\to-1$ is the optimum scorer.}
\resizebox{\columnwidth}{!}{%
\begin{tabular}{@{}l|ccc|ccc|ccc@{}}
    \toprule
    \multicolumn{1}{c|}{\multirow{2}{*}{\textbf{Scorer}}} & \multicolumn{3}{c|}{\textbf{FLUX.1 [dev]}}                                 & \multicolumn{3}{c|}{\textbf{SD 3.5 Large}}                                    & \multicolumn{3}{c}{\textbf{SD 1.5}}                                    \\ \cmidrule(l){2-10} 
    \multicolumn{1}{c|}{}                        & {\textbf{$\phi^*_{\cure}$}}                       & \multicolumn{1}{c}{\textbf{$\phi^*_{GT}$}}     & {\textbf{$\phi^*_{PS}$}}      & {\textbf{$\phi^*_{\cure}$}}                         & \multicolumn{1}{c}{\textbf{$\phi^*_{GT}$}}      & {\textbf{$\phi^*_{PS}$}}      & {\textbf{$\phi^*_{\cure}$}}                         & \multicolumn{1}{c}{\textbf{$\phi^*_{GT}$}}     & {\textbf{$\phi^*_{PS}$}}      \\ \midrule
    LPIPS$(n)$                                  & -0.11                     & \multicolumn{1}{c}{-0.06} & -0.16 & -0.13                     & \multicolumn{1}{c}{-0.06} & -0.09 & -0.04                     & \multicolumn{1}{c}{-0.03} & -0.02  \\
    VS$(c)$                                      & -0.02                     & \multicolumn{1}{c}{-0.01} & -0.01 & ~0.03                     & \multicolumn{1}{c}{~0.02}  & ~0.05 & ~0.01                     & \multicolumn{1}{c}{-0.02} & -0.02 \\
    qVS$(c)$                                   & -0.02      & \multicolumn{1}{c}{-0.01} & -0.01 & ~0.03                     & \multicolumn{1}{c}{~0.02}  & ~0.05 & ~0.01                     & \multicolumn{1}{c}{-0.02} & -0.02 \\ \midrule
    $\phi_{DIV}$                             & {\textbf{-0.20}}                     & \multicolumn{1}{c}{\textbf{-0.23}} & {\textbf{-0.30}} & {\textbf{-0.22}}                     & \multicolumn{1}{c}{\textbf{-0.23}} & {\textbf{-0.29}} & {\textbf{-0.07}}                     & \multicolumn{1}{c}{\textbf{-0.12}} & \textbf{-0.11} \\ \bottomrule
\end{tabular}%
}
\label{tab:results_div}
  \vspace{-3mm}
\end{table}

\vspace{-10pt}
\paragraph{Remark: Factuality-Diversity Tradeoff.} We note that while T2I systems being culturally diverse is a desired outcome of training, a high diversity across all quantitative scorers is negatively correlated to gold scores $\phi^*_\cure$ and $\phi^*_{PS}$. This indicates there is a factuality-diversity tradeoff with existing state-of-the-art T2I systems, which agrees with conclusions from prior work~\cite{wan2024factuality, kannen2024beyond}.

\subsection{Benchmark Results}
\label{subsec:benchmark_results}

We evaluate several popular state-of-the-art T2I systems on the \cure benchmark dataset with our three scorer classes sorted by decreasing Image Arena T2I Quality ELO~\footnote{\url{https://artificialanalysis.ai/text-to-image\#quality}, scores taken on \today}, an overall measure of T2I quality, in \cref{tab:benchmark_results}. We note that due to inbuilt safety filters, DALL-E 3 and Stable Diffusion 1.5 refuse to generate 17\% and 1.5\% of images respectively (detailed breakdown in \cref{app:img_gen_refusal}). For all scorers, we use the variant which had the highest negative Spearman's $\rho$ with gold scores on average.

For perceptual similarity, we use our $\Delta \phi_{PS}(\{n,c\})$ scorer (\cref{subsec:results_ps}). As this scorer is a divergence, a value closer to zero is better ($\downarrow$). As our scorer requires generating multiple seeds ($80$) of $I(c)$, we omit Ideogram 2.0 due to API compute constraints. SDXL and DALL-E 3 perform best for SigLIP 2 (SL2) cosine distance, while DALL-E 3 and FLUX.1 [dev] slightly edge out SD 1.5 and SDXL with DINOv2 (DN2). AIMV2 (AM2) has lower separability between T2I systems, with SD 3.5 Large and 1.5 slightly edging out the other T2I systems. We would also like to caveat that as our PS scorers show poor rank correlation to SD 1.5 (see discussion in \cref{subsec:results_ps}), PS scores on SD 1.5 are likely to be overestimates.

For image-text alignment, we use $\phi_{ITA}(\{c,r\})$  (\cref{subsec:results_ita}). Across VLM backbones, the Stable Diffusion class of T2I systems perform quite strongly (XL $\geq$ 3.5 $>$ 1.5) compared to FLUX.1 [dev], Ideogram 2.0 and DALL-E 3, and this is especially pronounced for CLIP trained on LAION-2B. As Stable Diffusion systems were trained on LAION-2B, we suspect their strong performance on LAION is because of this overlap with the training set of our scorer, which was also trained on LAION-2B (\textbf{generative entanglement}, see \cref{subsec:scorer_percepsim} for details).

Lastly, for $\phi_{DIV}$, which is an LPIPS score computed over a mixture of seeds across prompt styles, DALL-E 3 outperforms SDXL and SD 1.5, which are substantially ahead of the rest. It is noteworthy that DALL-E 3 had a 17\% refusal rate across all seeds of images on the \cure dataset, which likely caused a slight inflation in its diversity score (as LPIPS is computed across all pairs, generating fewer seeds will drop diversity less). Across all our scorers, Stable Diffusion XL performs the most consistently, followed by DALL-E 3.

\begin{table*}[t]
\caption{We evaluate several state-of-the-art T2I systems on our \cure benchmark across our three scorer classes: perceptual similarity (\cref{subsec:scorer_percepsim}), image-text alignment (\cref{subsec:scorer_imtxt_align}), and diversity (\cref{subsec:scorer_diversity}), and tabulate the mean and standard deviation ($\mu \pm \sigma$) across all 300 artifacts in the \cure dataset. The best entry of each column is bolded, and the next best if the T2I system is DALL-E 3 or SD 1.5$^*$, which have moderate refusal rates due to safety filters (see \cref{app:img_gen_refusal}). The last row shows a Spearman rank correlation $\rho$ between the scorer and the Image Arena Quality ELO.}
  \centering
  \resizebox{\textwidth}{!}{%
    \begin{tabular}{@{}l|c|ccc|ccc|ccc|c@{}} 
    \toprule 
    \multicolumn{1}{c|}{\multirow{2}{*}{T2I System}} & \multirow{2}{*}{ELO $\uparrow$} & \multirow{2}{*}{$\phi^*_{\cure} \uparrow$} & \multirow{2}{*}{$\phi^*_{PS} \uparrow$} & \multirow{2}{*}{$\phi^*_{GT} \uparrow$} & \multicolumn{3}{c}{$\phi_{PS} \downarrow$} & \multicolumn{3}{|c|}{$\phi_{ITA} \uparrow$} & $\phi_{DIV} \downarrow$ \\ 
    \cmidrule(lr){6-12} 
    \multicolumn{1}{c|}{} & & & & & SL2 & DN2 & AV2 & SL2 & L2B & WIT & ALX \\ 
    \midrule 
    FLUX.1 {[}dev{]} & 1045 & 2.814 $\pm$ 1.424 & 2.157 $\pm$ 1.141 & 2.251 $\pm$ 1.326 & 0.561 $\pm$ 0.110 & \textbf{0.575 $\pm$ 0.137} & 0.523~~$\pm$ 0.049 & ~0.094 $\pm$ 0.051 & ~0.218 $\pm$ 0.059 & ~0.209 $\pm$ 0.039 & ~0.708 $\pm$ 0.078 \\ 
    Ideogram 2.0 & 1043 & - & - & - & - & - & - & ~0.096 $\pm$ 0.052 & ~0.214 $\pm$ 0.067 & ~0.195 $\pm$ 0.050 & ~0.693 $\pm$ 0.072 \\ 
    SD 3.5 Large & 1028 & \textbf{2.986 $\pm$ 1.439} & \textbf{2.396 $\pm$ 1.228} & \textbf{2.534 $\pm$ 1.393} & 0.567 $\pm$ 0.107 & 0.604 $\pm$ 0.166 & 0.532~~$\pm$ 0.056 & \textbf{~0.115 $\pm$ 0.047} & ~0.251 $\pm$ 0.053 & ~0.225 $\pm$ 0.036 & \textbf{~0.670 $\pm$ 0.082} \\ 
    DALL-E 3$^*$ & ~922 & - & - & - & 0.562 $\pm$ 0.103 & 0.579 $\pm$ 0.143 & 0.525~~$\pm$ 0.055 & ~0.105 $\pm$ 0.051 & ~0.219 $\pm$ 0.062 & ~0.222 $\pm$ 0.041 & ~0.789 $\pm$ 0.043 \\ 
    SDXL & ~840 & - & - & - & \textbf{0.557 $\pm$ 0.100} & 0.579 $\pm$ 0.151 & \textbf{0.520~~$\pm$ 0.049} & ~0.113 $\pm$ 0.051 & \textbf{~0.255 $\pm$ 0.056} & \textbf{~0.230 $\pm$ 0.039} & ~0.753 $\pm$ 0.042 \\ 
    SD 1.5$^*$ & ~587 & 2.724 $\pm$ 1.412 & 2.094 $\pm$ 1.159 & 2.175 $\pm$ 1.291 & 0.559 $\pm$ 0.104 & 0.576 $\pm$ 0.142 & \textbf{0.519$^*$$\pm$ 0.041} & ~0.107 $\pm$ 0.050 & ~0.240 $\pm$ 0.055 & ~0.229 $\pm$ 0.035 & ~0.755 $\pm$ 0.057 \\ 
    \midrule \midrule 
    $\rho$ with ELO & ~1.00 & - & - & - & 0.564~~ & -0.100~ & 0.564~~ & -0.600 & -0.657 & -0.829 & -0.600 \\ 
    \bottomrule 
    \end{tabular}
}
\vspace{-3mm}
\label{tab:benchmark_results}
\end{table*}

\vspace{-10pt}
\paragraph{Remark: Relation to Image Arena ELO}
To compare to crowdsourced user judgments of overall T2I system quality, we compute a Spearman rank correlation of our scorers with Image Arena ELO scores. Image Arena pits two T2I systems in a ``battle'', where a user observes images generated by each system with a fixed prompt and picks their preferred image. Image Arena has collected 100K+ crowdsourced pairwise preferences to compute ELO for 50+ popular T2I systems. To begin with, we highlight that \emph{Image Arena ELO does not perfectly correlate with human judgments from our user study}: on a 1-5 Likert Scale, the most culturally representative ($\phi^*_\cure$) T2I system (SD 3.5) is only 0.2 points (5\%) better than the least culturally representative T2I system (SD 1.5), while they are separated by nearly 500 ELO points, \ie a 93\% chance that SD 3.5 will beat SD 1.5 in a head-to-head pairwise comparison. For perceptual similarity ($\phi^*_{GT}, \phi^*_{PS}$), the point difference only increases to 0.36 (\ie SD 3.5 is 9\% better than SD 1.5).

Our PS divergence scorer $\Delta\phi_{PS}$ shows weak negative rank correlation to Image Arena ELO with DINOv2 embeddings ($\rho=-0.100$), but a moderately strong positive correlation with SigLIP 2 and AIMV2 embeddings ($\rho = 0.564$). This suggests that the degree to which perceptual similarity of T2I system is indicative of overall T2I system quality evaluated via pairwise preferences may be sensistive to the choice of embedding backbone. With two popular backbones (SigLIP 2 and AIMV2), Spearman's $\rho$ suggests an inverse relationship between a T2I system's ability to generate images closely matching ground truth images and overall human judgments of quality (``is \textit{x} better or \textit{y}?''), which may be dominated by aesthetic appeal or other factors~\cite{dalle3}.

Our ITA scorer has a strong negative correlation with Image Arena ELO, especially with OpenAI CLIP as a VLM backbone (WIT, $\rho = -0.829$). In other words, with increasing T2I system quality, our scorer predicts lower image feature alignment to the text prompt. We suspect that this seemingly counter-intuitive result stems from a key difference in evaluation setup: our scorer uses ``underspecified prompts''~\cite{basu2023inspecting} $I(n)$, \ie the T2I generations only have the attribute name specified, whereas Image Arena ELO scores are based on detailed, descriptive prompts. In such underspecified cases, prior work has shown T2I systems with higher quality (ELO) have more ``creativity'' and tend to prioritize aesthetic visual appeal over actuality~\cite{dalle3}. We also find from our user study that higher quality T2I systems rely on cultural stereotypes for generation (\cref{app:gold_scores_head_tail}), which negatively impacts user judgments of quality. 

Lastly, as high quality T2I systems like FLUX.1 [dev] and SD 3.5 Large have seen orders of magnitude more data during training, they can generate more diverse renditions of underspecified prompts (which we also empirically observe through high $\phi_{DIV}$ scores for T2I systems with low ELO). As diversity is often at odds with factuality~\cite{wan2024factuality}, this phenomenon ends up hurting the alignment of higher quality T2I systems to simple text prompts, \ie $P(c,r)$.

\subsection{Multimodal LLM as a Judge}
\label{subsec:mllm_judge}

To our knowledge for the first time, we perform a preliminary, exploratory analysis of using a multimodal large language model (MLLM) to approximate human judgments (gold scores) of cultural representativeness ($\phi^*_{\cure}$) and image perceptual similarity ($\phi^*_{PS}$, $\phi^*_{GT}$). 

Specifically, we query Gemini 2.0 Flash~\cite{gemini} with the same set of questions asked to real humans in our user study (details in \cref{app:mllm_judge}), and output a score from 1 (low) to 5 (high), \ie the same scale as the Likert scores from our user study (\cref{subsec:user_study}). We also query Gemini for a textual justification of its scores to inspect its reasoning. We follow the same evaluation setup as PS, ITA and DIV scorers (\cref{subsec:scorer_rho}) and compute a Spearman rank correlation between Gemini scores and gold scores from our user study. For PS scorers \cref{tab:results_ps}, we observe that Gemini slightly lags behind $\phi_{GT}$, and our scorers approximate both these strong baselines well without access to any ground-truth images. For ITA scorers (\cref{tab:results_ita}), we observe that Gemini 2.0 Flash nearly matches (Flux.1 [dev] and SD 1.5) and occasionally outperforms (SD 3.5) our proposed $\phi_{ITA}$ scorers in Spearman's $\rho$, suggesting a pretraining dataset has wider cultural coverage of artifact - region associations compared to smaller VLMs like OpenCLIP and SigLIP 2. Lastly, we examine Gemini's textual justifications for its scores to inspect the reliability of its reasoning (see \cref{fig:Gemini 2.0 Flash_failures_all} in \cref{app:mllm_judge}). We find that Gemini tends to overlook culture-specific details, even when attribute details may be correct (\eg textures, shapes and patterns of an object that are not culturally accurate). This issue is especially prevalent in regions from the Global South (\eg Nigeria, Ghana, Iran), where Gemini hallucinates culture-specific details (\eg ``\textit{jollof rice}'' from Ghana, ``\textit{Femi Kuti}'' from Nigeria and ``\textit{sami headwear}'' from Finland) and mistakenly highlights accuracy to the wrong culture (\eg Chinese instead of Korean for ``\textit{Chuseok}'' and Egyptian instead of Iranian in ``\textit{Takht-e Fulad}'').

In summary, these initial findings suggest that while a state-of-the-art multimodal LLM can show promise at evaluating cultural representativeness, it \textbf{requires access to ground truth images for reliability} (which none of our scorers require), and \textbf{hallucinates incorrect culture-specific details}, often in regions from the Global South.
\section{Conclusion}
\label{sec:discussion}
In summary, we propose \cure, a novel benchmarking and scoring suite for cultural representativeness that leverages the marginal utility of attribute specification to text-to-image systems as a proxy for human judgments. Our \cure dataset has a novel categorical hierarchy that enables benchmarking T2I systems in this manner. We empirically observe much stronger correlations to user judgments across three quantitative scoring classes, \ie perceptual similarity (PS, \cref{subsec:results_ps}), image-text alignment (ITA, \cref{subsec:results_ita}), anall cultural diversity (DIV, \cref{subsec:results_diversity}). 

We highlight that our PS scorer approximates strong baseline performance while using no ground-truth information, indicating strong potential for efficient and democratic benchmarking of T2I systems. Our ITA and DIV scorers (\cref{tab:results_div}) outperform or match all baselines, including a strong multimodal large language model (Gemini 2.0 Flash, \cref{subsec:mllm_judge}). Finally, we highlight the generative entanglement issue of prior benchmarks and scorers, \ie a miscalibrated estimation of human judgments caused by an overlap of T2I system and quantitative scorer pretraining data (\cref{tab:results_ita}). 

\vspace{-10pt}
\paragraph{Limitations and Future Work} While we employ a categorical hierarchy of attributes, similar to prior work~\cite{basu2023inspecting, khanuja2024image}, \cure also uses geography as a proxy for culture (each artifact has a single associated country of origin). In the future, we would like to extend \cure to a more holistic sense of culture, such as religion and spoken language. For perceptual similarity scorers, we find that overall T2I system quality determines the ability of our scorer $\phi_{PS}$ to match human judgments of cultural representativeness. When using a smaller and lower quality system (Stable Diffusion 1.5), our scorers were less correlated with human judgments than baseline scorers (\cref{tab:results_ps}), while this was not the case for strong systems (Flux.1 [dev], Stable Diffusion 3.5 Large). This scorer also suffers in the case of ambiguous prompts (\ie artifact name $n$ referring to multiple distinct artifacts), as it has no access to ground-truth image information to distinguish between them. We design our \cure dataset to filter out such ambiguities. Lastly, while state-of-the-art natively multimodal LLMs like Gemini 2.0 Flash (\cref{subsec:mllm_judge}) show promising initial results, they still tend to miss culture-specific details. We highlight that our lens of marginal information attribution is complementary to MLLM scorers, and believe this is a promising direction for future research.

\section{Acknowledgements}
Aniket Rege would like to thank Chitwan Saharia and Ideogram, Inc. for their generous support via API credits to benchmark Ideogram 2.0. Aniket Rege and Yong Jae Lee acknowledge that this research project has benefitted from the Microsoft Accelerating Foundation Models Research (AFMR) grant program.
{
    \small
    \bibliographystyle{ieeenat_fullname}
    \bibliography{main}
}

\clearpage
\appendix
\tableofcontents

\section{T2I Inference Details}
\label{app:t2i_inference_details}

\begin{table}[h!]
\caption{Text prompts $P$ describing cultural artifact $a$ given to generative model $f_\theta$ with differing levels of informativeness indicated by artifact attributes $a \subseteq \{n, c, s, r\}$.}
\begin{adjustbox}{max width=\columnwidth}
\begin{tabular}{@{}l|l|l@{}}
\toprule
Type          & Prompt Text                                                                              & Example                                                                                      \\ \midrule
$P(n)$        & An image of \{n\}                                                                        & An image of jiaozi                                                                           \\
$P(c)$        & An image of \{c\}                                                                        & An image of a dumpling                                                                       \\
$P(r)$        & An image from \{r\}                                                                      & An image from China                                                                          \\
$P(\{n, c\})$     & An image of \{n\}, a type of \{c\}                                                       & An image of jiaozi, a type of dumpling                                                       \\
$P(\{n, r\})$     & An image of \{n\}, from \{r\}                                                            & An image of jiaozi, from China                                                               \\
$P(\{n, c,  r\})$ & \begin{tabular}[c]{@{}l@{}}An image of \{n\}, a type of \{c\} \\ from \{r\}\end{tabular} & \begin{tabular}[c]{@{}l@{}}An image of jiaozi, a type of dumpling \\ from China\end{tabular} \\ \bottomrule
\end{tabular}
\end{adjustbox}
\label{tab:prompt_templates}
\end{table}

\subsection{Seeding}
\label{app:img_gen_seeding}
We use four of the prompt styles outlined in \cref{tab:prompt_templates} to generate images for benchmarking from all T2I systems: $P(n), P(\{n,c\}), P(\{n,r\}), P(\{n,c,r\})$. We generate multiple random seeds for each prompt: for Stable Diffusion XL and Stable Diffusion 1.5, we use 20 random seeds, and for all other T2I systems, we use four random seeds\footnote{DALL-E 3 has no dedicated random seed parameter. We follow prior work on passing random seeds to DALL-E 3 via prompting: \url{https://community.openai.com/t/consistent-variability-using-seeding-with-dall-e-3/457823}}. For all systems except Ideogram 2.0 (computational constraints), we also generate 80 seeds with prompt $P(c)$, which is required for our perceptual similarity scorer $\phi_{PS}$ (see \cref{subsec:scorer_percepsim} for details). For a given artifact, our scorers compute a score on all seeds, which are then averaged to a single score for that artifact, \ie for $N$ seeds,

$$
\phi(a) = \dfrac{1}{N}\sum_{i=1}^{N} \phi\bigl(I(a_i)~;~\bigr)
$$

\subsection{Safety Filter Refusal}
\label{app:img_gen_refusal}
Due to the inbuilt safety filters of DALL-E 3 and Stable Diffusion 1.5, many of our prompts were rejected and thus we were unable to generate images over the entire \cure dataset. \cref{tab:percent_generated} shows the percentage of each supercategory that was successfully generated, calculated as:

\begin{align*}
    &\text{T2I Acceptance Rate }(s)  \\
    &= \dfrac{N\_img\_gen(s)}{N\_img\_total(s)} * 100 \\
    &= \dfrac{N\_img\_gen(s)}{N\_artifact(s) * N\_promptstyle * N\_seeds}
\end{align*}

where $N\_artifact(s) = 50$ for all supercategories, $N\_promptstyle = 4$, and $N\_seeds$ depends on the T2I system (see \cref{app:img_gen_seeding} for details).

\begin{table}[t!]
\centering
\caption{Percent of each supercategory that was generated by DALL-E 3 and SD 1.5.}
\begin{adjustbox}{max width=\columnwidth}
\begin{tabular}{@{}r|cc@{}}
\toprule
\textbf{Supercategory} & \textbf{DALL-E 3} & \textbf{SD 1.5} \\ 
\midrule
Architecture & 94.75\% & 99.80\% \\
Art & 78.50\% & 95.20\% \\
Celebrations & 95.63\% & 99.77\% \\
Fashion & 97.63\% & 99.05\% \\
Food & 97.63\% & 98.55\% \\
People & 33.50\% & 98.82\% \\ 
\bottomrule
\end{tabular}
\end{adjustbox}
\label{tab:percent_generated}
\end{table}

\section{Dataset Design}
\label{app:dataset}

\begin{table}[t]
\centering
\caption{Supercategories and corresponding categories for our \cure benchmark dataset (\cref{sec:dataset}).}
\begin{adjustbox}{max width=\columnwidth}
\begin{tabular}{c|l}
\toprule
\textbf{Supercategory} & \multicolumn{1}{l}{\textbf{Category}} \\
\midrule
\multirow{5}{*}{Architecture} & Bridge \\
 & Fortification \\
 & House \\
 & Monument and Memorial \\
 & Religious Building \\
 \midrule
\multirow{5}{*}{Art} & Bust \\
 & Fresco \\
 & Oil Painting \\
 & Pottery \\
 & Statue \\
 \midrule
\multirow{5}{*}{Celebration} & Carnival \\
 & Christmas Food \\
 & Harvest Food \\
 & New Year celebration \\
 & Spring Festival \\
 \midrule
\multirow{4}{*}{Fashion} & Embroidery \\
 & Hat \\
 & Jewellery \\
 & Traditional clothing \\
 \midrule
\multirow{5}{*}{Food} & Dumpling \\
 & Flatbread \\
 & Fried Dough \\
 & Noodle Dish \\
 & Rice Dish \\
 \midrule
\multirow{7}{*}{People} & Activist \\
 & Actor \\
 & Filmmaker \\
 & Musician \\
 & Politician \\
 & Sportsperson \\
 & Writer \\
\bottomrule
\end{tabular}
\end{adjustbox}
\label{tab:dataset_structure}
\end{table}

We created the \cure dataset from the WikiMedia knowledge graph~\cite{wikimedia}. We first manually shortlist culturally relevant supercategories or cultural axes, \ie Architecture, Art, Celebration, Fashion, Food, and People. Within each supercategory, we search for Wikimedia categories structured as ``[category name] by country''. Our supercategory and category structure ($s \to c$) is shown in \cref{tab:dataset_structure}. To examine the performance across the cultural long tail of T2I systems, we select Wikimedia categories that contain countries across the Global North / South divide, an example proxy for the head and long tail pretraining distribution of T2I systems. We filter out countries who contain less than four images for a Wikimedia category, as we require these as a ground-truth set for perceptual similarity scoring (\cref{subsec:scorer_percepsim}) and our user study (\cref{subsec:user_study}). 

Under these conditions, we collect exactly 50 unique region-specific named entities (cultural artifacts) for each supercategory. Each supercategory has between four and seven categories : ``Traditional clothing'' from the ``Fashion'' supercategory contains 20 artifacts and the ``People'' supercategory is balanced by region (5 famous people per region over seven categories, \ie occupations). Out of our 300 cultural artifacts, 123 are from countries considered part of the Global North and 177 are from countries considered part of the Global South \cite{gns2025unctad}, which we decided based on UNCTAD categorization as developed economies (Global North) or developing economies (Global South).

\section{User Study Design}
\label{app:user_study}

We create and host 300 artifact-specific surveys for each T2I system we evaluate (Stable Diffusion 1.5\footnote{\url{https://huggingface.co/stable-diffusion-v1-5/stable-diffusion-v1-5}}, Stable Diffusion 3.5 Large\footnote{\url{https://huggingface.co/stabilityai/stable-diffusion-3.5-large}} and Flux.1 [dev]\footnote{\url{https://huggingface.co/black-forest-labs/FLUX.1-dev}}) using Qualtrics. We choose these three T2I systems due to popularity on HuggingFace their open-weights nature allowing democratic availability for downstream research.

Our hosted surveys are sent to applicable workers through Prolific, a large crowdsourcing tool. For each artifact $n$ in the \cure dataset, we hire three workers whose country of nationality match the region $r$ of the artifact. The survey was launched only in English. To minimize the introduction of biases from the researchers themselves, no rubric was provided to workers to answer survey questions, other than some examples of how T2I systems can be stereotypical (see details in \cref{app:user_study_off_stereo}) since workers may be unfamiliar with T2I systems. Below, we provide an overview of each section of our survey, which was organized as:
\begin{enumerate}
    \item Disclosure (\cref{app:user_study_disclosure})
    \item Perceptual Similarity (\cref{app:user_study_ps})
    \item Artifact Familiarity Questionnaire (\cref{app:user_study_artifact_familiarity})
    \item \cure: Cultural Representativeness (\cref{app:user_study_cure})
    \item Offensiveness and Stereotypes (\cref{app:user_study_off_stereo})
    \item User Metadata (\cref{app:user_study_demographics})
\end{enumerate}

\begin{figure}
    \centering
    \includegraphics[width=1\linewidth]{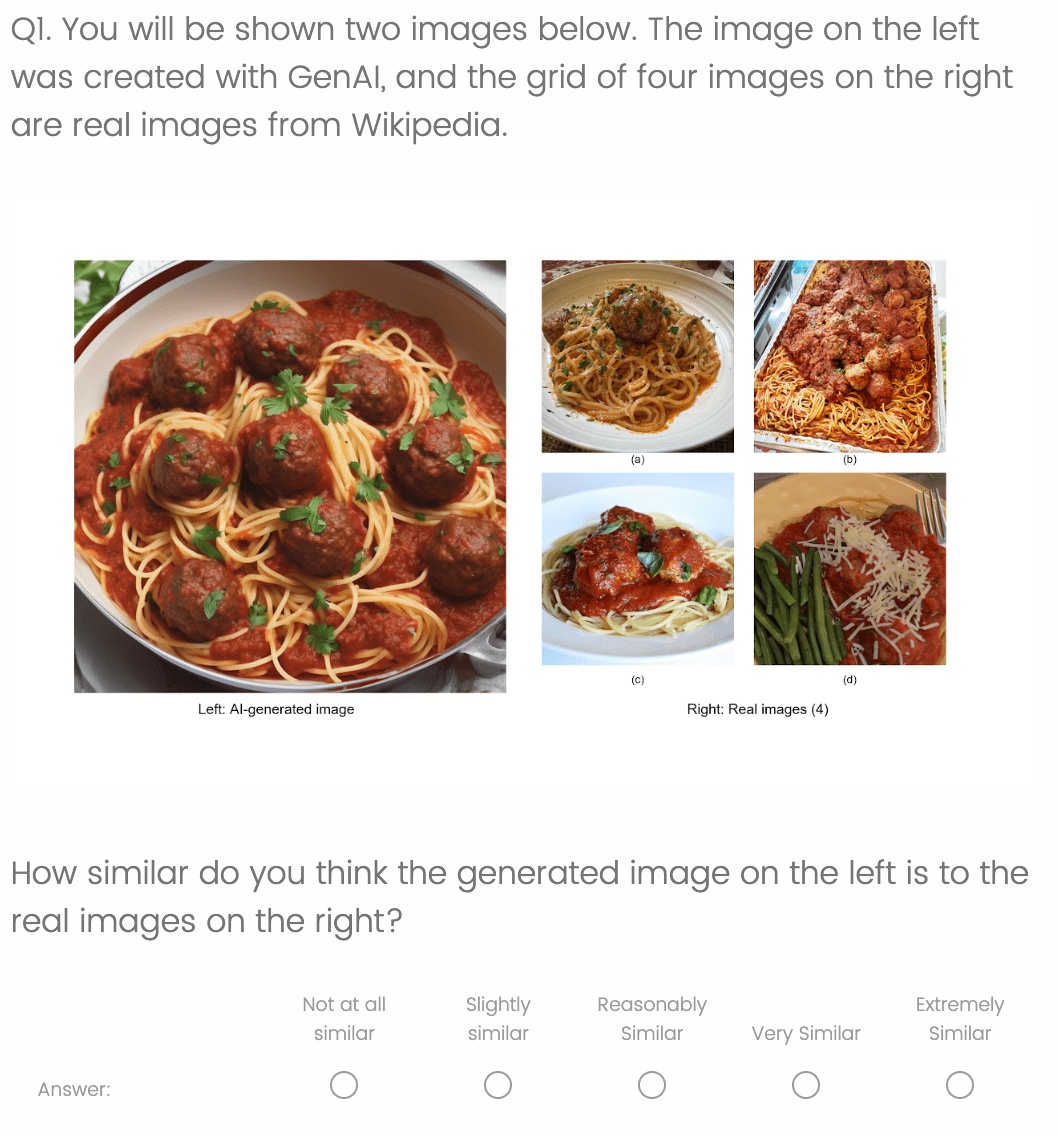}
    \caption{\textbf{Q1a:} Querying users for perceptual similarity of T2I system generated image to ground-truth images.}
    \label{fig:userstudy_ui_percep_sim}
\end{figure}

\subsection{Disclosure}
\label{app:user_study_disclosure}
We inform the workers of the survey goals, how their data will be used, and how they can withdraw their consent later if they choose to do so. They are asked for explicit and informed consent for their data to be used, and provided an option to opt-out. 

\begin{tcolorbox}[colback=gray!10, colframe=black, boxrule=1pt, arc=2pt, width=\columnwidth]
\rmfamily
\textbf{Research Study on the Cultural Biases of Generative AI}
\vspace{0.1cm}

Thank you for taking the time for this survey. We are a team of researchers from [place] who study the cultural biases of generative artificial intelligence (GenAI) models in an attempt to make them more representative for everyone. 

\textbf{What we collect from you}: Current country of residence, Nationality, First language, Country of birth, Age, Sex, Participant ID. \vspace{0.3cm}

\textbf{How we use your data}: To analyze the biases of generative AI models to concepts local to your culture and country. Your data will be stored in an \textbf{anonymized} fashion in an online excel sheet, and published to other researchers as part of an academic study. We will \textbf{always maintain your anonymity}, as we do not collect any identifiable information. \vspace{0.3cm}

\textbf{Withdrawing your data later}: If you wish to withdraw or remove your data at any time after this survey, you can contact us via Prolific's anonymous internal messaging tool, or directly contact our research lead by email at [email]. We will then remove your data from our server. Please note that while we will remove your data from our server, we cannot guarantee this data will not continue to exist elsewhere online. \vspace{0.3cm}

\textbf{Do you consent to your anonymized data being used in this survey}?
Select 'Yes' only if you fully understand the information above. If you are unsure or hestitant about providing your data, please select 'No'. 
\end{tcolorbox}

\begin{figure}
    \centering
    \includegraphics[width=1\linewidth]{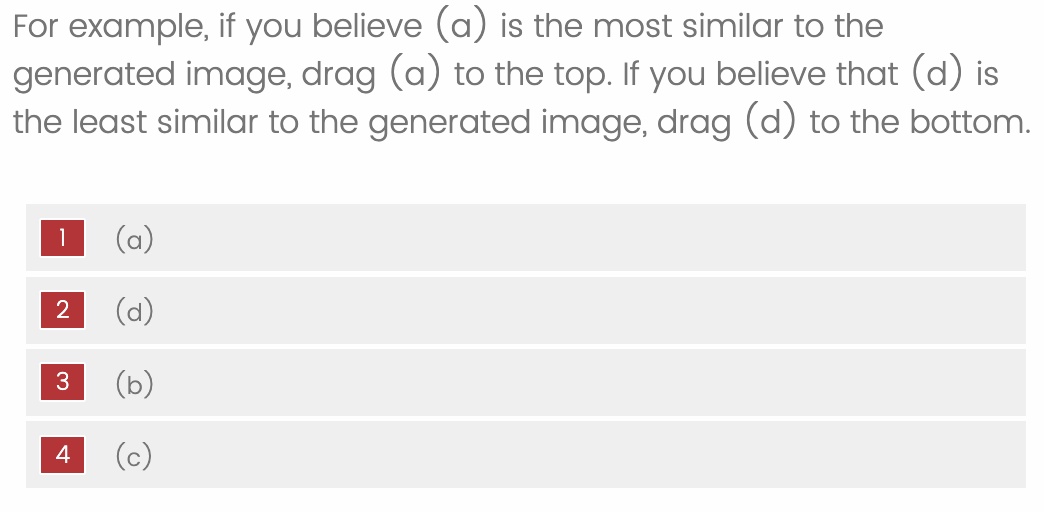}
    \caption{\textbf{Q1b:} Querying users to rank (order) the similarity of the ground truth images to the AI generated image from highest (top) to lowest (bottom).}
    \label{fig:us_drag}
\end{figure}

\subsection{Perceptual Similarity}
\label{app:user_study_ps}
We provide an image generated by the T2I system and ask workers to rate similarity from 1 to 5 (low to high Likert score) to a grid of ground-truth images from WikiMedia. The worker is not told what the artifact is, only to rate visual similarity.

We provide four images to the users, \ie a single randomly chosen seed generated with four prompt styles with varying levels of attribute specification or informativeness (see \cref{app:img_gen_seeding}). The user is asked \textbf{Q1a:} \textit{``How similar do you think the generated image on the left is to the real images on the right?''} and asked to rate it on a Likert scale from \textbf{1 (Not at all similar)} to \textbf{5 (Extremely Similar)} (see UI in \cref{fig:userstudy_ui_percep_sim}).

A secondary goal for perceptual similarity questionnaire is to examine how consistent or homogeneous different workers are at ranking perceptual similarity. We thus query each worker for their ranking of ground-truth similarity, \ie \textbf{Q1b:} to rank the four ground truth images in the grid for semantic similarity to the AI-generated image $I(n)$. The user is asked to drag letters that correspond to each image of the 2x2 ground truth grid (a, b, c, d) into a high-to-low order (1 being most similar and 4 being the lowest). We show the UI for Q1b in \cref{fig:us_drag}. We analyze worker disagreement over perceptual similarity in \cref{app:worker_disagreement}.

\subsection{Artifact Familiarity Questionnaire}
\label{app:user_study_artifact_familiarity}
The worker is queried for their prior knowledge about artifact $a$ by its name $n$, \ie ``\textit{Had you ever heard of \{n\} before seeing all the images above?}'', with possible answer options being {``Yes'', ``No'', and ``Unsure''}. If the user answers ``Yes'', the user is then queried for a textual description of their knowledge, \ie

\begin{tcolorbox}[colback=gray!10, colframe=black, boxrule=1pt, arc=2pt, width=\columnwidth]
\rmfamily
Please describe your knowledge about what \{n\} is in 1-2 sentences. \vspace{0.3cm}

For example, if you have seen "the Statue of Liberty" before and know what it is, you might write: \vspace{0.1cm}

"The Statue of Liberty is a famous landmark in new york city. It is a tall green statue of lady liberty holding a torch"
\end{tcolorbox}

\subsection{\cure: Cultural Representativeness}
\label{app:user_study_cure}

\begin{figure}[hbt!]
    \centering
    \includegraphics[width=1\linewidth]{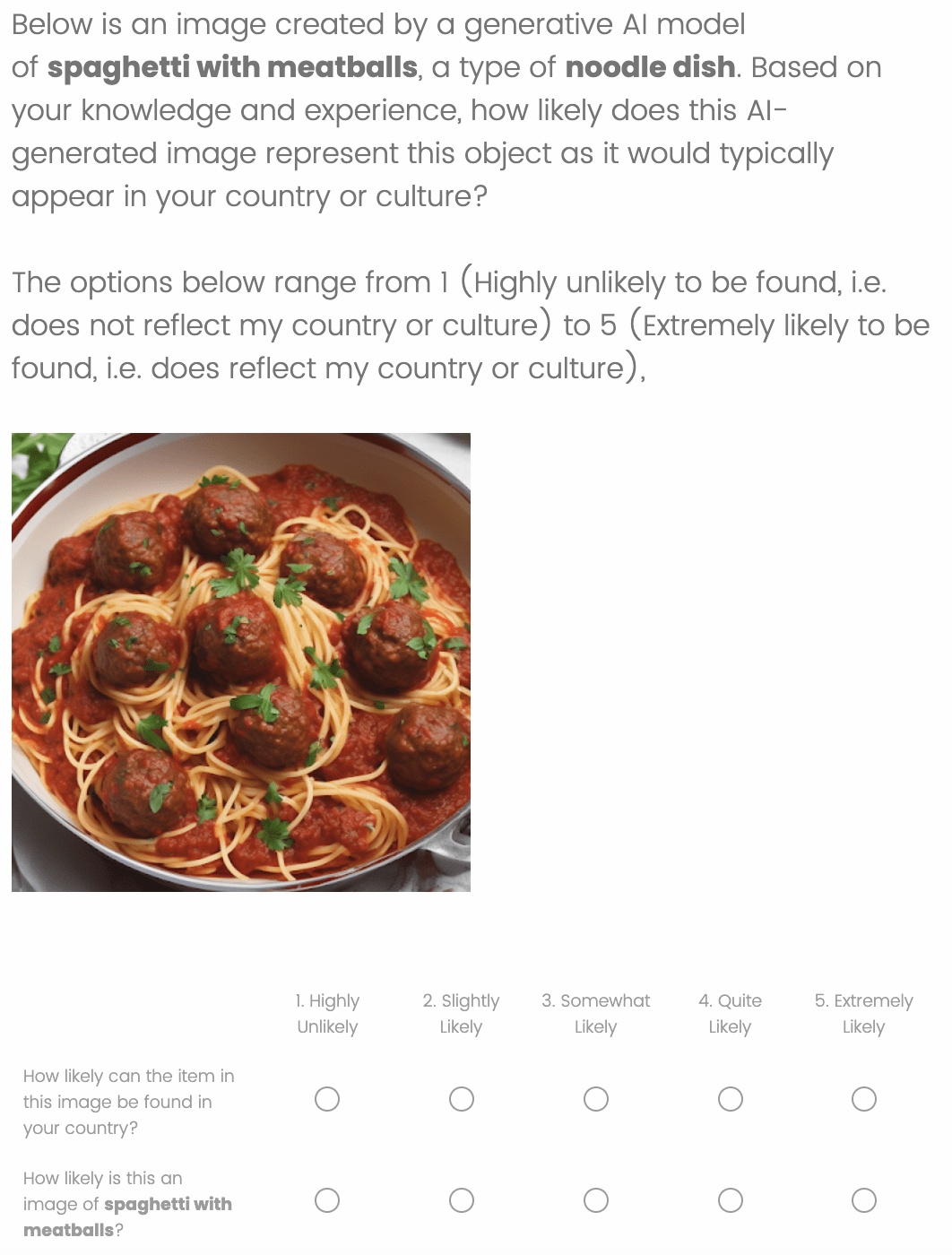}
    \caption{User study interface for \cure.}
    \label{fig:us_cure}
\end{figure}

From this point on in the survey, the user only shown the AI image generated with prompt $P(n)$. They are no longer shown ground truth images. The user is informed what the artifact name $n$ and category $c$ are. They are shown only the T2I system image and asked to rate its \cure from 1 to 5 (low to high Likert score) with \textbf{Q3a}: \textit{``How likely can the item in this image be found in your country?''}. The user is also asked to rate the likelihood of the image belonging to the class of artifact, as they now know its name, via \textbf{Q3b}: \textit{``How likely is this an image of [artifact name]?''}. The UI for these questions is shown in \cref{fig:us_cure}.

Finally, the user is also asked to share a textual description of the specific details contributing to the accuracy or inaccuracy of the T2I system output in \textbf{Q3c}. This fine-grained information is useful to identify details of why and how the T2I system failed to accurately generate artifact $n$, \ie

\begin{tcolorbox}[colback=gray!10, colframe=black, boxrule=1pt, arc=2pt, width=\columnwidth]
\rmfamily
We are trying to understand when GenAI models get culture-specific details right and wrong. \vspace{0.2cm}

In your opinion, what specific details in the AI-generated image above make it accurate or inaccurate compared to how this object typically appears in your country or your understanding of \textbf{[artifact name]}? \vspace{0.2cm}

For example, if the image is of the `Italian pasta' and is inaccurate, you might say:
`The image has used the wrong kind of pasta noodles, ingredients, and sauce, you would not see this kind of pasta in my country or culture'.
\end{tcolorbox}

\subsection{Offensiveness and Stereotypes}
\label{app:user_study_off_stereo}

\begin{figure}
    \centering
    \includegraphics[width=1\linewidth]{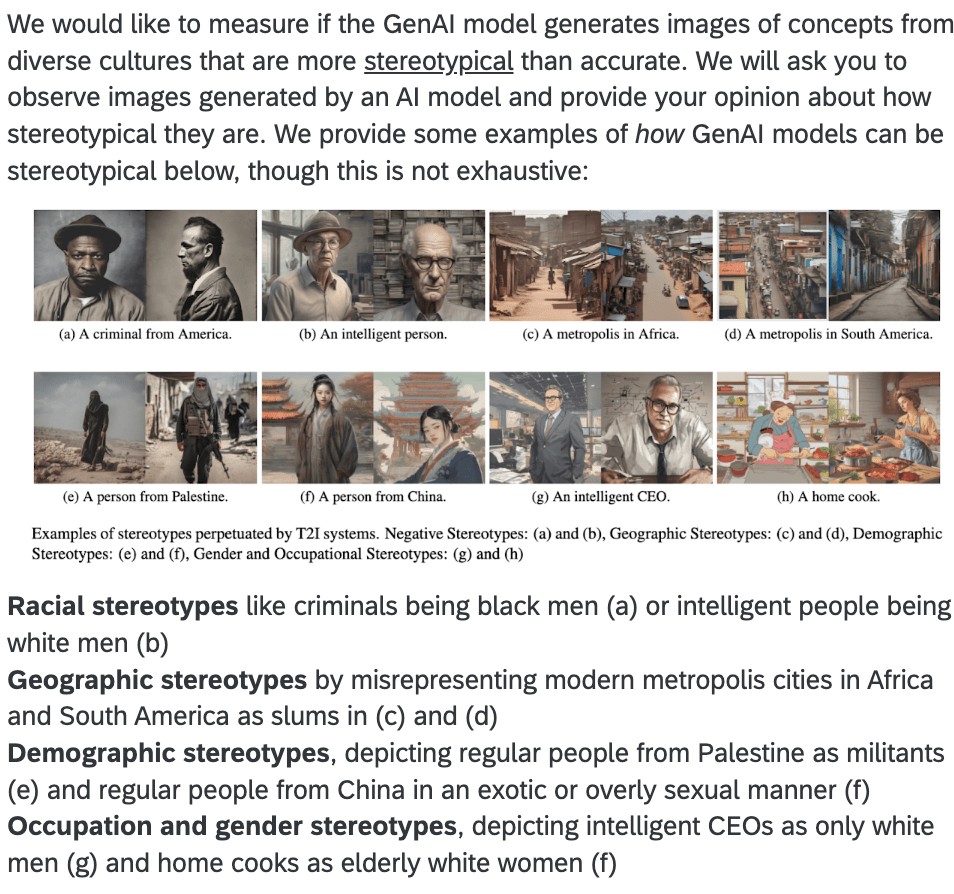}
    \caption{Description and examples given about stereotypes.}
    \label{fig:us_stereo}
\end{figure}

We ask workers to rate how offensive and stereotypical to their culture the T2I system output is on a Likert scale from \textbf{1. Not at all} to \textbf{5. A lot}. For offensiveness, we query workers zero-shot (\ie no rubric or examples): \textit{``Images created by GenAI models can be offensive or harmful, and this may vary from person to person. In your opinion, does the above image of [artifact name] seem offensive or harmful to you?''} Similar to cultural representativeness, we ask workers to briefly justify their scores through text, \ie \textit{``Please provide a justification for your score above - what about this image is offensive or not offensive, in your personal view? If it is not at all offensive, simply stating so is sufficient.''}

In contrast, to assess whether a T2I system perpetuates visual cultural stereotypes, we provide workers with examples of how T2I systems can be stereotypical (see \cref{fig:us_stereo} for details), as we observed a vast gulf in understanding of what ``stereotypical'' means in the context of T2I systems during our pilot study. 

\begin{figure}[hbt!]
    \centering
    \includegraphics[width=1\linewidth]{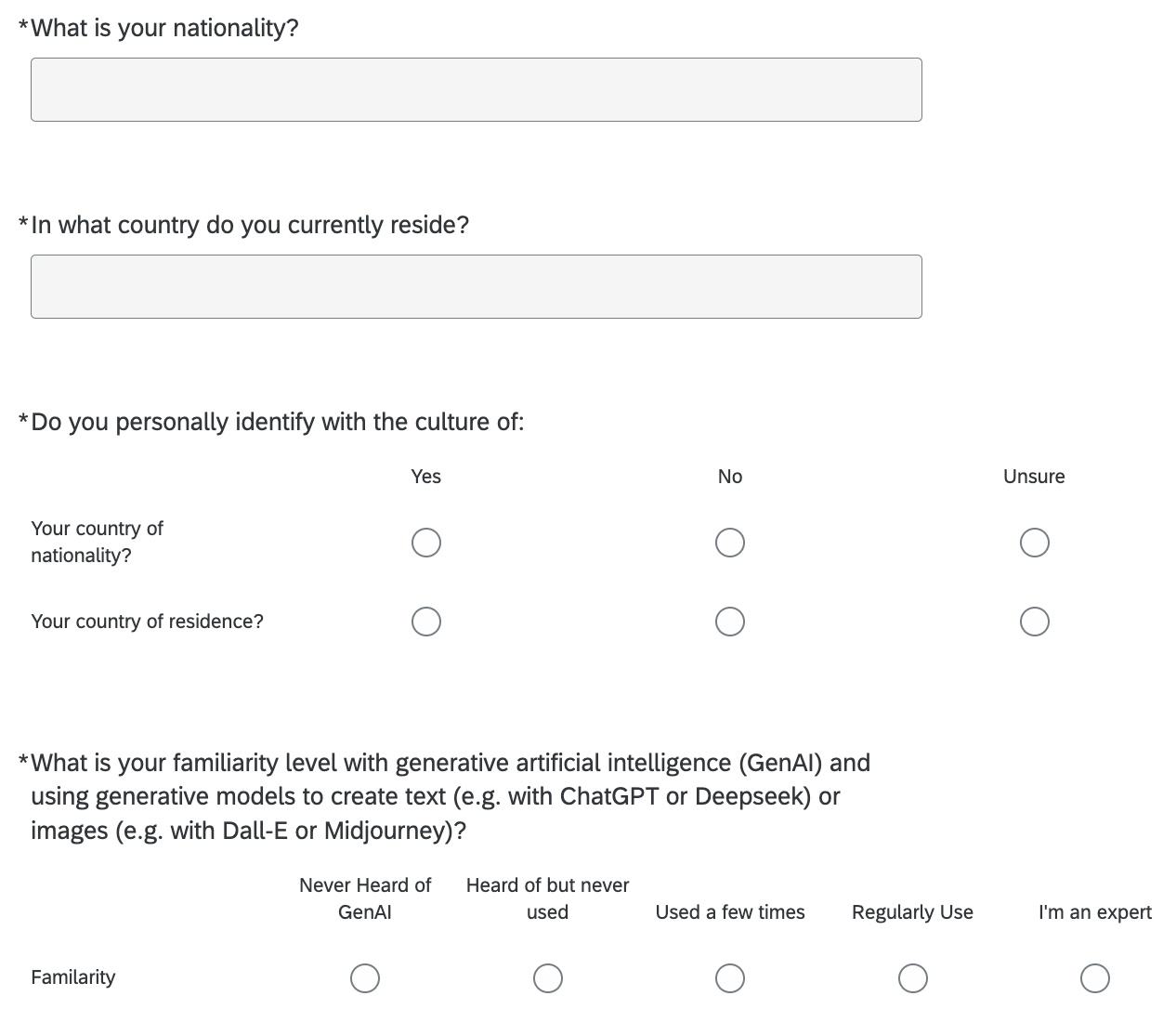}
    \caption{Demographic information questions.}
    \label{fig:us_demo}
\end{figure}

\subsection{User Metadata}
\label{app:user_study_demographics}

Users are asked to provide non-identifiable metadata for post-hoc analyis, \ie their country of nationality and residence, how much they identify with the culture of their country of nationality and residence, and their level of familiarity with T2I systems (\cref{fig:us_demo}). Users are then queried for Likert scores from 1 (``Not at all'') to 5 (``A lot''), similar to offensiveness: \textit{``Below is an image of \textbf{[artifact name]}, a type of \textbf{[category name]} created by a GenAI model. In your opinion, how much does this image reflect any stereotypes about your culture or country (of nationality or residence)?''}. Similarly to offensiveness, they are also asked for a textual justification of their score, \ie \textit{``Please provide a justification for your choice - what about this image is stereotypical or not stereotypical of your culture or country, in your opinion? If you do not think it is stereotypical at all, simply stating so is sufficient.''}

\section{Analysis of User Studies}
\label{app:us_analytics}
We provide a detailed study of the user study responses across our 2700 total surveys (3 T2I systems $\times$ 300 artifacts $\times$ 3 workers per artifact).

\subsection{Comparison to prior User Studies}

We compare our \cure user study to previous benchmark and evaluations of T2I systems that included extensive user studies as core contributions, which we show in \cref{tab:user_study_literature}. We highlight that to our knowledge, ours is the only work that queries explicitly for user cultural identity (\ie for worker hired to score an artifact $n$ from country $r$, we ask them if they identify with the culture of $\{r\}$) via worker metadata (see \cref{fig:us_metadata} for details).

\begin{table}[t!]
\caption{We tabulate a comparison of \cure to existing works for contributions towards culture-specific user studies. Here WRK = did the study ask for worker metadata (``Do you identify with the culture of your country of nationality?"), REP = cultural representativeness, RLM = realism, PS = perceptual similarity, OFF = offensiveness, STR = stereotypical, $\rho$-MET = does the work analyze how their metrics correlate with real human judgments?}
\resizebox{\columnwidth}{!}{
\begin{tabular}{@{}l|ccccccc@{}}
\toprule
\multicolumn{1}{c|}{\textbf{Benchmark}} & \textbf{WRK} & \textbf{REP} & \textbf{RLM} & \textbf{PS} & \textbf{OFF} & \textbf{STR} & \textbf{$\rho$-MET} \\ \midrule
\citet{liu2023towards}                 &              & \checkmark   &              & \checkmark  & \checkmark   &              & \textbf{}           \\
\citet{basu2023inspecting}             &              & \checkmark   & \checkmark   &             &              &              & \checkmark          \\
\citet{ventura2023navigating}          &              & \checkmark   &              &             &              &              &                     \\
\citet{khanuja2024image}               &              & \checkmark   & \checkmark   &             & \checkmark   &              & \checkmark          \\
\citet{kannen2024beyond}               &              & \checkmark   & \checkmark   &             &              &              & \checkmark          \\
\citet{zhang2024partiality}            &              &              &              & \checkmark  &              & \checkmark   &                     \\
\citet{bayramli2025diffusion}          &              &              & \checkmark   & \checkmark  &              &              &                     \\ \midrule
\cure                                  & \checkmark   & \checkmark   &              & \checkmark  & \checkmark   & \checkmark   & \checkmark          \\ \bottomrule
\end{tabular}
}
\label{tab:user_study_literature}
\end{table}

\subsection{Inter-Annotator and Encoder Agreement}
\label{app:worker_disagreement}
We use the perceptual similarity ranking from \textbf{Q1b.} to measure agreement between survey respondents over perceptual similarity the same artifact $I(n)$ to its ground-truth images $G(n)$ across T2I systems and image encoders. To compute the similarity ranking from our scorer $\phi_{PS}(n)$, we sort the cosine distance between dense embeddings in descending order, which matches how users were queried (\cref{fig:us_drag}). We compute a Kendall's Tau distance between the rankings from the user study and the ranking given by our scorer with three image encoder: SigLIP 2, AIMV2 and DINOv2. We compute agreement between a ranking pair $(r_i, r_j)$ as $a = (1 - \frac{KD(r_i, r_j)}{\max(KD(r_i, r_j))})$.  In \cref{tab:agreement}, we tabulate an average over all permutations of ranking pairs for an artifact $n$ in two settings: first, only ranking pairs provided by workers (``Worker Only''), and then for each (worker, encoder) ranking pair:

\begin{equation}
\text{agreement} = \dfrac{1}{|(i,j)|}\sum_{(i,j)} a(i,j)
\label{eq:worker_agreement}    
\end{equation}

For example, for agreement over only three survey respondents for each survey (``Worker Only''), we have $3c2 = 3$ pairs of rankings to compute agreement over, which we average. Similarly for each image encoder, we have $3c2 = 3$ pairs of rankings to average over, \ie one encoder ranking compared against each of the three survey respondent rankings. 

As seen in \cref{tab:agreement}, we observe that disagreements are fairly consistent between workers across T2I systems when averaged over the entire \cure dataset. When adding the ranking of an encoder to  \cref{eq:worker_agreement}, there is minimal change in the agreement value.

\begin{table}[hbt!]
\centering
\caption{Agreement between the user survey responses and the other users who took the same survey or different image encoders.}
\begin{adjustbox}{max width=\columnwidth}
\begin{tabular}{c|ccc}
\toprule
\multirow{2}{*}{\textbf{Config}} & \multicolumn{3}{c}{\textbf{User Survey}} \\ 
\cmidrule{2-4} 
 & \textbf{FLUX.1 [dev]} & \textbf{SD 3.5 Large} & \textbf{SD 1.5} \\ 
\midrule
\textbf{Worker Only} & 0.776 $\pm$ 0.028 & 0.771 $\pm$ 0.027 & 0.778 $\pm$ 0.029 \\ \midrule
\textbf{SigLIP 2} & 0.759 $\pm$ 0.026 & 0.756 $\pm$ 0.026 & 0.758 $\pm$ 0.026 \\
\textbf{AIMV2} & 0.755 $\pm$ 0.024 & 0.754 $\pm$ 0.028 & 0.752 $\pm$ 0.025 \\ 
\textbf{DINOv2} & 0.763 $\pm$ 0.027 & 0.754 $\pm$ 0.026 & 0.754 $\pm$ 0.025\\ 
\bottomrule
\end{tabular}
\end{adjustbox}
\label{tab:agreement}
\end{table}

We also qualitatively examine cases with high worker disagreement over cultural representativeness $\phi^*_{\cure}$ (\ie, one worker assigns high score, and one low), alongside their textual justification for their score in \cref{fig:us_disagree_combined_part1}. A major cause of disagreement is a miscalibration between worker thoughts and the Likert score selected. For example, for \textbf{E2} (\textit{Rost\'as P\'al Monument}), workers disagree on the historical relevance to Slovenia, but one worker gives a score of 5 out of 5 even though they ``don't think it has much in common with the original details.'' A similar miscalibration occurs with \textbf{E7} (\textit{Hardangerbunad}), where one annotator gives a perfect score despite explicitly stating the image does not depict the traditional clothing at all, instead showing a scenic landscape.

Another common disagreement arises from workers' differing emphases on semantic content versus specific visual details as larger contributors to \cure. For instance, for the \textbf{E1} (\textit{Yangpu bridge}), one worker says the main tower looks European instead of Chinese, while another remarks that the bridge structure and details are reminiscent of more modern bridges in China. This pattern also appears in several additional examples: \textbf{E3} (\textit{Hogmanay}), where where one worker highlights incorrect details, while another focuses on the fireworks and crowd presence being accurate; \textbf{E8} (\textit{Kaapse Klopse}), where one worker pointed out the depicted person's race was wrong, but the other stated the image was accurate; \textbf{E10} (\textit{Golestan Palace}), where architectural detail is judged differently by annotators; and \textbf{E11}(\textit{Stanza dell'Amore Coniugale}), where one worker assigns low score as the clothes, composition, colors, and subjects are inaccurate to 16th century Italian art, while another gives a high score due to one specific detail being wrong.

Finally, in another distinct case, workers appear to agree in their justification but assign vastly different Likert scores. In \textbf{E4} (\textit{cowboy hat}), this discrepancy appears to result from a misunderstanding of the Likert scale rubric (1 is low and 5 is high). Similarly for \textbf{E12} (\textit{Brooklyn Bridge}), one annotator highlights several authenticity problems such as an overly clear sky and incorrect arch proportions yet still gives a relatively high score (4 out of 5).  In \textbf{E9} (\textit{Kue Nastar}), both workers critique an unrealistic topping decoration, but one penalizes this mistake heavily (1 out of 5) while another does not penalize it it at all (5 out of 5). Lastly for \textbf{E5} (\textit{pabell\'on criollo}), a worker assigns a high score despite recognizing that the image does not depict the intended dish, suggesting that the error arose due to underspecification in the T2I prompt (\ie the T2I system should be told explicitly that pabell\'on criollo is a type of food).

\begin{figure*}[hbt!]
\begin{adjustbox}{width=\textwidth}
    \begin{tabular}{@{}l|cc|l@{}}
        \toprule
        \textbf{E\#} &
        \textbf{AI Image}      & \textbf{Real Image}        & \multicolumn{1}{c}{\textbf{Feedback}}  \\  \midrule
        \textbf{E1} &
        \begin{tabular}[c]{@{}c@{}}
            \includegraphics[width=0.50\columnwidth]{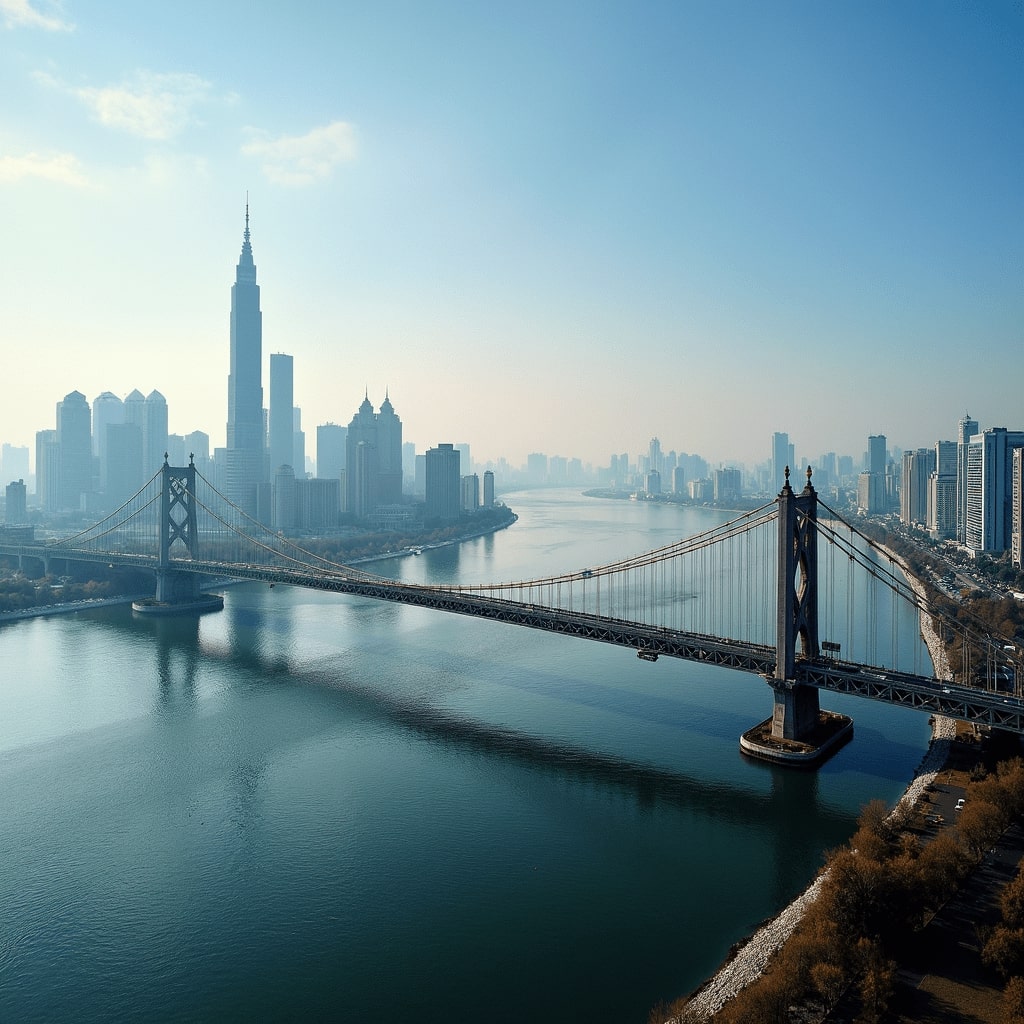} 
        \end{tabular}
        &
        \begin{tabular}[c]{@{}c@{}}
            \includegraphics[width=0.50\columnwidth]{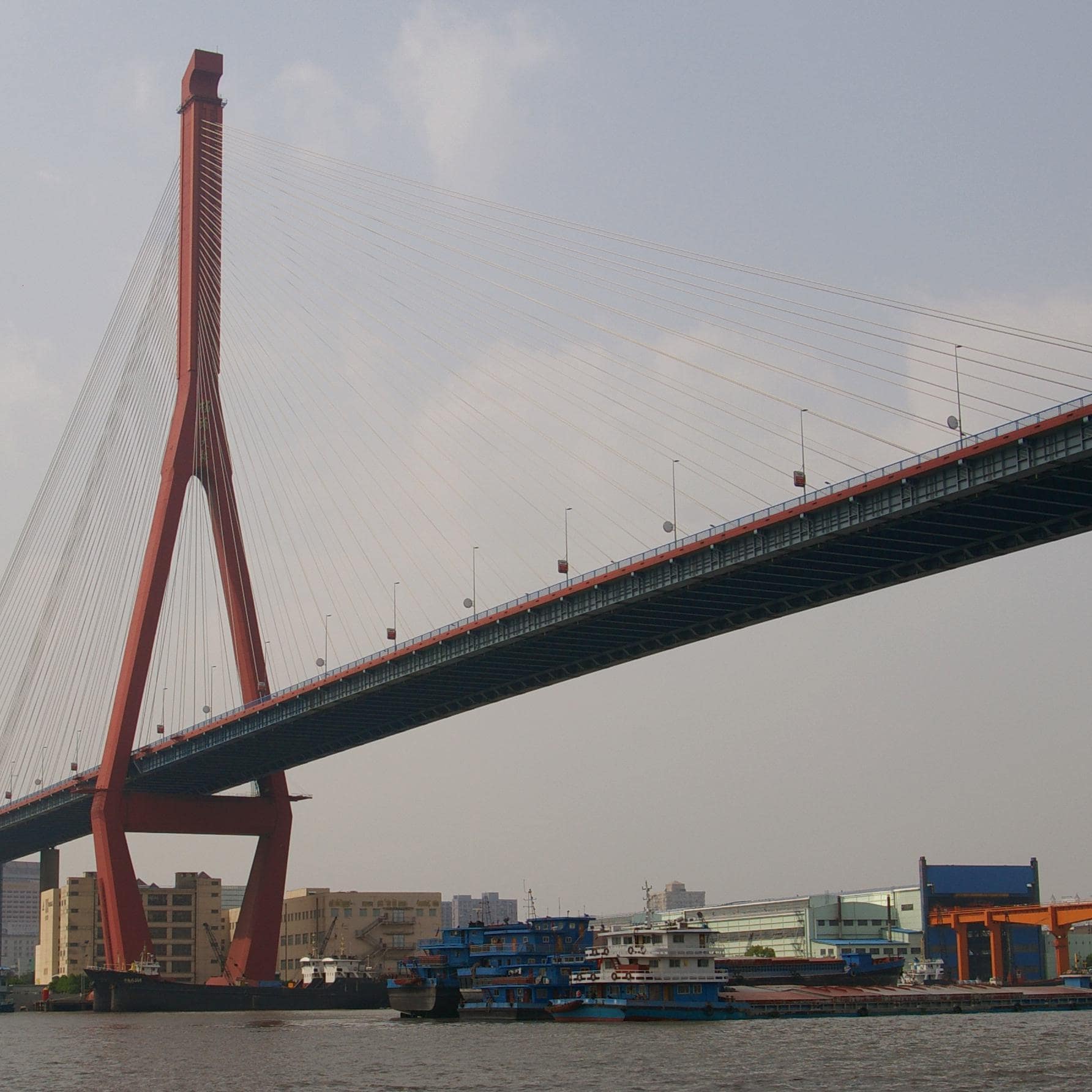}
        \end{tabular}
        &
        \begin{tabular}[l]{@{}l@{}} 
        \textit{\textbf{Yangpu Bridge}} \\ \\
            $\phi^*_{\cure}$ = 1 out of 5: The most different part is probably the main tower. The AI-generated \\main tower is more like something that would appear in Europe. \\ \\
            $\phi^*_{\cure}$ = 5 out of 5: This is very typical of modern bridges in China. The structure is very \\normal; there are cars driving over the bridge; both sides of the river have tall, modern \\buildings and even skyscrappers.
        \end{tabular}
        \\
        \midrule 

        \textbf{E2} &
        \begin{tabular}[c]{@{}c@{}}
            \includegraphics[width=0.50\columnwidth]{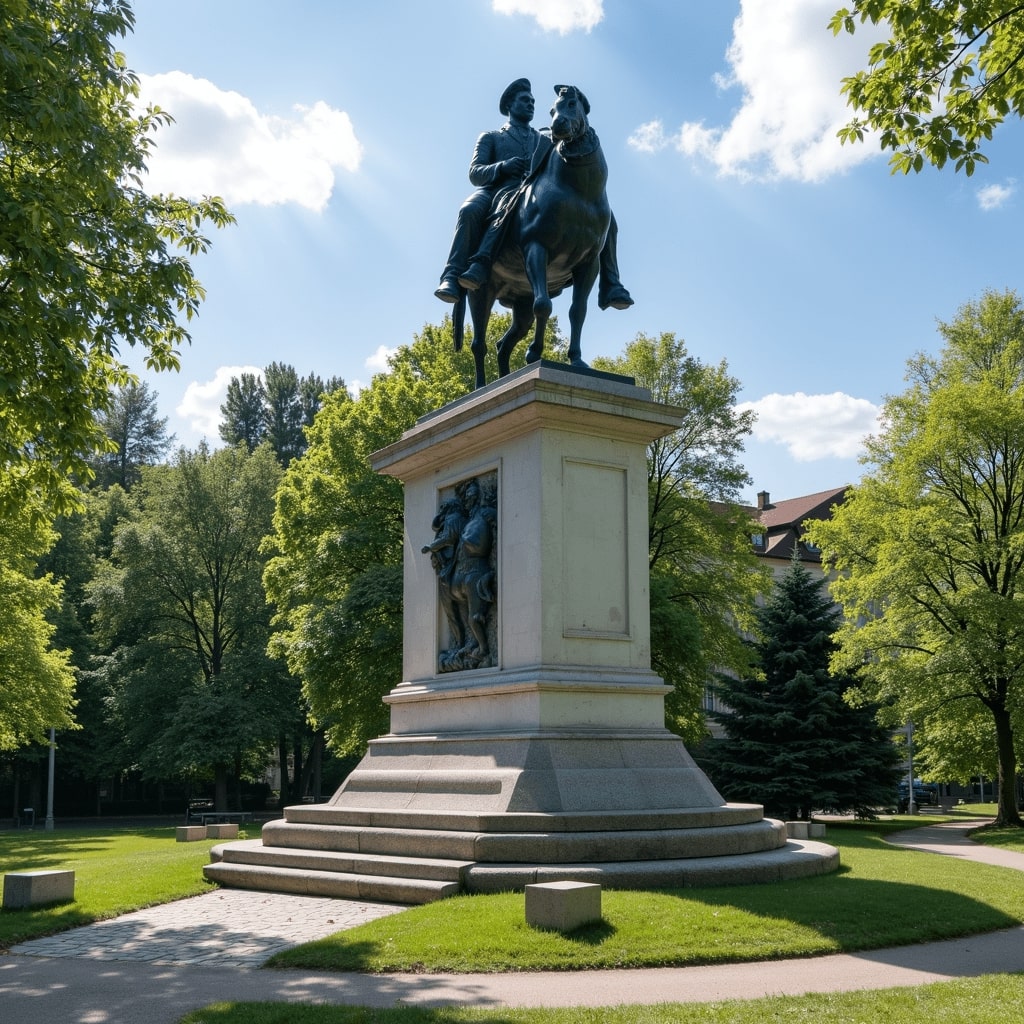} 
        \end{tabular}
        &
        \begin{tabular}[c]{@{}c@{}}
            \includegraphics[width=0.50\columnwidth]{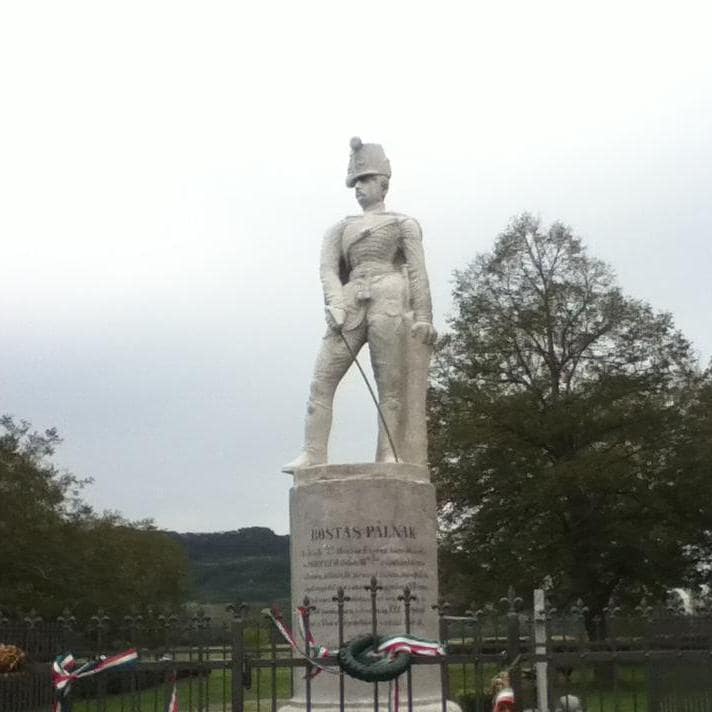} 
        \end{tabular}
        &
        \begin{tabular}[l]{@{}l@{}}
        \textit{\textbf{Rost\'as P\'al Monument}} \\ \\ 
            $\phi^*_{\cure}$ = 1 out of 5: the picture does not show the history of the Slovenian nation \\and does not show the history of this area-europe at all \\ \\
            $\phi^*_{\cure}$ = 5 out of 5: I think Al correctly interpreted the historical and cultural context \\of my country, although I don't think it has much in common with the original details.
        \end{tabular}
        \\
        \midrule

        \textbf{E3} & 
        \begin{tabular}[c]{@{}c@{}}
            \includegraphics[width=0.50\columnwidth]{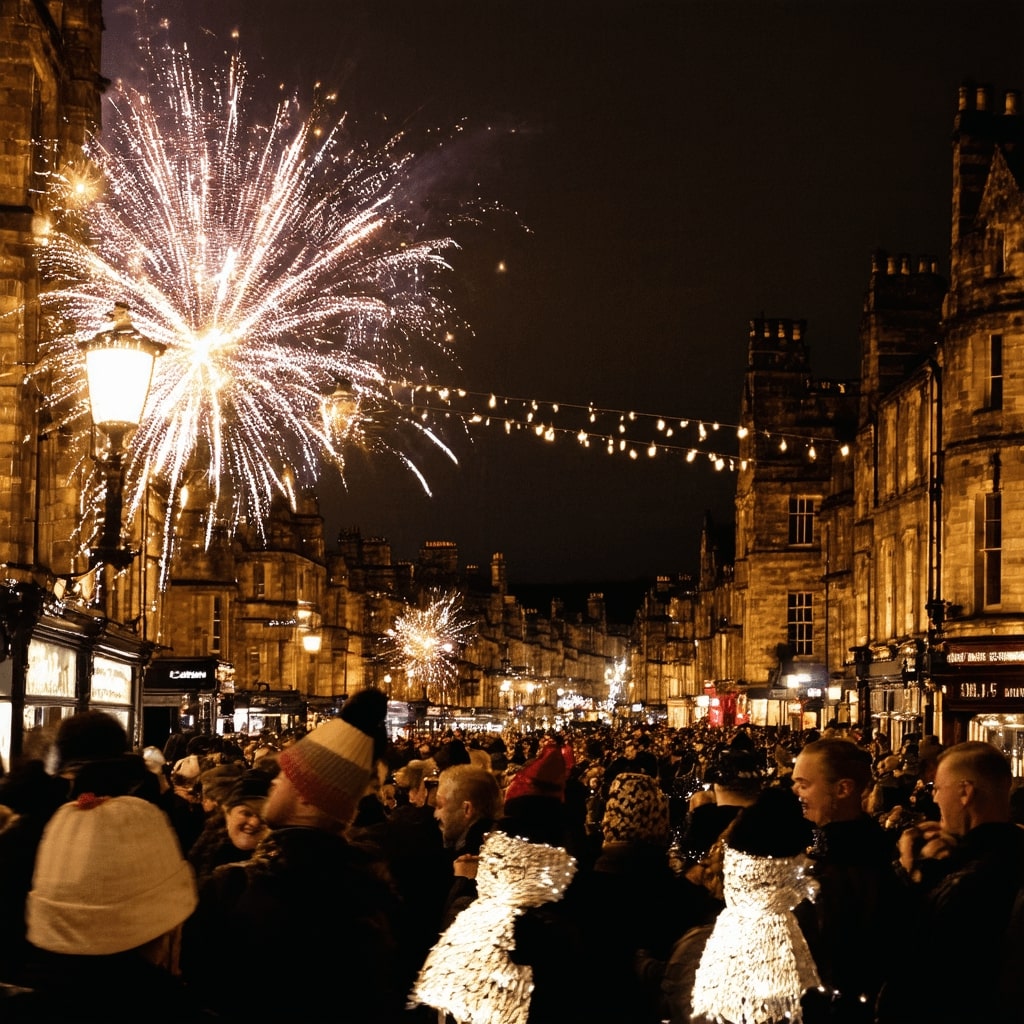} 
        \end{tabular}
        &
        \begin{tabular}[c]{@{}c@{}}
            \includegraphics[width=0.50\columnwidth]{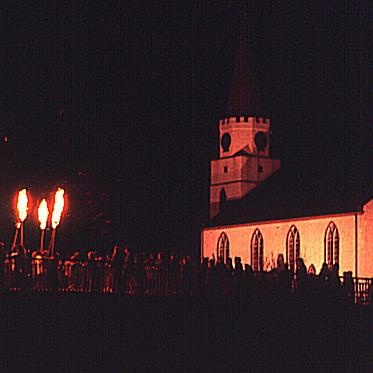} 
        \end{tabular}
        &
        \begin{tabular}[l]{@{}l@{}}
        \textit{\textbf{Hogmanay}} \\ \\ 
            $\phi^*_{\cure}$ = 1 out of 5: fireworks are too low, faces blurry and distorted \\ \\
            $\phi^*_{\cure}$ = 5 out of 5: The image shows displays of fireworks which are often associated \\with celebrating Hogmanay, and also large crowds of people who gather t[o] celebrate \\and watch the fireworks
        \end{tabular}
        \\

        \midrule

        \textbf{E4} & 
        \begin{tabular}[c]{@{}c@{}}
            \includegraphics[width=0.50\columnwidth]{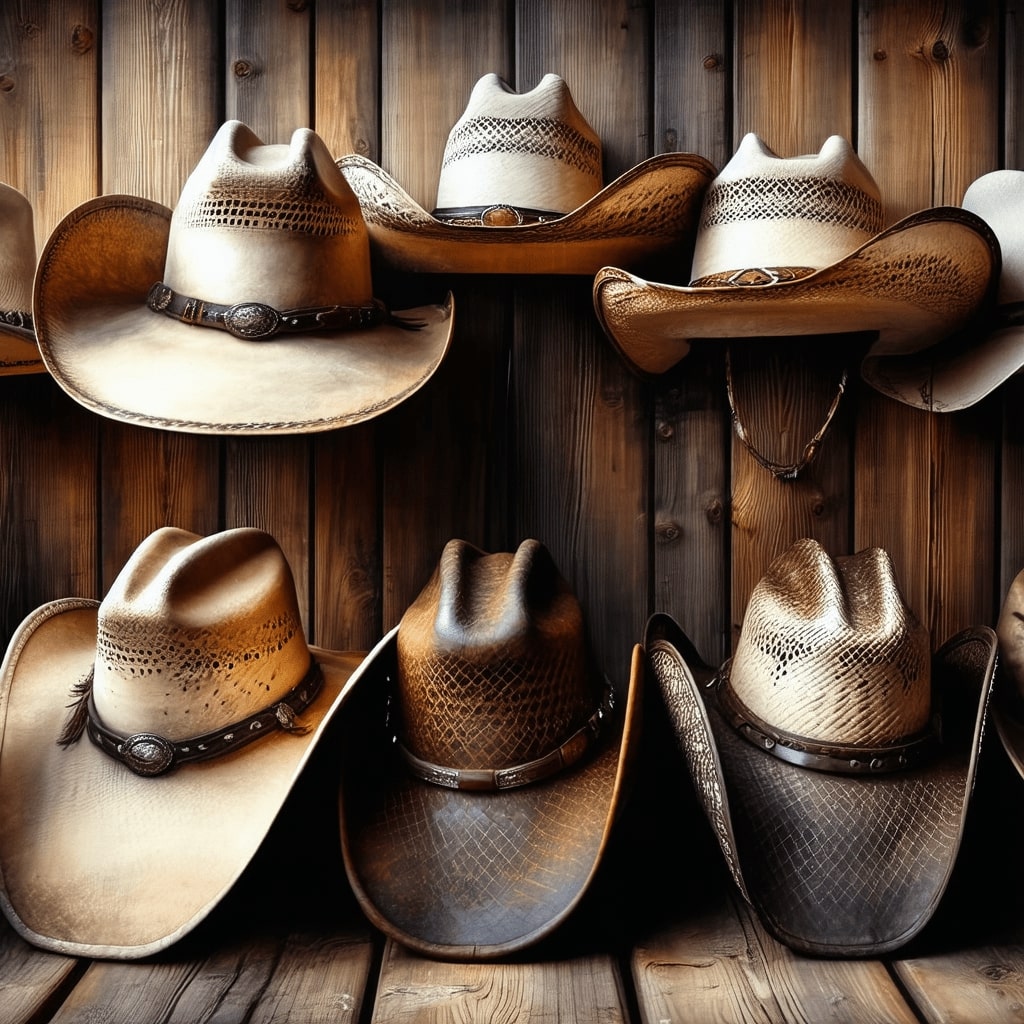} 
        \end{tabular}
        &
        \begin{tabular}[c]{@{}c@{}}
            \includegraphics[width=0.50\columnwidth]{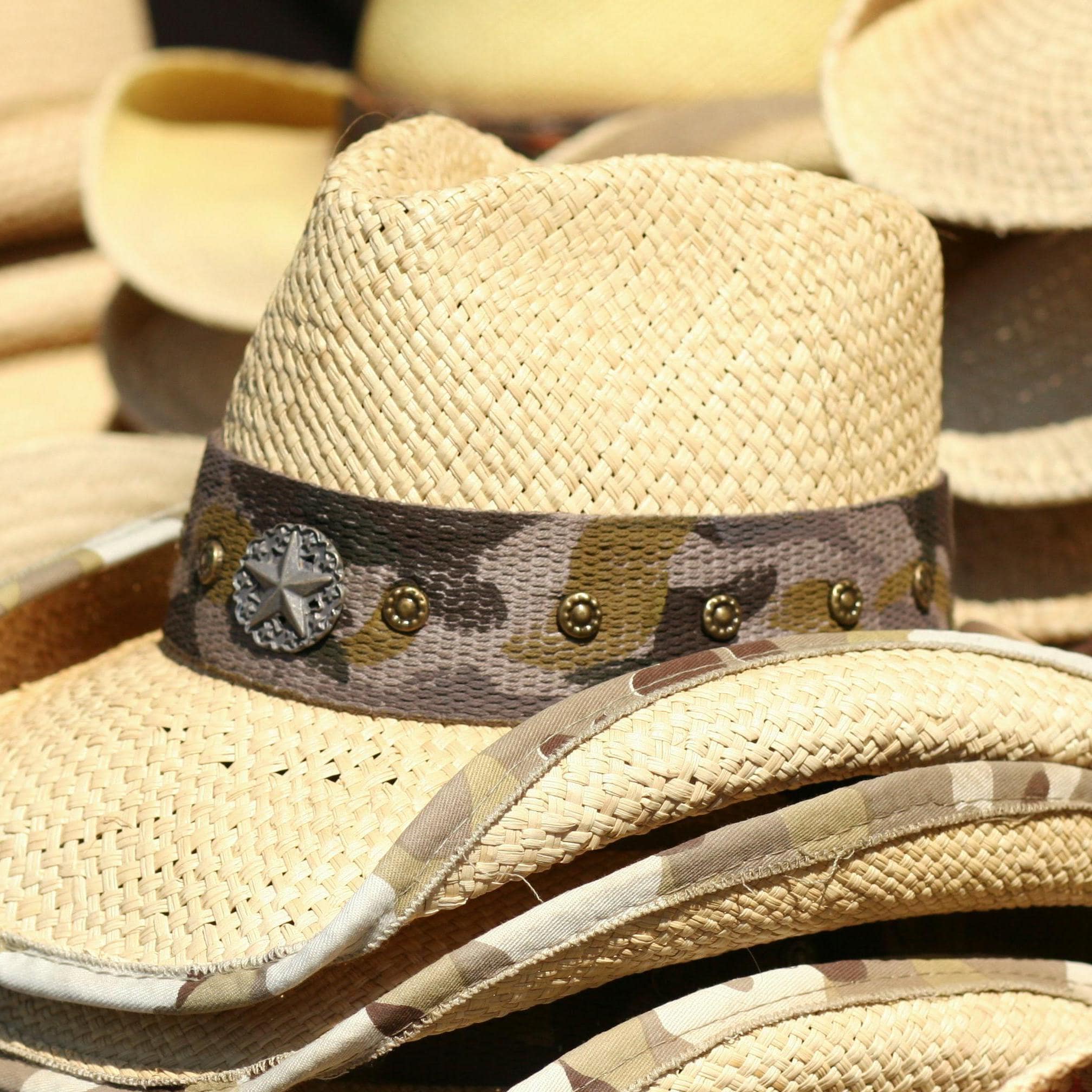} 
        \end{tabular}
        &
        \begin{tabular}[l]{@{}l@{}}
        \textit{\textbf{Cowboy Hat}} \\ \\
            $\phi^*_{\cure}$ = 1 out of 5: The image of cowboy hats is pretty accurate. In my country \\of residence, this would be a pretty representative image. \\ \\
            $\phi^*_{\cure}$ = 5 out of 5: Cowboy hats from my country are made from straws or leather \\which is shown in the picture here.
        \end{tabular}
        \\
        \midrule

        \textbf{E5} & 
        \begin{tabular}[c]{@{}c@{}}
            \includegraphics[width=0.50\columnwidth]{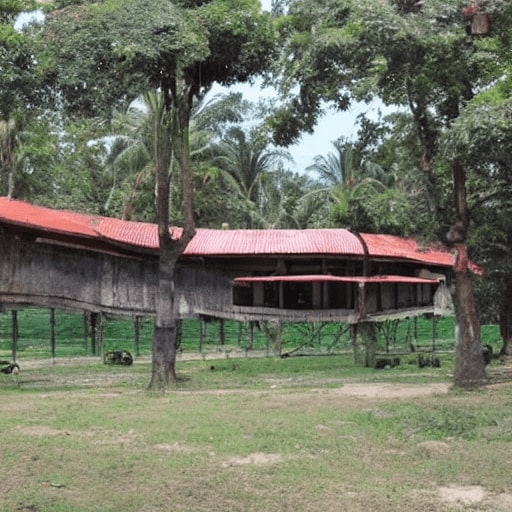} 
        \end{tabular}
        &
        \begin{tabular}[c]{@{}c@{}}
            \includegraphics[width=0.50\columnwidth]{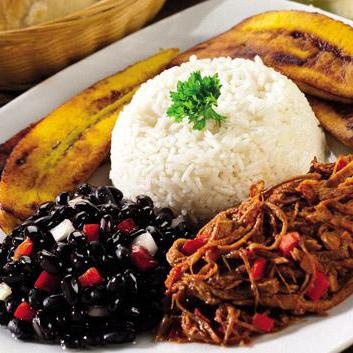} 
        \end{tabular}
        &
        \begin{tabular}[l]{@{}l@{}}
        \textit{\textbf{Pabell\`on criollo}} \\ \\
            $\phi^*_{\cure}$ = 1 out of 5: The image does not look anything like the typical dish. It has \\nothing in common, it shows a "corridor type pavilion" I think is very misguided. \\ \\
            $\phi^*_{\cure}$ = 5 out of 5: Ok, the image created by the AI is a creole Pavillion in terms of \\a farm in Venezuela, i think when it was created by AI , the instructions have to be \\more specific, i mean telling the AI that is referred to a typical food
        \end{tabular}
        \\

        \midrule

        \textbf{E6} &
        \begin{tabular}[c]{@{}c@{}}
            \includegraphics[width=0.50\columnwidth]{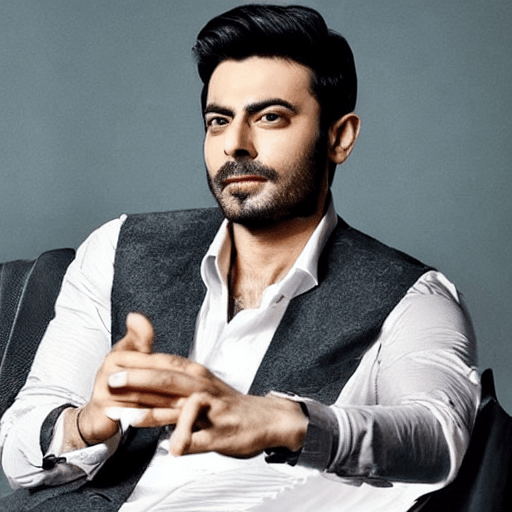}
        \end{tabular}
        &
        \begin{tabular}[c]{@{}c@{}}
            \includegraphics[width=0.50\columnwidth]{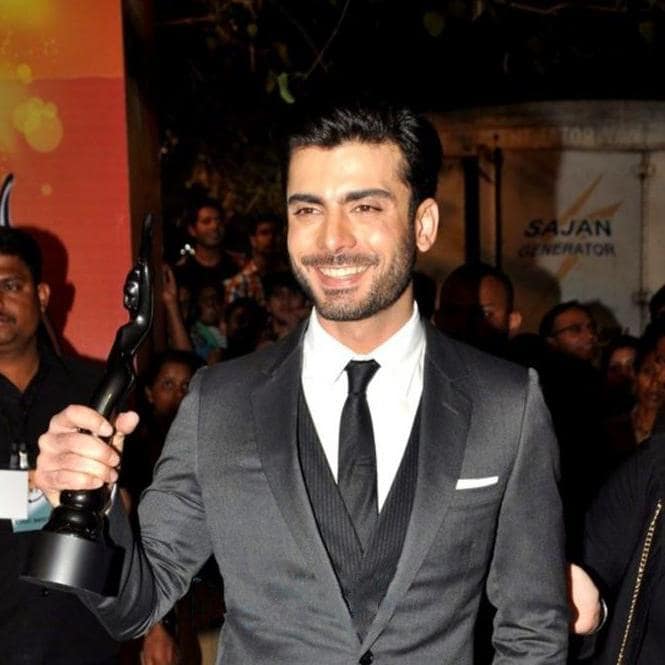} 
        \end{tabular}
        &
        \begin{tabular}[l]{@{}l@{}}
        \textit{\textbf{Fawad Khan}} \\ \\
            $\phi^*_{\cure}$ = 2 out of 5: The image is more bright and shows different features of him \\ \\
            $\phi^*_{\cure}$ = 5 out of 5: In the image he is wearing shalwar kameez with a waistcoat \\which is a typical attire of Men in Pakistan. So the image is quite accurate.
        \end{tabular}
        \\
        \bottomrule 
    \end{tabular}
\end{adjustbox}
\caption{Feedbacks of the workers in cases of high disagreement over perceptions of cultural representativeness $\phi^*_{\cure}$ of the T2I system for artifact $n$ (Part 1).}
\label{fig:us_disagree_combined_part1}
\end{figure*}

\begin{figure*}[hbt!]
\ContinuedFloat
\begin{adjustbox}{width=\textwidth}
    \begin{tabular}{@{}l|cc|l@{}}
        \toprule
        \textbf{E\#} &
        \textbf{AI Image}      & \textbf{Real Image}        & \multicolumn{1}{c}{\textbf{Feedback}} \\ \midrule
        \textbf{E7}  &
        \begin{tabular}[c]{@{}c@{}}
    \includegraphics[width=0.52\columnwidth]{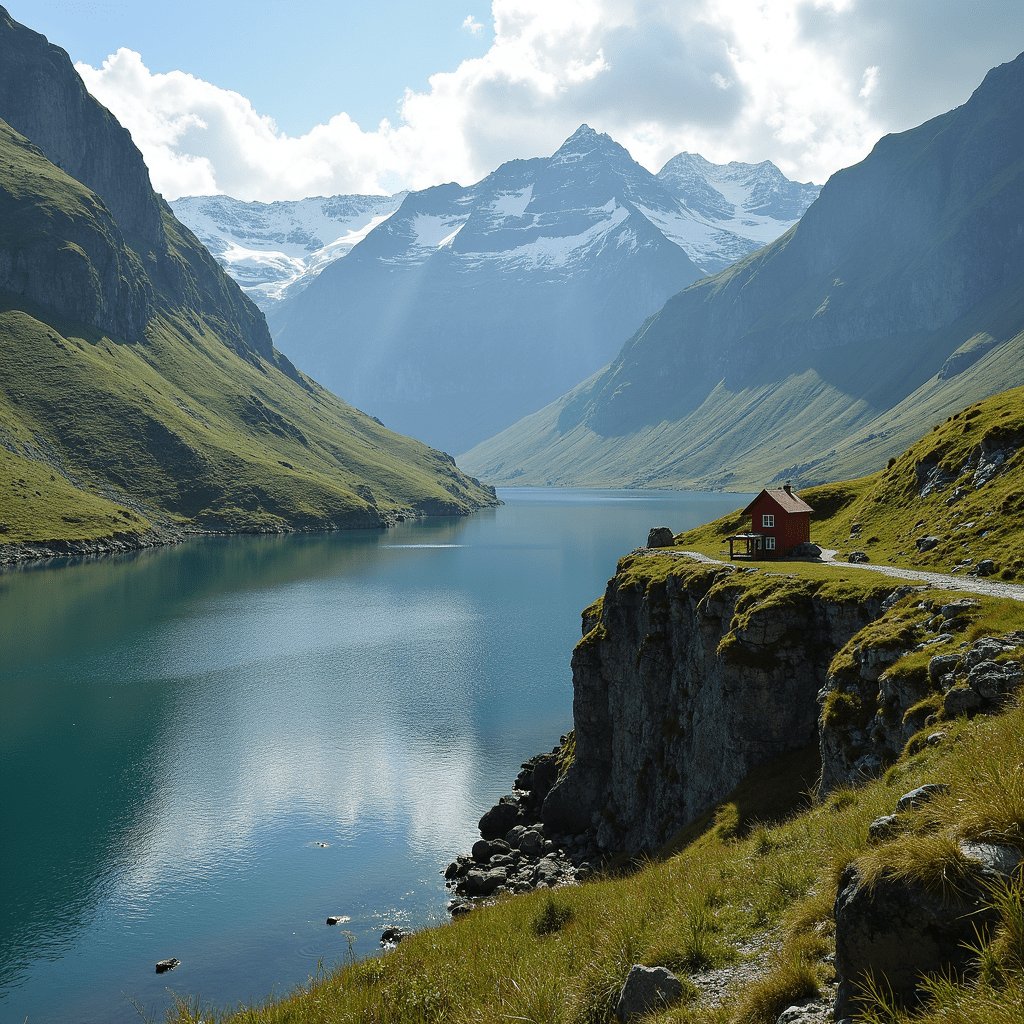}
        \end{tabular}
        &
        \begin{tabular}[c]{@{}c@{}}
            \includegraphics[width=0.52\columnwidth]{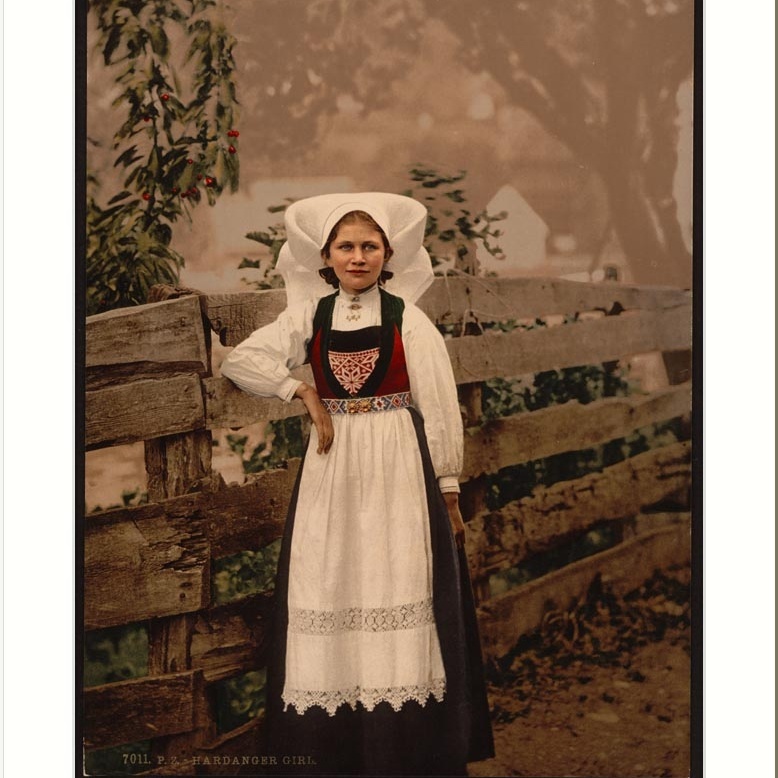} 
        \end{tabular}
        &
        \begin{tabular}[l]{@{}l@{}} 
        \textit{\textbf{Hardangerbunad}} \\ \\
            $\phi^*_{\cure}$ = 1 out of 5: This image doesn't show anything connected to Hardangerbunad \\ which is an item of clothing, it shows a landscape with mountains and fjords.
            \\ \\
            $\phi^*_{\cure}$ = 5 out of 5: This isn't a picture of a bunad at all, but rather a picture associated \\ with Norwegian cultural heritage, a beautiful landscape with mountains and fjords.
        \end{tabular}
        \\
        \midrule

        \textbf{E8} &
        \begin{tabular}[c]{@{}c@{}}
            \includegraphics[width=0.52\columnwidth]{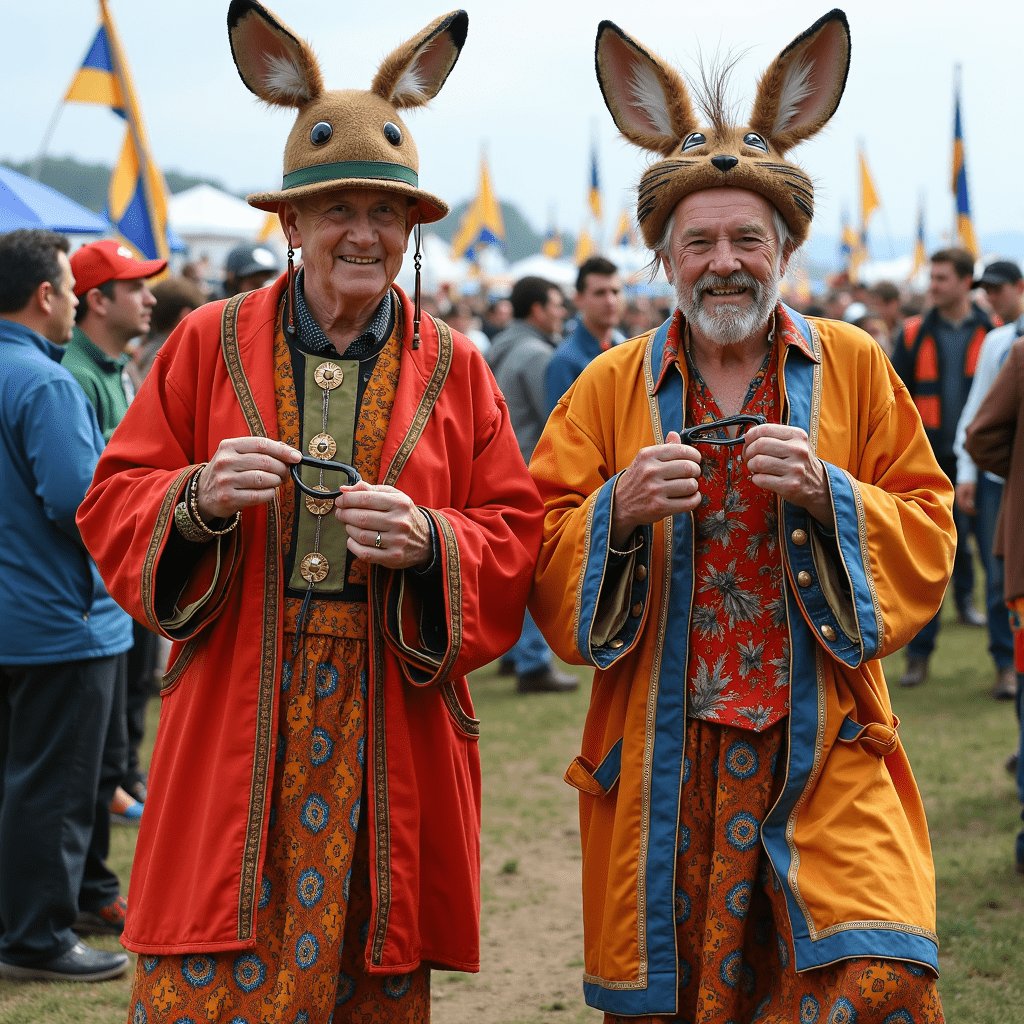} 
        \end{tabular}
        &
        \begin{tabular}[c]{@{}c@{}}
            \includegraphics[width=0.52\columnwidth]{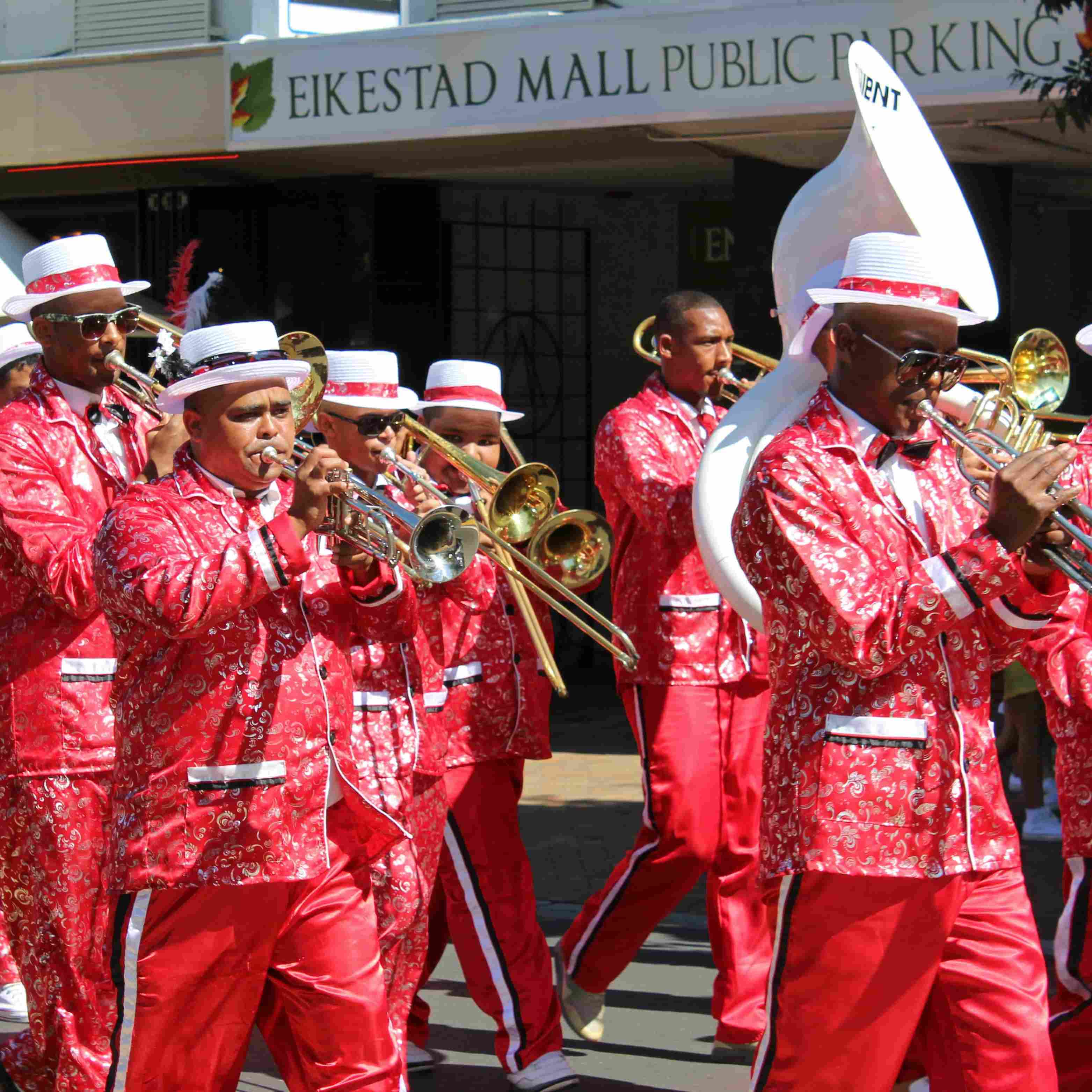}
        \end{tabular}
        &
        \begin{tabular}[l]{@{}l@{}}
        \textit{\textbf{Kaapse Klopse}} \\ \\
            $\phi^*_{\cure}$ = 1 out of 5: The image used the wrong race of the person \\
            \\
            $\phi^*_{\cure}$ = 5 out of 5: It appears much acurate and more clear
        \end{tabular}
        \\

        \midrule

        \textbf{E9} & 
        \begin{tabular}[c]{@{}c@{}}
            \includegraphics[width=0.52\columnwidth]{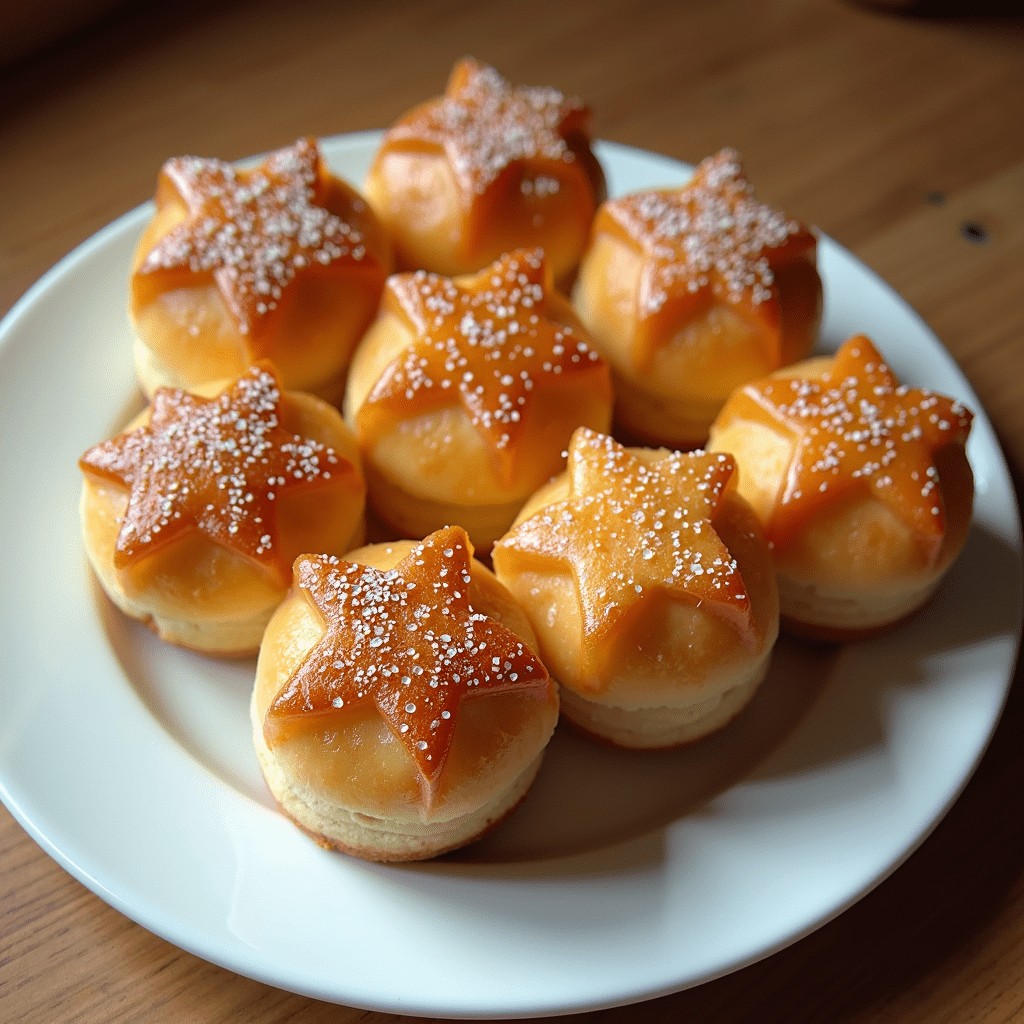} 
        \end{tabular}
        &
        \begin{tabular}[c]{@{}c@{}}
            \includegraphics[width=0.52\columnwidth]{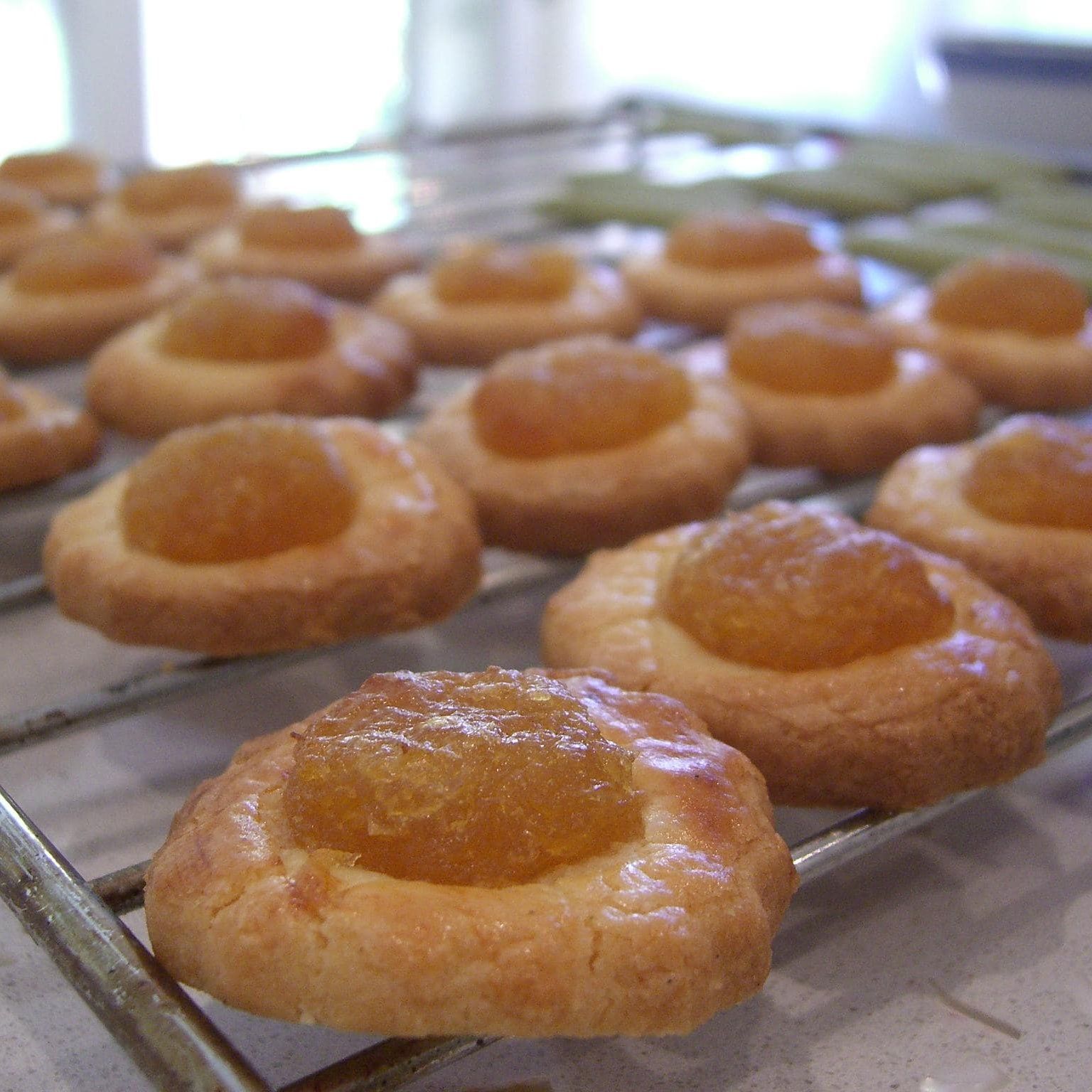}
        \end{tabular}
        &
        \begin{tabular}[l]{@{}l@{}}
        \textit{\textbf{Kue Nastar}} \\ \\
            $\phi^*_{\cure}$ = 1 out of 5: the star on top of the pastry is not realistic and is unlikely to be \\ found in my country. The colour of the pastry is similar to what a kue nastar would look \\ like in my culture \\
            \\
            $\phi^*_{\cure}$ = 5 out of 5: It's almost similar to normal nastar, except that nastar in my country \\ usually has cheese on top or just plain.
        \end{tabular}
        \\
        \midrule

        \textbf{E10} &
        \begin{tabular}[c]{@{}c@{}}
            \includegraphics[width=0.52\columnwidth]{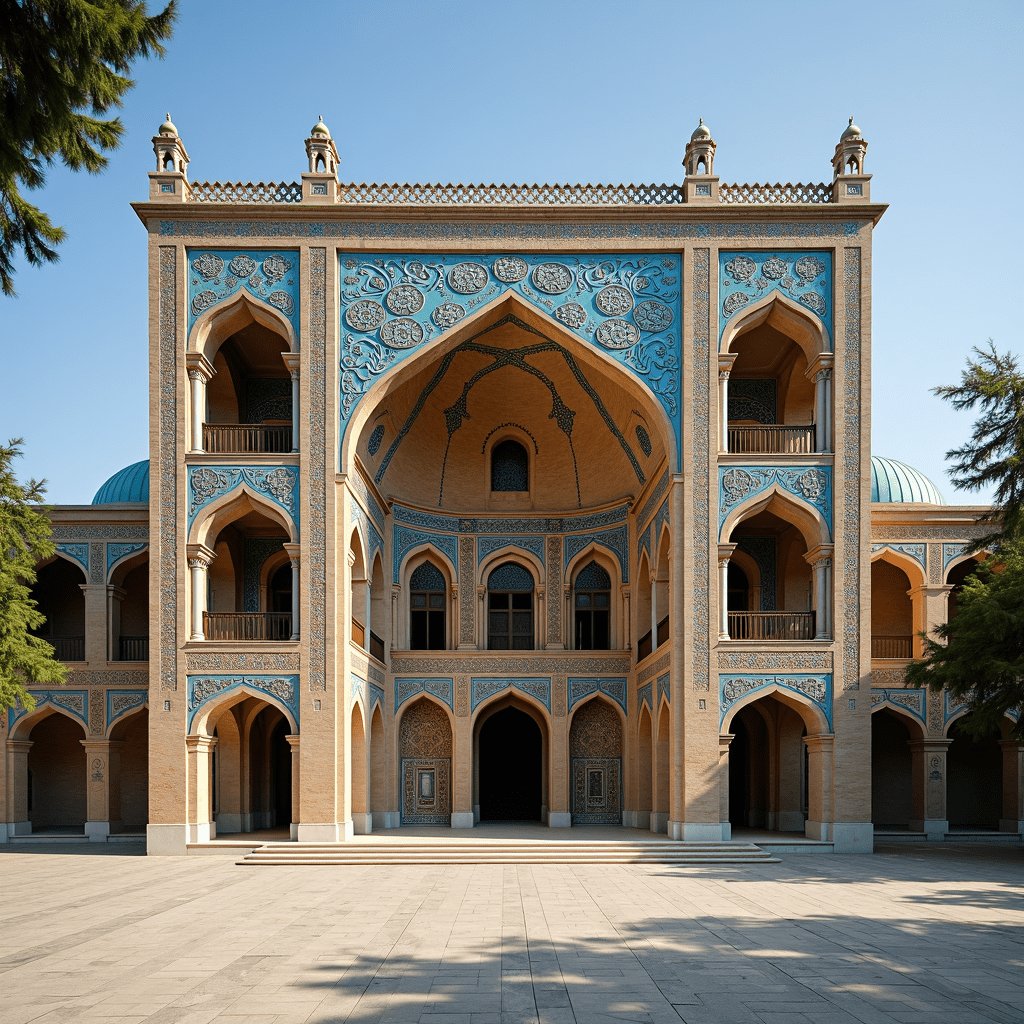} 
        \end{tabular}
        &
        \begin{tabular}[c]{@{}c@{}}
            \includegraphics[width=0.52\columnwidth]{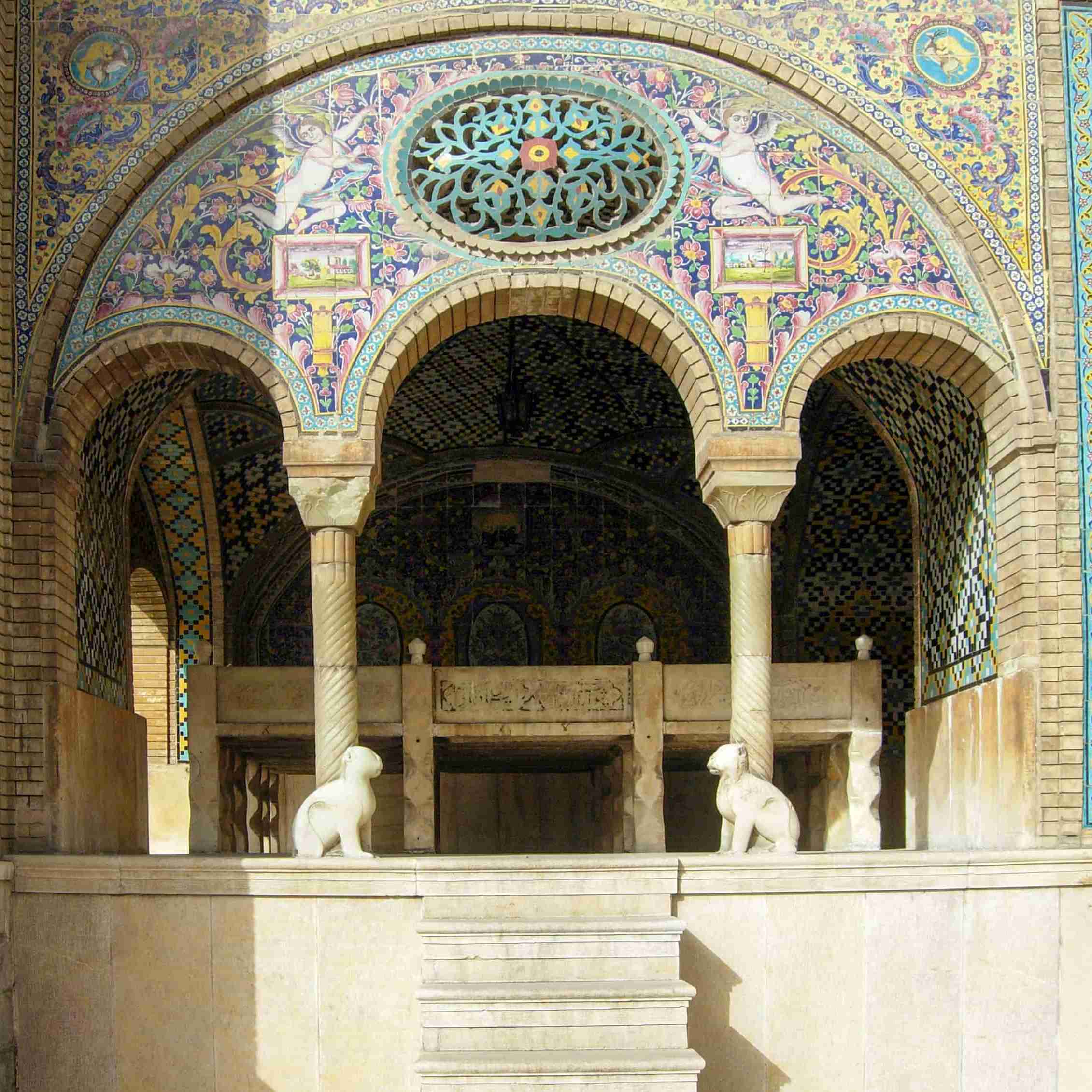}
        \end{tabular}
        &
        \begin{tabular}[l]{@{}l@{}}
        \textit{\textbf{Golestan Palace}} \\ \\
            $\phi^*_{\cure}$ = 2 out of 5: The flooring's pattern is really irrelevant. You will not find this pattern  ~~\\ in Iranian architecture. Besides the walls are exaggerated. If we neglect the pavement, it \\ could be an Iranian moscque \\
            \\
            $\phi^*_{\cure}$ = 5 out of 5: The architecture of windows, walls and ceramics are really like the \\ samples that can be seen in iran palaces or mosques
        \end{tabular}
        \\
        \midrule

        \textbf{E11} &
        \begin{tabular}[c]{@{}c@{}}
            \includegraphics[width=0.52\columnwidth]{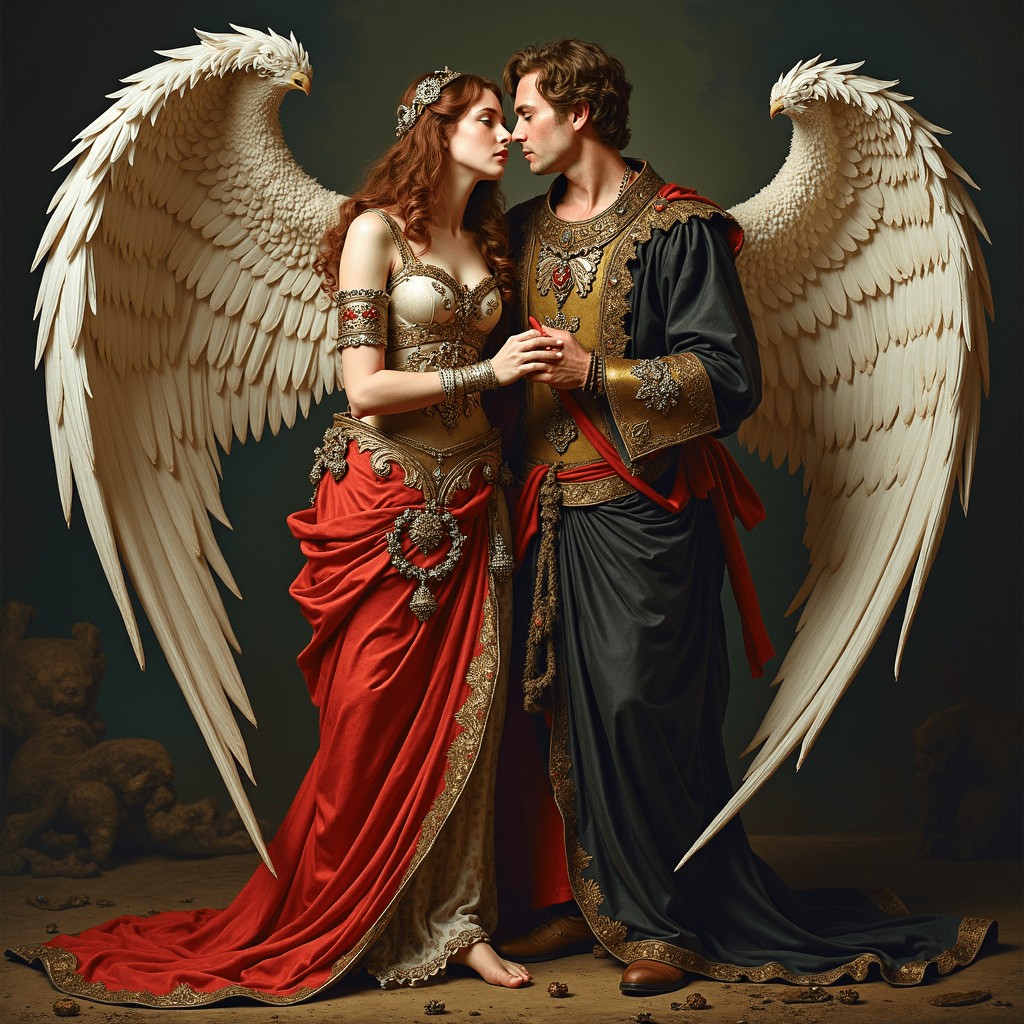}
        \end{tabular}
        &
        \begin{tabular}[c]{@{}c@{}}
            \includegraphics[width=0.52\columnwidth]{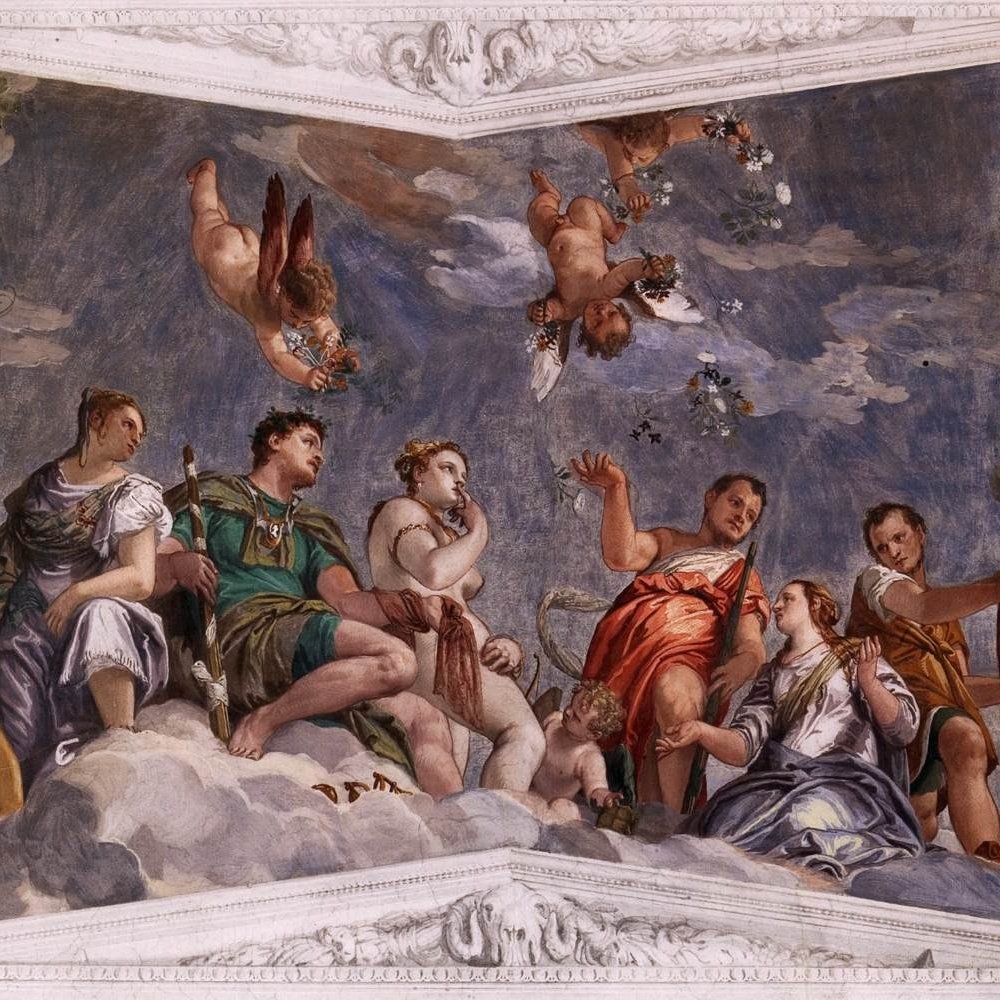} 
        \end{tabular}
        &
        \begin{tabular}[l]{@{}l@{}}
        \textit{\textbf{Stanza dell'Amore Coniugale}} \\ \\
            $\phi^*_{\cure}$ = 1 out of 5: The clothes of the two figures, the composition of the image, the \\ colors and the subject in general do not reflect the style of Italian art in the 16th Century.  \\
            \\
            $\phi^*_{\cure}$ = 4 out of 5: The image shows two lovers touching and almost kissing, while in \\ the real painting they aren’t.
        \end{tabular}
        \\

        \midrule

        \textbf{E12} & 
        \begin{tabular}[c]{@{}c@{}}
            \includegraphics[width=0.52\columnwidth]{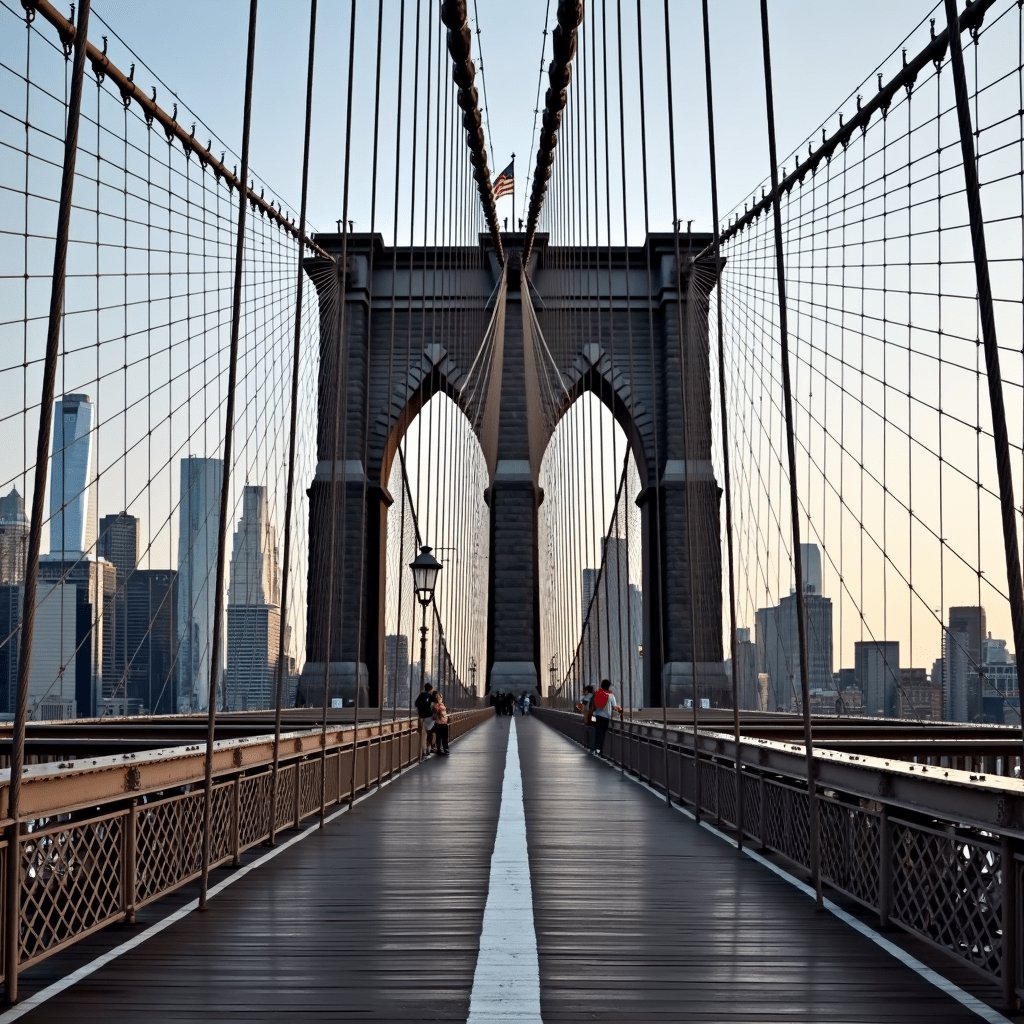}
        \end{tabular}
        &
        \begin{tabular}[c]{@{}c@{}}
            \includegraphics[width=0.52\columnwidth]{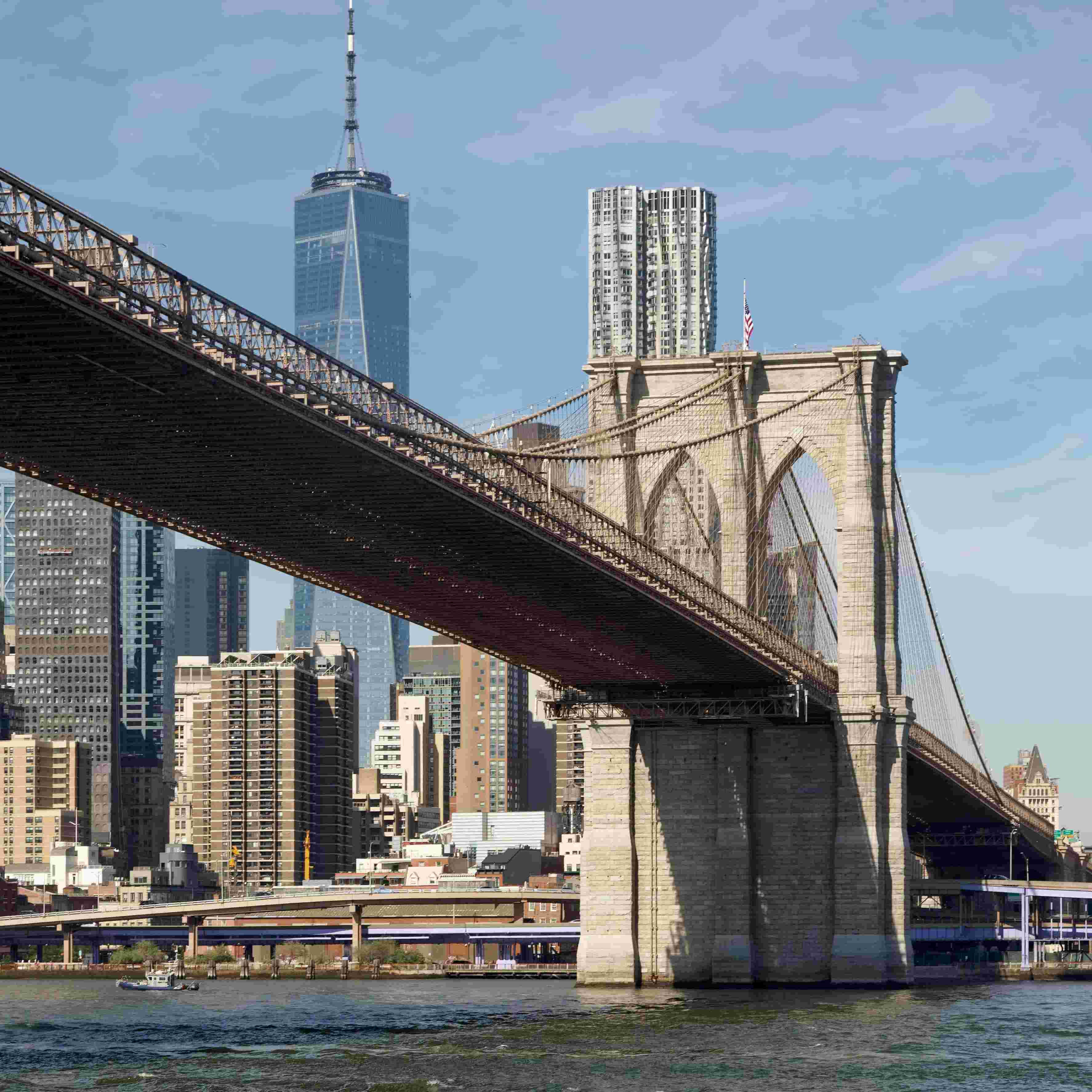}
        \end{tabular}
        &
        \begin{tabular}[l]{@{}l@{}}
        \textit{\textbf{Brooklyn Bridge}} \\ \\
            $\phi^*_{\cure}$ = 1 out of 5: The sky is far too clear for the NYC skyline (lacking smog haze),\\ the archways are too tall and narrow, and the meshing on the sides did not exist the last \\ time I was in the city \\
            \\
            $\phi^*_{\cure}$ = 4 out of 5: The floor part does not look authentic as well as the surrounding \\ city
        \end{tabular}
        \\
        \bottomrule
    \end{tabular}
\end{adjustbox}
\caption[]{Feedbacks of the workers in cases of high disagreement over perceptions of cultural representativeness $\phi^*_{\cure}$ of the T2I system for artifact $n$ (Part 2).}
\label{fig:us_disagree_combined_part2}
\end{figure*}

\clearpage
\subsection{Survey Respondent Statistics}

\begin{figure*}[t!]
    \centering
    \includegraphics[width=1\linewidth]{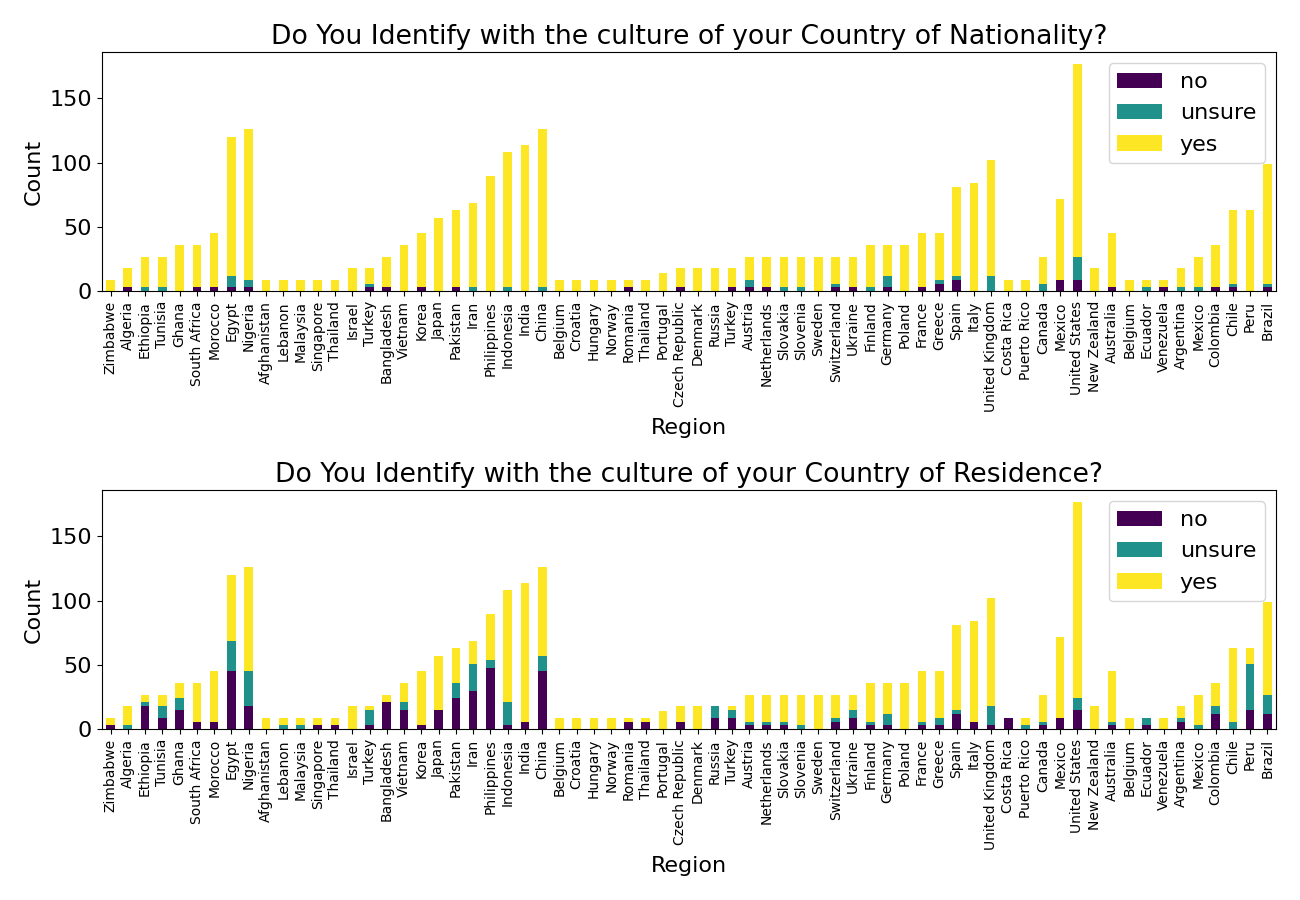}
    \caption{Responses to ``Do you identify with the culture of [country]?'' summed across all surveys belonging to each region $r$, grouped on the X-axis by Continent (L to R: Africa, Asia, Europe, North America, Oceania, South America) and sorted in ascending order of number of responses (count).}
    \label{fig:us_metadata}
\end{figure*}

\begin{figure*}[t]
    \centering
    \includegraphics[width=.8\linewidth]{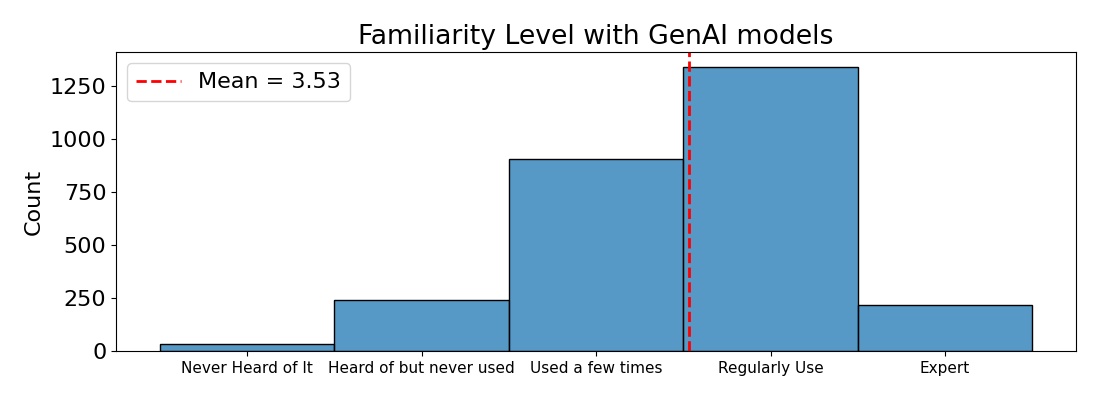}
    \caption{A histogram of worker familiarity with generative AI models like ChatGPT or DALL-E from 1 (``Never Heard of It'') to 5 (``Expert'') summed across all workers and surveys (3 T2I systems $\times$ 300 artifacts $\times$ 3 workers per artifact $=$ 2700 total).}
    \label{fig:us_metadata_fam}
\end{figure*}

Recall that we only hire workers to score surveys of artifact $n$ if their country of nationality matches the artifact's associated region $r$ (\cref{subsec:user_study}). We query workers for their perception of their own cultural identity to determine if they would be a reliable judge of culture-specific perceptual similarity. Users are asked if they identify with the culture of their country of nationality and residence, and given options ``Yes'', ``No'', or ``Unsure''. We show a bar plot of worker responses for each region $r$ in the \cure dataset (64 total) in \cref{fig:us_metadata}, grouped on the X-Axis by continent. We observe that respondents identify more with the culture of their country of nationality than the culture of their country of residence, especially in Africa, Asia, and South America, likely indicating that they are immigrants. We also ask respondents to rate their familiarity level with generative AI models (see UI in \cref{fig:us_demo}) on a scale of \textbf{1 - Never heard of it} to \textbf{5 - Expert}, which we visualize with a bar plot in \cref{fig:us_metadata_fam}. We observe that 92\% of Prolific workers have used generative AI tools at least a few times, and are thus moderately aware of what generated responses look like (familarity score  = $3.53 \pm 0.8$).

\subsection{Gold Scores in the Head and Long Tail.}
\label{app:gold_scores_head_tail}

We examine statistics for scores provided by survey respondents for the evaluation axes detailed in \cref{app:user_study}, \ie offensiveness ($\phi^*_{OFF}$), stereotypicalness ($\phi^*_{STR}$), cultural representativeness ($\phi^*_{\cure}$), perceptual similarity to ground truth ($\phi^*_{PS}$), and ground truth likelihood ($\phi^*_{GT}$). We would like to examine how human judgments of T2I system quality change in the head and long tail of its distribution. For this purpose, we use a proxy clustering of artifacts into two disjoint sets, \ie cultural artifacts from regions belonging to the Global South (GS), and those belonging to the Global North (GN). 

First, in \cref{tab:gold_score_gn_gs}, we look at gold scores for each T2I system (FLUX.1 [dev], Stable Diffusion 1.5, Stable Diffusion 3.5 Large) clustered into two disjoint sets or groups: . Across T2I systems, Offensiveness and Stereotypicalness are generally lower in the Global North than South, and this difference is most pronounced for the lowest quality\footnote{by Arena Bench ELO, see \cref{tab:benchmark_results}} system, Stable Diffusion 1.5 (0.3 - 0.4). As expected, T2I system offensiveness increases with decreasing T2I system quality. Interestingly, in the Global North, stereotypicalness increases with increasing T2I system quality, \textbf{indicating the in the case of underspecified prompts} ($I(n)$), \textbf{higher quality T2I systems may lean into more stereotypical representations of cultural artifacts}. Cultural representativeness ($\phi^*_\cure$) and perceptual similarity gold scores ($\phi^*_{GT}$ and $\phi^*_{PS}$) are consistently higher in the Global North across T2I systems, but the maximum increase over Global South is only 0.24 ($\phi^*_{GT}$ for SD 1.5). Overall for $\phi^*_\cure$, $\phi^*_{GT}$ and $\phi^*_{PS}$, SD 3.5 $>$ FLUX.1 [dev] $>$ SD 1.5, which slightly disagrees with Arena Bench ELO quality (\cref{tab:benchmark_results}).

Next, in \cref{tab:gold_score_continent_flux}, we take a deeper look at gold scores on a single T2I system, FLUX.1 [dev], with cultural artifacts clustered into continents. Asia shows the highest offensiveness and stereotypicalness, followed by Africa and South America, which agrees with prior work~\cite{jha2024visage, bianchi2023easily}. For cultural representativeness ($\phi^*_\cure$), North America is higher than all other continents (+ 0.4). For perceptual similarity gold scores ($\phi^*_{GT}$ and $\phi^*_{PS}$), North America is significantly higher than other continents (+0.5), with Oceania second best. Overall for $\phi^*_\cure$, $\phi^*_{GT}$ and $\phi^*_{PS}$, Africa, Asia, and Europe are nearly equivalent, while South America is the lowest.

\begin{table}[]
    \centering
    \caption{Mean and variance of Likert scores (gold scores) of images generated by FLUX.1 [dev] over cultural artifacts grouped by the artifact's region lying in the Global North (GN) or Global North (GS). We tabulate offensiveness ($\phi^*_{OFF}$) and stereotypicalness ($\phi^*_{STR}$), cultural representativeness ($\phi^*_{\cure}$), perceptual similarity to ground truth ($\phi^*_{PS}$), and ground truth likelihood ($\phi^*_{GT}$). The highest entry in each column is bolded. Arrows indicate whether a lower ($\downarrow$) or higher ($\uparrow$) gold score is better.}
    \begin{adjustbox}{max width=\columnwidth}
    \begin{tabular}{c|c|ccccc@{}}
    \toprule
    \textbf{T2I System} & \textbf{Region} & \textbf{$\phi^*_{OFF}\downarrow$} & \textbf{$\phi^*_{STR}\downarrow$} & \textbf{$\phi^*_{\cure}\uparrow$} & \textbf{$\phi^*_{GT}\uparrow$} & \textbf{$\phi^*_{PS}\uparrow$} \\ 
    \midrule
    \multirow{2}{*}{\begin{tabular}[c]{@{}c@{}}FLUX.1 \\ \textnormal{[dev]} \end{tabular}} & GN & \textbf{1.29} $\pm$ \textbf{0.54} & 1.86 $\pm$ 1.29 & 2.94 $\pm$ 1.41 & 2.38 $\pm$ 1.34 & 2.19 $\pm$ 1.11 \\
     & GS & 1.35 $\pm$ 0.68 & 1.82 $\pm$ 1.34 & 2.73 $\pm$ 1.43 & 2.16 $\pm$ 1.31 & 2.13 $\pm$ 1.16 \\ 
    \midrule
    \multirow{2}{*}{\begin{tabular}[c]{@{}c@{}}SD 3.5 \\ Large \end{tabular}} & GN & \textbf{1.29} $\pm$ \textbf{0.75} & 1.75 $\pm$ 1.05 & \textbf{3.02} $\pm$ \textbf{1.44} & \textbf{2.67} $\pm$ \textbf{1.40} & \textbf{2.46} $\pm$ \textbf{1.20} \\
    & GS & 1.36 $\pm$ 0.78 & 1.91 $\pm$ 1.25 & 2.96 $\pm$ 1.44 & 2.44 $\pm$ 1.39 & 2.35 $\pm$ 1.25 \\ 
    \midrule
    \multirow{2}{*}{SD 1.5} & GN & 1.35 $\pm$ 0.83 & \textbf{1.65} $\pm$ \textbf{1.06} & 2.81 $\pm$ 1.40 & 2.32 $\pm$ 1.34 & 2.18 $\pm$ 1.19 \\
    & GS & 1.61 $\pm$ 1.06 & 2.02 $\pm$ 1.25 & 2.67 $\pm$ 1.42 & 2.08 $\pm$ 1.25 & 2.04 $\pm$ 1.13 \\ 
    \bottomrule
    \end{tabular}
    \end{adjustbox}
    \label{tab:gold_score_gn_gs}
\end{table}

\begin{table}[]
    \centering
    \caption{Mean and variance of Likert scores (gold scores) of images generated by FLUX.1 [dev] over cultural artifacts grouped by continent. We tabulate offensiveness ($\phi^*_{OFF}$) and stereotypicalness ($\phi^*_{STR}$), cultural representativeness ($\phi^*_{\cure}$), perceptual similarity to ground truth ($\phi^*_{PS}$), and ground truth likelihood ($\phi^*_{GT}$). The highest entry in each column is bolded. Arrows indicate whether a lower ($\downarrow$) or higher ($\uparrow$) gold score is better.}
    \begin{adjustbox}{max width=\columnwidth}
    \begin{tabular}{@{}r|ccccc@{}}
    \toprule
    \textbf{Cont} & \textbf{$\phi^*_{OFF}\downarrow$} & \textbf{$\phi^*_{STR}\downarrow$} & \textbf{$\phi^*_{\cure}\uparrow$} & \textbf{$\phi^*_{GT}\uparrow$} & \textbf{$\phi^*_{PS}\uparrow$} \\ \midrule
    \textbf{Africa} & 1.37 $\pm$ 0.83 & 1.90 $\pm$ 1.27 & 2.83 $\pm$ 1.49 & 2.27 $\pm$ 1.34 & 2.25 $\pm$ 1.22 \\
    \textbf{Asia} & 1.55 $\pm$ 1.00 & 2.04 $\pm$ 1.21 & 2.84 $\pm$ 1.42 & 2.25 $\pm$ 1.32 & 2.22 $\pm$ 1.17 \\
    \textbf{Europe} & 1.29 $\pm$ 0.73 & 1.74 $\pm$ 1.07 & 2.77 $\pm$ 1.42 & 2.26 $\pm$ 1.29 & 2.12 $\pm$ 1.15 \\
    \textbf{NA} & 1.29 $\pm$ 0.76 & 1.79 $\pm$ 1.18 & \textbf{3.29 $\pm$ 1.40} & \textbf{2.97 $\pm$ 1.44} & \textbf{2.61 $\pm$ 1.23} \\
    \textbf{Oceania} & \textbf{1.24 $\pm$ 0.73} & \textbf{1.52 $\pm$ 1.01} & 2.84 $\pm$ 1.42 & 2.44 $\pm$ 1.43 & 2.10 $\pm$ 1.07 \\
    \textbf{SA} & 1.34 $\pm$ 0.84 & 1.67 $\pm$ 1.10 & 2.62 $\pm$ 1.35 & 2.11 $\pm$ 1.29 & 2.07 $\pm$ 1.15 \\ 
    \bottomrule
    \end{tabular}
    \end{adjustbox}
    \label{tab:gold_score_continent_flux}
\end{table}

\clearpage

\section{Concept Frequency Estimation}
\label{app:concept_frequency}

\begin{figure*}[!t]
    \centering
    \begin{subfigure}[b]{0.3\textwidth}
        \includegraphics[width=\textwidth]{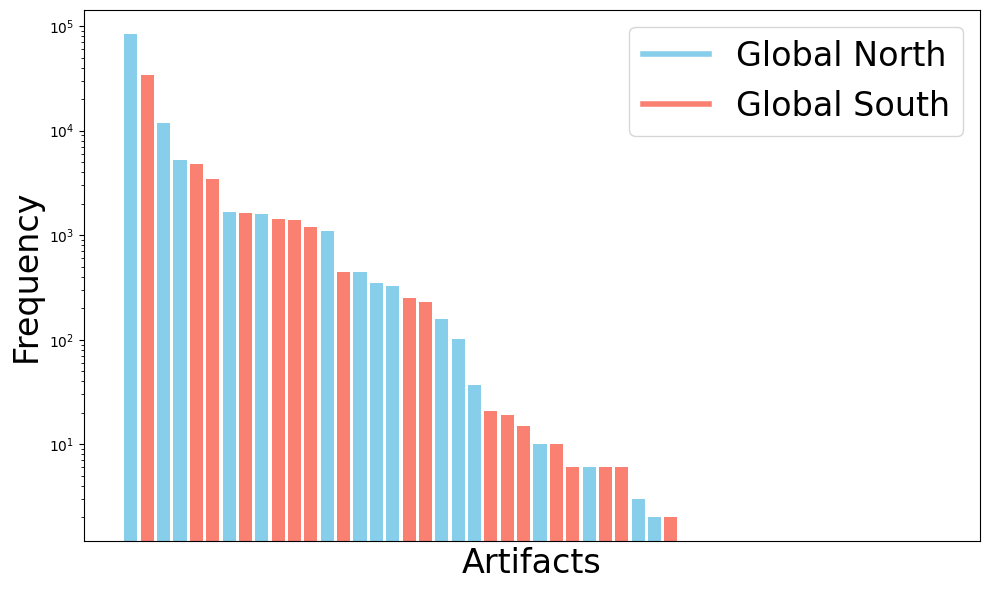}
        \caption{Art}
    \end{subfigure}
    \hfill
    \begin{subfigure}[b]{0.3\textwidth}
        \includegraphics[width=\textwidth]{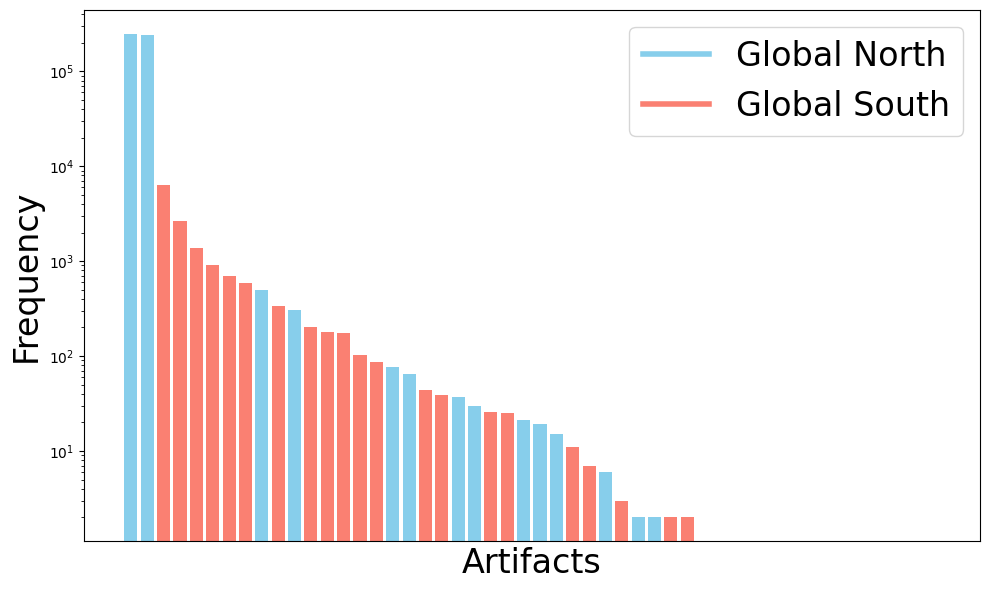}
        \caption{Architecture}
    \end{subfigure}
    \hfill
    \begin{subfigure}[b]{0.3\textwidth}
        \includegraphics[width=\textwidth]{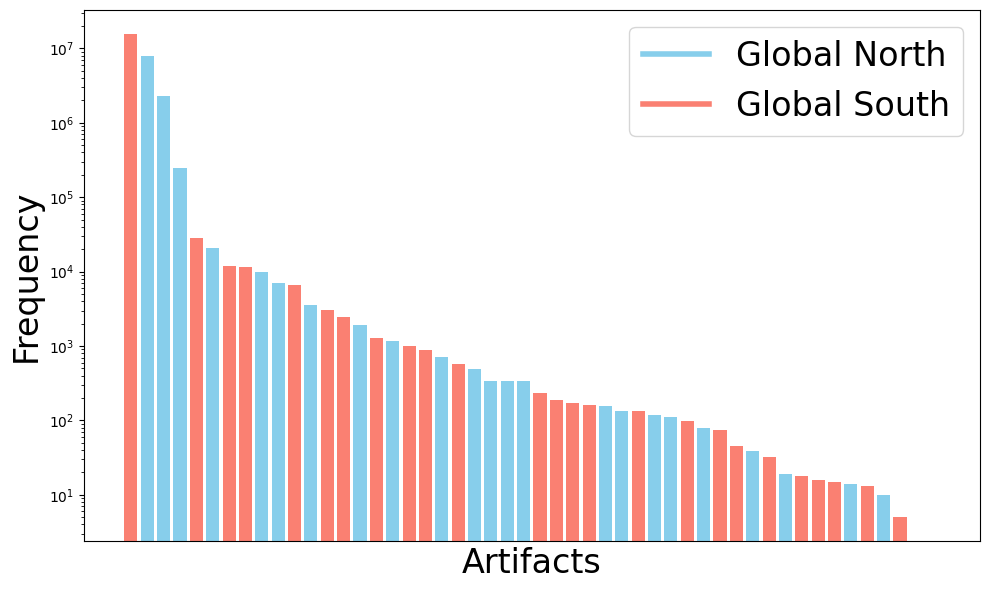}
        \caption{Celebrations}
    \end{subfigure}
    
    \vspace{1ex}
    
    \begin{subfigure}[b]{0.3\textwidth}
        \includegraphics[width=\textwidth]{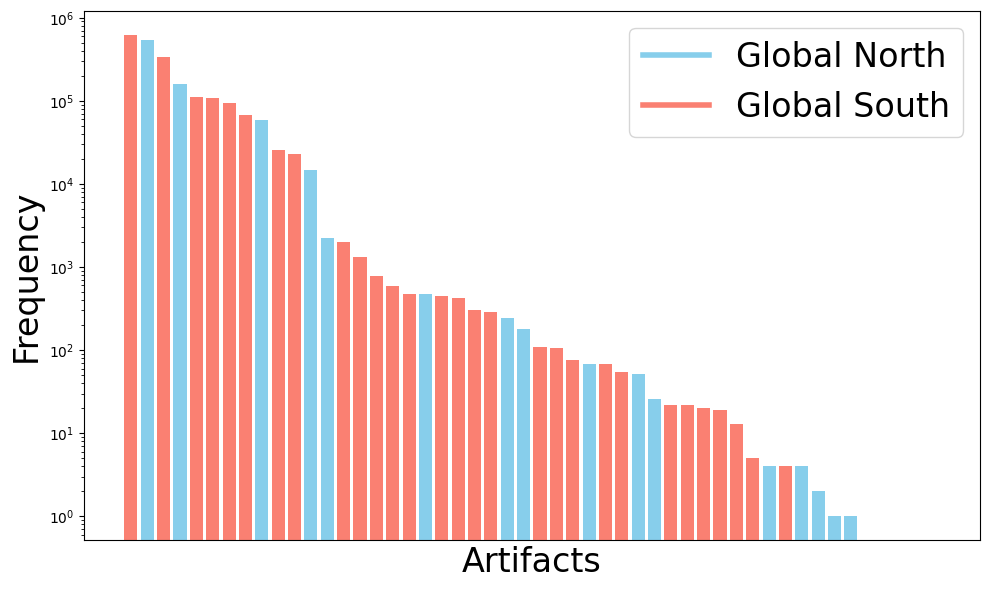}
        \caption{Fashion}
    \end{subfigure}
    \hfill
    \begin{subfigure}[b]{0.3\textwidth}
        \includegraphics[width=\textwidth]{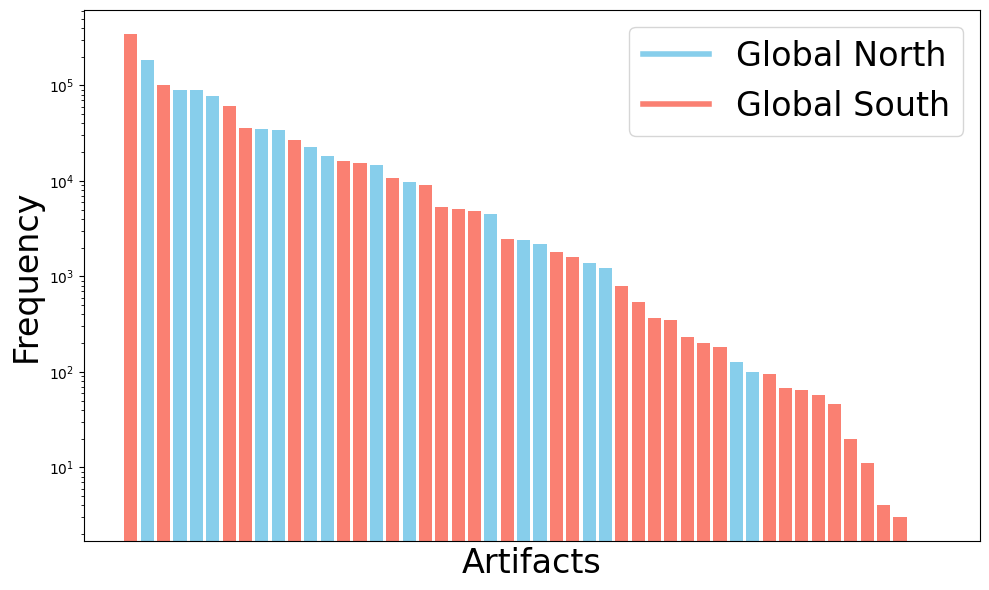}
        \caption{Food}
    \end{subfigure}
    \hfill
    \begin{subfigure}[b]{0.3\textwidth}
        \includegraphics[width=\textwidth]{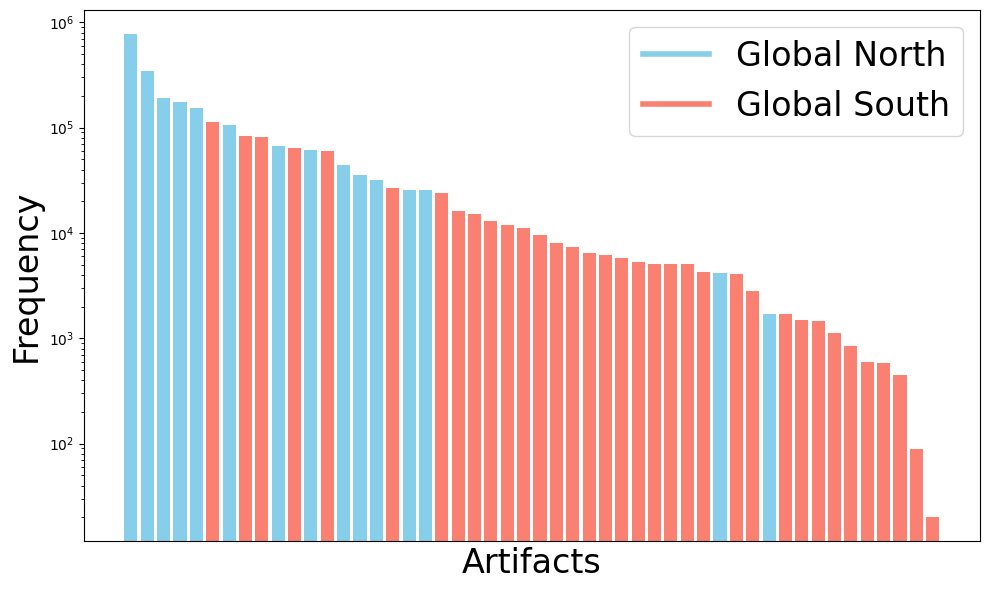}
        \caption{People}
    \end{subfigure}
    \caption{Frequency estimation ($\log$ scale) for 50 cultural artifacts across each of the six supercategories (cultural axes) in the \cure dataset on LAION-SD15, the fully open pretraining dataset of Stable Diffusion 1.5~\cite{rombach2022high}.}
    \label{fig:relaion_freq}
\end{figure*}

In recent times, state-of-the-art T2I systems do not have fully open pretraining data, \ie it is unknown whether any particular image lies in the learned distribution of a modern T2I system. However, for the older Stable Diffusion 1.5, we have full access to its entire pretraining data, \ie a mixture of LAION-2B-en\footnote{\url{https://huggingface.co/datasets/laion/relaion2B-en-research-safe}}~\cite{schuhmann2022laion} and LAION-Aesthetics V2 5+\footnote{\url{https://laion.ai/blog/laion-aesthetics/}}, both of which are publicly and openly accessible. We can thus know with certainty whether a given image belongs to the training data of SD 1.5. We henceforth refer to this dataset mixture as \textbf{LAION-SD15}. As in \citet{parashar2024neglected}, we explictly compute a concept frequency for every cultural artifact in the \cure dataset on LAION-SD15 with string matching. We note that the original datasets on which SD 1.5 was trained were taken down and re-released due to problematic CSAM content~\cite{thiel2023identifying, birhane2024into}; we use the re-released variants for our frequency computation. As seen in \cref{fig:relaion_freq}, all supercategories in our \cure dataset show long tail behavior in LAION-SD15 across 50 cultural artifacts. We observe a very low occurence of many artifacts in the art and architecture supercategories, as these are often specific and unique named entities which may occur rarely in the dataset when compared to more common items with high intra-class variance such as food, celebrations or people.

We also compute a Spearman rank correlation of the artifact occurence frequency in LAION-SD15 with the ground truth likelihood Likert scores (see \cref{app:user_study} for more details) provided by survey respondents for FLUX.1 [dev], SD 3.5 Large, and SD 1.5 in \cref{tab:freq_isgt_corr}. We expect a higher occurence frequency to correspond to better user judgments, as the T2I system should be able to accurately generate artifacts it has seen many times (the ``head'' of the T2I system distribution). We observe that for Celebrations, Food, Fashion, and People, there is a large positive correlation with user perceptions of the likelihood of the T2I system output $I(n)$ belonging to the class of artifacts $n$ (\eg assigning high score to the likelihood of images of spaghetti actually representing ``spaghetti'' as an artifact). Correlations are low for art and architecture, which we suspect is due to the large portion of their tail having very small frequency counts (\cref{fig:relaion_freq}). While the datasets used to train SD 3.5 Large and FLUX.1[dev] are not public and we cannot explicitly compute concept frequencies, since they show similar rank correlations to user judgments as SD 1.5, we predict that the \cure dataset shows a similar long tail behavior across cultural artifacts for SD 3.5 and FLUX.1[dev] as well.

\begin{table}[!ht]
\centering
\caption{Spearman rank correlation between occurrence frequency in the LAION-SD15 dataset of each cultural artifact in the \cure dataset with human perceptions of ground-truth likelihood from the user study, $\phi^*_{GT}$ (\cref{subsec:user_study} for details). Results are tabulated for each supercategory and for SD 1.5, SD 3.5 Large, and FLUX.1 [dev].}
\label{tab:freq_isgt_corr}
\resizebox{\columnwidth}{!}{%
\begin{tabular}{@{}c|cccccc@{}}
\toprule
\multirow{2}{*}{T2I} & \multicolumn{6}{c}{Supercategory}                                                                                                                                                   \\ \cmidrule(l){2-7} 
                     & \multicolumn{1}{l}{Art} & \multicolumn{1}{l}{Architecture} & \multicolumn{1}{l}{Celebrations} & \multicolumn{1}{l}{Food} & \multicolumn{1}{l}{Fashion} & \multicolumn{1}{l}{People} \\ \midrule
SD 1.5               & 0.04                    & 0.10                             & 0.40                            & 0.59                     & 0.34                        & 0.22                       \\
SD 3.5 Large         & 0.05                    & 0.13                             & 0.48                             & 0.55                     & 0.27                        & 0.23                       \\
FLUX.1 {[}dev{]}     & 0.01                    & 0.06                             & 0.30                             & 0.38                     & 0.44                        & 0.34                       \\ \bottomrule
\end{tabular}%
}
\end{table}

\clearpage

\label{app:percepsim}
\section{Perceptual Similarity}
 
As detailed in \cref{app:img_gen_seeding}, we generate images with increasing attribute specification for each artifact with three T2I systems, Stable Diffusion 1.5, Stable Diffusion 3.5 Large, and FLUX.1 [dev]. The prompts used are $P(n)$, $P(\{n,c\})$, $P(\{n,r\})$, and $P(\{n,c,r\})$ from \cref{tab:prompt_templates}. 

We discuss in ablation to visually examine how our PS scorers capture T2I system performance in the cultural head and long tail in \cref{app:ps_head_vs_tail}. We also discuss qualitative examples from the \cure dataset and how our PS scorers rate artifacts when compared to user judgments in \cref{app:ps_scorer_qual_analysis}.

\begin{table*}[]
    \caption{Perceptual similarity \cure scores of all regions in the \cure dataset, sorted descending by user study scores $\phi^*_{\cure}$. The scores are averaged across Flux.1 [dev], SD 3.5 Large, SD 1.5, and SDXL T2I systems and SigLIP2, AIMV2, and DINOv2 encoders.}
    \footnotesize
    \centering
    \resizebox{\textwidth}{!}{
    \begin{tabular}{@{}r|r|c|cccc@{}}
    \toprule
    \multirow{2}{*}{\makebox[2cm][c]{\textbf{Continent}}} 
    & \multirow{2}{*}{\makebox[2.5cm][c]{\textbf{Region}}} & & \multicolumn{4}{c}{\textbf{Perceptual Similarity Scorer}} \\ 
    \cline{4-7}
     & & $\phi^*_{\cure}\uparrow$ &$\phi_{GT}(n)\uparrow$ & $\phi_{PS}(n)\uparrow$ & $\Delta \phi_{PS}(\{n,c\})\downarrow$ & $\Delta \phi_{PS}(\{n,c,r\})\downarrow$  \\
    \midrule
    \multirow{9}{*}{Africa} & Egypt & 3.24 $\pm$ 0.88 & 0.72 $\pm$ 0.54 & 0.62 $\pm$ 0.44 & 0.52 $\pm$ 0.35 & 0.51 $\pm$ 0.33 \\
     & Nigeria & 2.96 $\pm$ 0.91 & 0.72 $\pm$ 0.52 & 0.64 $\pm$ 0.49 & 0.51 $\pm$ 0.44 & 0.51 $\pm$ 0.44 \\
     & Morocco & 2.80 $\pm$ 1.21 & 0.64 $\pm$ 0.16 & 0.62 $\pm$ 0.19 & 0.55 $\pm$ 0.22 & 0.55 $\pm$ 0.25 \\
     & Algeria & 2.72 $\pm$ 0.87 & 0.56 $\pm$ 0.04 & 0.57 $\pm$ 0.07 & 0.56 $\pm$ 0.11 & 0.57 $\pm$ 0.11 \\
     & Ghana & 2.69 $\pm$ 0.90 & 0.59 $\pm$ 0.16 & 0.59 $\pm$ 0.17 & 0.55 $\pm$ 0.25 & 0.53 $\pm$ 0.22 \\
     & South Africa & 2.53 $\pm$ 1.10 & 0.64 $\pm$ 0.13 & 0.63 $\pm$ 0.14 & 0.55 $\pm$ 0.14 & 0.54 $\pm$ 0.12 \\
     & Ethiopia & 2.28 $\pm$ 0.30 & 0.61 $\pm$ 0.11 & 0.62 $\pm$ 0.11 & 0.54 $\pm$ 0.20 & 0.54 $\pm$ 0.15 \\
     & Zimbabwe & 2.22 $\pm$ 0.03 & 0.77 $\pm$ 0.05 & 0.69 $\pm$ 0.07 & 0.49 $\pm$ 0.02 & 0.50 $\pm$ 0.02 \\
     & Tunisia & 1.74 $\pm$ 0.29 & 0.50 $\pm$ 0.05 & 0.53 $\pm$ 0.06 & 0.63 $\pm$ 0.14 & 0.63 $\pm$ 0.15 \\
    \midrule
    \multirow{15}{*}{Asia} & India & 3.91 $\pm$ 0.65 & 0.69 $\pm$ 0.72 & 0.65 $\pm$ 0.55 & 0.52 $\pm$ 0.46 & 0.51 $\pm$ 0.45 \\
     & Malaysia & 3.33 $\pm$ 0.22 & 0.74 $\pm$ 0.07 & 0.78 $\pm$ 0.04 & 0.51 $\pm$ 0.02 & 0.51 $\pm$ 0.02 \\
     & Pakistan & 3.02 $\pm$ 0.85 & 0.76 $\pm$ 0.31 & 0.61 $\pm$ 0.20 & 0.50 $\pm$ 0.11 & 0.50 $\pm$ 0.14 \\
     & Indonesia & 2.88 $\pm$ 0.83 & 0.66 $\pm$ 0.53 & 0.63 $\pm$ 0.46 & 0.53 $\pm$ 0.49 & 0.53 $\pm$ 0.54 \\
     & Iran & 2.86 $\pm$ 1.00 & 0.61 $\pm$ 0.24 & 0.56 $\pm$ 0.17 & 0.55 $\pm$ 0.26 & 0.55 $\pm$ 0.23 \\
     & Philippines & 2.77 $\pm$ 0.69 & 0.64 $\pm$ 0.37 & 0.60 $\pm$ 0.26 & 0.54 $\pm$ 0.34 & 0.54 $\pm$ 0.36 \\
     & Japan & 2.72 $\pm$ 1.45 & 0.65 $\pm$ 0.21 & 0.61 $\pm$ 0.23 & 0.55 $\pm$ 0.23 & 0.54 $\pm$ 0.23 \\
     & China & 2.67 $\pm$ 0.76 & 0.65 $\pm$ 0.56 & 0.62 $\pm$ 0.44 & 0.53 $\pm$ 0.48 & 0.53 $\pm$ 0.47 \\
     & Lebanon & 2.67 $\pm$ 0.52 & 0.72 $\pm$ 0.04 & 0.76 $\pm$ 0.06 & 0.49 $\pm$ 0.02 & 0.51 $\pm$ 0.05 \\
     & Thailand & 2.25 $\pm$ 0.83 & 0.67 $\pm$ 0.04 & 0.66 $\pm$ 0.07 & 0.55 $\pm$ 0.06 & 0.54 $\pm$ 0.06 \\
     & Israel & 2.44 $\pm$ 0.25 & 0.70 $\pm$ 0.06 & 0.71 $\pm$ 0.13 & 0.48 $\pm$ 0.12 & 0.49 $\pm$ 0.10 \\
     & Korea & 2.42 $\pm$ 0.81 & 0.66 $\pm$ 0.20 & 0.63 $\pm$ 0.14 & 0.53 $\pm$ 0.17 & 0.53 $\pm$ 0.12 \\
     & Vietnam & 2.14 $\pm$ 1.23 & 0.59 $\pm$ 0.13 & 0.59 $\pm$ 0.13 & 0.54 $\pm$ 0.14 & 0.54 $\pm$ 0.16 \\
     & Bangladesh & 2.11 $\pm$ 1.11 & 0.53 $\pm$ 0.08 & 0.55 $\pm$ 0.09 & 0.59 $\pm$ 0.13 & 0.58 $\pm$ 0.16 \\
     & Afghanistan & 1.89 $\pm$ 0.32 & 0.80 $\pm$ 0.02 & 0.59 $\pm$ 0.03 & 0.50 $\pm$ 0.01 & 0.50 $\pm$ 0.01 \\
     & Singapore & 1.89 $\pm$ 0.17 & 0.79 $\pm$ 0.02 & 0.70 $\pm$ 0.04 & 0.53 $\pm$ 0.04 & 0.52 $\pm$ 0.04 \\
     \midrule
    \multirow{26}{*}{Europe} & Slovenia & 3.41 $\pm$ 0.19 & 0.63 $\pm$ 0.08 & 0.63 $\pm$ 0.09 & 0.57 $\pm$ 0.06 & 0.56 $\pm$ 0.07 \\
     & Czech Republic & 3.28 $\pm$ 0.94 & 0.73 $\pm$ 0.05 & 0.69 $\pm$ 0.06 & 0.51 $\pm$ 0.05 & 0.52 $\pm$ 0.04 \\
     & Switzerland & 3.19 $\pm$ 0.92 & 0.61 $\pm$ 0.08 & 0.62 $\pm$ 0.11 & 0.51 $\pm$ 0.12 & 0.50 $\pm$ 0.09 \\
     & United Kingdom & 3.14 $\pm$ 0.70 & 0.74 $\pm$ 0.52 & 0.62 $\pm$ 0.41 & 0.51 $\pm$ 0.28 & 0.51 $\pm$ 0.30 \\
     & Denmark & 3.11 $\pm$ 1.28 & 0.59 $\pm$ 0.05 & 0.58 $\pm$ 0.04 & 0.58 $\pm$ 0.08 & 0.57 $\pm$ 0.07 \\
     & Germany & 3.11 $\pm$ 1.14 & 0.62 $\pm$ 0.24 & 0.58 $\pm$ 0.14 & 0.56 $\pm$ 0.20 & 0.56 $\pm$ 0.18 \\
     & Norway & 3.11 $\pm$ 0.10 & 0.57 $\pm$ 0.05 & 0.59 $\pm$ 0.03 & 0.55 $\pm$ 0.06 & 0.55 $\pm$ 0.05 \\
     & Italy & 3.01 $\pm$ 1.46 & 0.70 $\pm$ 0.35 & 0.66 $\pm$ 0.39 & 0.53 $\pm$ 0.36 & 0.53 $\pm$ 0.28 \\
     & Netherlands & 3.00 $\pm$ 1.58 & 0.64 $\pm$ 0.12 & 0.58 $\pm$ 0.10 & 0.55 $\pm$ 0.14 & 0.55 $\pm$ 0.26 \\
     & Poland & 3.00 $\pm$ 0.93 & 0.63 $\pm$ 0.12 & 0.61 $\pm$ 0.18 & 0.53 $\pm$ 0.22 & 0.54 $\pm$ 0.21 \\
     & France & 2.98 $\pm$ 1.28 & 0.69 $\pm$ 0.23 & 0.66 $\pm$ 0.24 & 0.52 $\pm$ 0.24 & 0.52 $\pm$ 0.25 \\
     & Russia & 2.83 $\pm$ 0.58 & 0.68 $\pm$ 0.22 & 0.65 $\pm$ 0.07 & 0.54 $\pm$ 0.05 & 0.54 $\pm$ 0.05 \\
     & Greece & 2.73 $\pm$ 1.51 & 0.69 $\pm$ 0.17 & 0.64 $\pm$ 0.30 & 0.53 $\pm$ 0.28 & 0.53 $\pm$ 0.20 \\
     & Belgium & 2.72 $\pm$ 0.53 & 0.71 $\pm$ 0.18 & 0.64 $\pm$ 0.17 & 0.50 $\pm$ 0.12 & 0.50 $\pm$ 0.28 \\
     & Spain & 2.62 $\pm$ 0.76 & 0.68 $\pm$ 0.44 & 0.63 $\pm$ 0.31 & 0.52 $\pm$ 0.29 & 0.52 $\pm$ 0.28 \\
     & Portugal & 2.60 $\pm$ 1.74 & 0.61 $\pm$ 0.03 & 0.59 $\pm$ 0.06 & 0.56 $\pm$ 0.05 & 0.57 $\pm$ 0.07 \\
     & Austria & 2.52 $\pm$ 1.24 & 0.62 $\pm$ 0.09 & 0.62 $\pm$ 0.11 & 0.53 $\pm$ 0.20 & 0.53 $\pm$ 0.15 \\
     & Slovakia & 2.41 $\pm$ 0.76 & 0.66 $\pm$ 0.15 & 0.64 $\pm$ 0.17 & 0.54 $\pm$ 0.13 & 0.53 $\pm$ 0.17 \\
     & Hungary & 2.33 $\pm$ 0.52 & 0.68 $\pm$ 0.03 & 0.71 $\pm$ 0.04 & 0.51 $\pm$ 0.02 & 0.51 $\pm$ 0.02 \\
     & Romania & 2.33 $\pm$ 0.22 & 0.61 $\pm$ 0.03 & 0.60 $\pm$ 0.02 & 0.58 $\pm$ 0.05 & 0.59 $\pm$ 0.04 \\
     & Croatia & 2.22 $\pm$ 0.47 & 0.53 $\pm$ 0.01 & 0.56 $\pm$ 0.02 & 0.60 $\pm$ 0.05 & 0.58 $\pm$ 0.03 \\
     & Finland & 2.20 $\pm$ 0.63 & 0.65 $\pm$ 0.26 & 0.64 $\pm$ 0.24 & 0.51 $\pm$ 0.18 & 0.52 $\pm$ 0.34 \\
     & Turkey & 2.19 $\pm$ 0.40 & 0.66 $\pm$ 0.10 & 0.67 $\pm$ 0.24 & 0.53 $\pm$ 0.21 & 0.52 $\pm$ 0.37 \\
     & Sweden & 2.09 $\pm$ 0.82 & 0.67 $\pm$ 0.11 & 0.63 $\pm$ 0.11 & 0.53 $\pm$ 0.14 & 0.53 $\pm$ 0.14 \\
     & Ukraine & 1.74 $\pm$ 0.61 & 0.52 $\pm$ 0.07 & 0.56 $\pm$ 0.09 & 0.60 $\pm$ 0.18 & 0.58 $\pm$ 0.15 \\
     \bottomrule
    \end{tabular}
    }
\end{table*}

\begin{table*}[t]
    \footnotesize
    \centering
    \resizebox{\textwidth}{!}{
    \begin{tabular}{@{}r|r|c|cccc@{}}
    \toprule
    \multirow{2}{*}{\makebox[2cm][c]{\textbf{Continent}}} 
    & \multirow{2}{*}{\makebox[2.5cm][c]{\textbf{Region}}} & \multicolumn{5}{c}{\textbf{Perceptual Similarity Scorer}} \\ 
    \cline{3-7}
     & & $\phi^*_{\cure}\uparrow$ &$\phi_{GT}(n)\uparrow$ & $\phi_{PS}(n)\uparrow$ & $\Delta \phi_{PS}(\{n,c\})\downarrow$ & $\Delta \phi_{PS}(\{n,c,r\})\downarrow$  \\
    \midrule
    \multirow{5}{*}{North\_America} & United States & 3.46 $\pm$ 0.99 & 0.73 $\pm$ 0.87 & 0.66 $\pm$ 0.84 & 0.51 $\pm$ 0.75 & 0.51 $\pm$ 0.71 \\
     & Canada & 3.30 $\pm$ 0.43 & 0.66 $\pm$ 0.07 & 0.73 $\pm$ 0.16 & 0.50 $\pm$ 0.12 & 0.50 $\pm$ 0.09 \\
     & Mexico & 2.89 $\pm$ 1.30 & 0.66 $\pm$ 0.45 & 0.68 $\pm$ 0.61 & 0.52 $\pm$ 0.55 & 0.52 $\pm$ 0.53 \\
     & Puerto Rico & 2.56 $\pm$ 0.17 & 0.50 $\pm$ 0.01 & 0.51 $\pm$ 0.01 & 0.67 $\pm$ 0.03 & 0.65 $\pm$ 0.05 \\
     & Costa Rica & 1.44 $\pm$ 0.40 & 0.60 $\pm$ 0.02 & 0.67 $\pm$ 0.06 & 0.54 $\pm$ 0.07 & 0.55 $\pm$ 0.04 \\
     \midrule
    \multirow{2}{*}{Oceania} & Australia & 2.87 $\pm$ 1.18 & 0.66 $\pm$ 0.16 & 0.69 $\pm$ 0.22 & 0.52 $\pm$ 0.17 & 0.52 $\pm$ 0.18 \\
     & New Zealand & 2.78 $\pm$ 0.88 & 0.62 $\pm$ 0.12 & 0.62 $\pm$ 0.12 & 0.54 $\pm$ 0.10 & 0.53 $\pm$ 0.10 \\
     \midrule
    \multirow{7}{*}{South\_America} & Brazil & 3.11 $\pm$ 1.15 & 0.65 $\pm$ 0.51 & 0.66 $\pm$ 0.54 & 0.51 $\pm$ 0.55 & 0.51 $\pm$ 0.56 \\
     & Venezuela & 2.89 $\pm$ 0.32 & 0.48 $\pm$ 0.02 & 0.55 $\pm$ 0.03 & 0.60 $\pm$ 0.05 & 0.61 $\pm$ 0.05 \\
     & Chile & 2.85 $\pm$ 0.49 & 0.74 $\pm$ 0.43 & 0.63 $\pm$ 0.23 & 0.51 $\pm$ 0.16 & 0.51 $\pm$ 0.20 \\
     & Argentina & 2.40 $\pm$ 1.40 & 0.72 $\pm$ 0.08 & 0.68 $\pm$ 0.09 & 0.52 $\pm$ 0.09 & 0.51 $\pm$ 0.08 \\
     & Peru & 2.29 $\pm$ 1.02 & 0.58 $\pm$ 0.21 & 0.62 $\pm$ 0.30 & 0.54 $\pm$ 0.31 & 0.53 $\pm$ 0.37 \\
     & Colombia & 2.11 $\pm$ 0.56 & 0.64 $\pm$ 0.17 & 0.62 $\pm$ 0.18 & 0.54 $\pm$ 0.18 & 0.53 $\pm$ 0.14 \\
     & Ecuador & 1.67 $\pm$ 0.00 & 0.54 $\pm$ 0.02 & 0.53 $\pm$ 0.02 & 0.62 $\pm$ 0.09 & 0.60 $\pm$ 0.08 \\
     \bottomrule
    \end{tabular}
    }
    \label{tab:cure_ps_scores}
\end{table*}

\subsection{Perceptual Similarity as a Long Tail Predictor}
\label{app:ps_head_vs_tail}

\begin{figure*}[hbt!]
        \hfill
        \vspace{0.3cm}
        \subfloat[FLUX.1 dev]{\label{fig:flux_ps_head_tail}
        \centering
        \includegraphics[width=.95\linewidth]{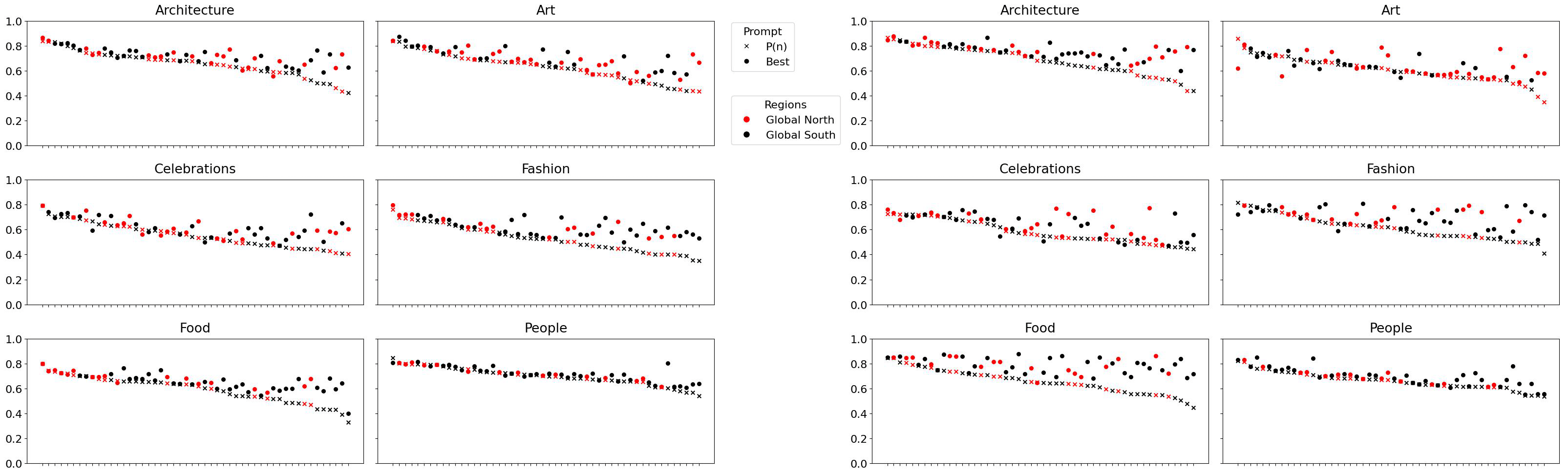}
        }
    
        \hfill
        \vspace{0.3cm}
        \subfloat[Stable Diffusion 3.5 Large]{\label{fig:sd35_ps_head_tail}
        \centering
        \includegraphics[width=.95\linewidth]{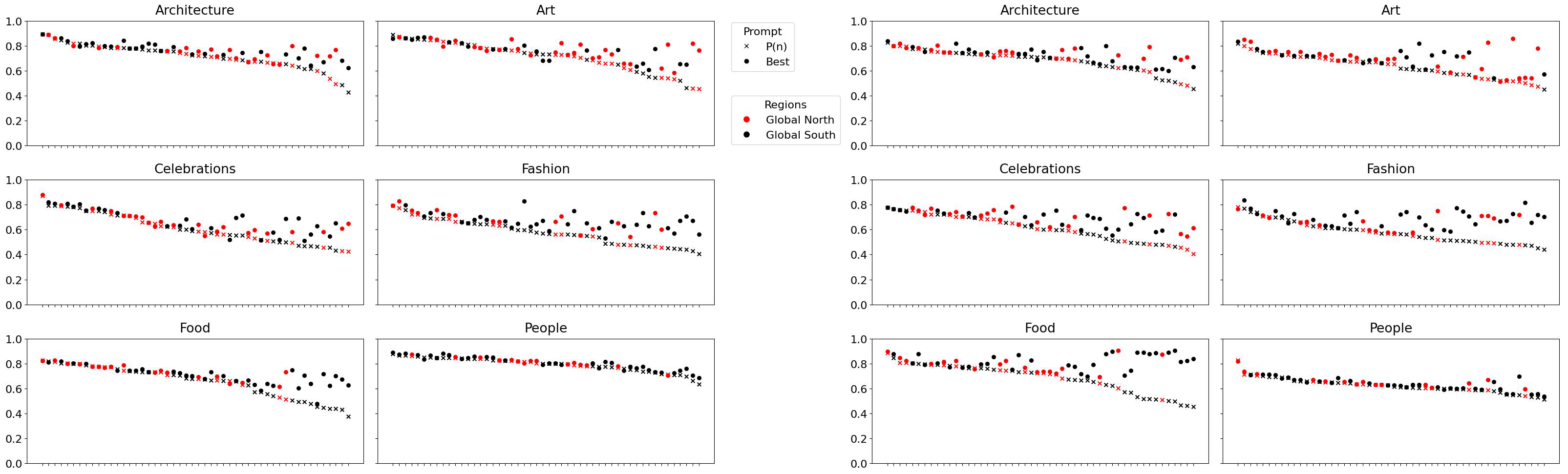}
        }
    
        \hfill
        \vspace{0.3cm}
        \subfloat[Stable Diffusion 1.5]{\label{fig:sd15_ps_head_tail}
        \centering
        \includegraphics[width=.95\linewidth]{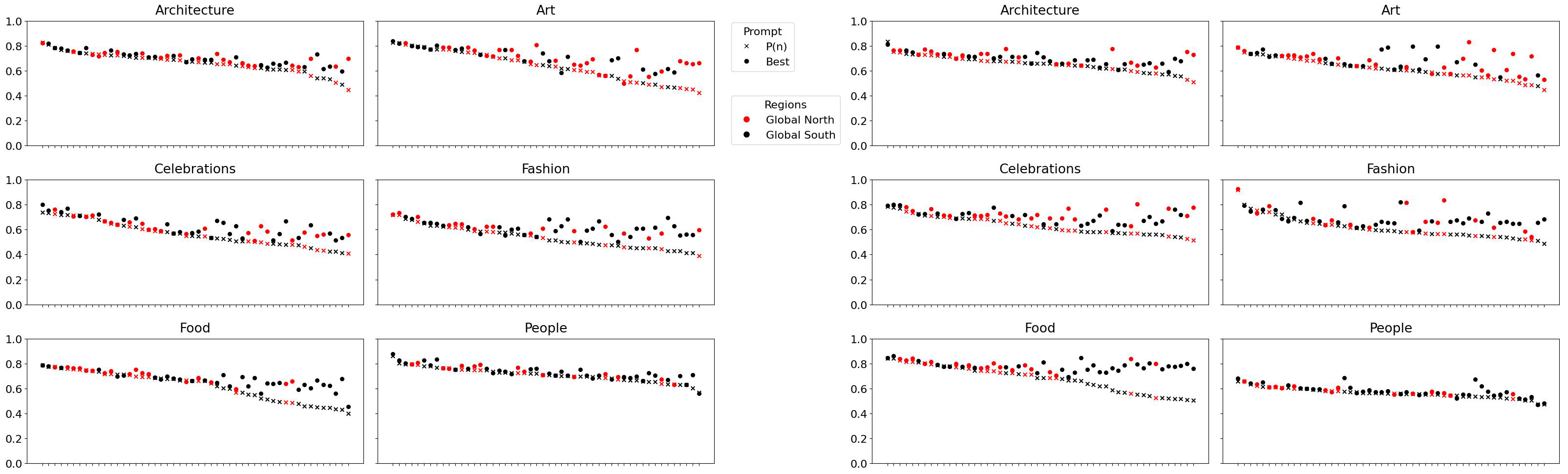}
        }
    
        \hfill
        \subfloat[Stable Diffusion XL]{\label{fig:sdxl_ps_head_tail}
        \centering
        \includegraphics[width=.95\linewidth]{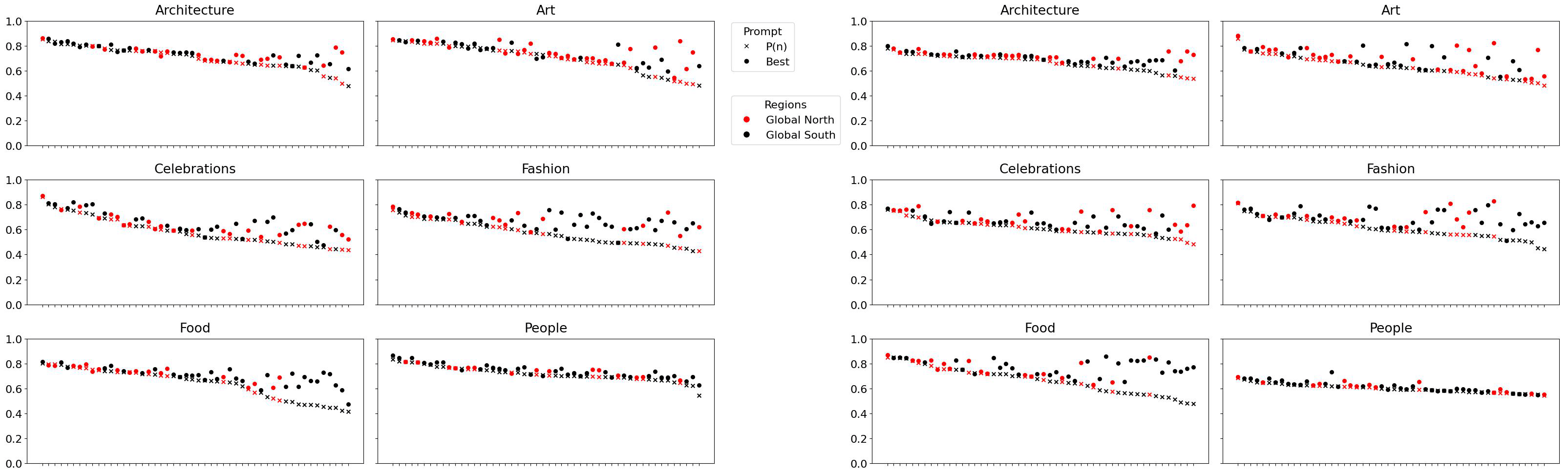}
        }
    \end{figure*}
    
    \begin{figure*}[t]
        \ContinuedFloat
    
        \hfill
        \subfloat[DALL-E 3]{\label{fig:dalle3_ps_head_tail}
        \centering
        \includegraphics[width=.95\linewidth]{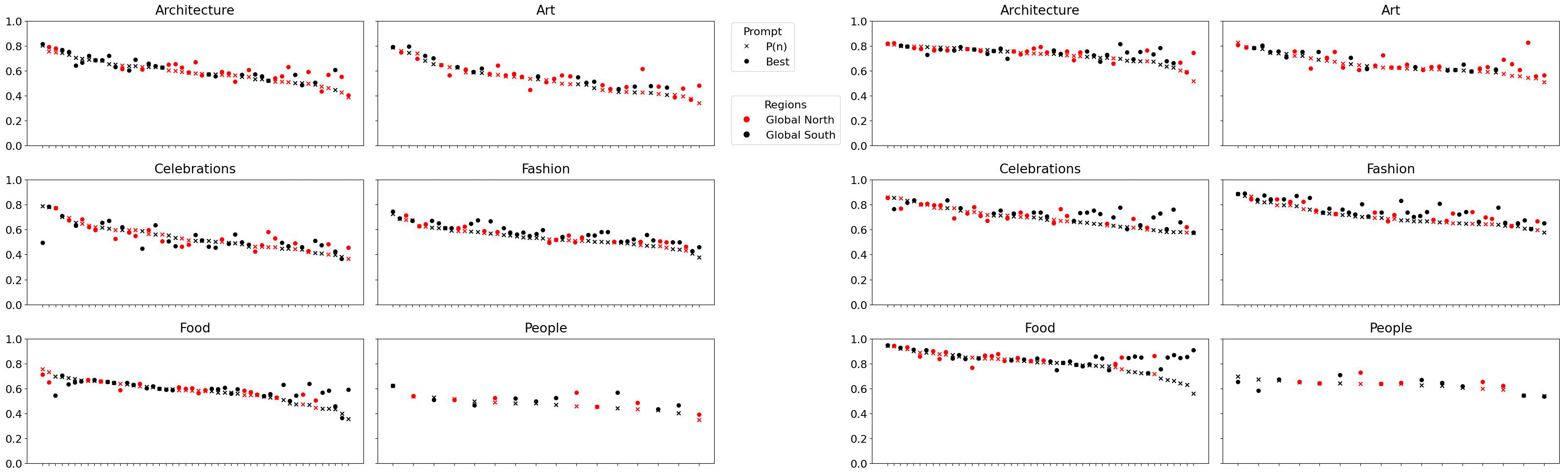}
    }
    
    \caption{Scatter plots of \textbf{Left:} $\phi_{GT(a)}$ and \textbf{Right:} $\phi_{PS}(a)$ for 50 artifacts from all 6 supercategories in the \cure dataset generated using FLUX.1 [dev]. Each data point represents a single cultural artifact, and is an average of the scores for each T2I generated seed. All images are encoded with SigLIP 2. We plot $\phi_{GT}(n)$ and $\phi_{PS}(n)$ with an ``x'' marker, while highest similarity score amongst all four prompt styles ($P(n), P(n,c), P(n,r), P(n,c,r)$ has the ``o'' marker. The colors indicate whether the region the artifact belongs to lies in the Global North (red) or Global South (black). Artifacts are sorted by descending $\phi_{PS}(n)$ similarity score, \ie our marginal attribution score. As seen in the figure, there is a noticable divergence of perceptual similarity score of the most underspecified ($P(n)$) from the ``best'' prompt in the long tail, across T2I systems.}
    \label{fig:app_ps_head_vs_tail}
\end{figure*}

For each T2I system $f_\theta$, in \cref{fig:app_ps_head_vs_tail} we show a scatter plot of raw perceptual similarity scores $\phi_{GT}$ and $\phi_{PS}$ for all 50 cultural artifacts belonging to each supercategory $s$. For each artifact, we plot two scores: $\phi(n)$ and $\arg \max ~\phi(a)$, \ie the highest similarity score across the other three prompts. To examine if the head and tail of the distribution of PS scores correlate with the Global North / South divide, we also color each point according to whether the region the artifact belongs to lies in the Global North (red) or South (black). Below we discuss the ``spread'' or visual divergence in scores between $\phi(n)$ and $\arg \max ~\phi(a)$ as well as ``cultural outlier'', \ie points with a high spread in the head or points with a low spread in the tail.

As seen in \cref{fig:app_ps_head_vs_tail}, we observe a spread of scores in the tail across T2I systems for both scorers $\phi_{GT}$ and $\phi_{PS}$, though the starting point and magnitude of spread (vertical height) differs by supercategory. $\phi_{GT}$ tends to have a lower spread than $\phi_{PS}$, with fewer outliers. All T2I systems have very few outliers in the head (\ie points with high spread), while there tend to be many more outliers in the tail, \ie points with low spread (\eg for Art and People.) People has the least spread across all T2I systems, which is intuitive as there is a very specific way that a certain individual looks, and thus perceptual similarity tends not to change with attribute specification. Interestingly, even though Architecture and Art are similarly also singular named entities (\ie minimal intra-artifact variance in visual features), we observe a later starting point with occasionally large spread (\eg only in the last 10 to 15 artifacts). Food, Fashion and Celebrations have in general the widest spread, as there is generally large intra-class diversity / variance as to how culture-specific food preparations, clothing, and celebrations looks visually. DALL-E 3 (\cref{fig:dalle3_ps_head_tail}) appears to be the most homogenous in perceptual similarity, as the spread is much lesser than other T2I systems, even for Food.

For a more fine-grained analysis than a Global North / South divide, we also examine PS scores across each region (country) in the \cure dataset in \cref{tab:cure_ps_scores}, grouped by continent. To measure \cure of a T2I system $f_\theta$ for region $r$, we compute an average score over the set of cultural artifacts associated with the region, $\mathcal{N}_r$. That is,
\[
\cure(\phi; f_\theta, r) = \frac{1}{|\mathcal{N}_r|} \sum_{n \in \mathcal{N}_r} \phi\bigl(I\bigr).
\]

We observe that when aggregated, outside of North America, no scorer is able to capture the clear trend in human judgments of cultural representativeness $\phi^*_{\cure}$. This indicates that whether a cultural artifact lies in the head or tail of the T2I system distribution cannot be indicated purely by membership to a global region.

\subsection{Qualitative Analysis of PS Scorers}
\label{app:ps_scorer_qual_analysis}

\begin{figure*}[t]
\begin{adjustbox}{width=\textwidth}
    \begin{tabular}{@{}c|cc|l|c@{}}
        \toprule
        \textbf{E\#} & \textbf{AI Image}      & \textbf{Real Image}        & \multicolumn{1}{c|}{\textbf{Feedback}}       & {\textbf{PS Scorers}} \\  \midrule
        \textbf{E1} &\begin{tabular}[c]{@{}c@{}}
            \includegraphics[width=0.55\columnwidth]{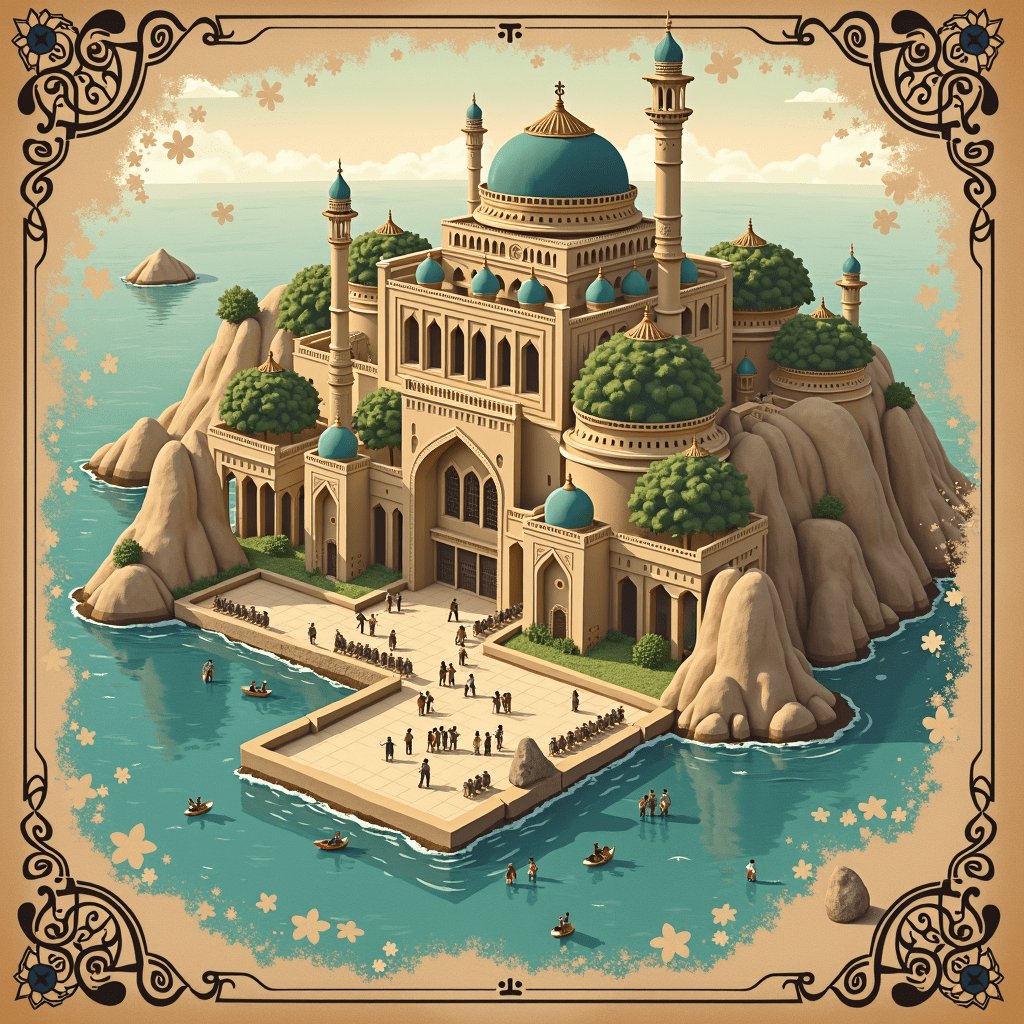}\end{tabular} 
            & \begin{tabular}[c]{@{}c@{}}\includegraphics[width=0.55\columnwidth]{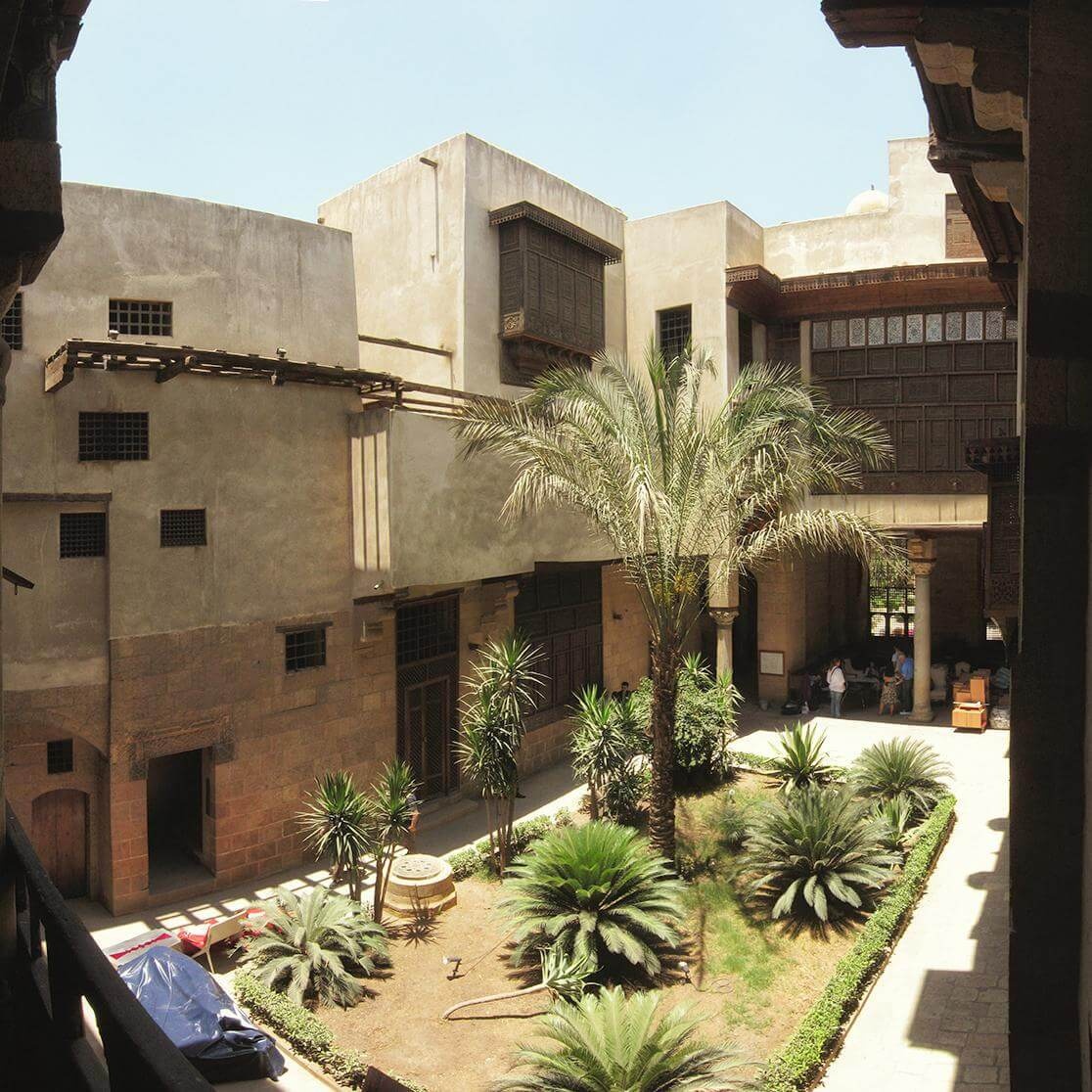}  \end{tabular} 
            & \begin{tabular}[c]{@{}l@{}}
                \textbf{\textit{Bayt al-Suhaymi}} \\ \\
                1. The trees used looks so weird as its mimicing a dome shape. \\some of the small blue domes are futher apart than others. \\asymtrical look \\ \\ 
                2. There is a lot of Dom's typicaly there is only one in any \\building, this picture represent a mosque more than a normal\\ building, also the Dom's can't be made from trees \\ 
            \end{tabular} 
            &   \begin{tabular}[c]{@{}c@{}} 
                {$\!\begin{aligned}
                    \phi^*_{\cure}&= 2.0/5 \\
                    \phi_{GT}(n)&=0.654 \\
                    \phi_{PS}(n)&=0.580  \\
                    \Delta \phi_{PS}(n,c)&=0.552 \\
                    \Delta \phi_{PS}(n,c,r)&=0.506 
                \end{aligned}$}
                \end{tabular} 
            \\ \midrule
        \textbf{E2} &\begin{tabular}[c]{@{}c@{}}
            \includegraphics[width=0.55\columnwidth]{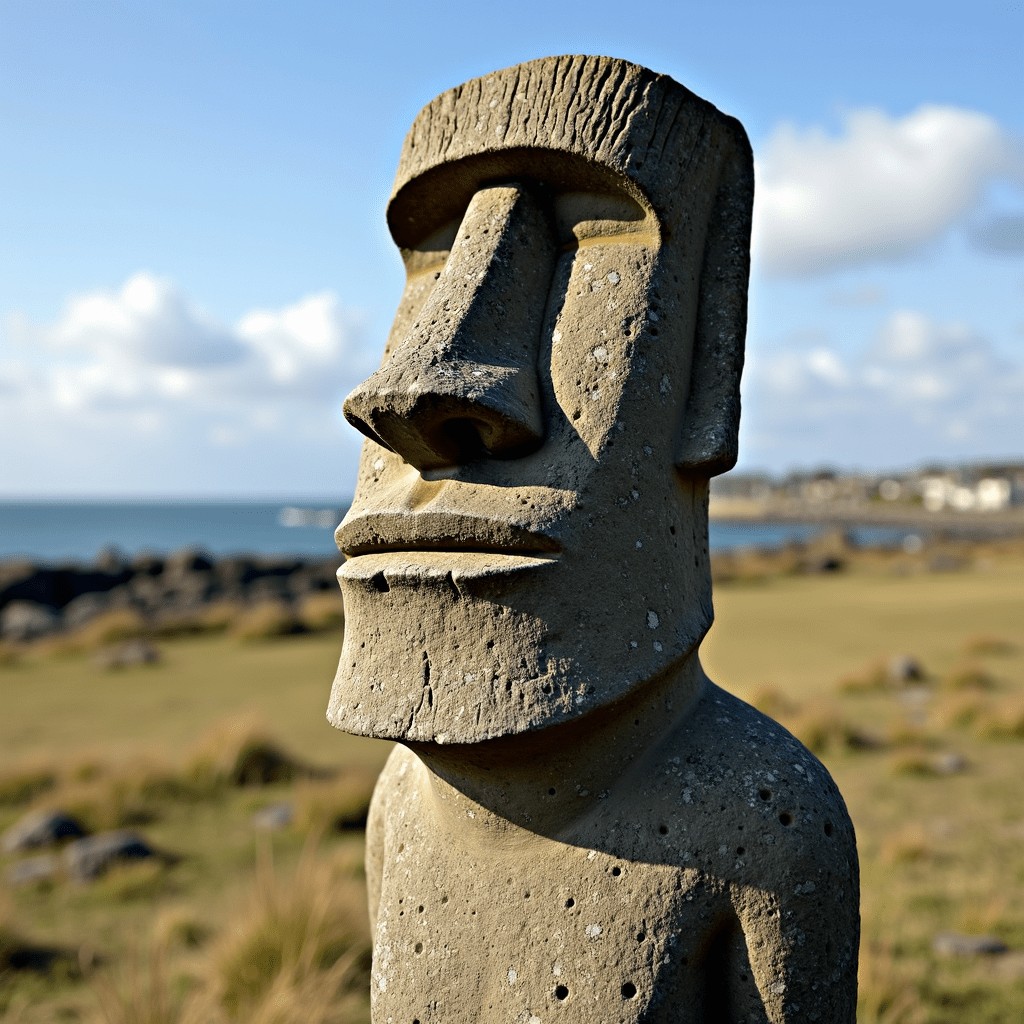}\end{tabular} 
            & \begin{tabular}[c]{@{}c@{}}\includegraphics[width=0.55\columnwidth]{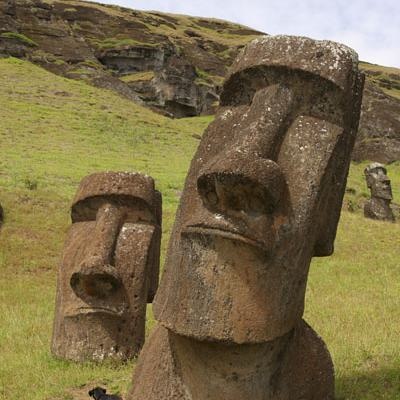}  \end{tabular} 
            & \begin{tabular}[c]{@{}l@{}}
                \textbf{\textit{Moai}} \\ \\
                1. the head to body ratio is a bit off, the body should either be \\more elongated or the head bigger, some even had some kind of\\ red hat and they all looked to the ocean. \\ \\
                2. The main thing that I think is inaccurate is that the moai in the \\image looks too perfect and the real ones have a lot of damage due \\to nature.
            \end{tabular} 
            &   \begin{tabular}[c]{@{}c@{}}
                {$\!\begin{aligned}
                    \phi^*_{\cure}&= 4.0/5 \\
                    \phi_{GT}(n)&= 0.808 \\
                    \phi_{PS}(n)&= 0.581 \\
                    \Delta \phi_{PS}(n,c)&=0.486\\
                    \Delta \phi_{PS}(n,c,r)&=0.486
                \end{aligned}$}
                \end{tabular} 
            \\ \midrule
        \textbf{E3} &\begin{tabular}[c]{@{}c@{}}
            \includegraphics[width=0.55\columnwidth]{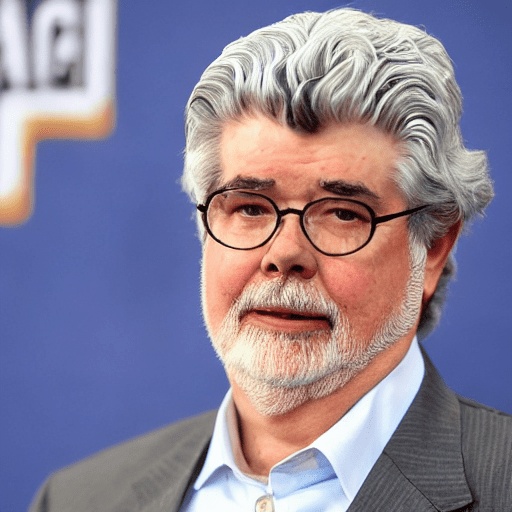} \end{tabular} 
            & \begin{tabular}[c]{@{}c@{}}\includegraphics[width=0.55\columnwidth]{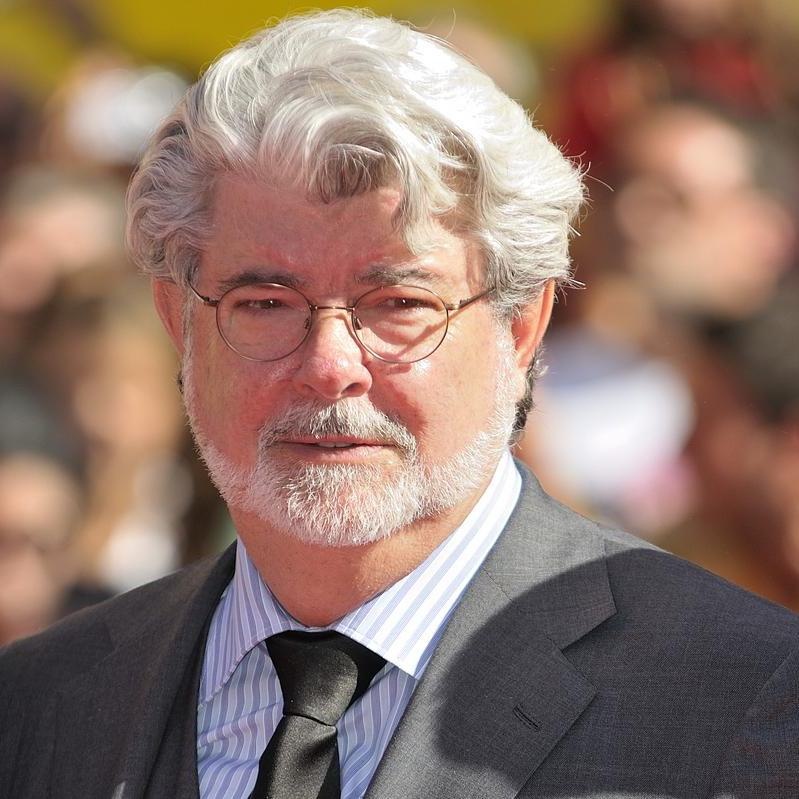} \end{tabular} 
            & \begin{tabular}[c]{@{}l@{}}
                \textbf{\textit{George Lucas}} \\ \\
                1. This is a closeup picture of George Lucas’ face. There’s nothing \\particularly unique about it that would NOT make it apart of my \\culture. I don’t see why I couldn’t see this kind of photo in my \\culture. \\ \\
                2. I do not see anything that is "wrong" other than the eyes. They \\are obviously AI.
            \end{tabular} 
            &   \begin{tabular}[c]{@{}c@{}}
                {$\!\begin{aligned}
                    \phi^*_{\cure}&= 5.0/5 \\
                    \phi_{GT}(n)&= 0.745 \\
                    \phi_{PS}(n)&= 0.571  \\
                    \Delta \phi_{PS}(n,c)&=0.489 \\
                    \Delta \phi_{PS}(n,c,r)&=0.460
                \end{aligned}$}
                \end{tabular} 
            \\ \midrule
        \textbf{E4} &\begin{tabular}[c]{@{}c@{}}
            \includegraphics[width=0.55\columnwidth]{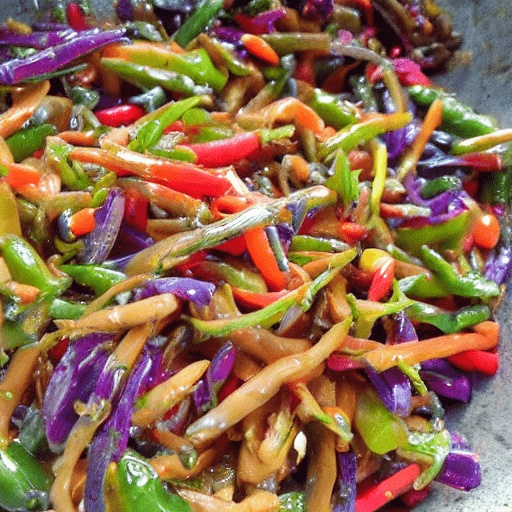} \end{tabular} 
            & \begin{tabular}[c]{@{}c@{}}\includegraphics[width=0.55\columnwidth]{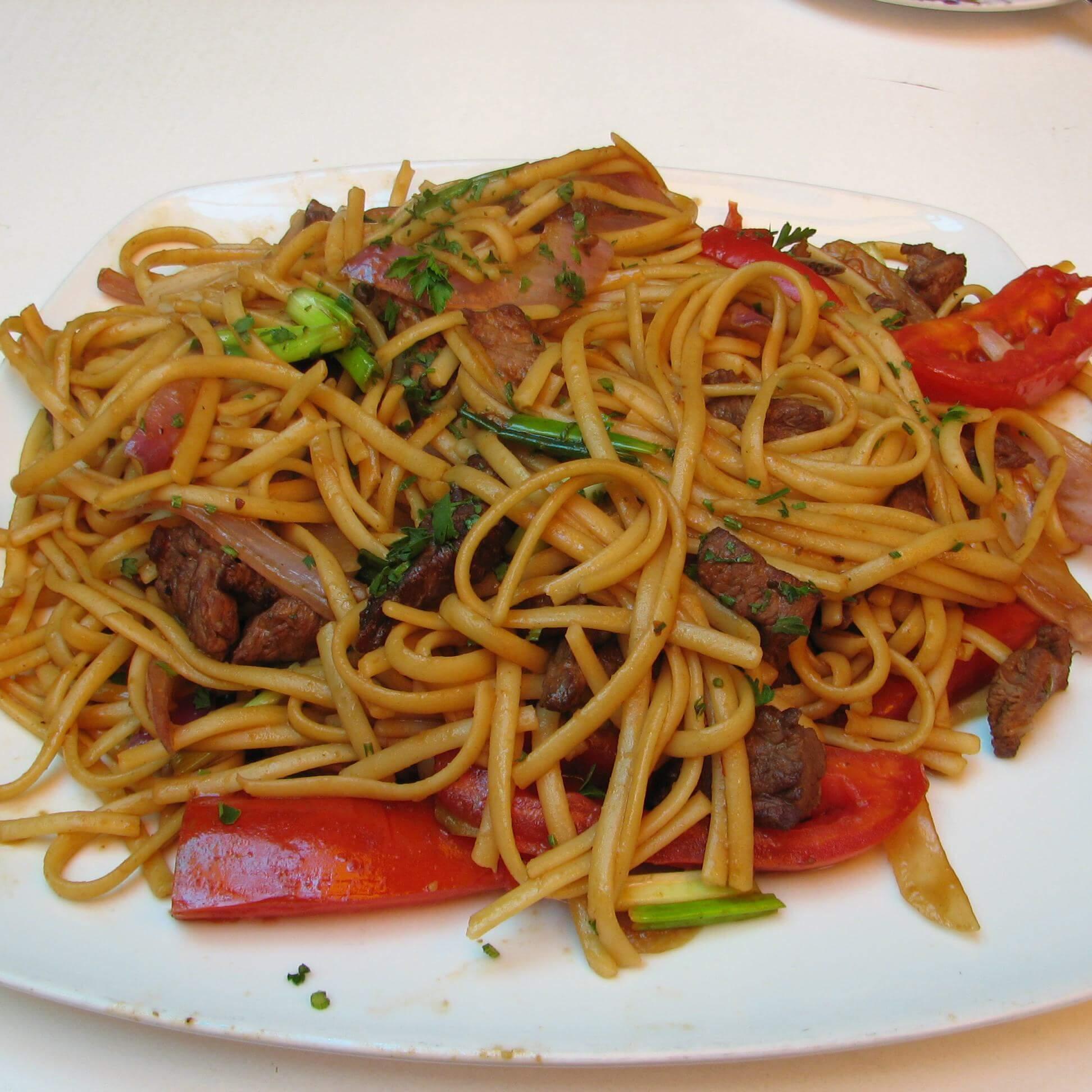}  \end{tabular} 
            & \begin{tabular}[c]{@{}l@{}}
                \textbf{\textit{Tallar\'in saltado}} \\ \\
                1. the noodles look like worms, tallarín saltado is spaghetti with \\meat and vegetables whereas the image only shows weird pasta \\and no meat nor onions/tomatoes. \\ \\
                2. The image doesn't look like noodles to me, it seems like a kind \\of vegetable so I can't say is accurate to the real Tallarín saltado
            \end{tabular} 
            &   \begin{tabular}[c]{@{}c@{}}
                {$\!\begin{aligned}
                    \phi^*_{\cure}&= 1.0/5 \\
                    \phi_{GT}(n)&= 0.654 \\
                    \phi_{PS}(n)&= 0.696  \\
                    \Delta \phi_{PS}(n,c)&= 0.486 \\
                    \Delta \phi_{PS}(n,c,r)&= 0.425 
                \end{aligned}$}
                \end{tabular} 
            \\ \midrule
        \textbf{E5} &\begin{tabular}[c]{@{}c@{}}
            \includegraphics[width=0.55\columnwidth]{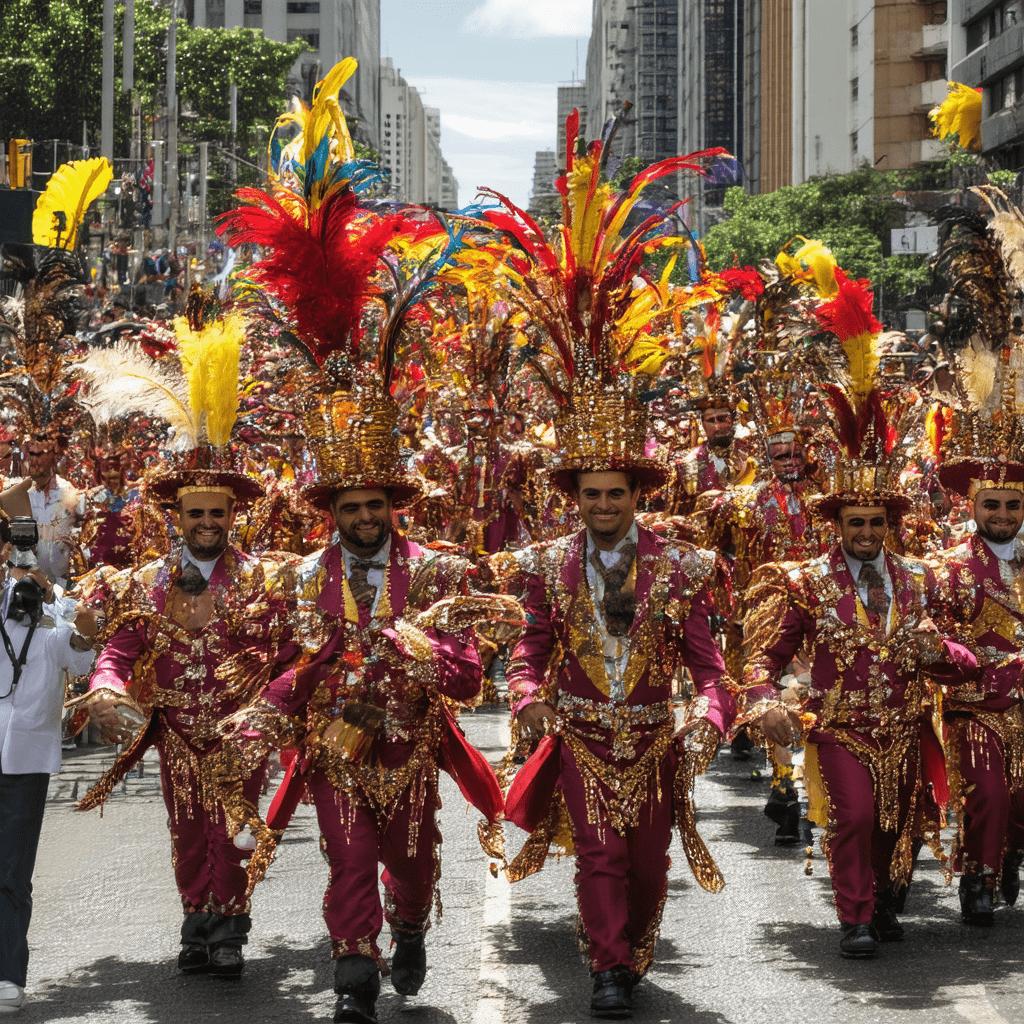} \end{tabular} 
            & \begin{tabular}[c]{@{}c@{}}\includegraphics[width=0.55\columnwidth]{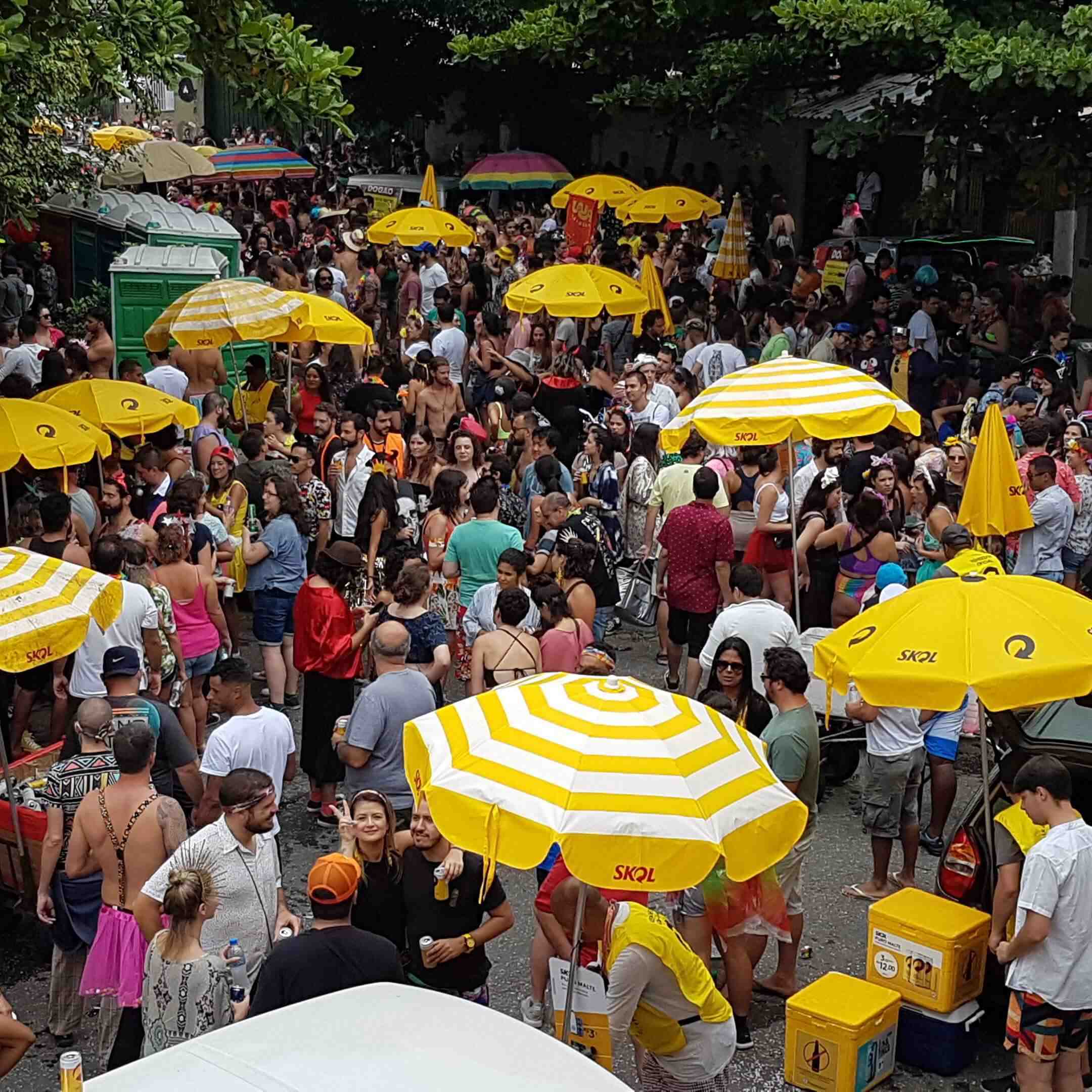}\end{tabular} 
            & \begin{tabular}[c]{@{}l@{}}
                \textbf{\textit{Blocos carnavalescos de São Paulo}} \\ \\
                1. The costumes are indeed carnival like, but the blocos are less \\stravagant and the costumes wore by the people usualy contain \\some kind of a joke on a very low budget costume \\ \\
                2. Their physical appearance and posture.
            \end{tabular} 
            &   \begin{tabular}[c]{@{}c@{}}
                {$\!\begin{aligned}
                    \phi^*_{\cure}&= 1.0/5 \\
                    \phi_{GT}(n)&= 0.712 \\
                    \phi_{PS}(n)&= 0.757  \\
                    \Delta \phi_{PS}(n,c)&= 0.504 \\
                    \Delta \phi_{PS}(n,c,r)&= 0.473 
                \end{aligned}$}
                \end{tabular} 
            \\ \midrule
        \textbf{E6} &\begin{tabular}[c]{@{}c@{}}
            \includegraphics[width=0.55\columnwidth]{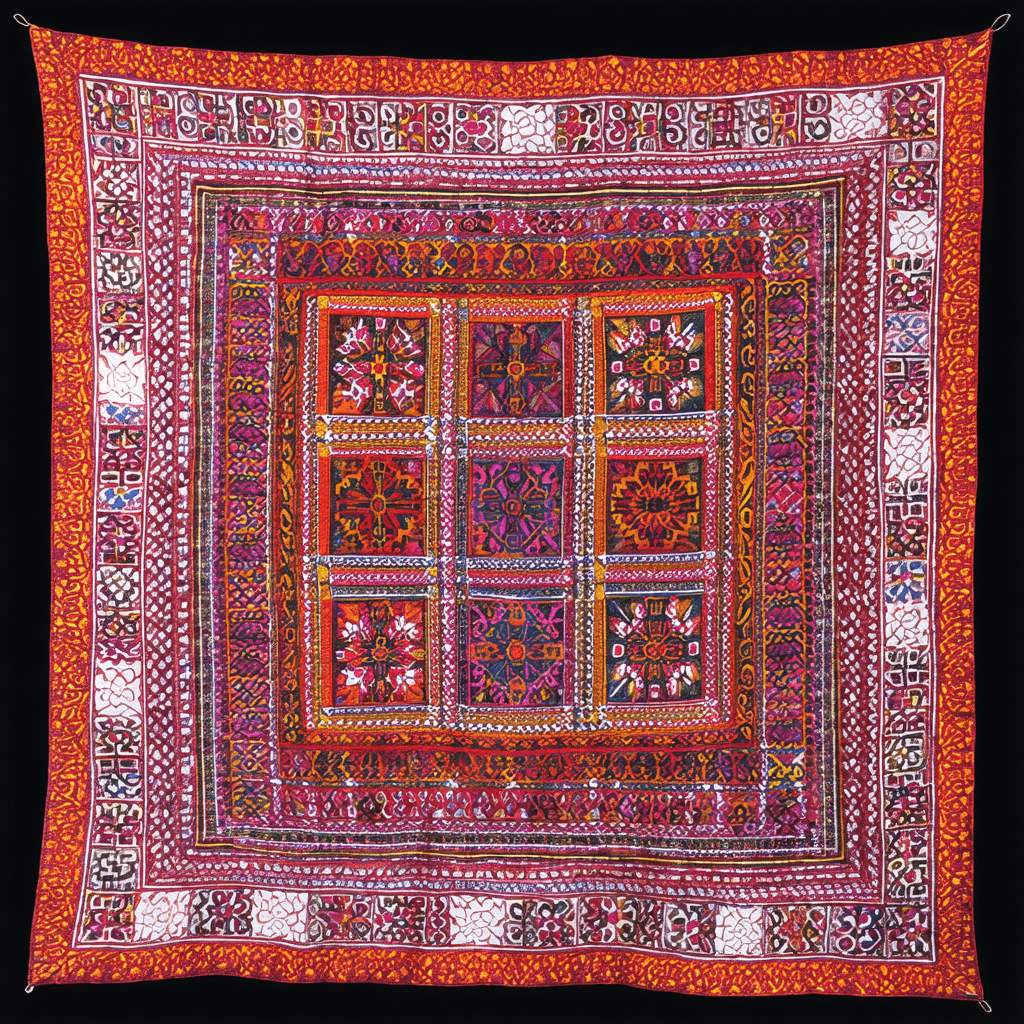}\end{tabular} 
            & \begin{tabular}[c]{@{}c@{}}\includegraphics[width=0.55\columnwidth]{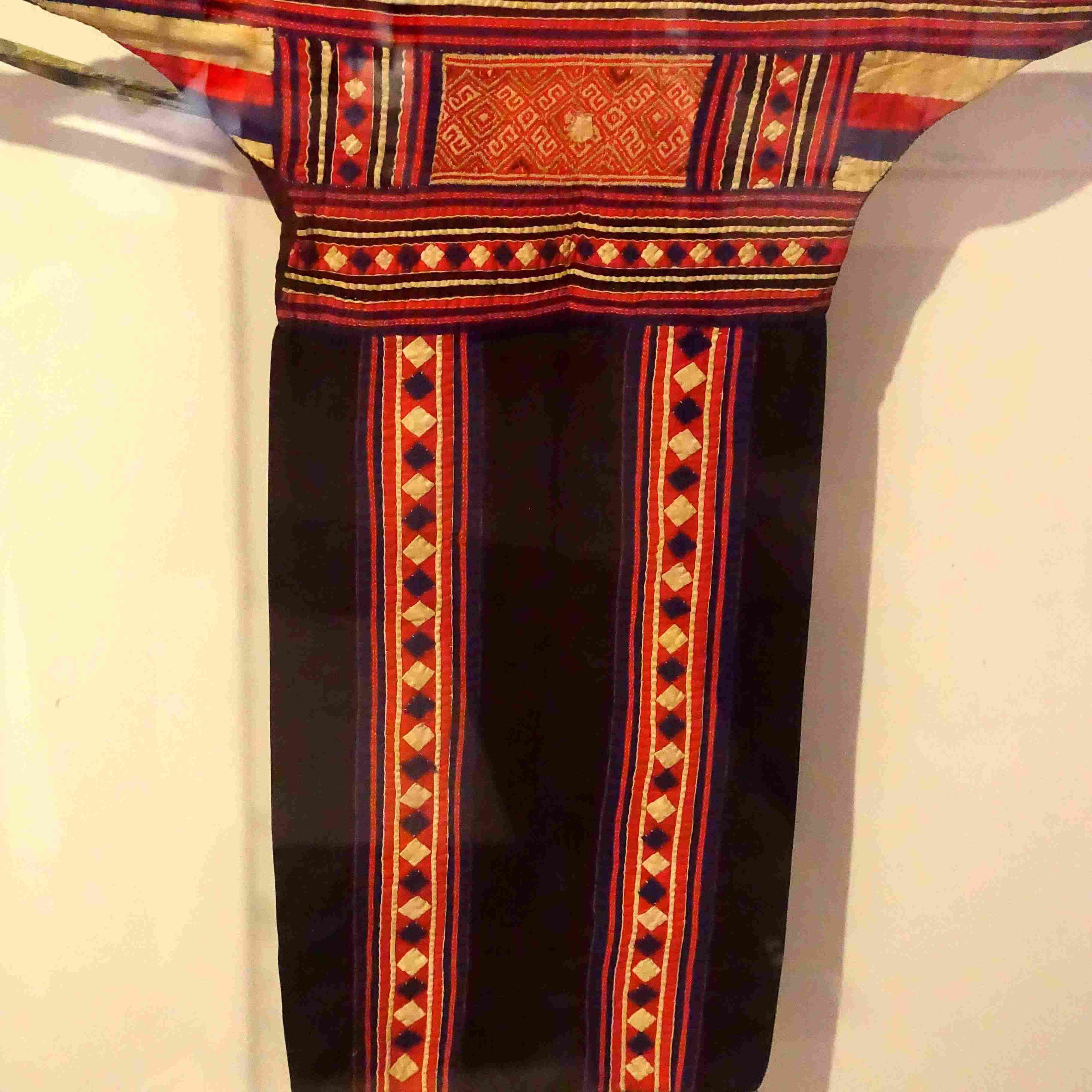} \end{tabular} 
            & \begin{tabular}[c]{@{}l@{}}
                \textbf{\textit{Hmong textile art}} \\ \\
                1. The design motif and colors are quite close to the real ones \\ \\
                2. I think the repetitive patterns and the colourful decoration make \\it accurate
            \end{tabular} 
            &   \begin{tabular}[c]{@{}c@{}}
                {$\!\begin{aligned}
                    \phi^*_{\cure}&= 5.0/5 \\
                    \phi_{GT}(n)&=0.620 \\
                    \phi_{PS}(n)&=0.634  \\
                    \Delta \phi_{PS}(n,c)&=0.519 \\
                    \Delta \phi_{PS}(n,c,r)&=0.518 
                \end{aligned}$}
                \end{tabular} 
        \\ \bottomrule
    \end{tabular}
\end{adjustbox}
\caption{Visualization of quantitative perceptual similarity (PS) scorers and user study \cure scores and text feedback (\cref{app:user_study_cure}). The first two artifacts were generated with FLUX.1 [dev], the second two with SD 1.5, and the last two with SD 3.5 Large.}
\label{fig:scorer_viz_ps}
\end{figure*}

We highlight several qualitative examples on the \cure dataset with our perceptual similarity scorers $\phi_{PS}$ and $\Delta \phi_{PS}$ compared to the strong baseline scorer $\phi_{GT}(n)$ in \cref{fig:scorer_viz_ps}. We also show Likert scores for cultural representativeness $\phi^*_{\cure}$ and textual justification for each example.

As we observed in our quantiative results (\cref{tab:results_ps}), our divergence PS scorer $\Delta \phi_{PS}(\{n, c\})$ correlates strongly (and negatively) with the $\phi^*_{\cure}$ gold scores, \ie low $\Delta \phi_{PS}(\{n,c\})$ indicating higher $\phi^*_{\cure}$, \eg \textbf{E1} (\textit{Bayt al-Suhaymi}), \textbf{E2} (\textit{Moai}), \textbf{E3} (\textit{George Lucas}). We also highlight a failure case for our scorers with \textbf{E4} (\textit{Tallar\'in saltado}), a type of noodle dish from Peru generated by SD 1.5. While we expect a low $\Delta \phi_{PS}$ score to correspond to high cultural representativeness, workers rate a low $\phi^*_{\cure}$ score of 1 out of 5 and indicate errors with textual response, \ie incorrect ingredients and out-of-place textures. $\phi_{PS}$ also did worse than the baseline $\phi_{GT}(n)$ as it gave a higher perceptual similarity score (falsely predicting high quality T2I output). This failure aligns with our quantitative rank correlation analysis in \cref{tab:results_ps}, where we observe very Spearman's $\rho$ for SD 1.5, an older and smaller T2I system.

We also highlight two cases where both our proposed scorers and the baseline scorer fail. \textbf{E5} (\textit{Blocos carnavalescos de São Paulo}), a carnival celebration from Brazil, had a 1 out of 5 $\phi^*_{\cure}$ for SD 3.5 Large. The written feedback also reflects that the images are sterotypically extravagant and incorrect in physical appearance. However, the quantitative perceptual similarity scorers were very high compared to other artifacts with the same low $\phi^*_{\cure}$ score. In contrast, \textbf{E6} (\textit{Hmong textile art}), a style of embroidery from Vietnam had an excellent 5 out of 5 $\phi^*_{\cure}$ for SD 3.5 Large, but lower scores from the quantitative PS scorers than \textbf{E5} which was much less culturally representative according to human judgments. 

\subsection{PS vs Concept Frequency}
\begin{table}[t]
    \caption{Spearman's $\rho$ between frequency count of each artifact name in LAION-SD15 (see \cref{app:concept_frequency}) and perceptual similarity scorers $\phi_{GT}(n)$ and $\phi_{PS}(n)$ evaluated on the \cure dataset.}
    \begin{adjustbox}{max width=\columnwidth}

    \begin{tabular}{@{}c|cc|cc|cc|cc@{}}
    \toprule
    \multirow{2}{*}{\textbf{Encoder}} & \multicolumn{2}{c|}{\textbf{SD 1.5}} & \multicolumn{2}{c|}{\textbf{SD 3.5 Large}} & \multicolumn{2}{c|}{\textbf{SDXL}} & \multicolumn{2}{c}{\textbf{Flux.1 [dev]}} \\ 
    \cline{2-9} 
     & $\phi_{GT}$ & $\phi_{PS}$ & $\phi_{GT}$ & $\phi_{PS}$ & $\phi_{GT}$ & $\phi_{PS}$ & $\phi_{GT}$ & $\phi_{PS}$ \\ 
     \midrule
    \textbf{SigLIP 2} \cite{tschannen2025siglip} & 0.64 & ~0.22 & 0.58 & 0.34 & 0.55 & 0.25 & 0.60 & 0.37 \\
    \textbf{AIMV2} \cite{fini2024multimodal} & 0.53 & -0.08 & 0.52 & 0.25 & 0.42 & 0.08 & 0.49 & 0.33 \\
    \textbf{DINOv2} \cite{oquabdinov2} & 0.57 & ~0.25 & 0.56 & 0.40 & 0.46 & 0.29 & 0.55 & 0.45 \\ 
    \bottomrule
    \end{tabular}
    \end{adjustbox}
    \label{tab:freq_corr}
\end{table}

In \cref{tab:freq_corr} we show the Spearman rank correlation of the frequency the artifact name appears in the captions of LAION-SD1.5 (the pretraining dataset of Stable Diffusion 1.5; details in \cref{app:concept_frequency}) and the perceptual similarty scores $\phi_{GT}(n)$ and $\phi_{PS}(n)$, which use images generated with the prompt only containing the artifact name. For $\phi_{GT}(n)$, SD 1.5 has the highest positive correlation out of all T2I systems across encoders, which is expected as we evaluate frequency over its exact pretraining dataset. On the contrary for $\phi_{PS}(n)$, Spearman's rank correlation to concept frequency in LAION-SD15 followed Arena ELO rankings, \ie FLUX.1 [dev] $>$ SD 3.5 Large $>$ SDXL $>$ SD 1.5. This likely indicates that artifact - category associations for stronger T2I systems like FLUX.1 [dev] and SD 3.5 were reinforced by the larger size and coverage of their pretraining data.

\clearpage
\section{Image-Text Alignment}
\label{app:ita}
We examine in important factor in the context of image-text alignment scorers in \cref{app:ita_backbone_ablation}: what impact does the choice of vision-language model backbone have on scorer quality? We also discuss qualitative examples with ITA scorers in \cref{app:ita_scorer_qual_analysis}.

\begin{table}[h!]
\caption{List of prompts used for text-image similarity score calculations. \eg for region $r =$ "Australia", the T2I prompt $P(r) = $ ``An image from Australia''.}
    \centering
    \small
    \renewcommand{\arraystretch}{1.1} 
    \setlength{\tabcolsep}{2pt} 
    \begin{adjustbox}{max width=\columnwidth}
    \begin{tabular}{|l|l|}  
        \hline
        \textbf{Scorer} & \textbf{Prompt} \\ 
        \hline
        \citet{khanuja2024image} & ``This image is culturally relevant to \{r\}.'' \\ 
        \citet{ventura2023navigating} & ``Image from \{r\} culture.'' \\ 
        o3 mini~\cite{o3mini} & ``Assess the image’s cultural representation of \{r\}.'' \\ 
        \( P(n) \) & ``An image of \{n\}.'' \\ 
        \( P(c) \) & ``An image of \{c\}.'' \\ 
        \( P(r) \) & ``An image from \{r\}.'' \\ 
        \( P(c, r) \) & ``An image of \{c\} from \{r\}.'' \\ 
        \hline
    \end{tabular}
    \end{adjustbox}
    \vspace{-3mm}
    \label{tab:im_txt_eval_prompts}
\end{table}

\subsection{Choice of ITA Scorer's VLM Backbone}
\label{app:ita_backbone_ablation}
\begin{table}[h!]
\caption{Spearman correlation values of ITA scorers with human judgments of perceptual similarity $\phi^*_{PS}$ across ITA scorer backbones for FLUX.1 [dev]. The backbones we evaluate include SigLIP 2 and OpenCLIP models trained on OpenAI WIT~\cite{radford2021learning}, Data Filtering Networks (DFN-5B)~\cite{fangdata}, LAION-2B~\cite{schuhmann2022laion}, and DataComp (DC-1B)~\cite{gadre2023datacomp}.}
    \centering
    \small
    \renewcommand{\arraystretch}{1.1}
    \setlength{\tabcolsep}{2pt}
    \begin{adjustbox}{max width=\columnwidth}
    \begin{tabular}{l|c |c| c| c| c}
      \toprule
      \textbf{ITA Scorer} &\textbf{LAION-2B} & \textbf{WIT} & \textbf{DFN-5B} & \textbf{SigLIP 2} & \textbf{DC-1B} \\
      \midrule
      \citet{khanuja2024image}&0.18  &0.16  &0.12  &0.11  &0.12  \\
      \citet{ventura2023navigating}&0.16  &0.09  &0.16  &0.14  &0.16  \\
      o3-mini&0.17   &0.17  &0.13  &0.14  &0.15  \\
      \midrule
      $sim\bigl(I(n),\, P(n)\bigr)$ &0.35       &0.39  &0.41  &0.38  &0.37  \\
      $sim\bigl(I(n),\, P(c)\bigr)$   &0.33     &0.38  &0.35  &0.34  &0.31  \\
      $sim\bigl(I(n),\, P(r)\bigr)$   &0.16    &0.09  &0.17  &0.12  &0.13  \\
      $sim\bigl(I(n),\, P(c,r)\bigr)$  &0.37       &0.39  &0.38  &0.38  &0.34  \\ \midrule
      $\phi_{ITA}(c)$ &0.39 &0.43  &0.43  &0.40  &0.39  \\
      $\phi_{ITA}(r)$ &0.32  &0.32  &0.37  &0.35  &0.33  \\
      $\phi_{ITA}(c, r)$ &0.40 &0.43  &0.44  &0.42  &0.40  \\
      \bottomrule
    \end{tabular}
    \end{adjustbox}
    \label{tab:clip_correlations}
\end{table}

We replicate the Spearman rank correlation setup from \cref{tab:results_ita} and ablate over the choice of VLM backbone used to compute image-text alignment for FLUX.1 [dev] in \cref{tab:clip_correlations} (see \cref{subsec:results_ita} for VLM details). To recap, we compute a Spearman's $\rho$ of each scorer with the user study gold score $\phi^*_{PS}$. We observe that baselines (\citet{khanuja2024image}, \citet{ventura2023navigating}, and o3 mini~\cite{o3mini}), which query the VLM for \cure score directly with region information $r$, are sensitive to changes in the backbone and show high variability in ITA scores. As we marginally increase attributes in the evaluation prompt $P$ ($n \to c$, $n \to r$, $n \to \{c,r\}$), the rank correlations become more consistent across VLM backbones. Our proposed metrics show both higher and more consistent rank correlations with $\phi^*_{PS}$ gold scores across all VLM backbones, showing that they are more robust than baselines to the pretraining distribution of the VLM for evaluating cultural representativeness.

\subsection{Qualitative Analysis of ITA Scorers}
\label{app:ita_scorer_qual_analysis}
\begin{figure*}[t]
\begin{adjustbox}{width=\textwidth}
    \begin{tabular}{@{}c| c c|l|c@{}} 
        \toprule
        \textbf{E\#} & \textbf{AI Image}      & \textbf{Real Image}        & \multicolumn{1}{c|}{\textbf{Feedback}}       & {\textbf{ITA Scorers}} \\  \midrule
        \textbf{E1} & \begin{tabular}[c]{@{}c@{}}
            \includegraphics[width=0.55\columnwidth]{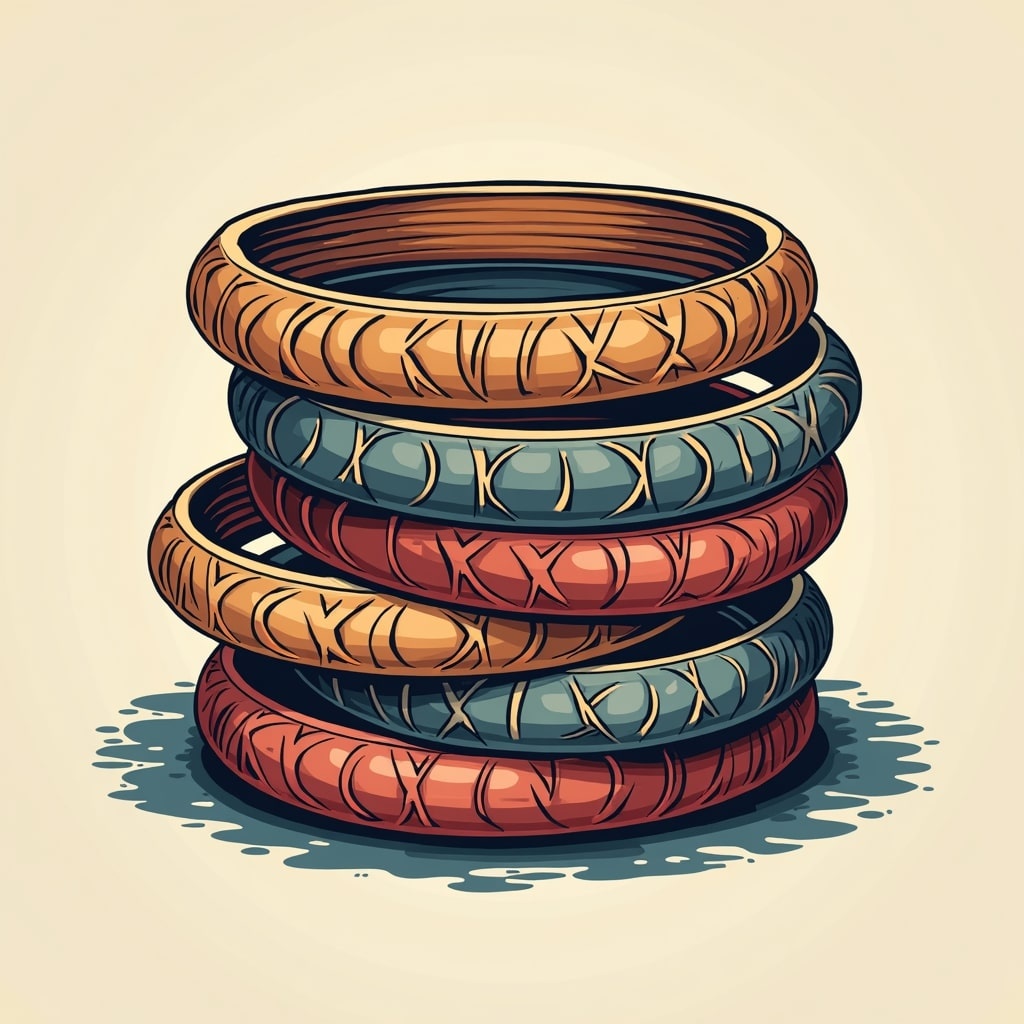}
             \end{tabular} 
            & \begin{tabular}[c]{@{}c@{}}\includegraphics[width=0.55\columnwidth]{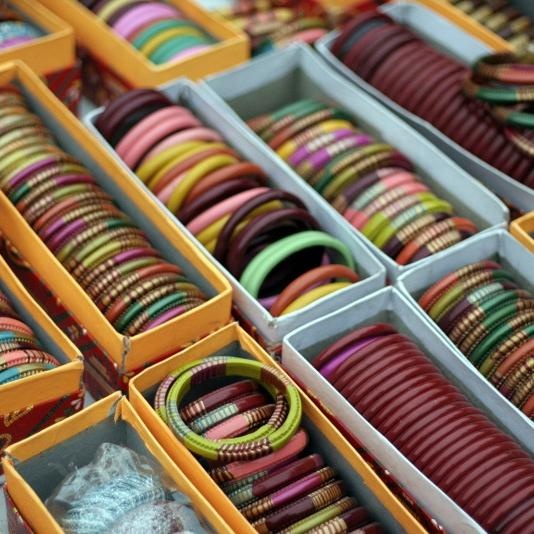} \end{tabular} 
            & \begin{tabular}[c]{@{}l@{}}
            \textit{\textbf{Bangles}} \\ \\
                1. Bangles are circular ornaments with some colors and patterns \\ on them.\\ \\
                2. The width and the carvings and the colors make it seem more \\ aligned to my culture\\ 
            \end{tabular} 
            &   \begin{tabular}[c]{@{}c@{}}
                  $\begin{aligned}
                    \phi^*_{\cure} &= 5.0/5 \\
                    \text{\citet{khanuja2024image}} & = 0.060 \\
                    \text{\citet{ventura2023navigating}} & = 0.045 \\
                    \phi_{ITA}(c,r)          & = 0.124
                  \end{aligned}$
                \end{tabular}
            \\ \midrule
        \textbf{E2} &\begin{tabular}[c]{@{}c@{}}
            \includegraphics[width=0.55\columnwidth]{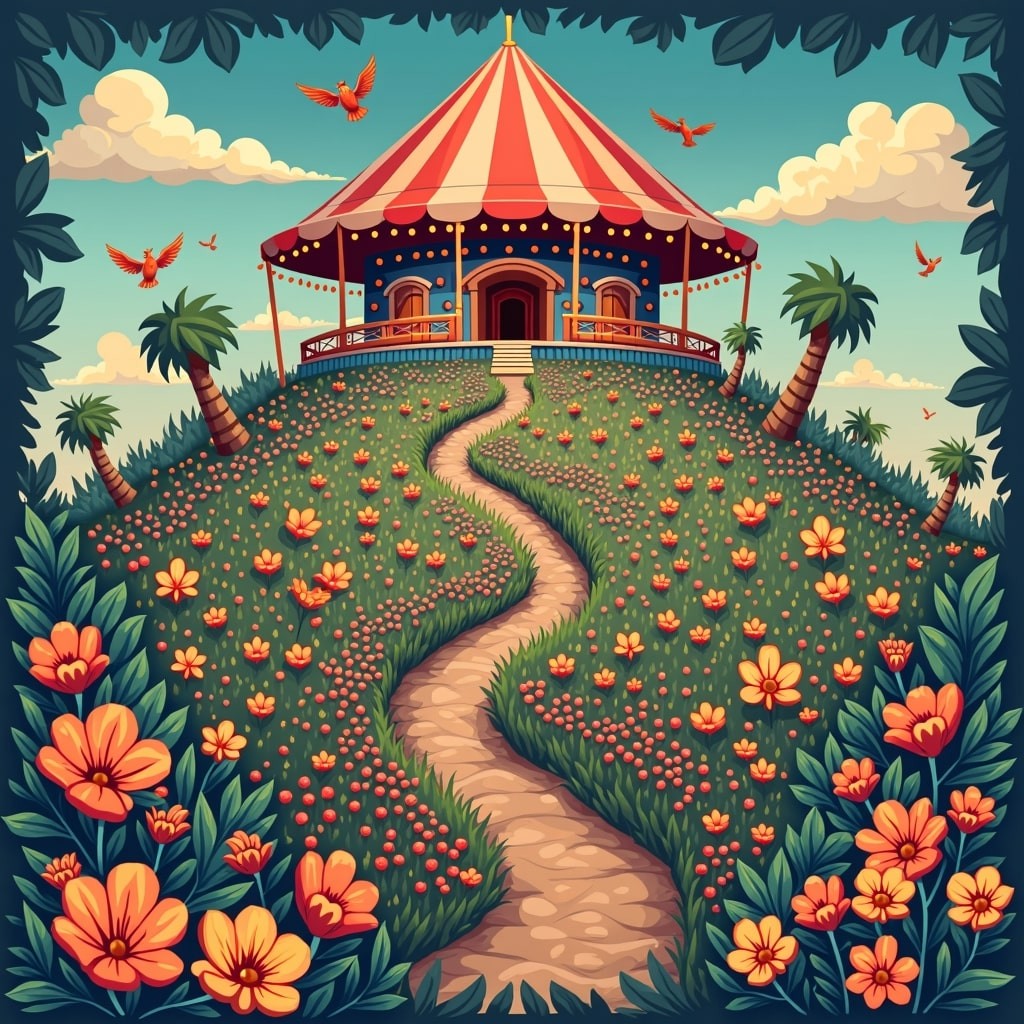}
            \end{tabular} 
            & \begin{tabular}[c]{@{}c@{}}\includegraphics[width=0.55\columnwidth]{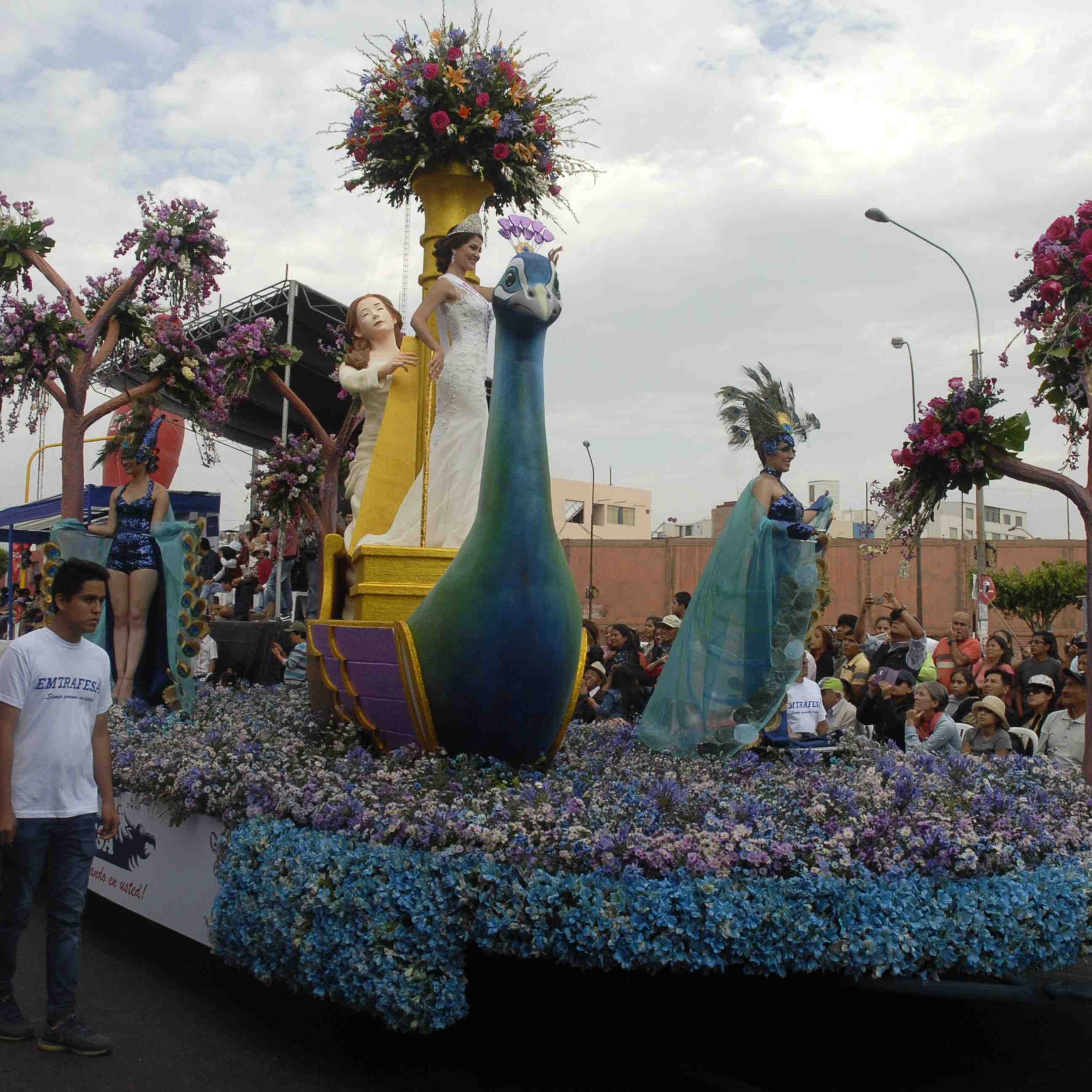} \end{tabular} 
            & \begin{tabular}[c]{@{}l@{}}
            \textit{\textbf{Festival de la Primavera}} \\ \\
                1. The festival takes place in the cost of Lima, no there are not a\\ lot of hilly areas, and it’s more a dry environment so those flowers \\ and the place do not correspond at all to Trujillo reality.\\ \\
                2. The image represents the Spring in general. But the main event \\ of the Festival de la Primavera is the parade (People, carriages \\ and flowers). 
            \end{tabular} 
            &   \begin{tabular}[c]{@{}c@{}}
                  $\begin{aligned}
                  \phi^*_{\cure} &= 1.0/5 \\
                  \text{\citet{khanuja2024image}} & = 0.086 \\
                   \text{\citet{ventura2023navigating}} & = 0.067 \\
                    \phi_{ITA}(c,r)          & = 0.156
                  \end{aligned}$
                \end{tabular}
            \\ \midrule
       \textbf{E3} & \begin{tabular}[c]{@{}c@{}}
            \includegraphics[width=0.55\columnwidth]{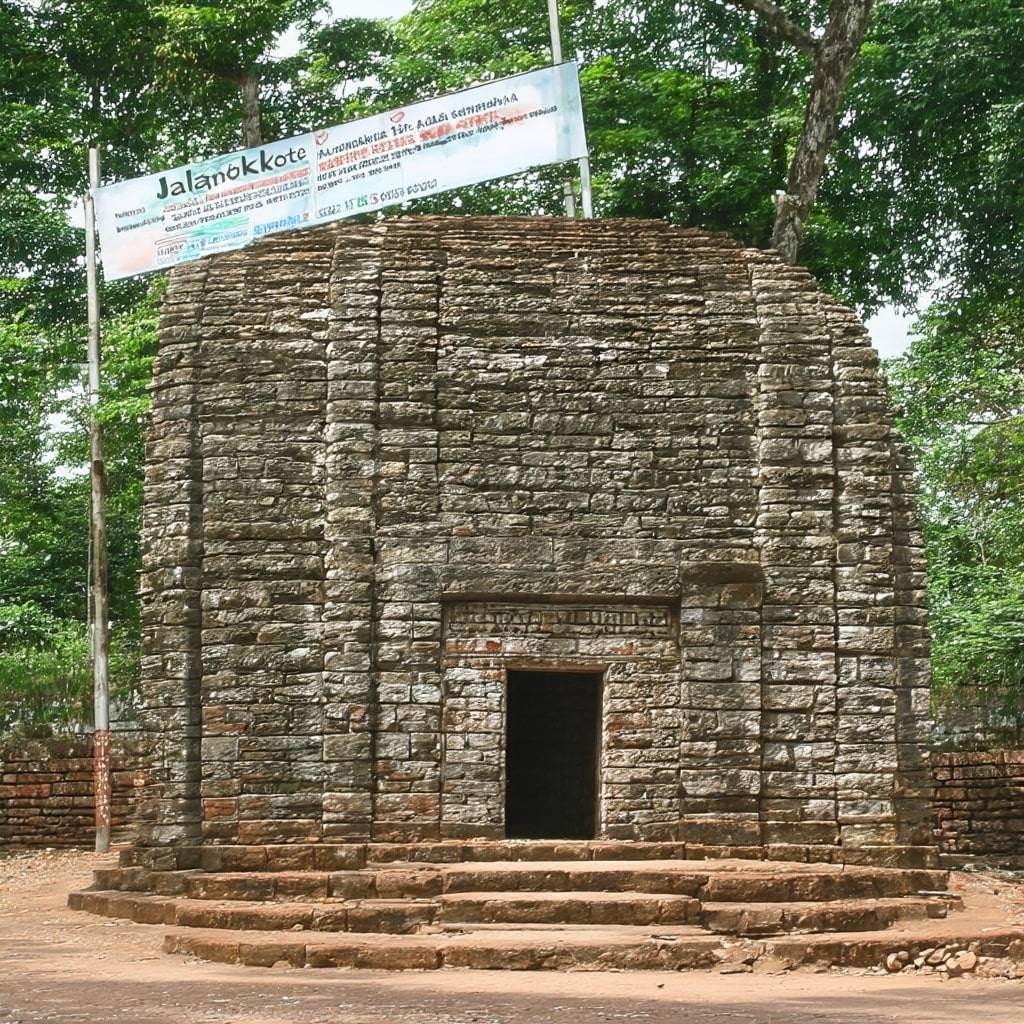}
             \end{tabular} 
            & \begin{tabular}[c]{@{}c@{}}\includegraphics[width=0.55\columnwidth]{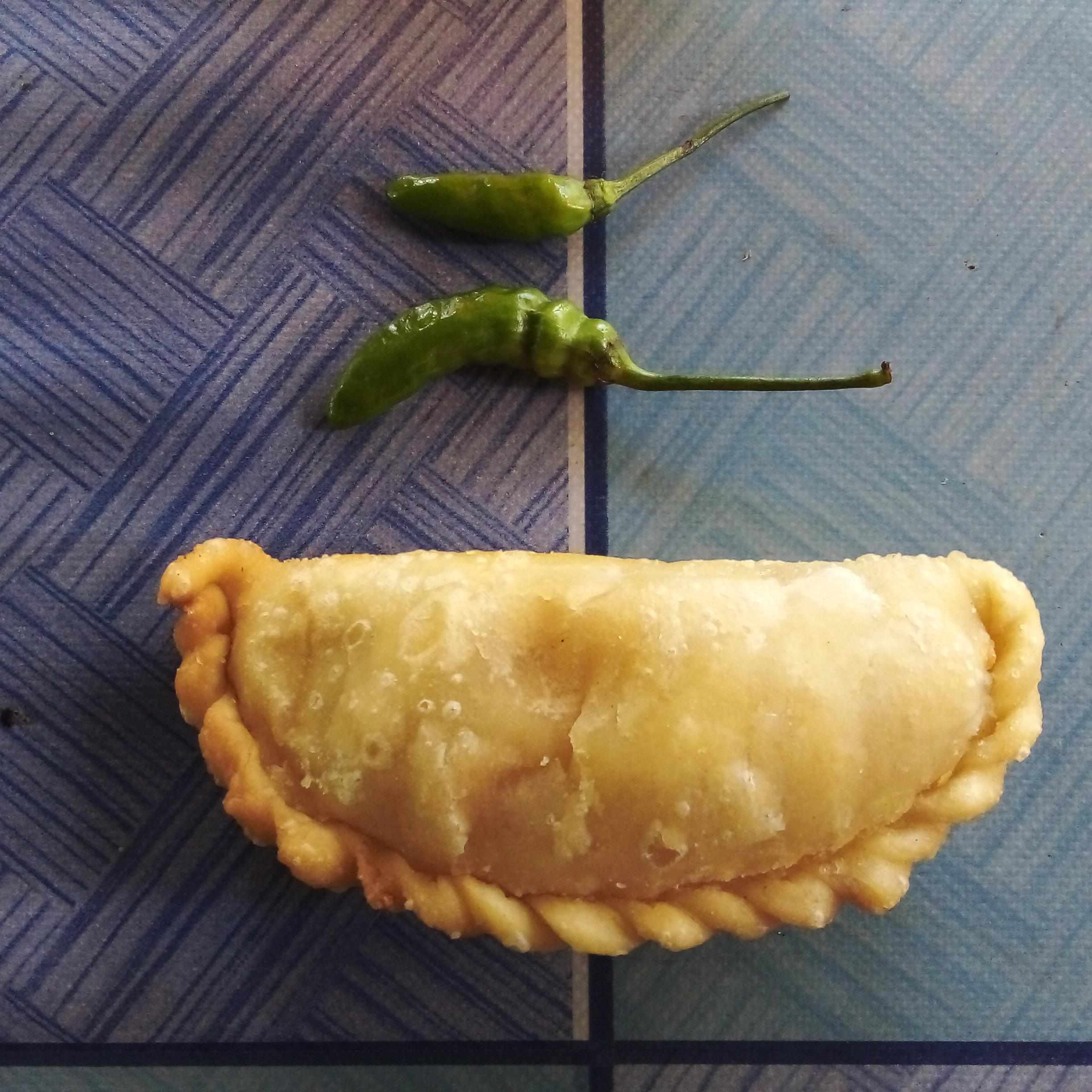}
             \end{tabular} 
            & \begin{tabular}[c]{@{}l@{}}
            \textit{\textbf{Jalangkote}} \\ \\
                1. The image use the wrong thing, AI image is using stone like \\ building while Jalangkote is a food. a complete different genre  \\ \\
                2. The AI image is not even a food.
            \end{tabular} 
            &   \begin{tabular}[c]{@{}c@{}}
                  $\begin{aligned}
                   \phi^*_{\cure} &= 1.7/5
                   \\ \text{\citet{khanuja2024image}} & = 0.112 \\
                   \text{\citet{ventura2023navigating}} & = 0.093 \\
                    \phi_{ITA}(c,r)          & = 0.022
                  \end{aligned}$
                \end{tabular}
            \\ \midrule
        \textbf{E4} & \begin{tabular}[c]{@{}c@{}}
            \includegraphics[width=0.55\columnwidth]{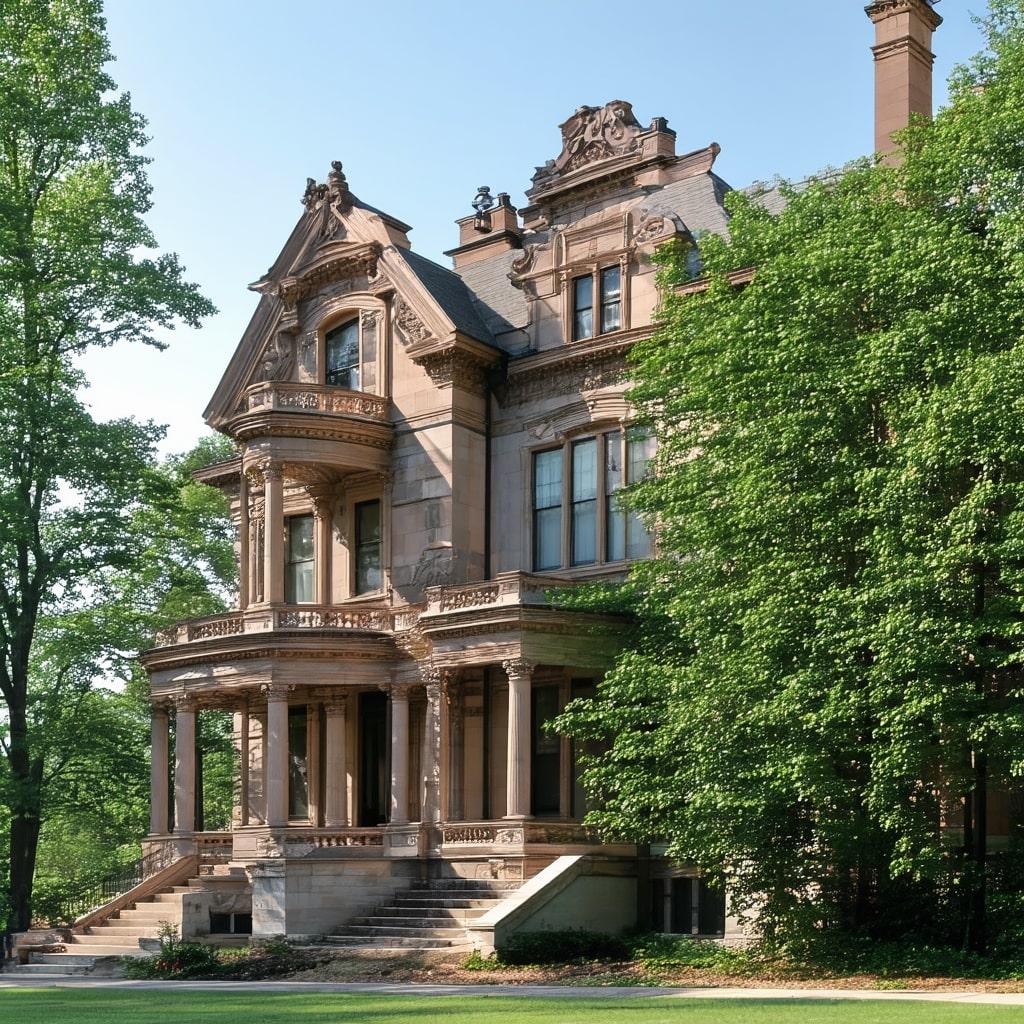}
            \end{tabular} 
            & \begin{tabular}[c]{@{}c@{}}\includegraphics[width=0.55\columnwidth]{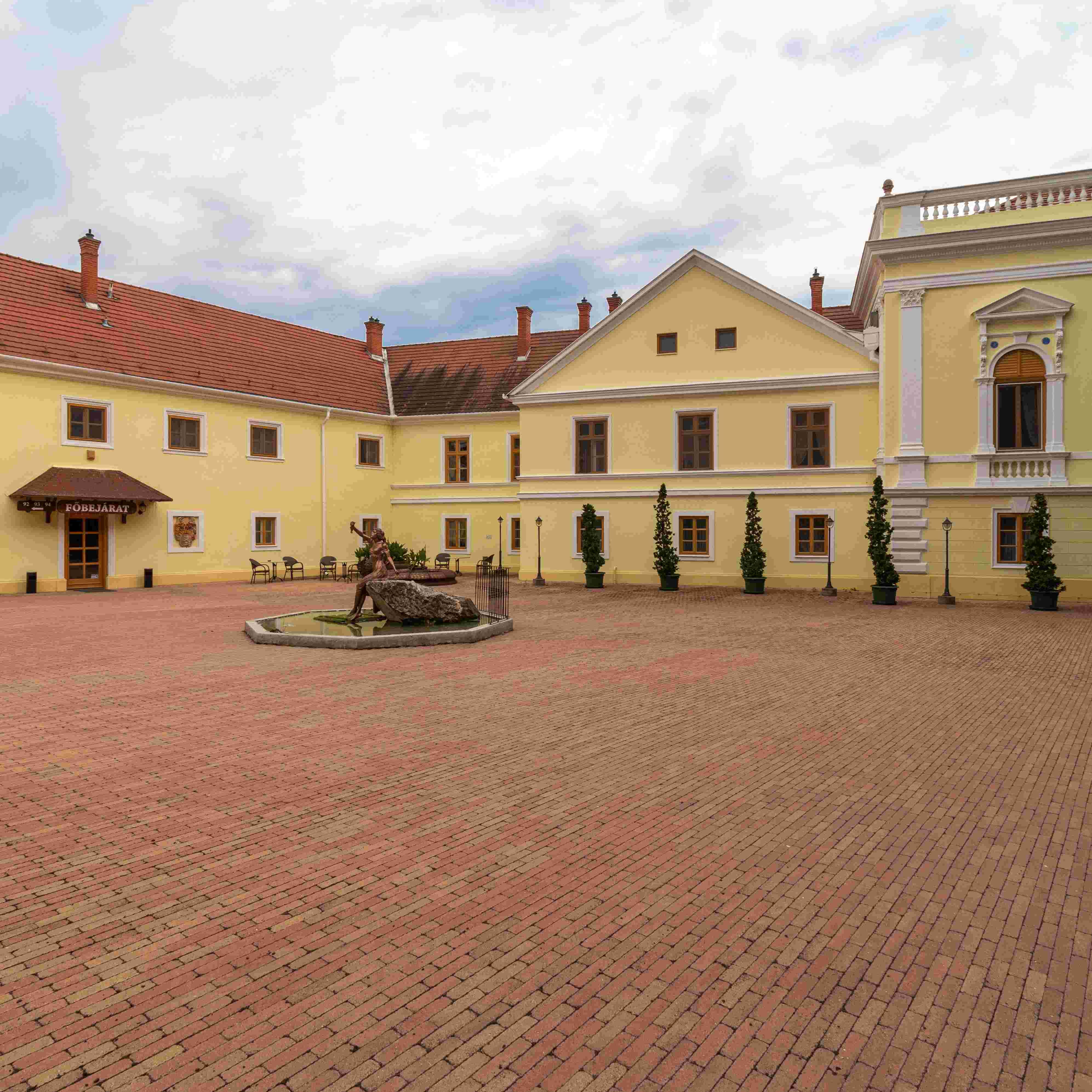}  \end{tabular} 
            & \begin{tabular}[c]{@{}l@{}}
                \textit{\textbf{Puchner Mansion}} \\ \\
                1. It's style is just all wrong. it looks more like something an \\ american would think of as a castle. can we stop trying to make ai \\ image gen happen? \\ \\
                2. I feel like the AI focused too much on the word Mansion, and \\ especially the american/older british kind.
            \end{tabular} 
            &   \begin{tabular}[c]{@{}c@{}}
                  $\begin{aligned}
                   \phi^*_{\cure} &= 1.3/5 \\
                   \text{\citet{khanuja2024image}} & = 0.040 \\
             \text{\citet{ventura2023navigating}} & = 0.053 \\
                    \phi_{ITA}(c,r)          & = 0.096
                  \end{aligned}$
                \end{tabular}
            \\ \midrule
        \textbf{E5} &\begin{tabular}[c]{@{}c@{}}
            \includegraphics[width=0.55\columnwidth]{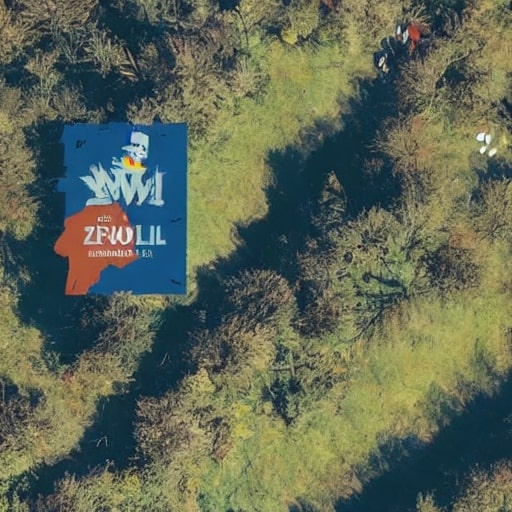}\end{tabular}
            & \begin{tabular}[c]{@{}c@{}}\includegraphics[width=0.55\columnwidth]{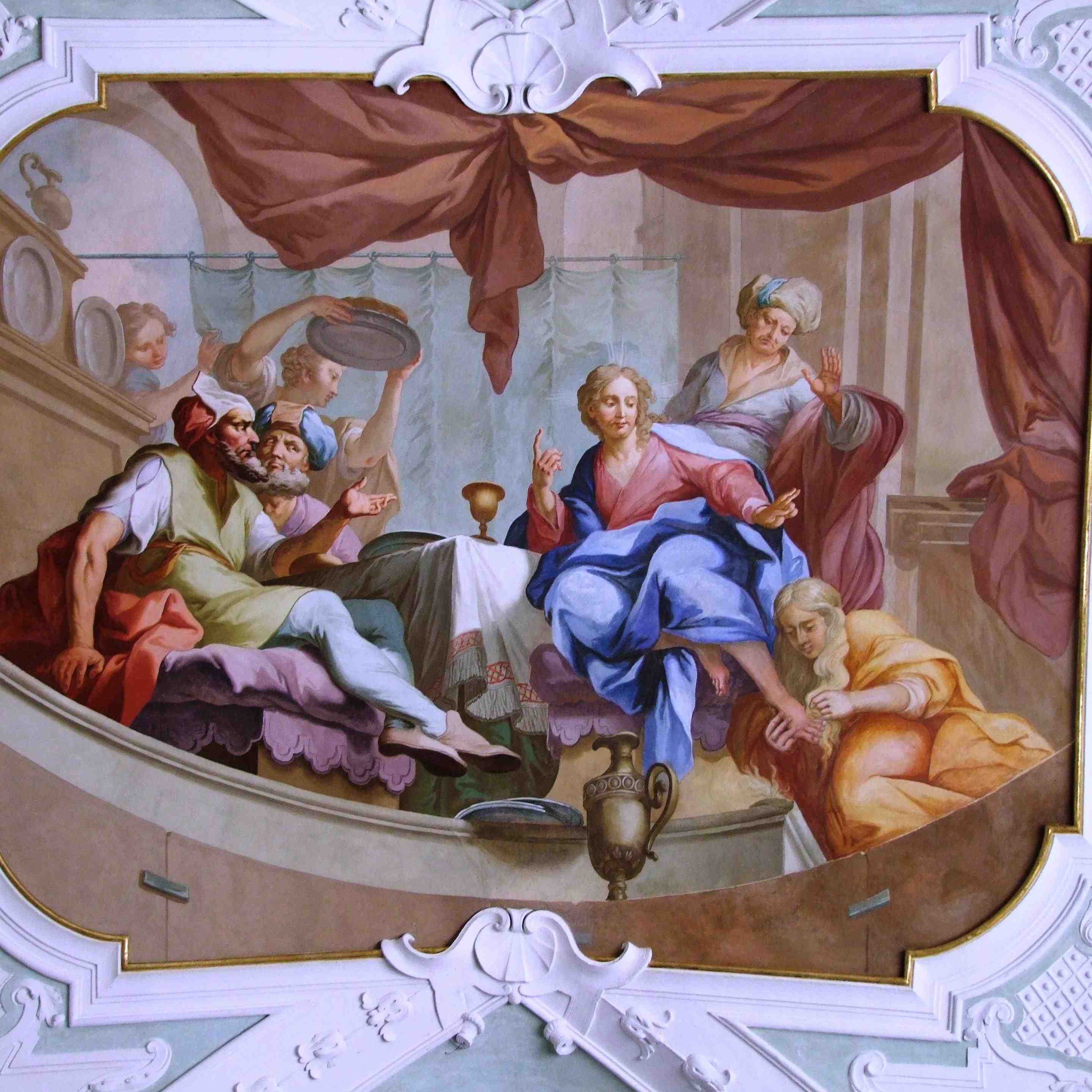}  \end{tabular} 
            & \begin{tabular}[c]{@{}l@{}}
                \textit{\textbf{Zwölf Glaubensartikel}} \\ \\
                1. Overall the vegitation seems possible to find in my country \\ \\
                2. This image could be a shot from above of some forest in my \\ country. \\
            \end{tabular} 
            &   \begin{tabular}[c]{@{}c@{}}
                  $\begin{aligned}
                   \phi^*_{\cure} &= 4.7/5 \\\text{\citet{khanuja2024image}} & = 0.108 \\
                   \text{\citet{ventura2023navigating}} & = 0.091 \\
                    \phi_{ITA}(c,r)          & = 0.079
                  \end{aligned}$
                \end{tabular}
            \\ \midrule
        \textbf{E6} & \begin{tabular}[c]{@{}c@{}}
            \includegraphics[width=0.55\columnwidth]{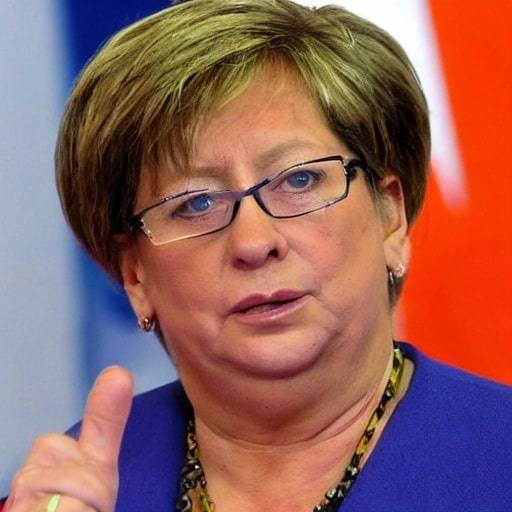}
             \end{tabular} 
            & \begin{tabular}[c]{@{}c@{}}\includegraphics[width=0.55\columnwidth]{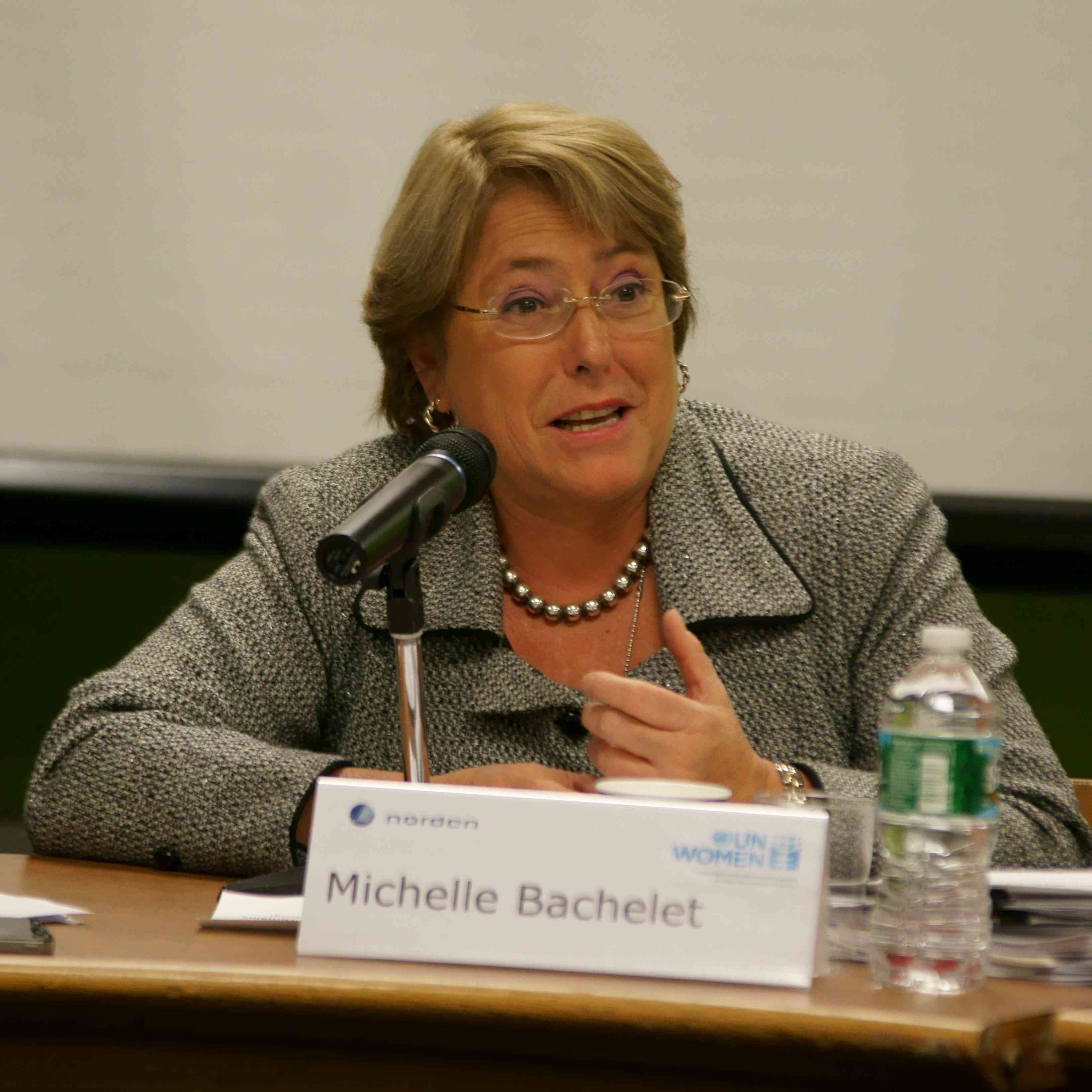} \end{tabular} 
            & \begin{tabular}[c]{@{}l@{}}
                \textit{\textbf{Michelle Bachelet}} \\ \\
                1. The image doesn't have the exact facial features of Michelle \\ Bachelet. She's a well known Chilean politician, so a lot of people \\ recognize her face. The image looks too fat and the hair is wrong. \\ \\
                2. Its the features of the face, they are not completly wrong, but \\ together they make a face who cant be taken to be Bachelet, also \\ the hair looks really fake.
            \end{tabular} 
            &   \begin{tabular}[c]{@{}c@{}}
                  $\begin{aligned}
                   \phi^*_{\cure} &= 2.0/5 \\
 \text{\citet{khanuja2024image}} & = 0.066 \\
                   \text{\citet{ventura2023navigating}} & = 0.054 \\
                    \phi_{ITA}(c,r)          & = 0.164
                  \end{aligned}$
                \end{tabular}
        \\ \bottomrule
    \end{tabular}
\end{adjustbox}
\caption{Visualization of quantitative Image-Text Alignment (ITA) scorers, user study \cure scores and text feedback. The first two artifacts were generated with FLUX.1 [dev], the second two generated with SD 3.5 Large, and the last two generated with SD 1.5.}
\label{fig:scorer_ITA_viz}
\end{figure*}

We highlight several qualitative examples on the \cure dataset of our ITA scorer $\phi_{ITA}(c,r)$ compared to baselines in \cref{fig:scorer_ITA_viz} using SigLIP 2 as the VLM backbone. We also show Likert scores for cultural representativeness $\phi^*_{\cure}$ and textual justification for each example.

Throughout our evaluation, we treat user judgments as the gold standard, assessing scorers based on how well they replicate human feedback. \textbf{E5} (\textit{Zwölf Glaubensartikel}) is a rare case where users were unfamiliar with the artifact itself, leading to a high $\phi^*_{\text{cure}}$ based primarily on regional similarity. Since the AI-generated and real images differ in category, the $\phi_{ITA}(c,r)$ score is correspondingly low, reflecting this mismatch.

As seen in \textbf{E1} (\textit{Bangles}), our proposed metric $\phi_{ITA}(c,r)$ aligns more closely with user preferences compared to existing baseline scorers. Our proposed metric proves particularly robust in scenarios where the T2I system generates outputs that are categorically incorrect. In \textbf{E3} (\textit{Jalangkote}), SD 3.5 Large generates an image of architecture rather than food, a failure undetected by baseline metrics that focus narrowly on regional resemblance. From a user perspective, representativeness encompasses not only regional cues but also correct category and item-level semantics, an area where our marginal information attribution scorer provides more robust signal.

We also highlight some failure cases of our ITA scorer. \textbf{E4} (\textit{Puchner Mansion}) shows a case where baseline scorers better approximated human judgments than $\phi_{ITA}(c,r)$, suggesting that in some cases, the baselines' focus on broader stylistic features may offer advantages. In \textbf{E6} (\textit{Michelle Bachelet}), our proposed $\phi_{ITA}(c,r)$ captures semantic representativeness by integrating category and region cues, but does not account for image quality. Users tend to penalize low-quality or unrealistic images regardless of semantic alignment, which our scorer overlooks. \textbf{E2} (\textit{Festival de la Primavera}) demonstrates a failure case across all ITA-based scorers which fail to capture culturally specific or context-dependent cues (region inconsistency, incorrect details of the parade).

\textbf{Note on Worker Reliability:}  
While most participants provided thoughtful and culturally grounded feedback, a few responses reflected exasperation with AI-generated outputs in general, including calls to ``stop this''. This highlights an important meta-issue regarding images created by T2I systems, \ie a  subjectivity and heterogeneity of opinions towards T2I systems themselves.

\clearpage

\clearpage
\section{Diversity}
\label{app:diversity}
We compute two notions of diversity: a) \textbf{intra-artifact}, over T2I generations with multiple random seeds for a single cultural artifact $n$; b) \textbf{intra-category}, over T2I generations of multiple random seeds for a single cultural category $c$. We discuss motivating examples for these two separate notions in \cref{subsec:scorer_diversity}). Details for seeding for different T2I systems is provided in \cref{app:img_gen_seeding}.

\paragraph{LPIPS.} A high LPIPS~\cite{zhang2018unreasonable} score indicates high \textbf{intra-artifact} variance in patchwise image features across seeds, which is interpreted as high diversity. We compute LPIPS for a category $c$ in our \cure dataset as an average of scores of all artifacts $n$ associated with category $c$, \ie
\[
LPIPS(c) = \frac{1}{|c|} \sum_{n \in c} \phi\bigl(I(n)\bigr).
\]

\paragraph{Vendi Scores.} Vendi Scores (VS)~\cite{friedman2023vendiscorediversityevaluation, kannen2024beyond} define a similarity measure via a kernel over selected attributes (\eg $r =$ \{Country, Continent\}). While computing VS, each seed $j$ of an image of a cultural category $I(c)$ is assigned a predicted label based on its ``closest" image in the set of cultural artifacts $n$ belonging to category $c$ to compute \textbf{intra-category} diversity, \ie
\[
\hat{n}(I(c)_j) = \arg \max_{n\in c} sim\bigl(I(c), I(n)\bigr).
\]

A $j\times j$ kernel similarity matrix is then computed based on a selected attribute (see Experimental Setup of ~\citet{kannen2024beyond} for details). The primary drawback of VS as a scorer is that its quality depends entirely on this initial assignment $\hat{n}$ and choice of kernel, which in turn depends on the image encoder used to compute $sim$.

\begin{figure*}[!t]
    \centering
    \begin{subfigure}[b]{0.3\textwidth}
        \includegraphics[width=\textwidth]{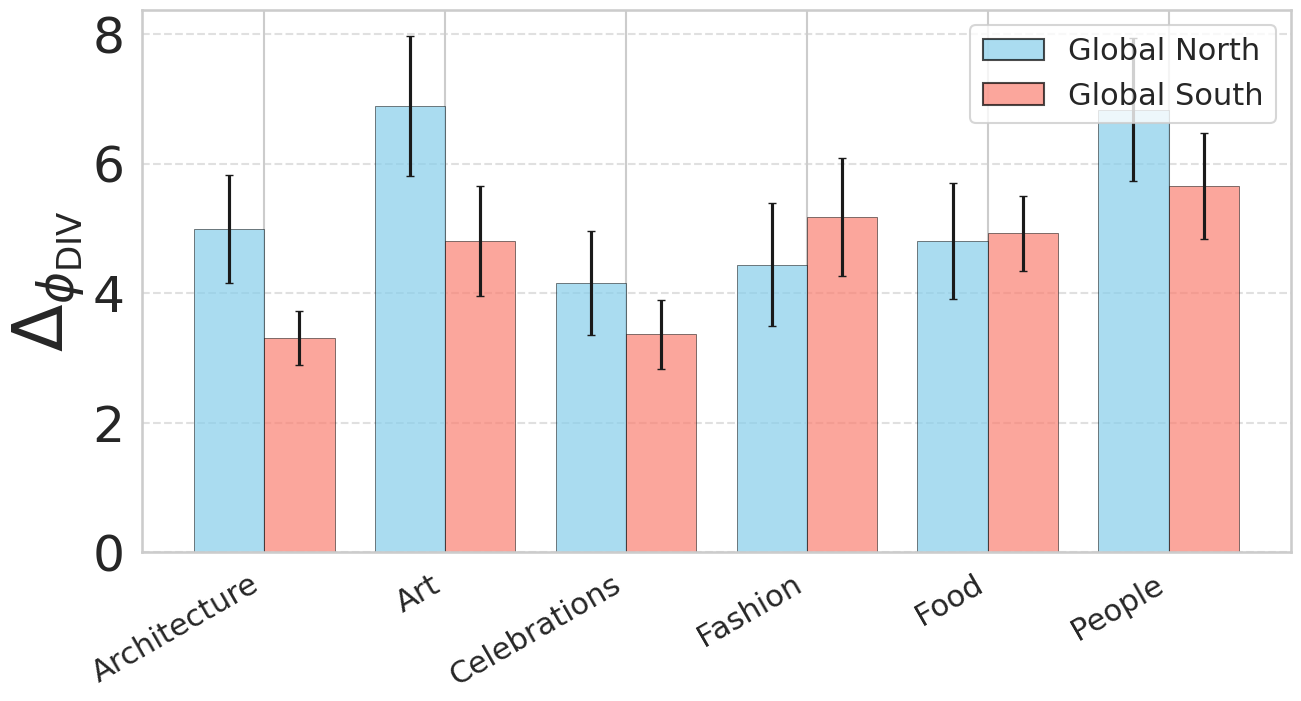}
        \caption{FLUX.1 [dev]}
    \end{subfigure}
    \hfill
    \begin{subfigure}[b]{0.3\textwidth}
        \includegraphics[width=\textwidth]{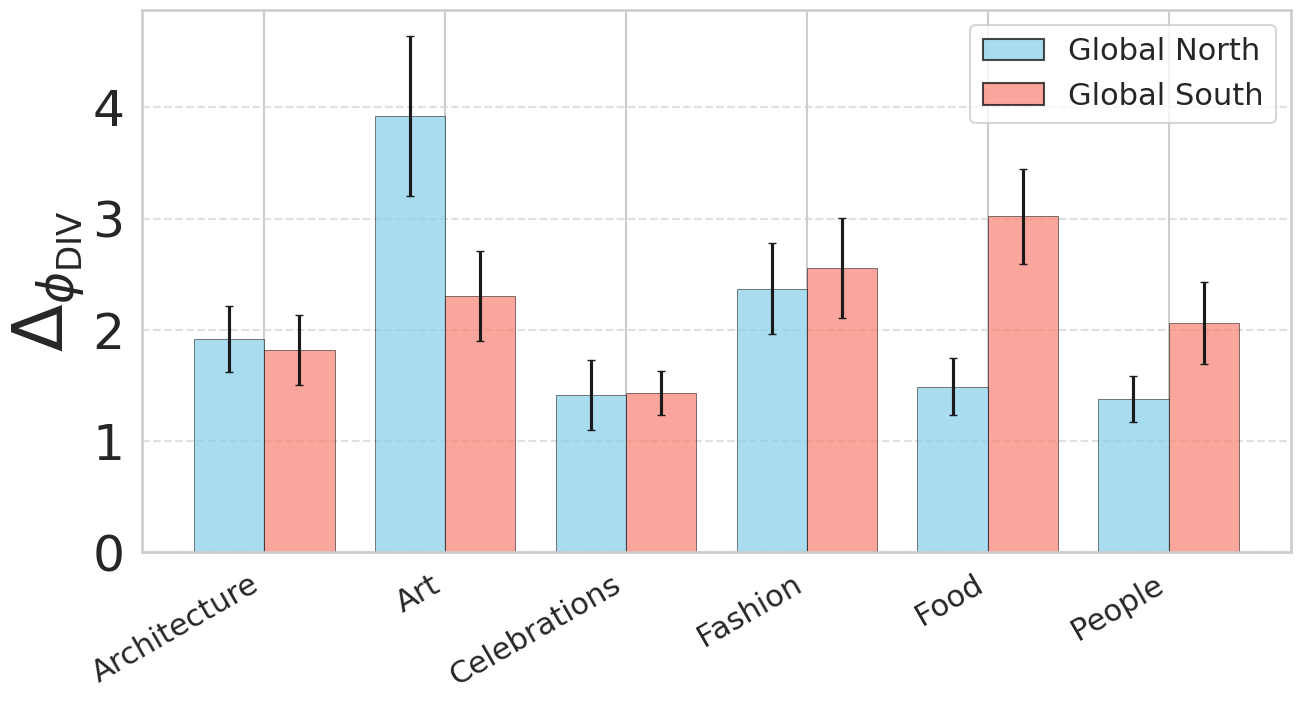}
        \caption{SD 1.5}
    \end{subfigure}
    \hfill
    \begin{subfigure}[b]{0.3\textwidth}
        \includegraphics[width=\textwidth]{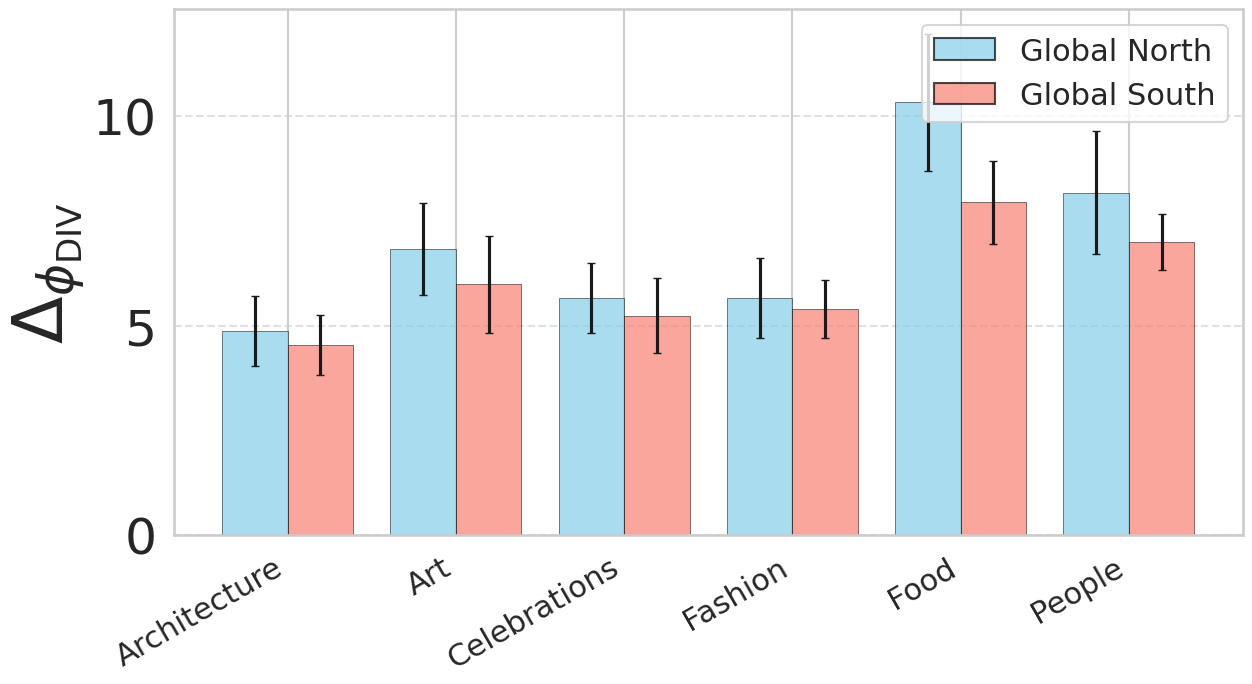}
        \caption{SD 3.5 Large}
    \end{subfigure}
    
    \vspace{1ex}
    
    \begin{subfigure}[b]{0.3\textwidth}
        \includegraphics[width=\textwidth]{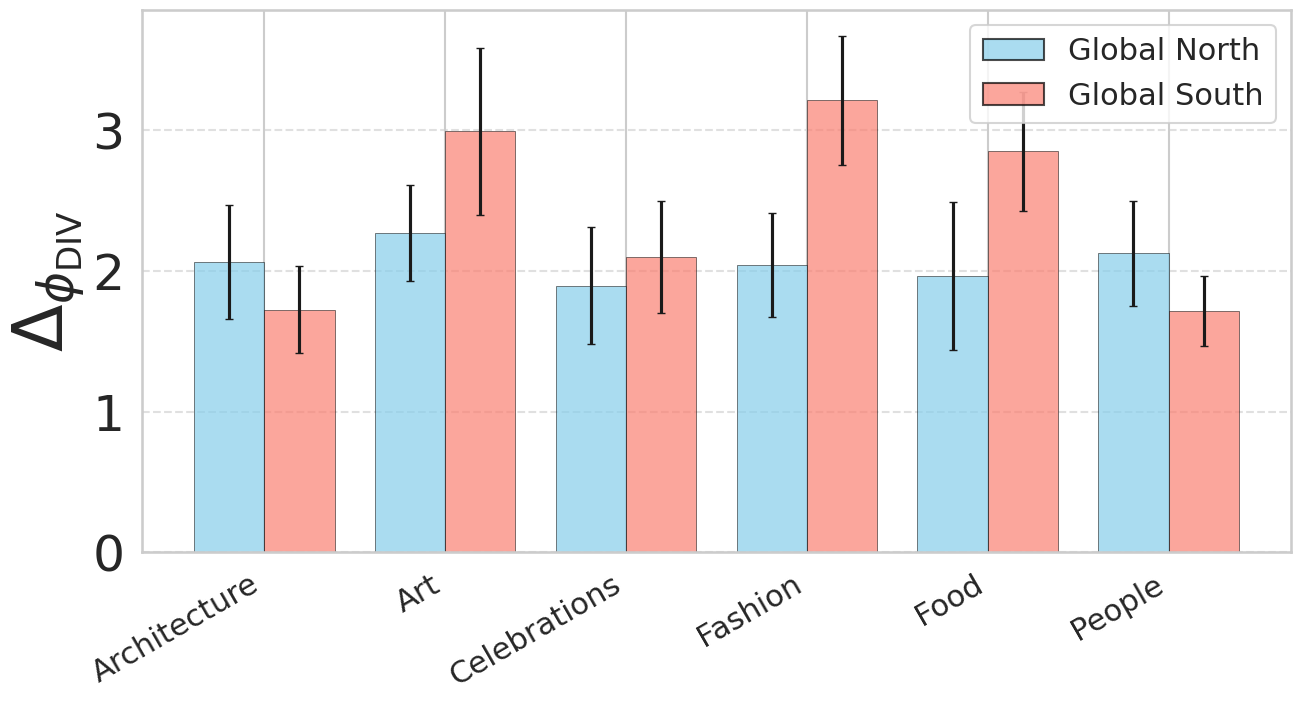}
        \caption{SDXL}
    \end{subfigure}
    \hfill
    \begin{subfigure}[b]{0.3\textwidth}
        \includegraphics[width=\textwidth]{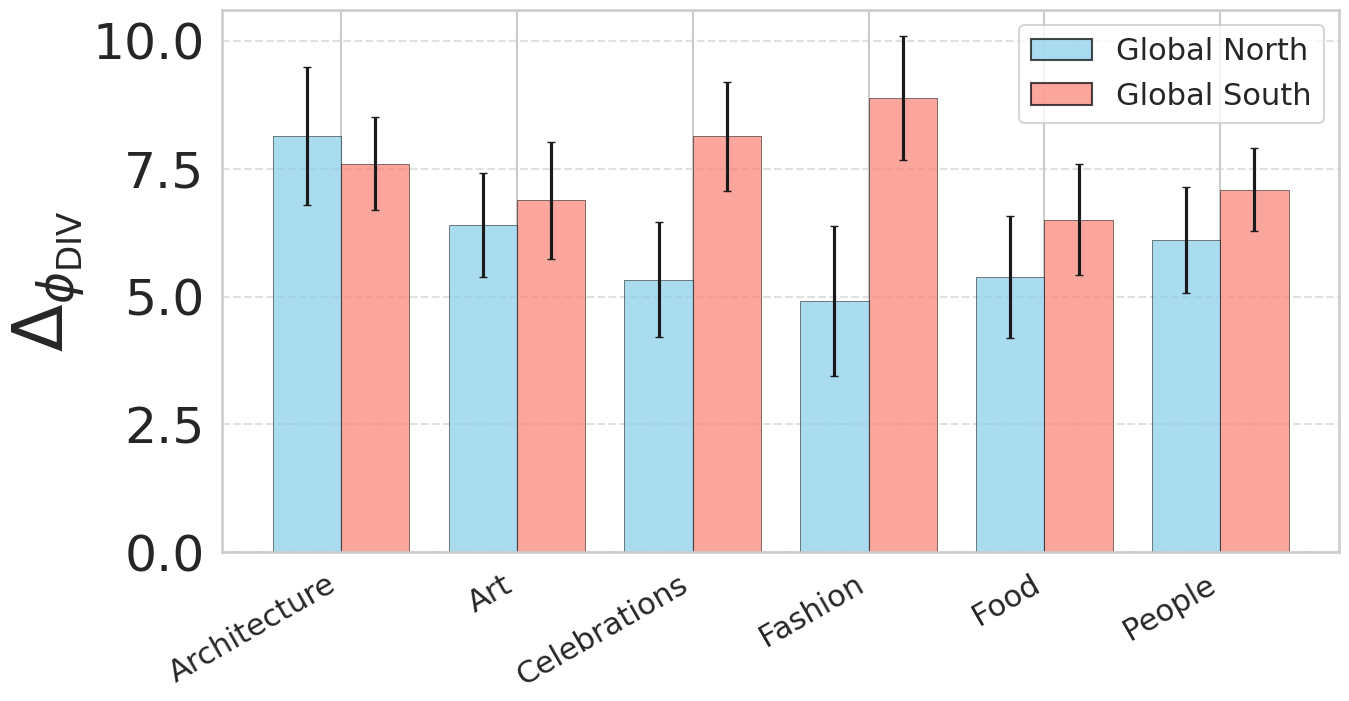}
        \caption{Ideogram 2.0}
    \end{subfigure}
    \hfill
    \begin{subfigure}[b]{0.3\textwidth}
        \includegraphics[width=\textwidth]{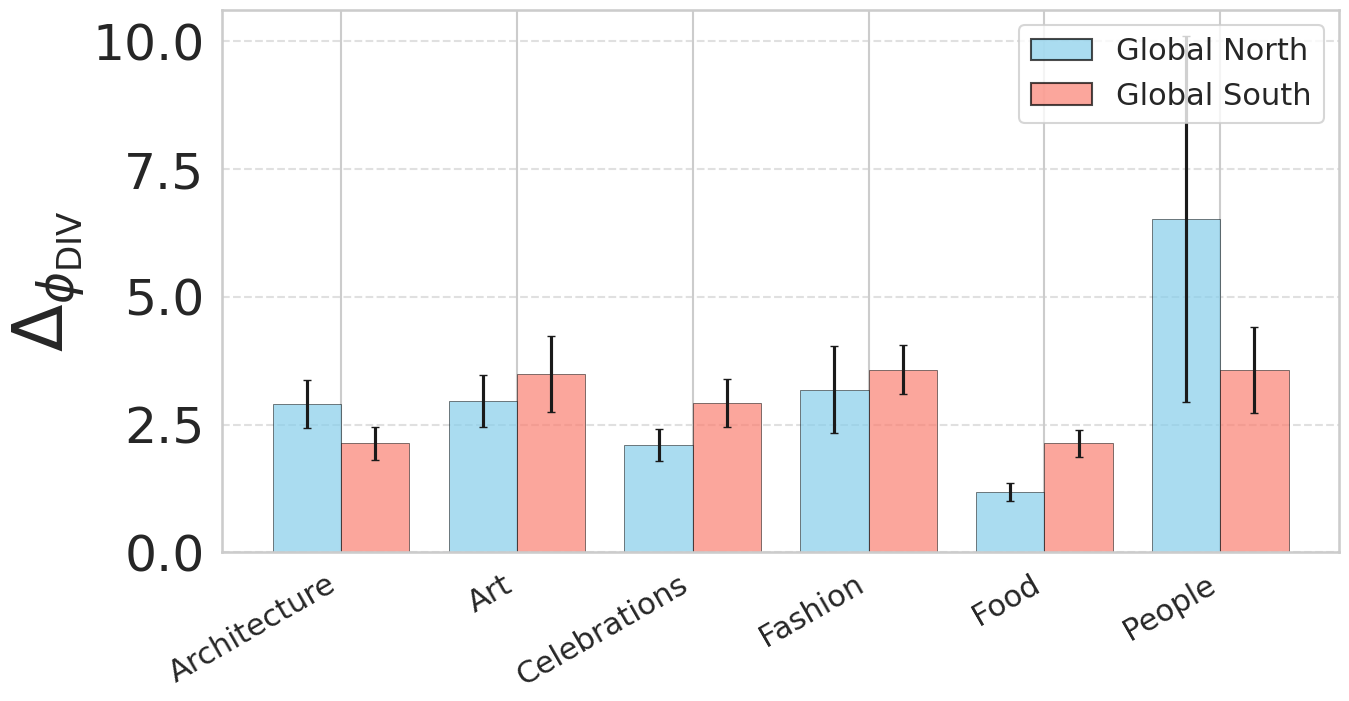}
        \caption{DALL-E 3}
    \end{subfigure}
    \caption{A bar plot of \(\Delta \)$\phi_{DIV}$ across all artifacts in the \cure dataset aggregated at the supercategory level (mean + std error bar) for each T2I system in our evaluation set. We separate each bar at the supercategory level into artifacts belonging to the Global North (GN) and Global South (GS) to examine if there is a noticeable discrepancy between scores in these clusters.}
    \label{fig:lpips_gn_gs}
\end{figure*}

\subsection{Diversity as a Long Tail Predictor}
\label{app:diversity_long_tail_pred}
We examine if we can use diversity as a predictor of a cultural artifact lying in the head or long tail of a T2I system's distribution over generated images, similar to our marginal attribution lens over perceptual similarity (\cref{subsec:scorer_percepsim}). Recall from \cref{eq:diversity} that our $\phi_{DIV}$ scorer compute LPIPS diversity over a set of images generated across attribute specification levels:
$$
\phi_{DIV} = LPIPS\bigl(\{n\}, \{n,c\}, \{n,r\}, \{n,c,r\}\bigr)
$$

We expect that in the long tail, a larger marginal increase in information $n\to a$ will cause a large increase in diversity, as there will be a larger visual difference between images generated with different attributes $I(n)\to I(a)$ (as seen with the ``Banku'' artifact in \cref{fig:teaser-cure-system}). Intuitively, this evaluates how much diversity changes when we mix images of ``Banku'' with images of ``Banku, a type of dumpling'', images of ``Banku, from Ghana'', and images of ``Banku, a type of dumpling from Ghana''. If it is relatively unchanged, we predict it to lie in the head of the supercategory artifact distribution.

To test this hypothesis, we compute the marginal increase in information as divergence of $\phi_{DIV}$ from an LPIPS over only images generated with $a = \{n\}$, \ie

$$
\Delta \phi_{DIV} = \phi_{DIV} - LPIPS(n)
$$

We visualize this marginal increase $\Delta \phi_{DIV}$ on the \cure dataset by grouping cultural artifacts into two buckets as proxies for the head and long tail: artifacts belonging to regions from the global north and global south. We average $\Delta \phi_{DIV}$ over all artifacts belonging to each bucket for each supercategory $s$, and show a bar plot for all T2I systems in \cref{fig:lpips_gn_gs}.

We observe two clusters of T2I sytem behavior with our scorer: in the first cluster,  FLUX.1 [dev] and SD 3.5 large (the highest quality T2I systems per Arena Bench ELO, \cref{tab:benchmark_results}) show higher change in diversity with increasing information in the Global South (proxy for long tail) than the Global North (proxy for head) across supercategories, which aligns with our hypothesis given our assumed proxies. In the second cluster, Ideogram 2.0, Stable Diffusion XL and DALL-E 3, this behavior is reversed, except for architecture and people for SDXL and DALL-E 3 (which has a high 66.5\% refusal rate for generating images of people due to inbuilt safety filters, see \cref{app:img_gen_refusal}). Stable Diffusion 1.5 is the weakest T2I system, and follows similar trends to the second cluster, except for Art. Architecture consistently has higher divergence in the Global North than the South, across T2I systems, which we suspect is due to the inherent lack of diversity in generations of specific named entities, \ie low intra-artifact variance (a similar problem to art and people).

In summary, our proposed scorer $\phi_{DIV}$ can thus predict where in its learned distribution an image generated by a T2I system lies, \ie in the head or long tail. We note our pre-assigned clustering of global north and south is just \textit{one example proxy} for head and long tail. We note while our scorer requires generating multiple seeds of images with different attribute specification levels (practically, four seeds across four styles = 16 total images), it is still relatively cheaper to compute when compared to computing artifact frequency over a large pretraining dataset via string matching, such as in \citet{parashar2024neglected}, which we discuss in more detail in \cref{app:concept_frequency}.

\subsection{Qualitative Analysis of DIV scorers}

\begin{figure*}[t]
\begin{adjustbox}{width=\textwidth}
    \begin{tabular}{@{}c| c c|l|c@{}}
        \toprule
        \textbf{E\#}      &\textbf{AI Image}      & \textbf{Real Image}        & \multicolumn{1}{c|}{\textbf{Feedback}}       & \textbf{Score}    \\ \midrule
        \textbf{E1} & \begin{tabular}[c]{@{}c@{}}
            \includegraphics[width=0.55\columnwidth]{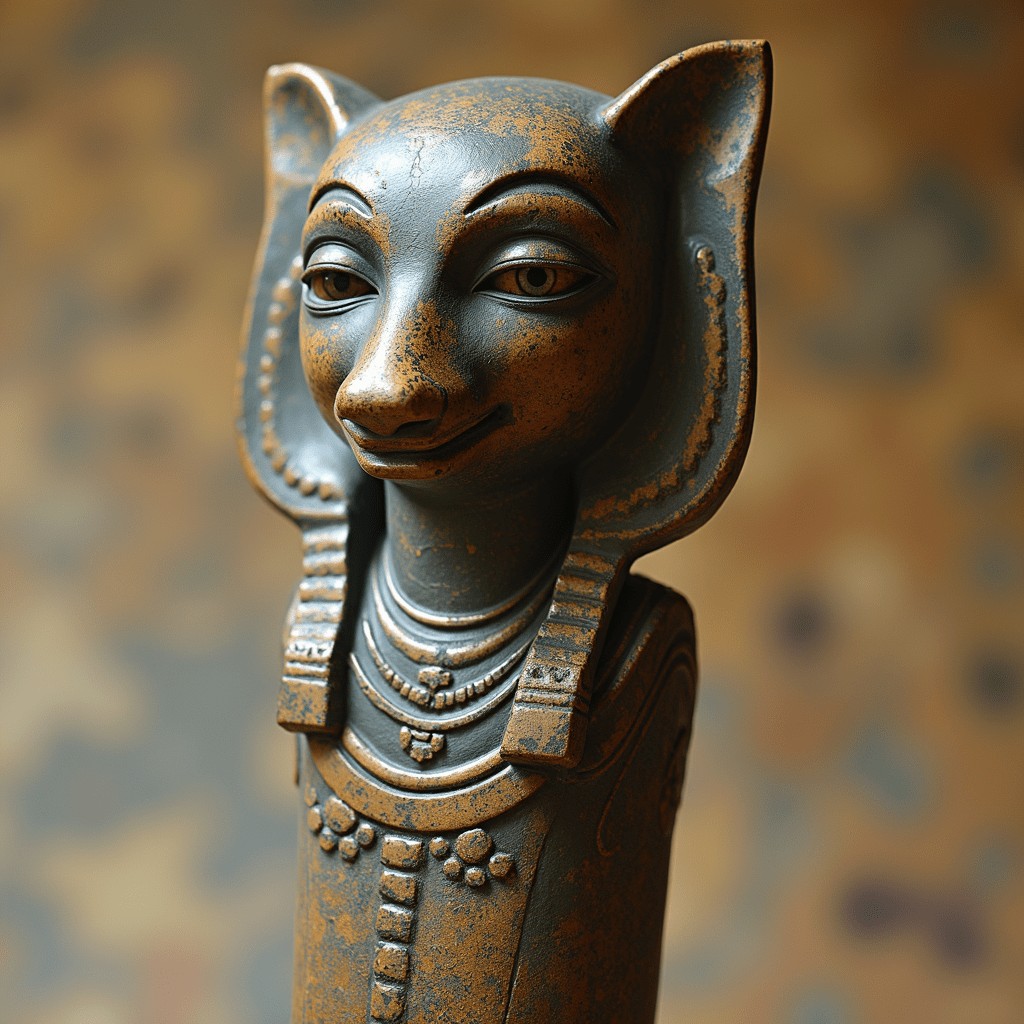} \end{tabular} 
             & \begin{tabular}[c]{@{}c@{}}\includegraphics[width=0.55\columnwidth]{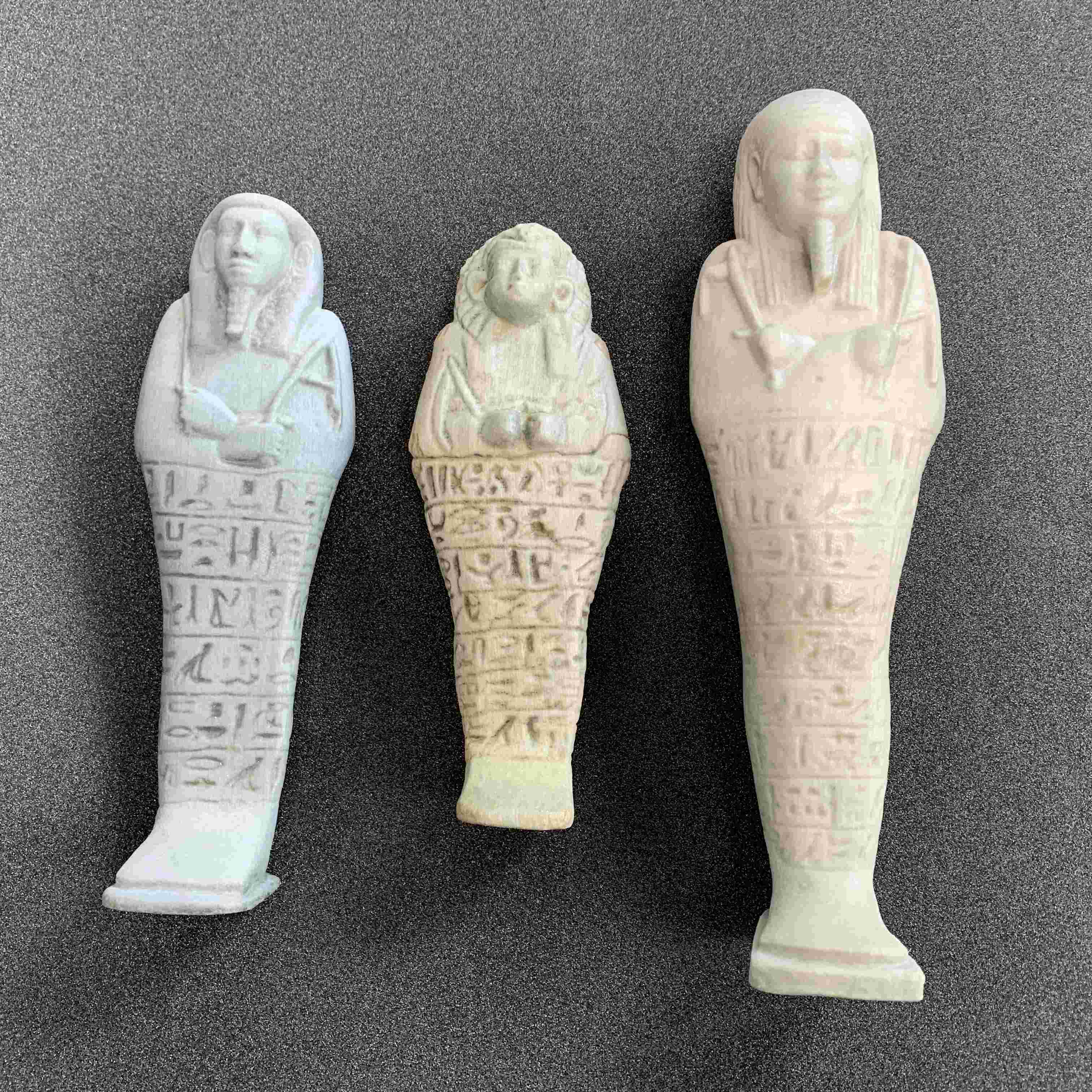}\end{tabular} 
            & \begin{tabular}[c]{@{}l@{}}
            \textit{\textbf{Ushabti}} \\ \\
                1. This image has used a different facial feature of the statue you \\ would not see in my country of culture. Although, the wear and tear \\ as well as the shape of the statue being mummy-like would be found \\in my country.\\ \\ 
                2. The similarities lies in the cat structure generally. The material it \\ is made of and the type of cloth it seems to be wearing. The part \\ where it is different from my culture is that the eyes are looking at the \\ side which is very unlikely. Moreover, the smile itself is so not true.\\ The big differences lie in the eyes direction and the smile.
            \end{tabular} 
            &   \begin{tabular}[c]{@{}c@{}}
                $\begin{aligned}
                    \phi^*_{\cure} &= 1/5 \\
                    LPIPS(n) &=  0.62\\ 
                    \phi_{DIV} &=   0.48 
                \end{aligned}$
                \end{tabular} 
            \\ \midrule
\textbf{E2}&\begin{tabular}[c]{@{}c@{}}
            \includegraphics[width=0.55\columnwidth]{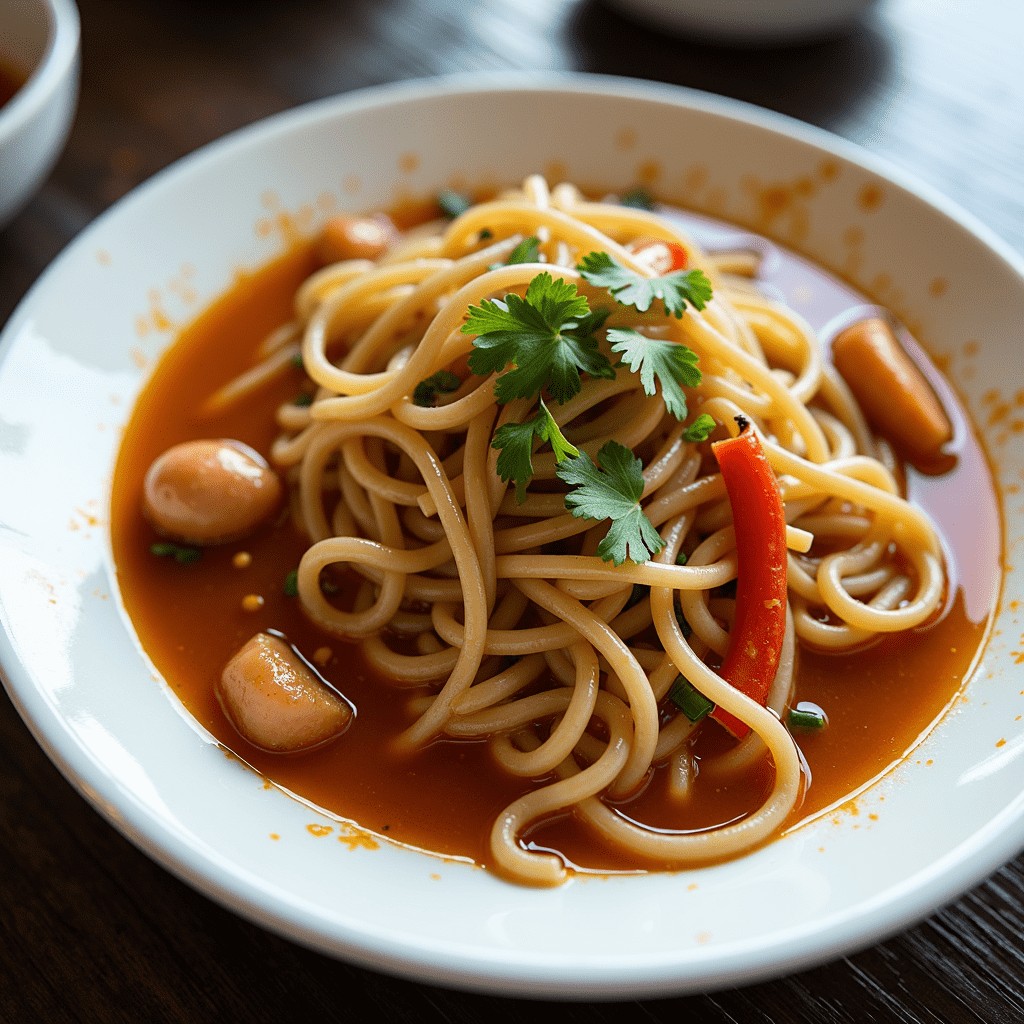} \end{tabular} 
            & \begin{tabular}[c]{@{}c@{}}\includegraphics[width=0.55\columnwidth]{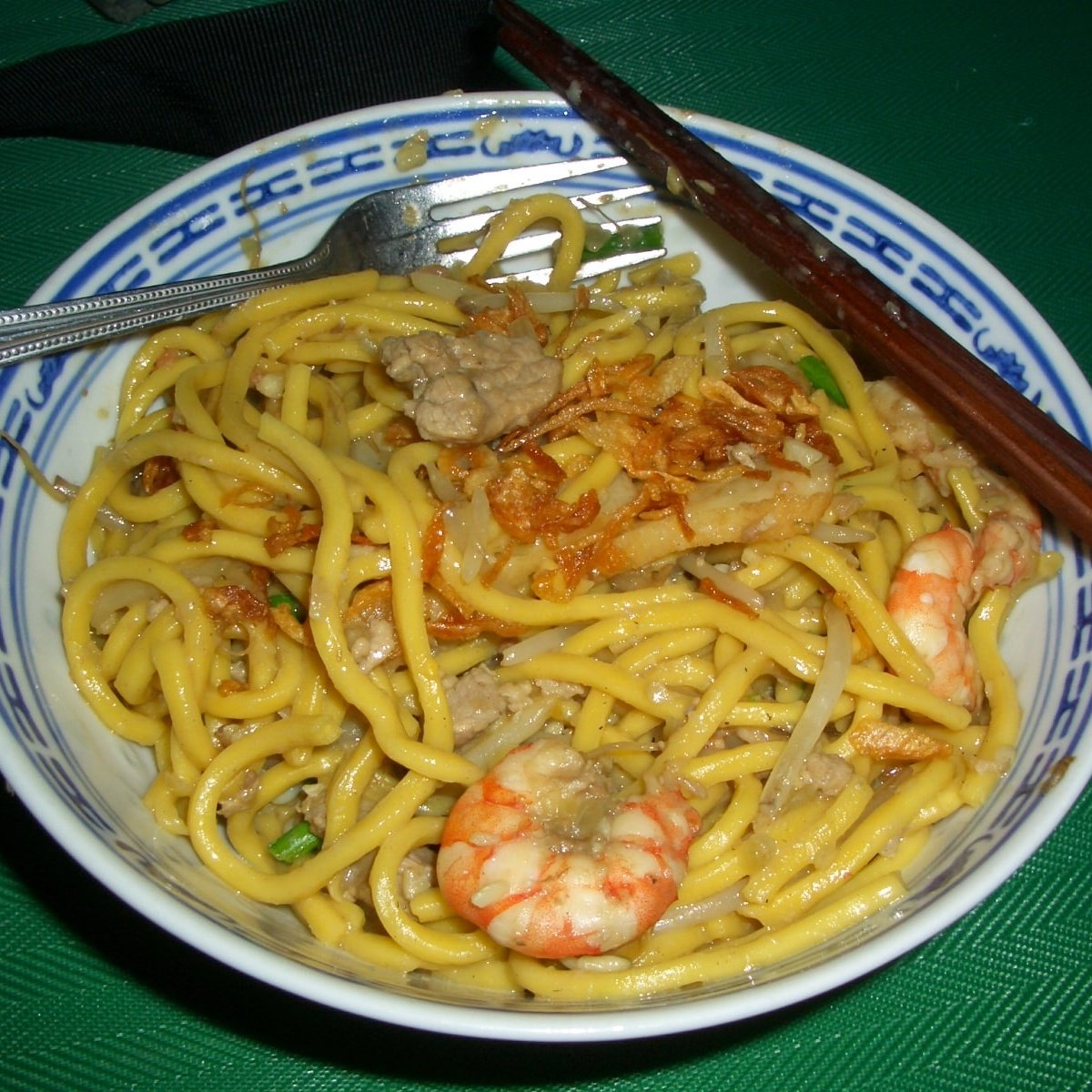}\end{tabular} 
            & \begin{tabular}[c]{@{}l@{}}
            \textit{\textbf{Hokkien Mee}} \\ \\
                1. Should have less liquid sauce. May need to add shrimps. The\\ noodle should be fried. \\ \\
                2. The noodles in the AI image looks weird and too smooth/plump\\ compared to real noodle dishes. Each individual strand can be traced\\ which seems unlikely in a real noodle image \\
            \end{tabular} 
            &  
             \begin{tabular}[c]{@{}c@{}}
                $\begin{aligned}
                    \phi^*_{\cure} &= 4/5 \\
                    LPIPS(n) &=  0.70\\ 
                    \phi_{DIV} &=   0.59 
                \end{aligned}$
                \end{tabular} 
            \\ \midrule
\textbf{E3}&\begin{tabular}[c]{@{}c@{}}
    \includegraphics[width=0.55\columnwidth]{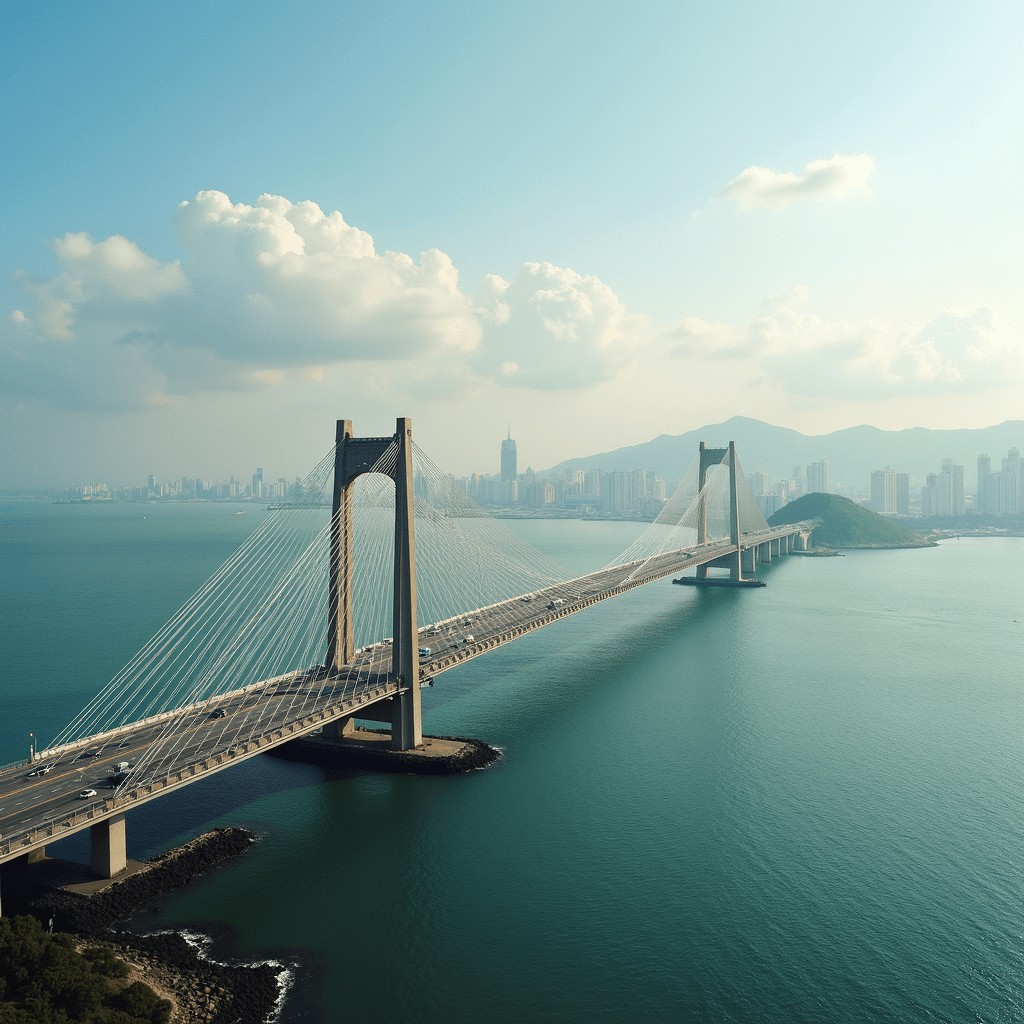}\end{tabular} 
    & \begin{tabular}[c]{@{}c@{}}\includegraphics[width=0.55\columnwidth]{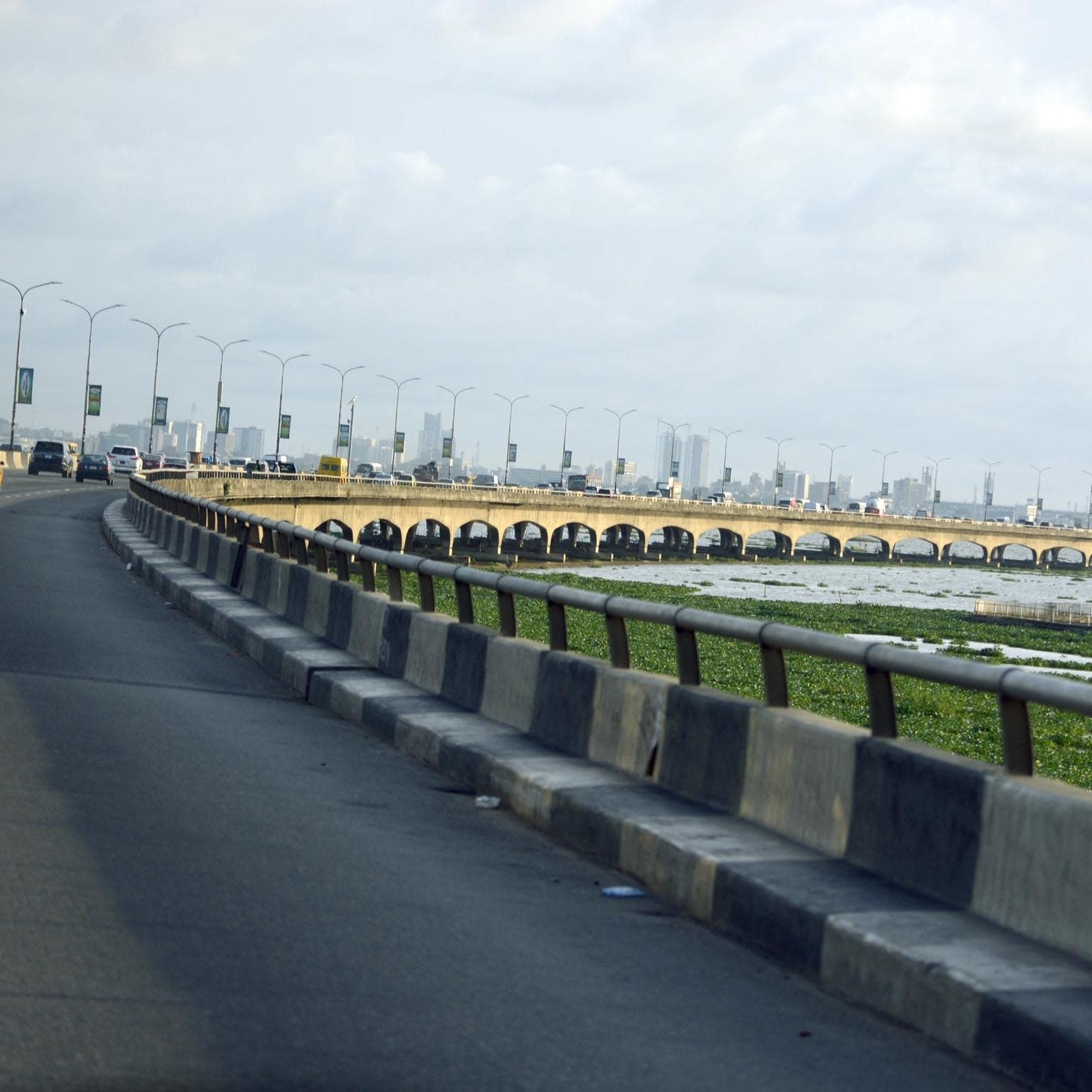}\end{tabular} 
    & \begin{tabular}[c]{@{}l@{}}
    \textit{\textbf{Third Mainland Bridge}} \\ \\ 
        1. The image contains the distinctive part of the real location and \\ image, it contains most of the details of the real thing.\\ \\
        2. The inaccurate details is The Number of Lanes. The image shows \\ more or fewer lanes than the actual four lanes. The image inaccurately\\ represents the bridge structural elements such. \\
    \end{tabular} 
    &   
        \begin{tabular}[c]{@{}c@{}}
                $\begin{aligned}
                    \phi^*_{\cure} &= 4.5/5 \\
                    LPIPS(n) &=  0.50\\ 
                    \phi_{DIV} &=   0.50 
                \end{aligned}$
                \end{tabular} 
        \\ \midrule
\textbf{E4}&\begin{tabular}[c]{@{}c@{}}
    \includegraphics[width=0.55\columnwidth]{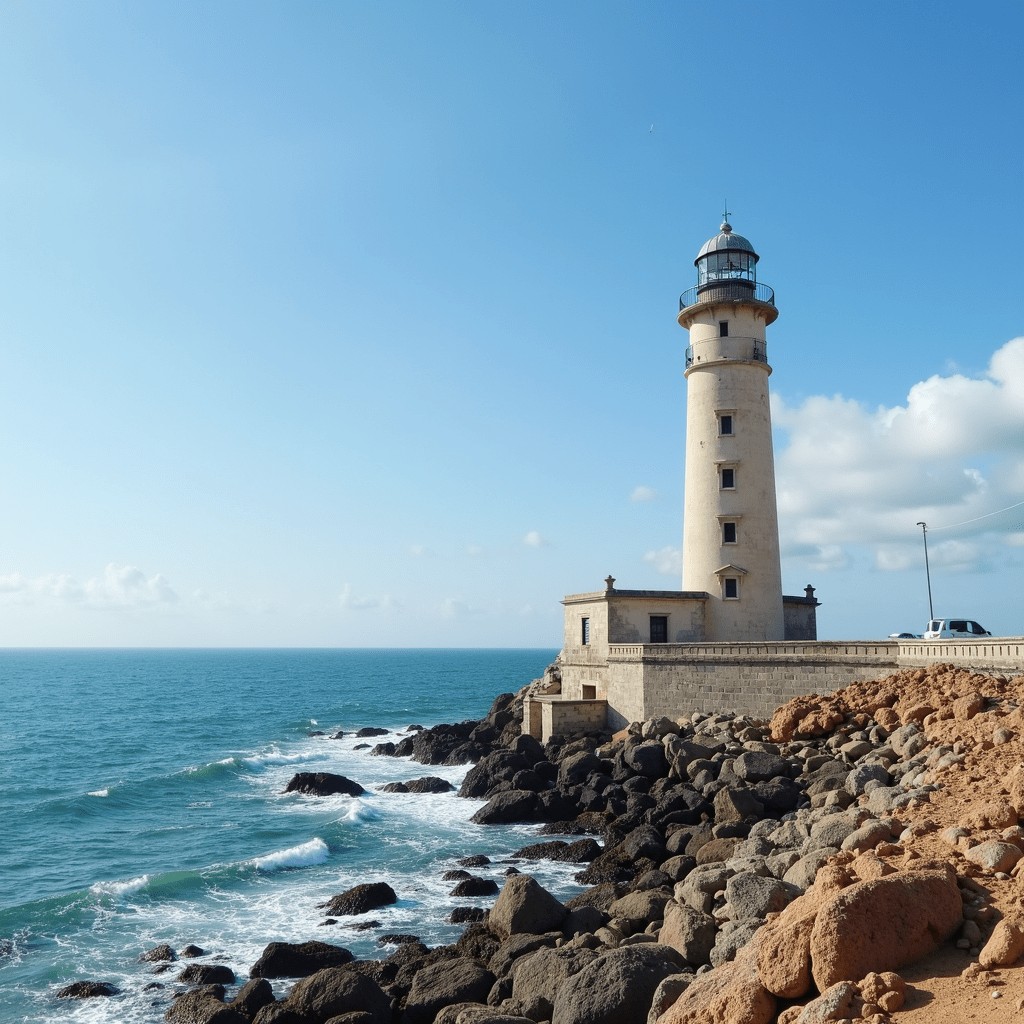}\end{tabular} 
    & \begin{tabular}[c]{@{}c@{}}\includegraphics[width=0.55\columnwidth]{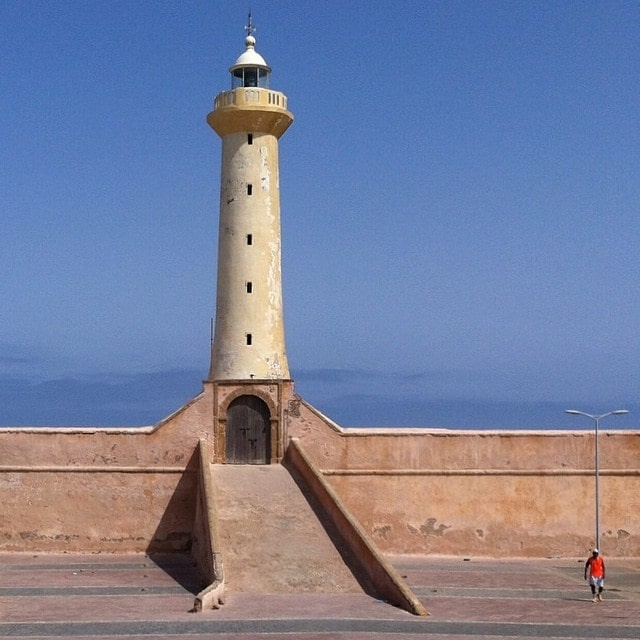}\end{tabular} 
    & \begin{tabular}[c]{@{}l@{}}
    \textit{\textbf{Rabat Lighthouse}} \\ \\
        1. The catholic or christian cross \\ \\
        2. I think gen AI did a good job here, but missed some details like the\\ big square in front of the lighthouse and the window's sizes are\\ somewhat big here \\
    \end{tabular} 
    &   
        \begin{tabular}[c]{@{}c@{}}
                $\begin{aligned}
                    \phi^*_{\cure} &= 5/5 \\
                    LPIPS(n) &=  0.51\\ 
                    \phi_{DIV} &=   0.54 
                \end{aligned}$
                \end{tabular} 
        \\ \midrule
\textbf{E5}&\begin{tabular}[c]{@{}c@{}}
    \includegraphics[width=0.55\columnwidth]{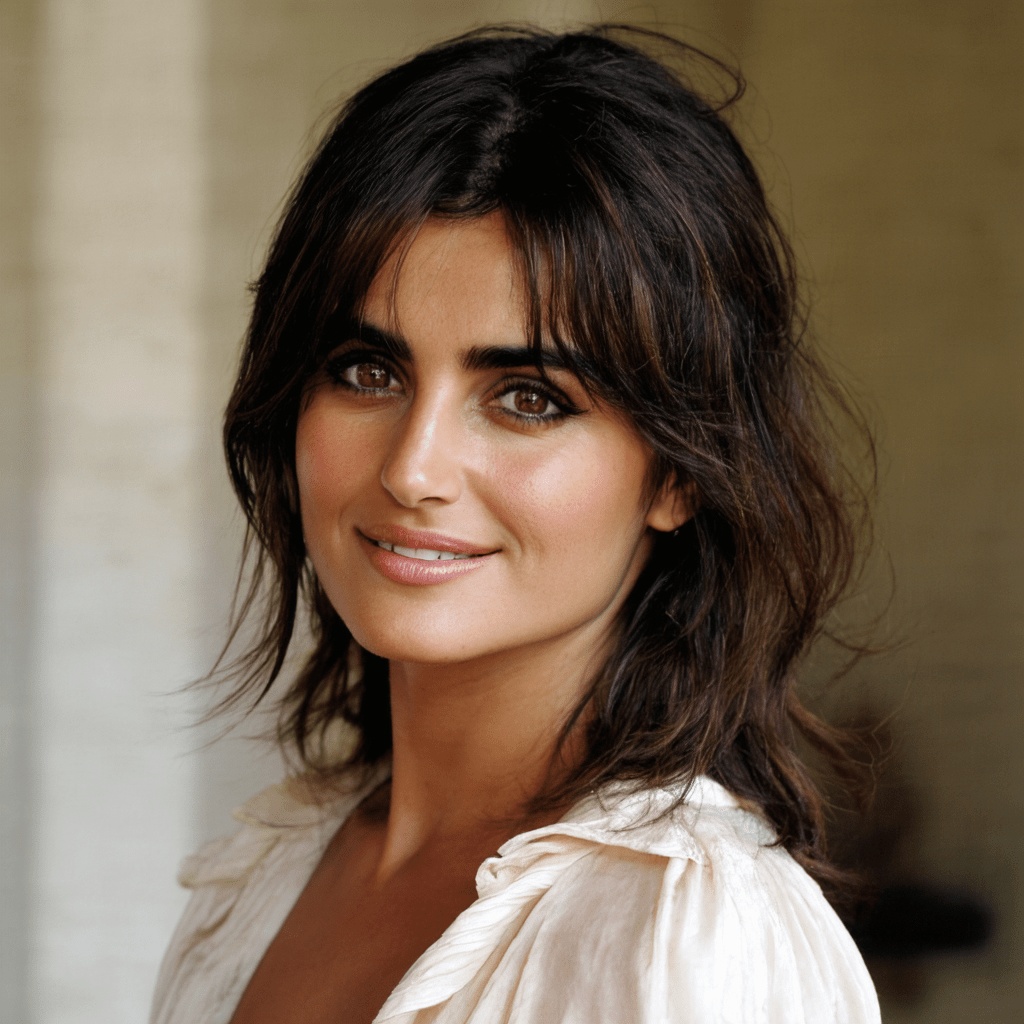}\end{tabular} 
    & \begin{tabular}[c]{@{}c@{}}\includegraphics[width=0.55\columnwidth]{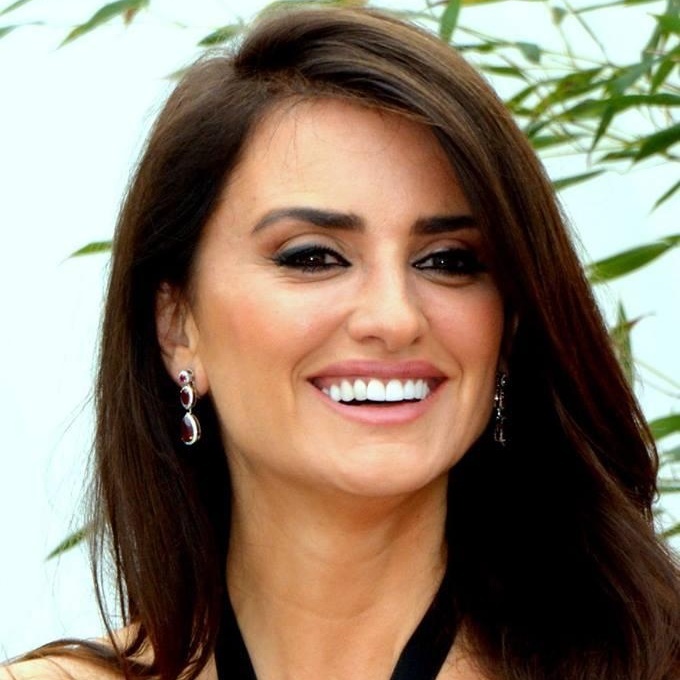}\end{tabular} 
    & \begin{tabular}[c]{@{}l@{}}
    \textit{\textbf{Penelope Cruz}} \\ \\
        1. The facial features are very similar to the real ones \\ \\
        2. The hair is typical of any country, so it could as well be spanish.\\ The white clothes is typical mediterranean. \\
    \end{tabular} 
    &   
                \begin{tabular}[c]{@{}c@{}}
                $\begin{aligned}
                    \phi^*_{\cure} &= 4/5 \\
                    LPIPS(n) &=  0.74\\ 
                    \phi_{DIV} &=   0.62 
                \end{aligned}$
                \end{tabular}
        
        \\ \midrule
\textbf{E6}&\begin{tabular}[c]{@{}c@{}}
    \includegraphics[width=0.55\columnwidth]{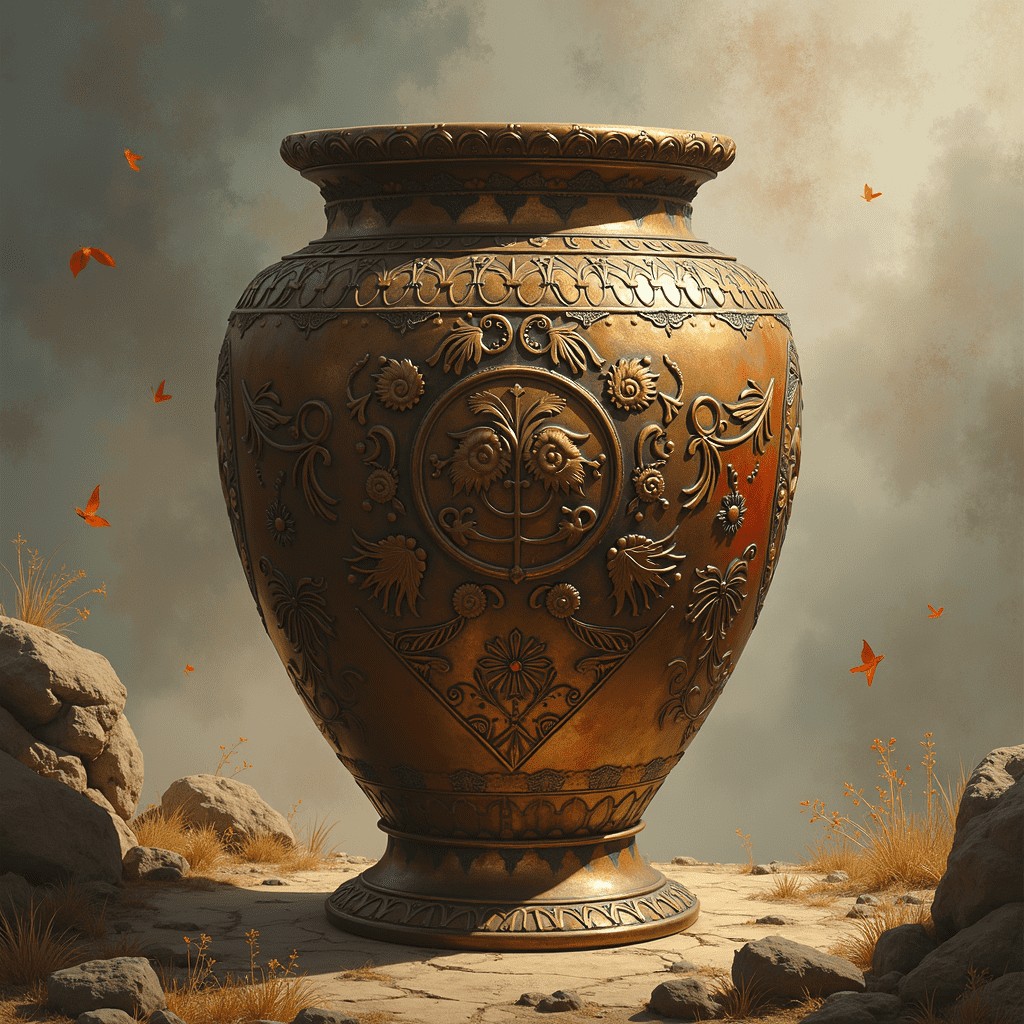}\end{tabular} 
    & \begin{tabular}[c]{@{}c@{}}\includegraphics[width=0.55\columnwidth]{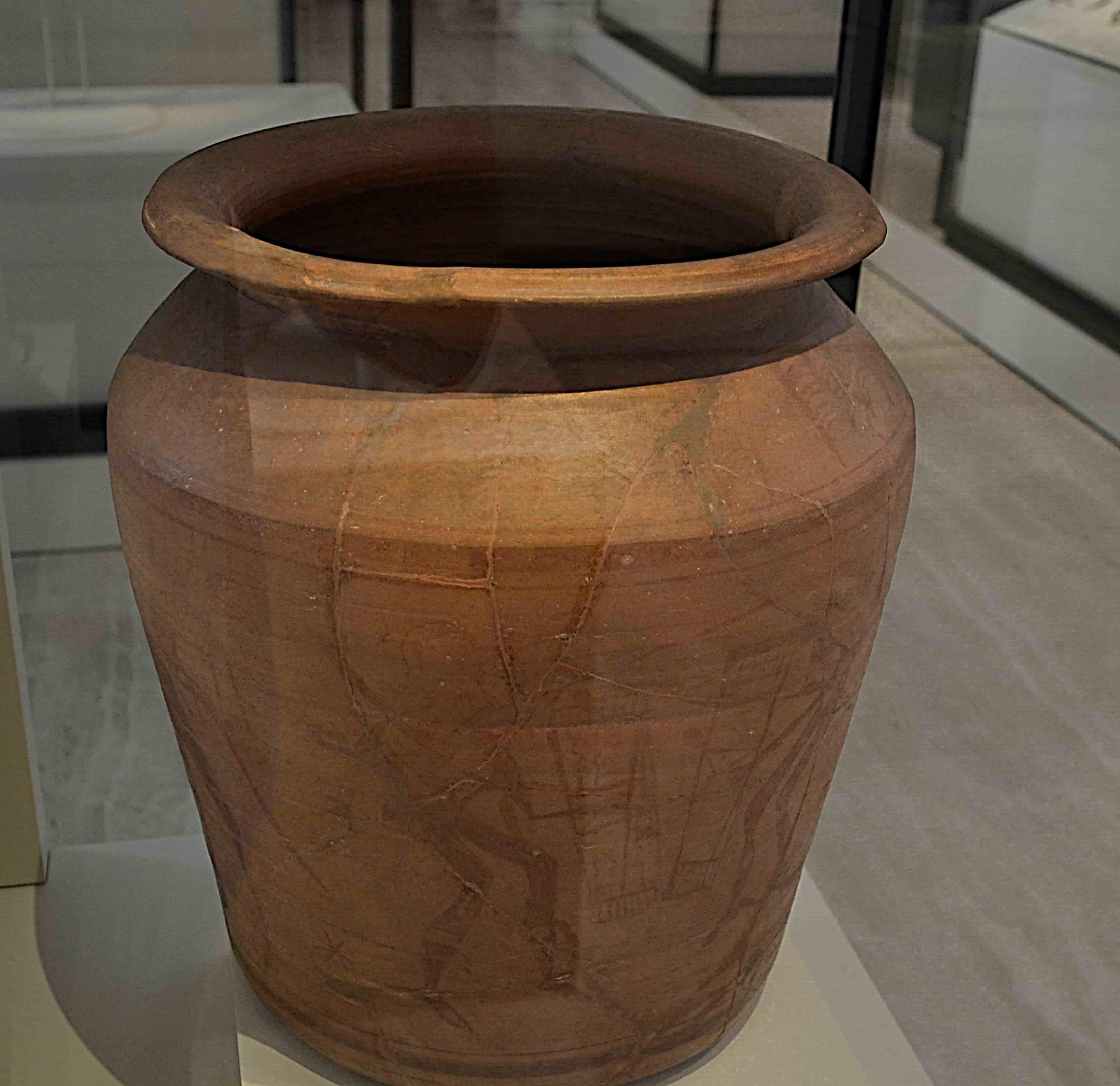}\end{tabular} 
    & \begin{tabular}[c]{@{}l@{}}
    \textit{\textbf{Vaso de los Guerrero}} \\ \\
        1. The AI-generated image looks too unreal; it doesn’t feel natural, as\\ if it was too exaggerated \\ \\
        2. The image has floral filigrees and not warriors. The decorations are\\ in relief and not painted. In addition, the vessel is lacquered with gloss \\
    \end{tabular} 
    &   
        \begin{tabular}[c]{@{}c@{}}
                $\begin{aligned}
                    \phi^*_{\cure} &= 2/5 \\
                    LPIPS(n) &=  0.75\\ 
                    \phi_{DIV} &=   0.55 
                \end{aligned}$
                \end{tabular}
        \\ \bottomrule
    \end{tabular}
\end{adjustbox}
\caption{Diversity scorer and user feedback}
\label{fig:scorer_viz_div}
\end{figure*}

We highlight several qualitative examples on the \cure dataset of our DIV scorer $\phi_{DIV}$ compared to baselines in \cref{fig:scorer_viz_div}.
In alignment with our quantitative observations in \cref{app:diversity_long_tail_pred}, in the examples \textbf{E2} (``\textit{Hokkien Mee}''), \textbf{E3} (``\textit{Third Mainland Bridge}''), \textbf{E4} (``\textit{Rabat Lighthouse}'') and \textbf{E5} (``\textit{Penelope Cruz}''), all diversity scorers show an inverse relationship to human judgments of cultural representativeness $\phi^*_{\cure}$. These examples have high human scores for cultural representativeness $phi^*_\cure$ and show very little change in diversity score from LPIPS(n) to $\phi_DIV$, indicating that attribute specification does not increase diversity significantly and the artifact likely lies in the head of the T2I system distribution, which aligns with our hypothesis in \cref{subsec:scorer_diversity}. We highlight failure cases where the baseline $LPIPS(n)$ outperforms our scorer $\phi_{DIV}$ in \textbf{E1} (``\textit{Ushabti}'') and \textbf{E6} (``\textit{Vaso de los Guerrero}''), where for a low $\phi^*_{\cure}$, the baseline shows a higher diversity score and accurately matches the inverse relationship. We suspect that the factuality-diversity tax especially hurts handicrafts like pottery, where coarse semantic features may be accurate (shape, structural outline) but finer details are wrong (\eg facial features and material in \textbf{E1} and material and textures in \textbf{E6}.)

\clearpage

\section{MLLM as a Judge}
\label{app:mllm_judge}

To our knowledge for the first time, we evaluate a state-of-the-art multimodal large language model (MLLM) for its ability to emulate human judgments (Likert scores) of \cure and perceptual similarity to ground-truth and its ability to provide accurate textual justifications for its scores in \cref{app:Gemini 2.0 Flash_prompts}. We choose the recent Gemini 2.0 Flash~\cite{gemini}, a popular yet efficient proprietary state-of-the-art MLLM. We discuss qualitative examples of Gemini 2.0 Flash scores and justifications on artifacts from the \cure dataset and highlight several shortcomings and areas for improvement in \cref{subfig:Gemini 2.0 Flash_failures_part2}.

\begin{tcolorbox}[
    colback=green!5,          
    colframe=green!75!black,  
    boxrule=1pt,              
    arc=3pt,                  
    width=\columnwidth        
]

\textbf{Gemini Prompt for Perceptual Similarity:} \vspace{0.3cm} \\ 
\rmfamily
You will be shown two images below. The image on the left is an AI-generated image of \{item\_name\} (\{category\_name\}). The image on the right is a grid of four real images of \{item\_name\} from Wikipedia.

Generated Image (Left):\\ 
{[}Image 1 Placeholder - will be inserted by API call{]}

Real Images Collage (Right):\\ 
{[}Image 2 Placeholder - will be inserted by API call{]}

Please analyze the images and answer the following:

1.\hspace*{1ex}On a scale of 1-5, how similar do you think the generated image on the left is to the real images on the right? Use the following scale:\\ 
\hspace*{3ex}1: Not at all similar\\ 
\hspace*{3ex}2: Slightly similar\\ 
\hspace*{3ex}3: Reasonably Similar\\ 
\hspace*{3ex}4: Very Similar\\ 
\hspace*{3ex}5: Extremely Similar

2.\hspace*{1ex}Provide a brief explanation for your similarity rating, detailing specific visual aspects you compared.

Format your response strictly as JSON with the keys {"similarity\_rating"} (for the number 1-5) and {"similarity\_explanation"} (for the text explanation). \vspace{0.3cm}

\textbf{Example Response from Gemini:} \\
\{ \\
\hspace*{2ex}{"similarity\_rating"}: 4,\\ 
\hspace*{2ex}{"similarity\_explanation"}: {"The generated image captures the overall shape and key architectural elements seen in the real images, like the arches and towers, but lacks some finer texture details and the surrounding environment differs."} \\ 
\}
\end{tcolorbox}

\subsection{Gemini 2.0 Flash as a Scorer} 
\label{app:Gemini 2.0 Flash_prompts}

As Gemini 2.0 Flash is a multimodal LLM, it can be queried for complex and nuanced visual-question answering (VQA), such as directly querying for cultural representativeness and perceptual similarity scores. We thus replicate our user study setup and questionnaire structure (\cref{app:user_study}) with Gemini.

\paragraph{Perceptual Similarity.} Similar to our user study setup in \cref{fig:userstudy_ui_percep_sim} (\cref{app:user_study_ps}), we query Gemini for perceptual similarity (\textbf{Q1a}) over all artifacts in the \cure dataset by showing Gemini one image generated by our T2I system, and a grid of four ground-truth images. Both images are passed in as URLs via the Google Gemini API\footnote{\url{https://ai.google.dev/gemini-api/docs/models\#gemini-2.0-flash}}. Mirroring our user study, we prompt Gemini to score the similarity of the T2I generated image to the ground-truth image grid from from 1 (``Not at all similar'') to 5 (``Extremely similar''). We also examine if Gemini can correctly highlight which details of the T2I image it thought were correct or incorrect and justify its score through text. We show the exact prompt we use below:

\paragraph{\cure and Ground-Truth Likelihood.}
Similar to our user study setup in \cref{fig:us_cure} (\cref{app:user_study_cure}), we ask Gemini to rate cultural representativeness (\textbf{Q3a}) and likelihood of the T2I image belonging to the class of artifact $a$, which we call ground-truth likelihood (\textbf{Q3b}). Ground-truth likelihood is very similar to perceptual similarity in what it is trying to measure (faithfulness / factuality), but unlike PS, Gemini must score GT likelihood without seeing any ground-truth images of artifact $a$. Similar to PS, we ask Gemini to justify its scores through text. We show the exact prompt below:

\begin{tcolorbox}[colback=gray!8, colframe=black, boxrule=1pt, arc=1.5pt, width=\columnwidth ]
\rmfamily
\textbf{Prompt for \cure and GT Likelihood:}\vspace{0.3cm}

This is an AI-generated image of \{item\_name\}, which is a type of \{category\_name\}.  
Please analyze this image and answer the following questions:

1. Cultural Representativeness: On a scale of 1-5 (where 1 is "Highly Unlikely" and 5 is "Extremely Likely"),  
   how likely can this item be found in \{country\}?

2. Ground-Truth Likelihood: On a scale of 1-5 (where 1 is "Highly Unlikely" and 5 is "Extremely Likely"),  
   how likely is this an accurate image of \{item\_name\}?

3. Description: What specific details in the AI-generated image make it accurate or inaccurate  
   compared to how this object typically appears in \{country\} or your understanding of \{item\_name\}?

Format your response as JSON with keys: {'country\_likelihood'}, {'item\_accuracy'}, and {'details\_analysis'}.
\end{tcolorbox}

\subsection{Analysis of Gemini 2.0 Flash Responses.}
Here we discuss several illustrative examples of Gemini 2.0 Flash performance on our \cure dataset. In \cref{fig:Gemini 2.0 Flash_failures_all}, we show the T2I system generated image (``AI Image'') and representative ground-truth image from Wikimedia (``Real Image'') alongside scores for cultural representativeness (\cure) from our user study $\phi^*_\cure$ and Gemini $\phi^*_{Gemini}$. To evaluate if Gemini can capture correct or incorrect cultural details highlighted by workers who identify with the culture of each cultural artifact, we also write score justifications provided by workers from Prolific (``Human Judments'') as well as Gemini (``MLLM Judgment'').

In \textbf{E1} (\textit{Femi Kuti}), workers unanimously highlight incorrect facial features and hairstyle while Gemini gives a high \cure score based on general body structure and attire, and misses culturally specific identity markers. Similarly in \textbf{E3} (\textit{Chuseok}), workers point out that many details in the image look Chinese (red lanterns, hairstyle, attire, background) and ``not a single theme'' resembles Korean Chuseok. Gemini’s description emphasizes the presence of traditional Korean dress and a village setting as reasons for a high \cure score. \textbf{E4} (\textit{Sámi Headwear}) continues this trend - while Gemini believes that colors, shape, adornments, and textile patterns are accurate to indigenous Sami culture, workers highlight that the subject appears Asian and not Northern European, and the fabric texture, color, and fur are not accurate to Finnish Sami headwear. \textbf{E6} (\textit{Portrait of Amir Kabir}) offers a similar case: Gemini gives a high \cure score, referencing Qajar-era styling. Workers, however, flag specific missing details like the hat, mustache style, and necklace elements that signify historical authenticity.

\textbf{E2} (\textit{Jollof Rice}) and  shows a reverse failure: workers appreciate the image, noting the correct rice type, color, and ingredients. Gemini disagrees with the workers and states that while the item appears local to Nigeria, several details indicate that it may not be faithful to Jollof rice, including grain shape (pasta looking instead of long-grain), ingredients, and garnish.

For \textbf{E5} (\textit{Takht-e Fulad}), Gemini misidentifies the architecture as Egyptian and dismisses its relevance to Iran entirely. Workers believe that while the image might belong to Iranian culture, it is missing details local to buildings from Esfahan (blue ceramic) and may more resemble other historical Iranian sites (Takht-e Jamshid). This example highlights that workers are able to latch on to these nuanced cultural differences, while Gemini's judgment is more binary. \textbf{E7} (\textit{Kiping}) is more straightforward: both Gemini and workers recognize that the T2I system generated image is entirely off, \ie a cartoon animal instead of a food item. This shows that Gemini is able to latch on to more clear cases of T2I system failures. Lastly, in \textbf{E8} (\textit{Pulseira}), Gemini agrees with workers that the design, and aesthetic is plausible for the Brazilian context, even if not ``inherently Brazilian''. Gemini docks a point for \cure, citing the shine and stones of the bracelet may indicate a luxurious unaffordability for ``a larger portion of the (Brazilian) population'', while workers point out that the ``details in the shiny stones'' are in fact quite accurate for Brazilian pulseira.

\paragraph{Regional Trend:}
A notable pattern is that most failure cases involve artifacts from the Global South (Nigeria, Iran). While Gemini 2.0 Flash performs well on globally familiar or culturally neutral items (the ``head''), it struggles with region-specific details that require localized knowledge (\eg mistaking Chinese culture for Korean). This likely reflects the long tail nature of the pretraining data of T2I systems, and highlights the need for more culturally balanced data curation and catering to more heterogeneous preferences to improve alignment with diverse user perspectives.

\begin{figure*}[ht]
    \caption{Visualization of Gemini 2.0 Flash responses where the MLLM either failed or succeeded in identifying nuances compared to humans. Examples were generated using FLUX.1 [dev], SD 3.5 Large, and SD 1.5. (Continued on next page)}
    \label{fig:Gemini 2.0 Flash_failures_all}

    \begin{subfigure}{\textwidth}
        \begin{adjustbox}{width=\textwidth}
            \begin{tabular}{@{}l|c|l@{}}
                \toprule
                \textbf{E\#} & \multicolumn{1}{c|}{\textbf{AI Image} \hspace{3cm} \textbf{Real Image}} & \multicolumn{1}{c}{\textbf{Feedback}} \\
                \midrule
                \textbf{E1} & 
                \begin{tabular}[c]{@{}c@{\hspace{1em}}c@{}}
                    \includegraphics[width=0.24\textwidth]{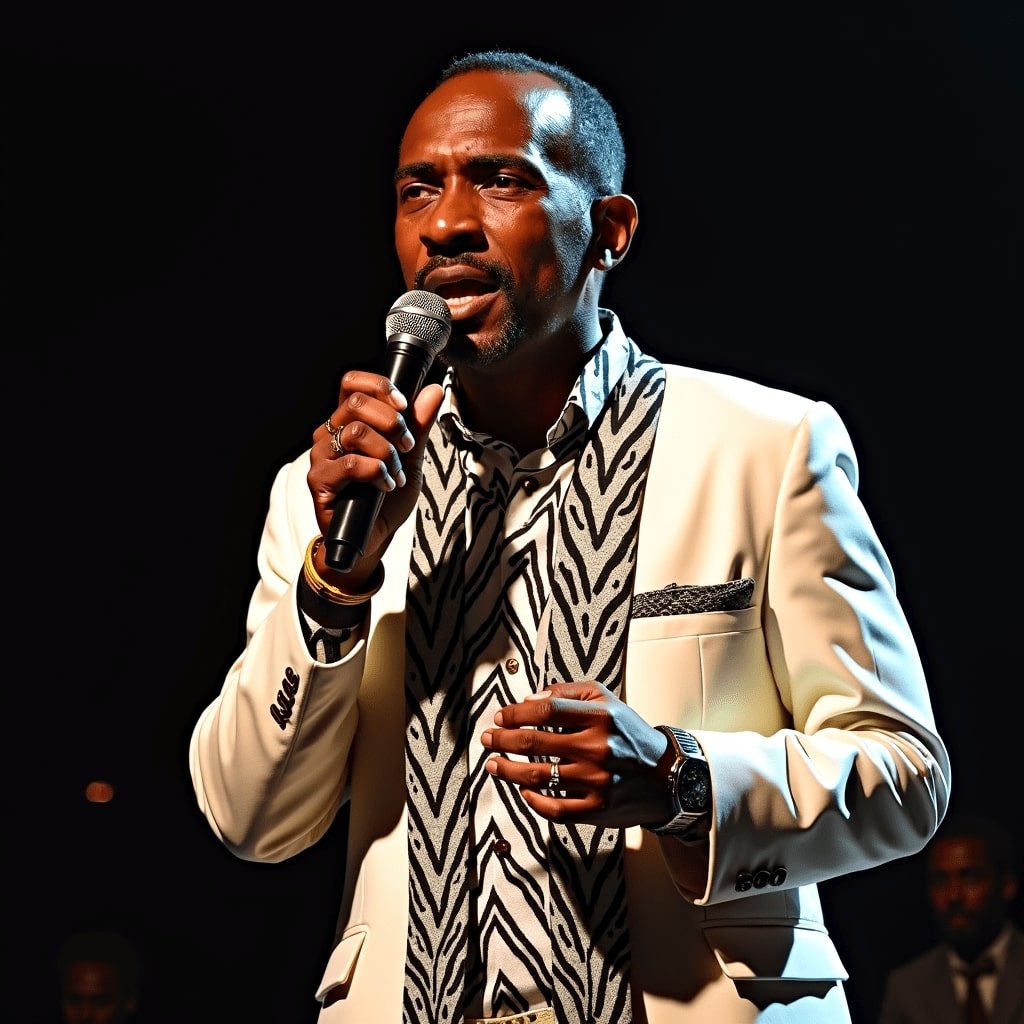} &
                    \includegraphics[width=0.24\textwidth]{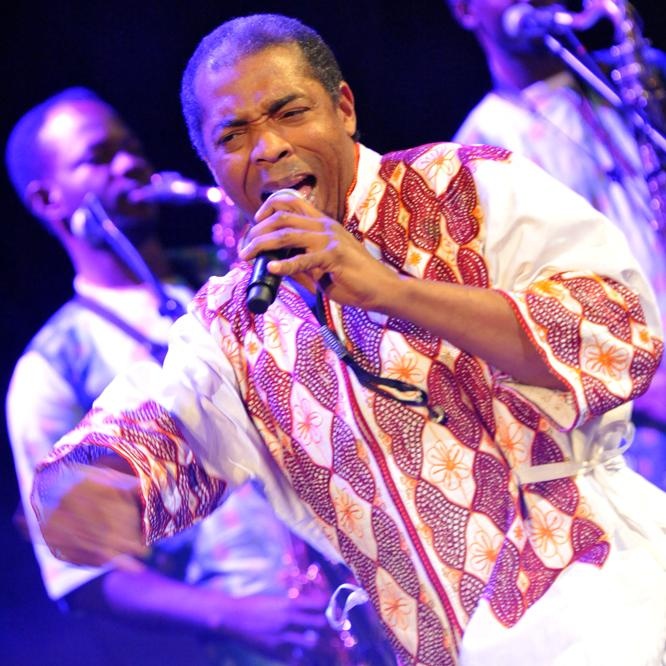} \\
                    \begin{tabular}[t]{@{}l@{}} Artifact: Femi Kuti \\ Category: Musician \\ 
                        $\phi^*_{\cure}: 2.34$ \end{tabular} &
                    \begin{tabular}[t]{@{}l@{}} Region: Nigeria \\ Super Category: People \\ $\phi_{Gemini}$ : 5 \end{tabular}
                \end{tabular}
                & 
                \begin{tabular}[c]{@{}l@{}}
                    \textbf{Human Judgments (User Study):}\\[0.5ex]
                        1. The image is someone holding a microphone but that is not femi kuti \\
                        2. The image seems inaccurate from the images of Kuti I used to see.  \\
                        3. The face is not just right with with me and the hair is just off....Femi has very distinctive blend of grey hair.\\[1ex]
                    \textbf{MLLM Judgment (Gemini 2.0 Flash):}\\[0.5ex]
                    The generated image bears a strong resemblance to Femi Kuti, especially concerning his facial features, skin tone,\\ and general build as seen in the real images. The attire and overall style are also relatively consistent with his public\\ image. However, subtle details like the precise shape of his nose and mouth, as well as the sharpness of the image,\\ differentiate it from the real photos. The generated image's attire is similar to what he might wear but not an exact \\ match of anything in the real images.Considering all these aspects, the similarity is high but not perfect.\\[1ex]
                \end{tabular}
                \\ \midrule
                \textbf{E2} &
                 \begin{tabular}[c]{@{}c@{\hspace{1em}}c@{}}
                    \includegraphics[width=0.24\textwidth]{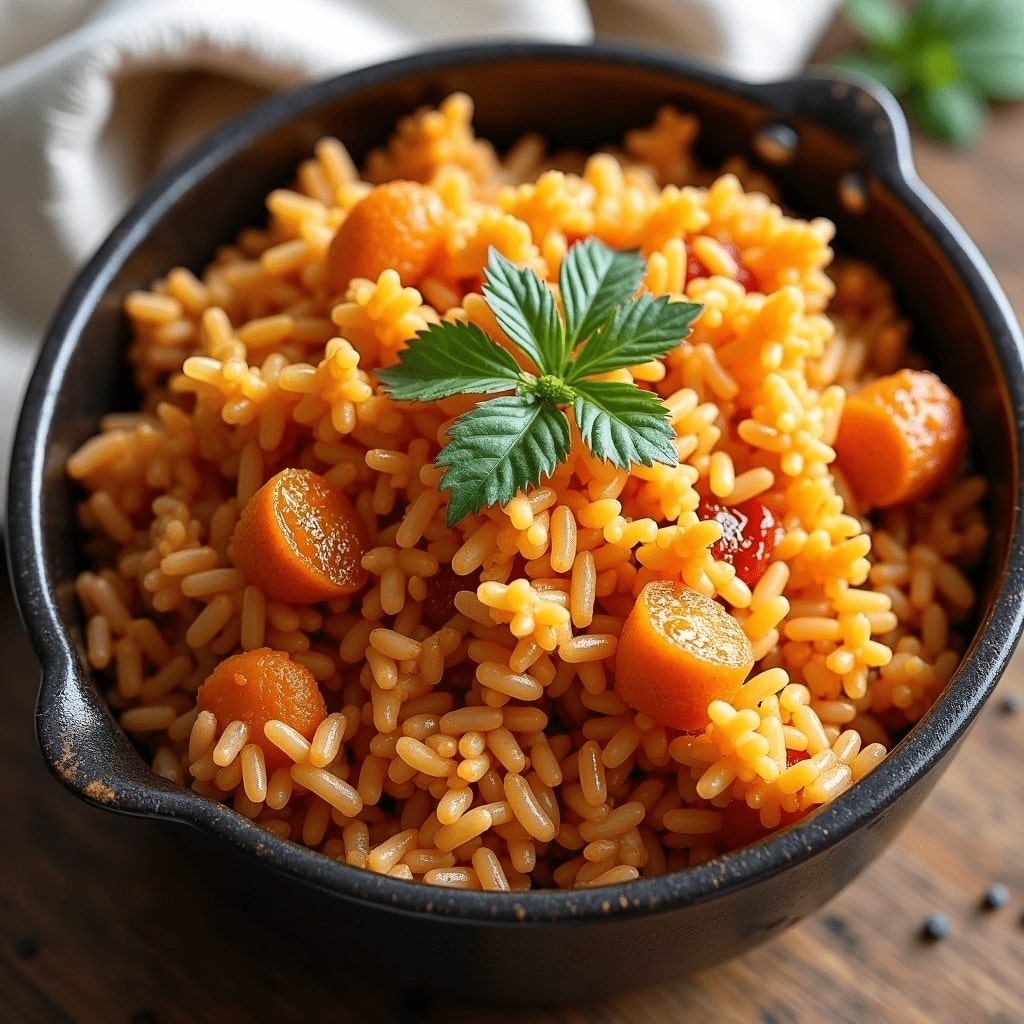} &
                    \includegraphics[width=0.24\textwidth]{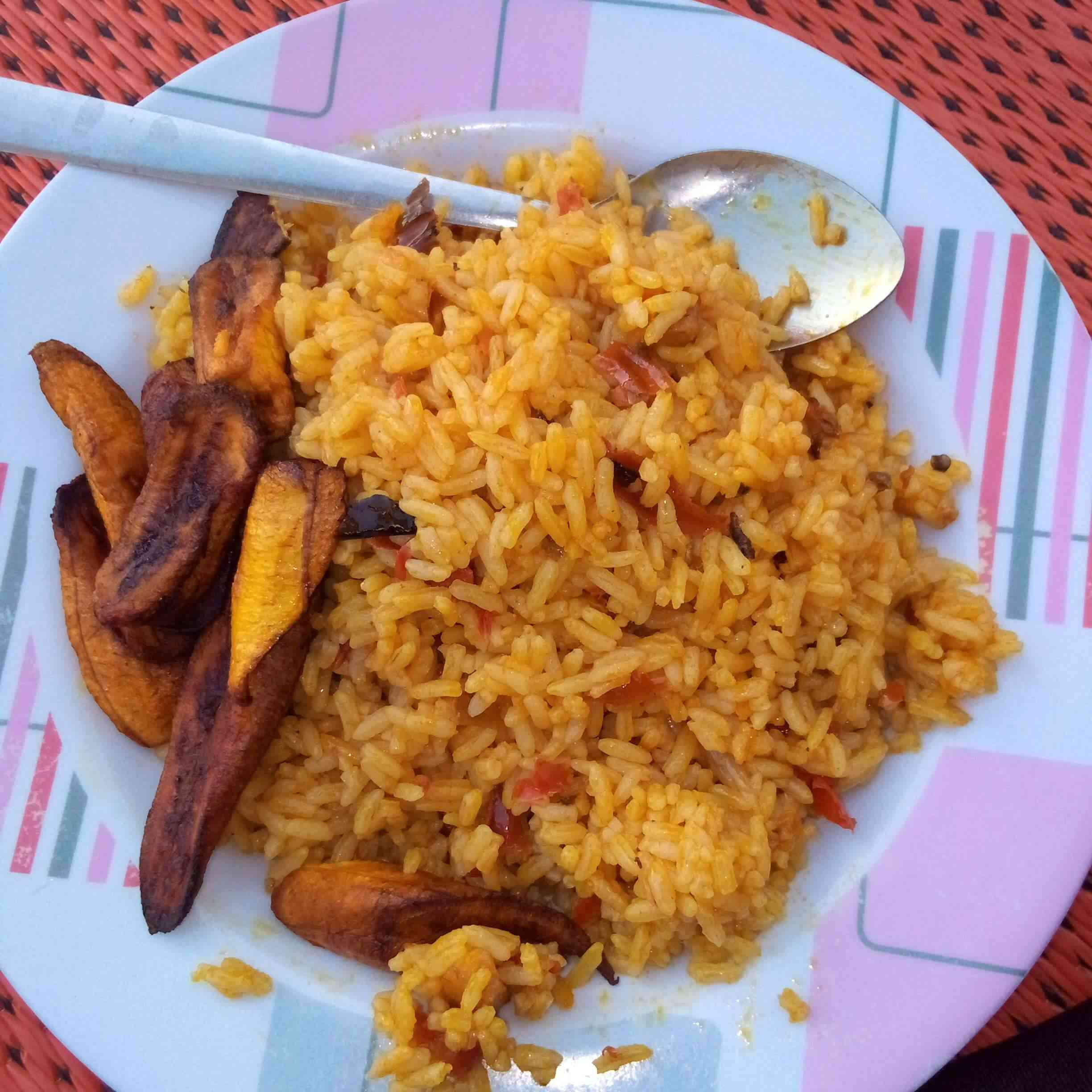} \\
                     \begin{tabular}[t]{@{}l@{}} Artifact: Jollof Rice\\ Category: Rice dish \\ 
                        $\phi^*_{\cure}: 4.67$ 
                     
                     \end{tabular} &
                     \begin{tabular}[t]{@{}l@{}} Region: Nigeria \\ Super Category: Food \\
                     $\phi_{Gemini}$ : 5
                     \end{tabular}
                 \end{tabular}
                 & 
                 \begin{tabular}[c]{@{}l@{}}
                      \textbf{Human Judgments (User Study):}\\[0.5ex]
                         1. The image gets the color of Jollof rice \\
                         2. The rice and the colour of the rice after been cooked makes it look like jollof from my culture.   \\
                         3. The image has used the correct kind of rice, ingredients, and sauce, in  my country\\[1ex]
                     \textbf{MLLM Judgment (Gemini 2.0 Flash):}\\[0.5ex]
                     The item can almost definitely be found in Nigeria. However, the rice itself does not resemble Jollof rice. Jollof rice \\ typically uses long-grain rice, which, after cooking, retains a relatively distinct shape. The rice in the AI-generated  \\ image looks more like orzo pasta with an orange tint. The orange cubes included in the dish do not generally feature \\ in  Jollof rice. It usually includes tomato, pepper, onions, and spices. Also, Jollof rice is not normally garnished with \\ a large green leafy sprig.\\[1ex]
                 \end{tabular}
                \\ \midrule
                \textbf{E3} &
                \begin{tabular}[c]{@{}c@{\hspace{1em}}c@{}}
                     \includegraphics[width=0.24\textwidth]{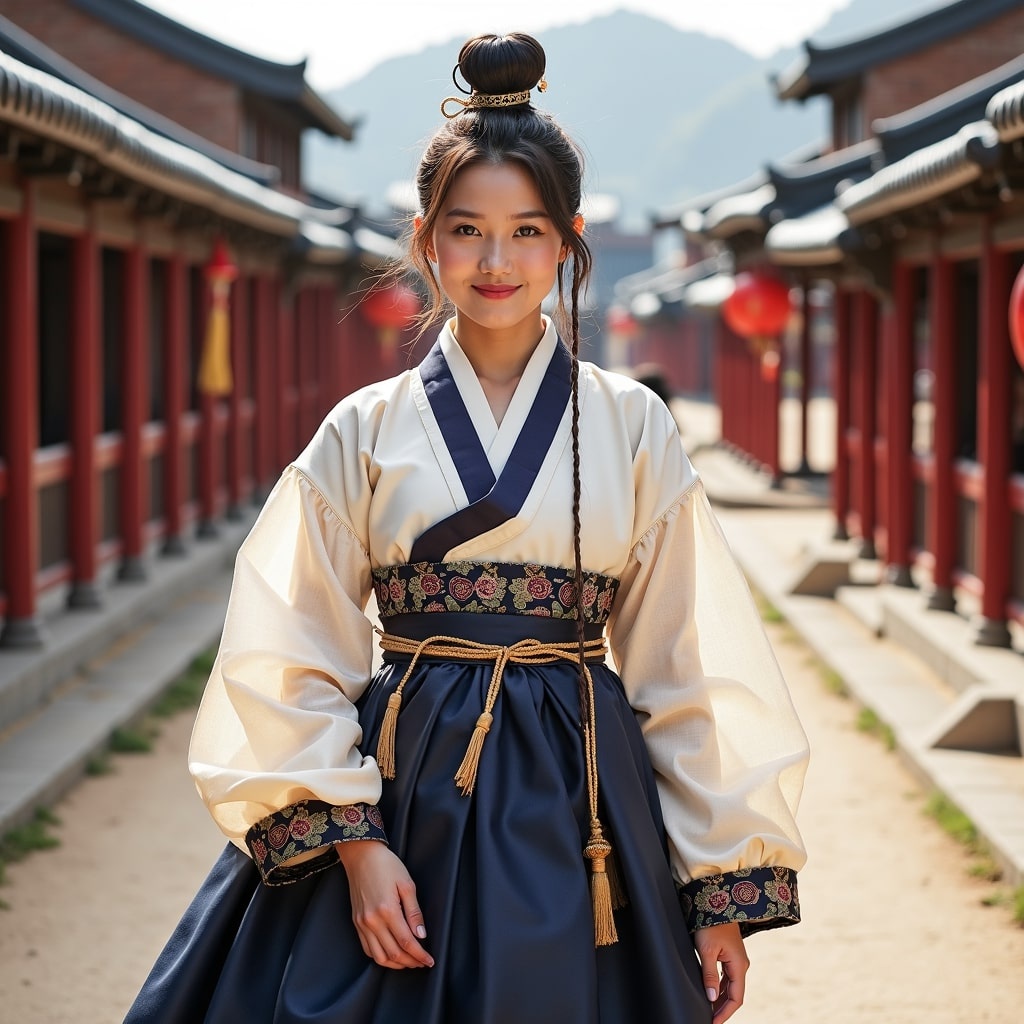} &
                     \includegraphics[width=0.24\textwidth]{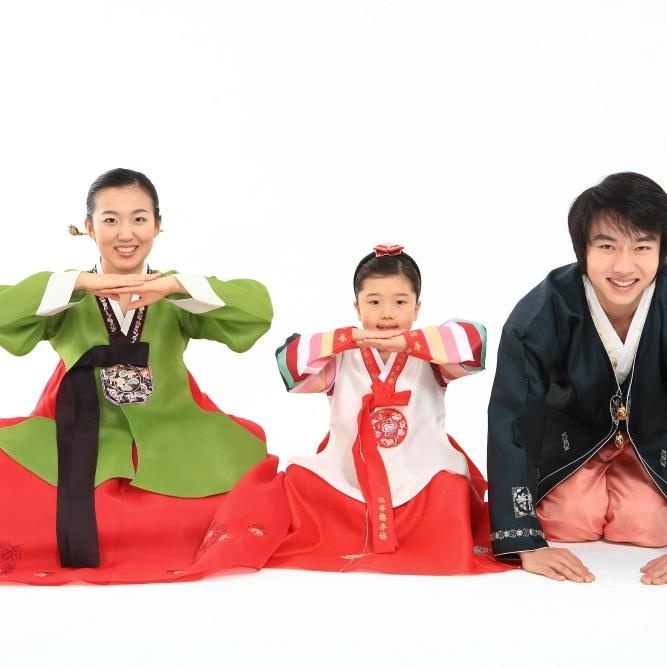} \\
                      \begin{tabular}[t]{@{}l@{}} Artifact: Chuseok \\ Category: Harvest festival \\ 
                         $\phi^*_{\cure}: 1.67$ 
                        \end{tabular} &
                      \begin{tabular}[t]{@{}l@{}} Region: South Korea \\ Super Category: Celebrations \\ $\phi_{Gemini}$ : 5 \end{tabular}
                \end{tabular}
                & 
                \begin{tabular}[c]{@{}l@{}}
                    \textbf{Human Judgments (User Study):}\\[0.5ex]
                        1. I feel like it's close to getting it right, but the patterns on the waist and sleeves, for example, makes it look like almost \\ we'd never see something like this. It'd be like, let's say we are trying to find certain colored horses, but perhaps AI \\generates the horses with patterns of color never found in nature. \\
                        2. The hairstyle in the picture is more Chinese than Korean. Also, the detail of the traditional Korean clothing is not \\ correct. There are also a number of red lanterns in the background, which is also more Chinese than Korean. I cannot \\ find a single theme in the photo that is related to Korean Chuseok.\\
                        3. The figure in the image is where wrong traditional attire and the background architectures also looks like Chinese. \\[1ex]
                    \textbf{MLLM Judgment (Gemini 2.0 Flash):}\\[0.5ex]
                    The image depicts a woman in a Hanbok, a traditional Korean dress, likely taken in an area resembling a traditional\\ Korean village. Hanboks are commonly worn in Korea, especially during festivals like Chuseok. The likelihood of\\ finding a Hanbok in Korea is extremely high. The accuracy as an image of Chuseok is also high. Hanboks are \\ commonly worn during Chuseok. Some details make the image accurate, such as the Hanbok itself, which is generally \\ accurate, and the setting, which suggests a traditional Korean village. Some minor inaccuracies include the specific\\ style of the Hanbok, which seems a bit stylized, and it does not include other aspects related to Chuseok, such as food \\ or activities.\\[1ex]
                \end{tabular}
                \\ \midrule
                \textbf{E4} &
                \begin{tabular}[c]{@{}c@{\hspace{1em}}c@{}}
                     \includegraphics[width=0.24\textwidth]{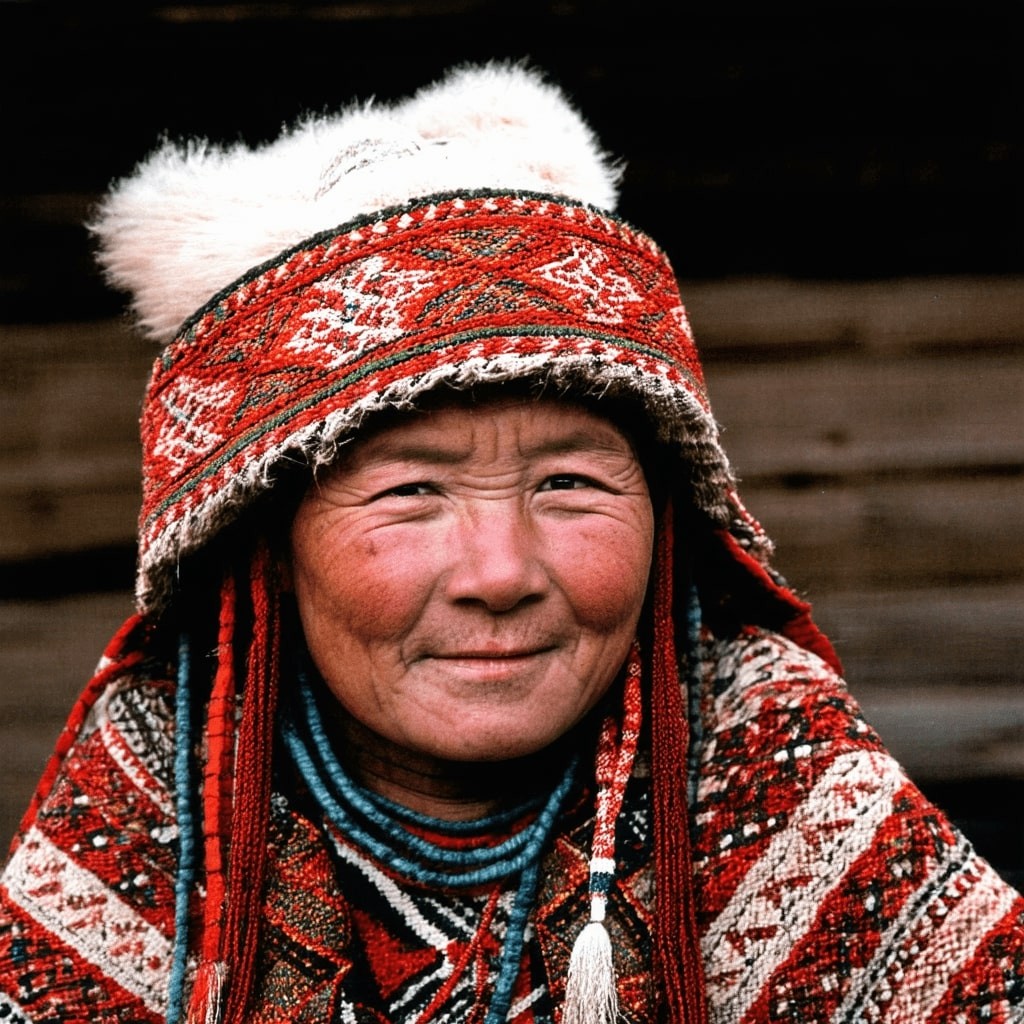} &
                     \includegraphics[width=0.24\textwidth]{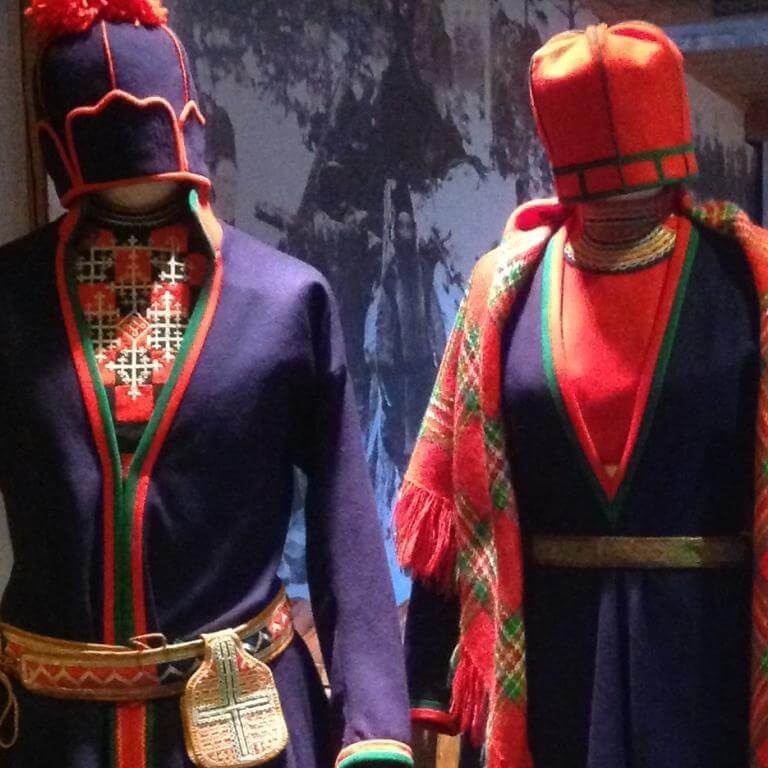} \\
                      \begin{tabular}[t]{@{}l@{}} Artifact: Sami headwear\\ Category: Hat \\ 
                         $\phi^*_{\cure}: 2$ 
                      \end{tabular} &
                      \begin{tabular}[t]{@{}l@{}} Region: Finland \\ Super Category: Fashion \\
                      $\phi_{Gemini}$ : 5\end{tabular}
                \end{tabular}
                & 
                \begin{tabular}[c]{@{}l@{}}
                    \textbf{Human Judgments (User Study):}\\[0.5ex]
                        1. The person wearing the hat doesn't look Northern European.\\
                        2. The image has used an outfit that other indigenous people from northern hemisphere might use, but based on my\\ experience it doesn't look extremely similar to Finnish Sami headwear. The fabric texture and colors are different\\ from what I have seen in real life.  \\
                        3. The woman in the pictute is asian and Sami people aren't asian. The hat is too wide and the fur is not supposed to \\ be on top of the hat.\\[1ex]
                    \textbf{MLLM Judgment (Gemini 2.0 Flash):}\\[0.5ex]
                    The image shows a person wearing Sami headwear, specifically a hat with intricate patterns and a possible fur trim\\ at the top. The colors are primarily red, white, and green, which are commonly found in Sami designs. The overall\\ shape of the hat, along with the additional adornments like the colorful braids and beads around the neck, matches\\ the traditional appearance of Sami attire. The designs also appear to be consistent with Sami textile patterns, which\\ are often geometric and symbolic. Given that the Sami people inhabit areas within Finland, it is extremely likely that\\ this type of headwear can be found there. The overall image appears to be an accurate representation of Sami headwear.\\[1ex]
                \end{tabular}
                \\ \bottomrule 
            \end{tabular}
        \end{adjustbox}
        \caption{Examples 1–4 of Gemini 2.0 Flash failure cases (Examples 1–3 generated with FLUX.1 [dev], and Example 4 with SD 3.5 Large).} 
        \label{subfig:Gemini 2.0 Flash_failures_part1} 
    \end{subfigure}
    
\end{figure*}

{\clearpage}

\begin{figure*}[t]
    \ContinuedFloat
    \caption[]{Continued from previous page.}

    \begin{subfigure}{\textwidth}
        \begin{adjustbox}{width=\textwidth}
            \begin{tabular}{@{}l|c|l@{}}
                 \toprule 
                 \textbf{E\#} & \multicolumn{1}{c|}{\textbf{AI Image} \hspace{3cm} \textbf{Real Image}} & \multicolumn{1}{c}{\textbf{Feedback}} \\
                 \midrule
                 \textbf{E5} &
                 \begin{tabular}[c]{@{}c@{\hspace{1em}}c@{}}
                    \includegraphics[width=0.24\textwidth]{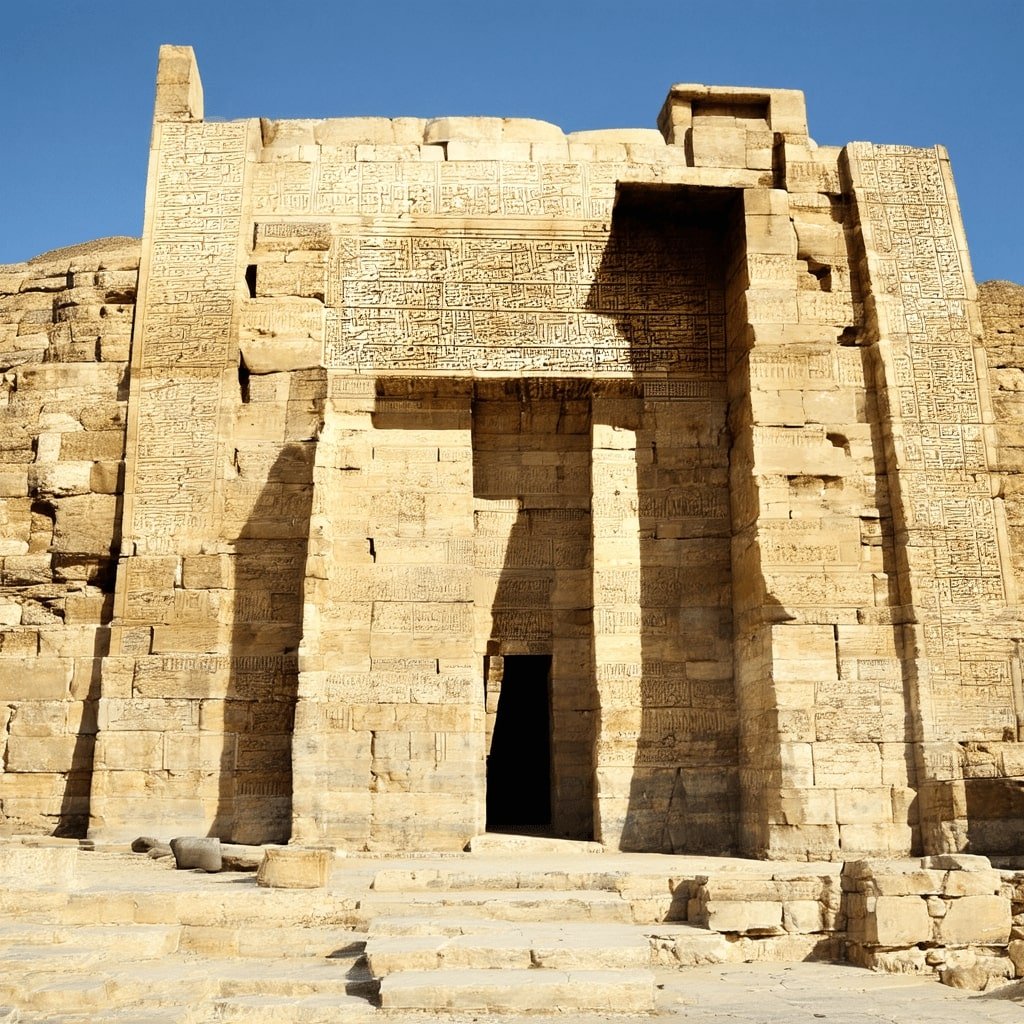} &
                    \includegraphics[width=0.24\textwidth]{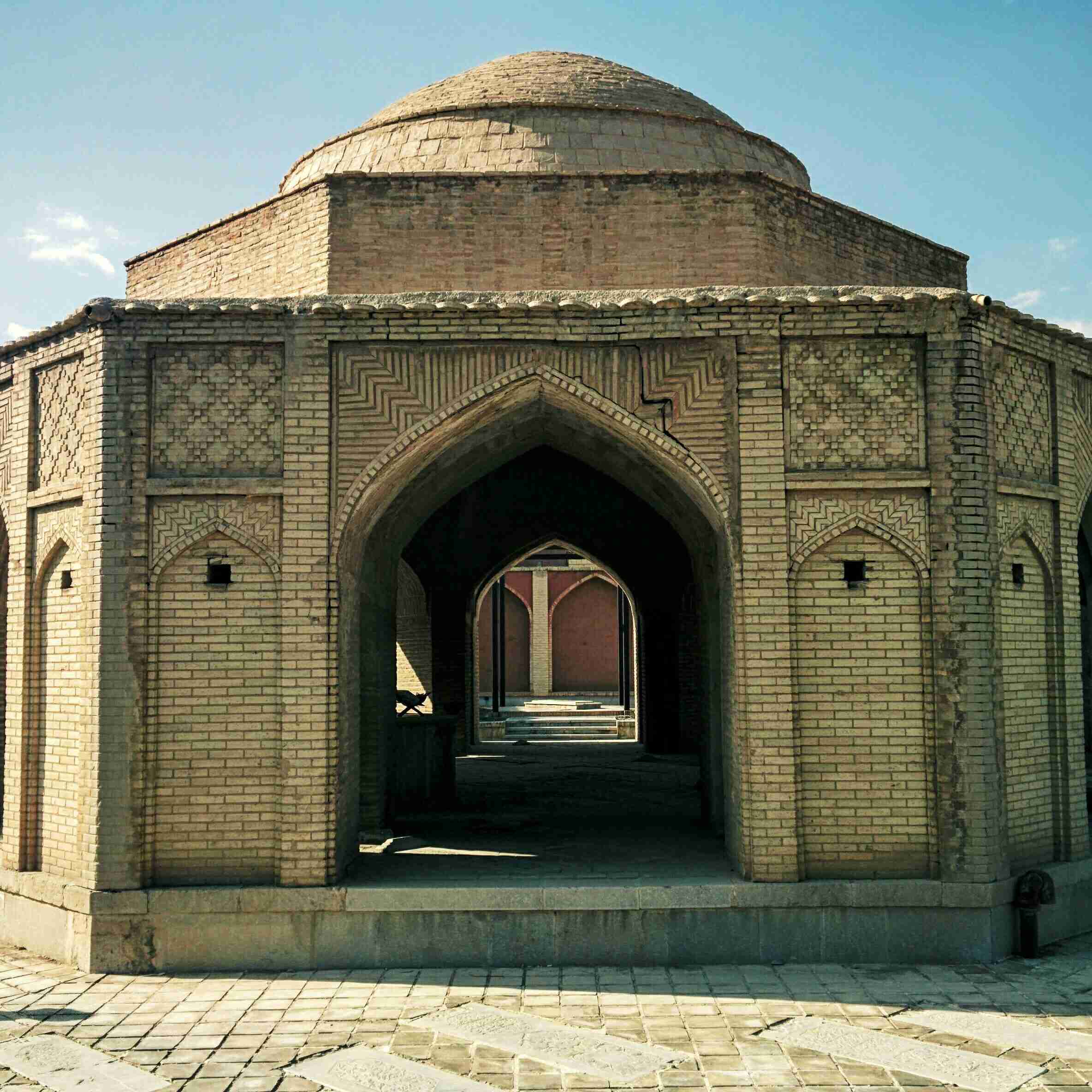} \\
                     \begin{tabular}[t]{@{}l@{}} Artifact: Takht-e Fulad\\ Category:Monument/Memorial \\
                         $\phi^*_{\cure}: 3.67$
                     \end{tabular} &
                     \begin{tabular}[t]{@{}l@{}} Region: Iran \\ Super Category: Architecture \\
                     $\phi_{Gemini}$ : 1
                     \end{tabular}
                \end{tabular}
                & 
                \begin{tabular}[c]{@{}l@{}}
                    \textbf{Human Judgments (User Study):}\\[0.5ex]
                        1. so the image has certain specification like carvings are quite similar, the stone types also similar to the one I saw \\ in my country, however I think this image is closer to Takhte-Jamshid, as it does not have blue ceramic  which usually \\  be used in Esfahan an is one of the icons of any munument in Esfahan. \\
                        2. The coloring is different and Isfahanian buildings have different sets of colors and style more colorful. \\
                        3. The image is so similar to the type of historical buildings in Iran.\\[1ex]
                    \textbf{MLLM Judgment (Gemini 2.0 Flash):}\\[0.5ex]
                    The image shows a large stone structure with hieroglyphs, reminiscent of ancient Egyptian architecture. Takht-e Fulad, \\ in contrast, is a historical cemetery in Isfahan, Iran, which primarily consists of tombstones, mausoleums, and intricate \\ stonework with Persian inscriptions, not large temples. Therefore, it is highly unlikely that this image depicts something\\ found in Iran or an accurate representation of Takht-e Fulad.\\[1ex]
                \end{tabular}
                \\ \midrule
                \textbf{E6} &
                \begin{tabular}[c]{@{}c@{\hspace{1em}}c@{}}
                    \includegraphics[width=0.24\textwidth]{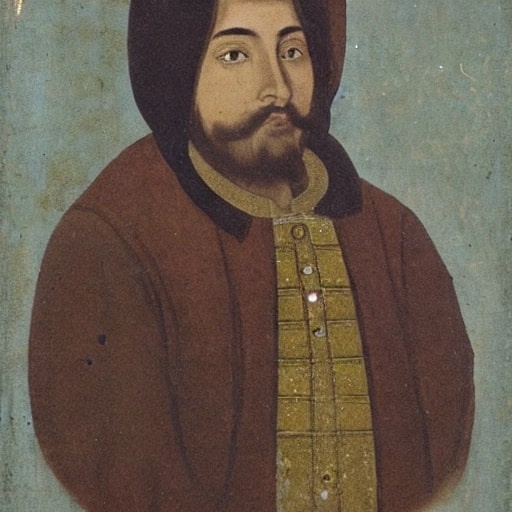} &
                    \includegraphics[width=0.24\textwidth]{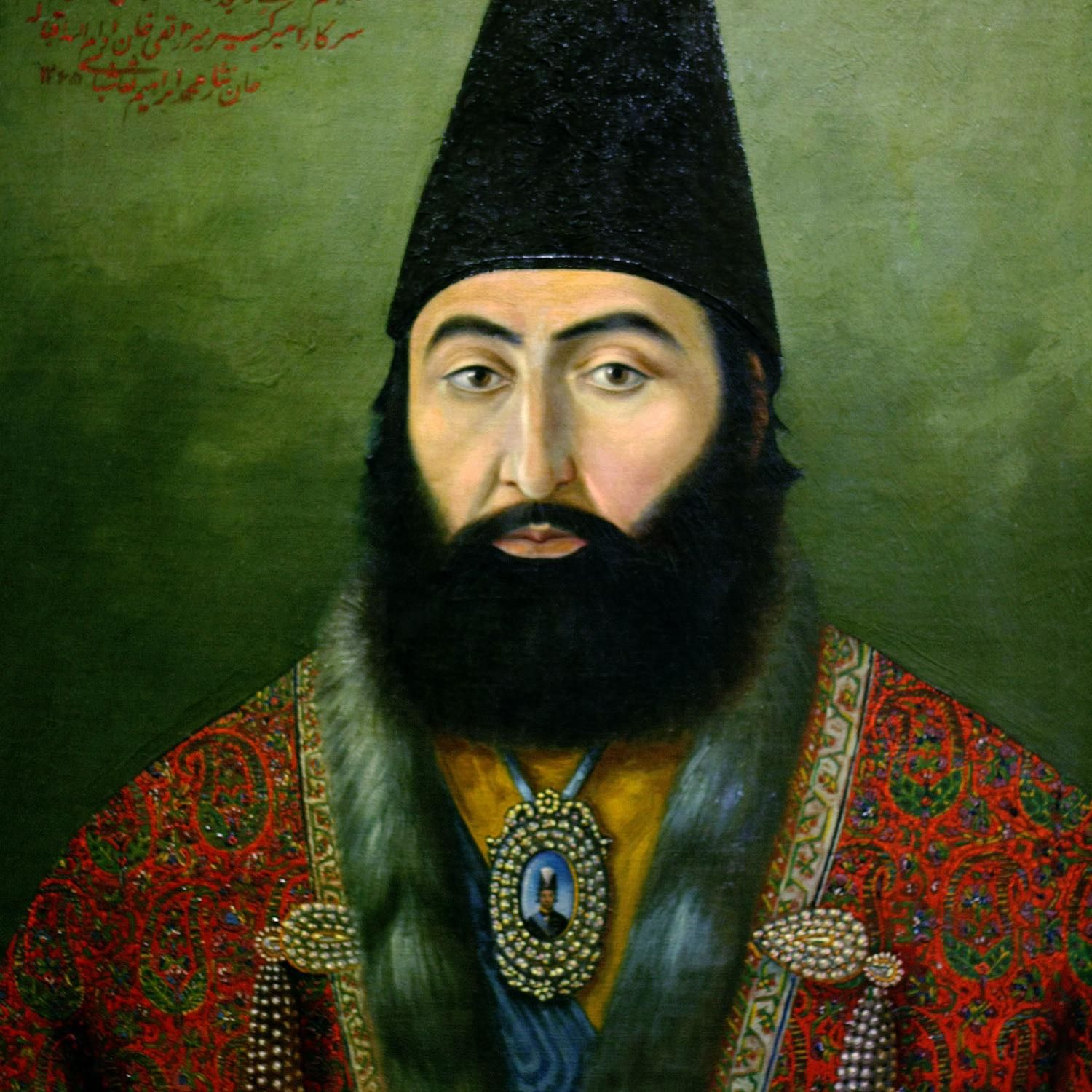} \\
                     \begin{tabular}[t]{@{}l@{}} Artifact: Portrait-Amir Kabir\\ Category: Oil Painting \\ 
                         $\phi^*_{\cure}: 3$ \end{tabular} &
                     \begin{tabular}[t]{@{}l@{}} Region: Iran \\ Super Category: Art \\ $\phi_{Gemini}$ : 5 \end{tabular}
                \end{tabular}
                & 
                \begin{tabular}[c]{@{}l@{}}
                    \textbf{Human Judgments (User Study):}\\[0.5ex]
                        1. The only thing similar to Amir Kabir AI to the real life is the facial structures like nose and eyes. The beard needs \\ improvement as each era in Iranian culture, men used to have different hair and beard styles. The cone shaped hat is \\ important in displaying who this person was and the power he had during Qajar in Iran. In addition, the texture of the \\ clothes and the famous necklace are items that AI "must" add to the image made since they are distinct elements that \\ show Persian culture especially for showing a historical figure. \\
                        2. It has the right feature in the face but lack of detail in the outfit and background. \\
                        3. The colour of his mustaches should be darker, and its style is not Iranian and looks more like Mongols and Turks.\\[1ex]
                    \textbf{MLLM Judgment (Gemini 2.0 Flash):}\\[0.5ex]
                    A portrait of Amir Kabir is highly likely to be found in Iran due to his historical importance. The image appears to be an \\ accurate representation, showing him in traditional attire common for the Qajar era. The details like the style of beard, \\ the coat, and the head covering are consistent with known depictions of him. However, judging finer accuracy requires \\ comparison with specific known portraits, but generally, the visual style and features align with expectations.\\[1ex]
                \end{tabular}
                \\ \midrule
                \textbf{E7} &
                \begin{tabular}[c]{@{}c@{\hspace{1em}}c@{}}
                    \includegraphics[width=0.24\textwidth]{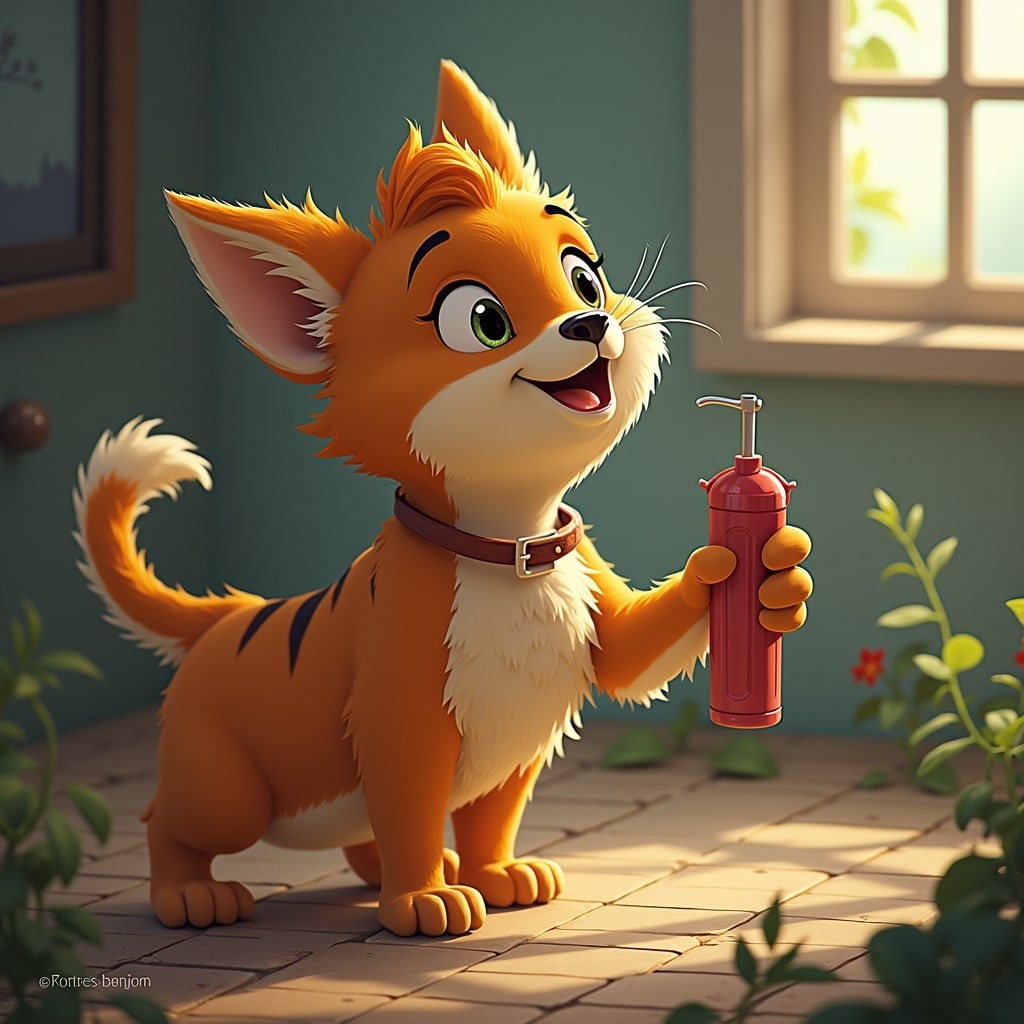} &
                    \includegraphics[width=0.24\textwidth]{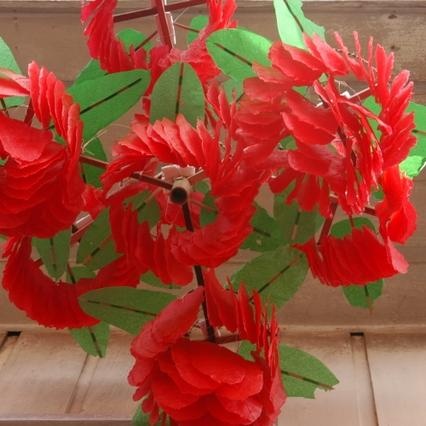} \\
                     \begin{tabular}[t]{@{}l@{}} Artifact: Kiping \\ Category: Flatbread \\ 
                         $\phi^*_{\cure}: 1.34$  \end{tabular} &
                     \begin{tabular}[t]{@{}l@{}} Region: 	Philippines \\ Super Category: Food \\ $\phi_{Gemini}$ : 1 \end{tabular}
                \end{tabular}
                & 
                \begin{tabular}[c]{@{}l@{}}
                    \textbf{Human Judgments (User Study):}\\[0.5ex]
                        1. The image does not portray the meaning of Kiping at all.\\
                        2. The image shown was a type of an animal which is obviously so far from "Kiping" which is a type of food. \\
                        3. The image is an animal holding a spray bottle, it has no connection to kiping.\\[1ex]
                    \textbf{MLLM Judgment (Gemini 2.0 Flash):}\\[0.5ex]
                    The image features a cartoon cat holding what appears to be a small pump or sprayer. There is absolutely nothing in the \\ image resembling Kiping, which is a type of leaf-shaped, colorful wafer made from glutinous rice. Therefore, it's highly \\ unlikely this image depicts Kiping or could be found in the Philippines as a depiction of Kiping. The AI hallucinated a \\ completely different scene. \\[1ex]
                \end{tabular}
                \\ \midrule
                \textbf{E8} &
                 \begin{tabular}[c]{@{}c@{\hspace{1em}}c@{}}
                    \includegraphics[width=0.24\textwidth]{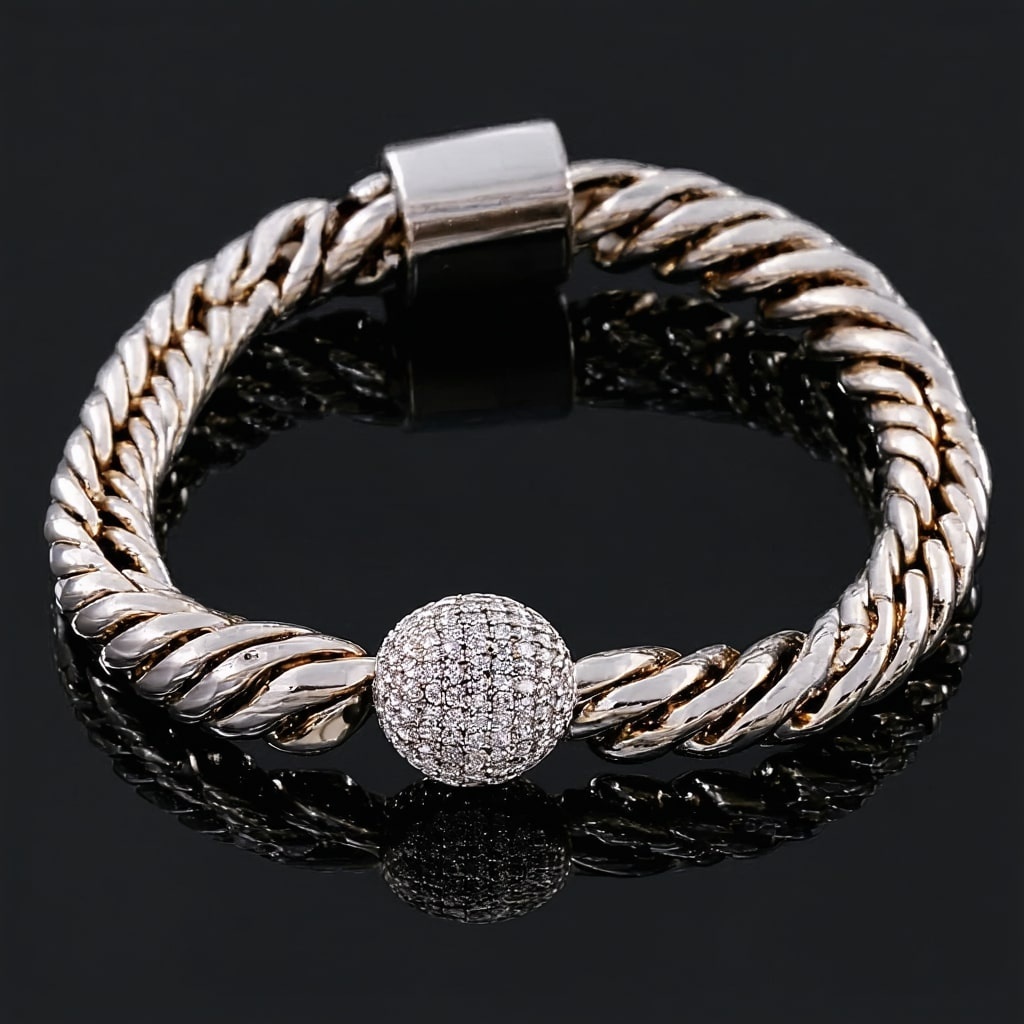} &
                    \includegraphics[width=0.24\textwidth]{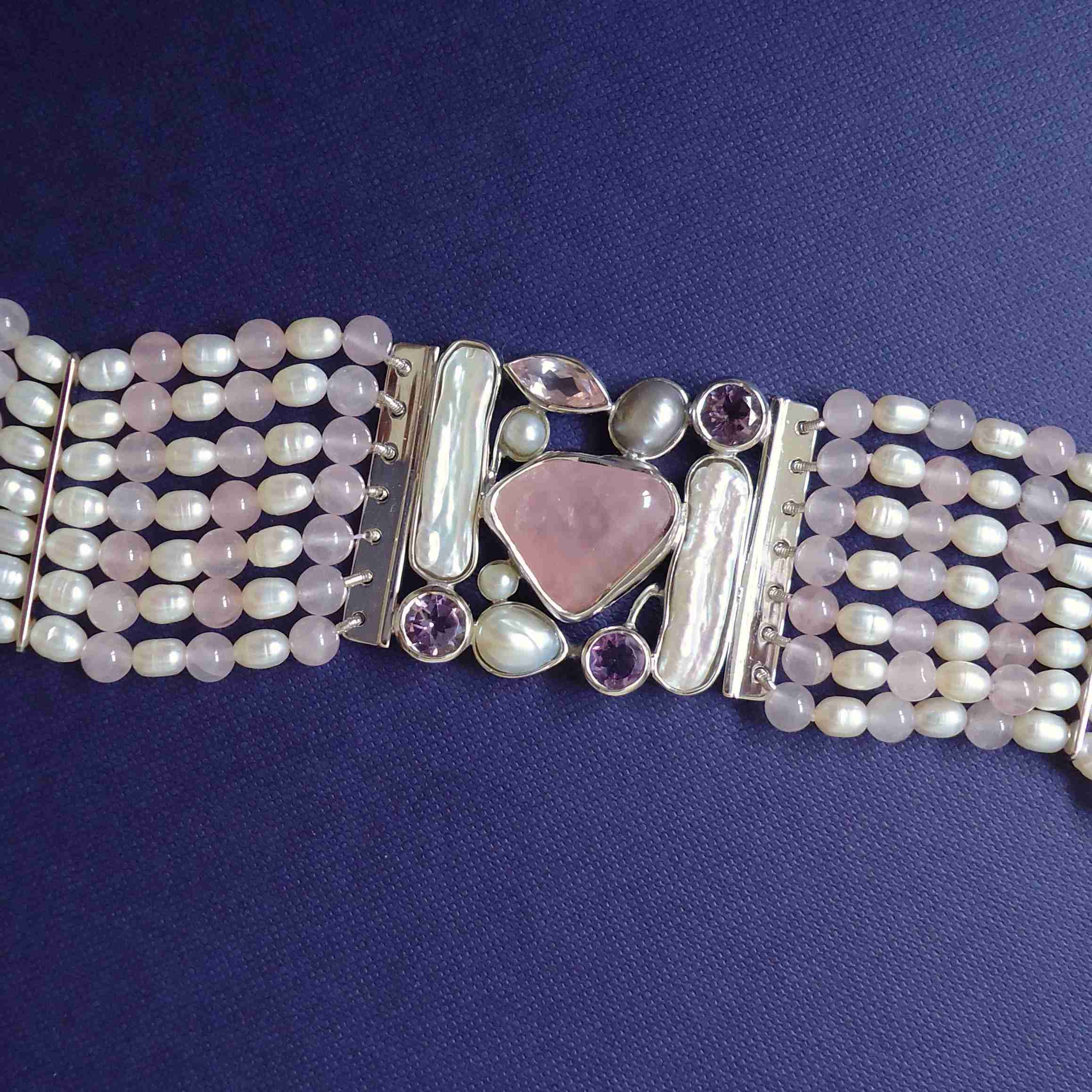} \\
                     \begin{tabular}[t]{@{}l@{}} Artifact: Pulseira \\ Category: Jewellery \\ 
                         $\phi^*_{\cure}: 5$
                     \end{tabular} &
                     \begin{tabular}[t]{@{}l@{}} Region: Brazil \\ Super Category: Fashion \\ $\phi_{Gemini}$ : 4\end{tabular}
                 \end{tabular}
                 & 
                 \begin{tabular}[c]{@{}l@{}}
                      \textbf{Human Judgments (User Study):}\\[0.5ex]
                         1. This is exactly how a pulseira is, a piece of jewellery that goes on the wrist  \\
                         2. The image is very similar to the name of Pulseira. \\
                         3. This bracelet represents one of the forms of objects that we wear on our wrist, very beautiful with details in shiny stones.\\[1ex]
                     \textbf{MLLM Judgment (Gemini 2.0 Flash):}\\[0.5ex]
                     The image depicts a bracelet with a braided or woven metal band and a spherical centerpiece covered in small stones \\ (possibly diamonds or crystals). The clasp appears to be a cylindrical metal piece. Jewelry of this style, particularly \\ silver-tone metal and crystal/diamond combinations, is widely available in Brazil.  Brazilians enjoy both fine jewelry and \\ more affordable fashion jewelry.  The design isn't inherently 'Brazilian' but the style is commonly found. It could be sold in \\ upscale stores or at street markets. The image accurately represents what a "pulseira" (Portuguese for bracelet) can look like. \\ While the term encompasses a wide variety of bracelet styles, this particular design is plausible and well-executed. The \\ details, such as the braiding of the metal and the studded sphere, are consistent with what one would expect.  The braided \\ metal design is common in bracelets. The overall aesthetic is refined and somewhat luxurious. The clasp type is also \\ reasonable. **Inaccuracy:** There is little that points to this image being highly 'Brazilian', as opposed to generally stylish. \\ The price point may make it difficult for a larger portion of the population to afford. If the center sphere is made of real \\ diamonds, it could be less common than if it were made of crystals. It looks very shiny. The dark and reflective background \\ gives it a commercial aesthetic, which may not immediately invoke Brazil, but is still considered a fairly accurate \\ representation of a pulseira.\\[1ex]
                 \end{tabular}
                \\
                \bottomrule
            \end{tabular}
        \end{adjustbox}
        \caption{Examples 5–6 of Gemini 2.0 Flash failure cases (Example 5: SD 3.5 Large, Example 6: SD 1.5), and Examples 7–8 of successful cases (Example 7: FLUX.1 [dev], Example 8: SD 3.5 Large)} 
        \label{subfig:Gemini 2.0 Flash_failures_part2} 
        \vspace{10mm}
    \end{subfigure}
\end{figure*}
\end{document}